\documentclass[lettersize,journal]{IEEEtran}

\usepackage{graphicx}%
\usepackage{multirow}%
\usepackage{amsmath,amssymb,amsfonts}%
\usepackage{amsthm}%
\usepackage{mathrsfs}%
\usepackage{xcolor}%
\usepackage{textcomp}%
\usepackage{manyfoot}%
\usepackage{booktabs}%
\usepackage{algorithm}%
\usepackage{algorithmicx}%
\usepackage{algpseudocode}%
\usepackage{listings}%

 \usepackage{hyperref}

\usepackage{xcolor,xspace}
\usepackage{amssymb}
\usepackage{nicefrac} 
\usepackage{booktabs}
\usepackage{arydshln}
\usepackage{caption}
\usepackage{subcaption}

\newcommand{\diff}[2]{\frac{\partial #1}{\partial #2}}
\newcommand{\vct}[1]{\ensuremath{\boldsymbol{#1}}}

\newcommand{\set}[1]{\ensuremath{\mathcal{#1}}}

\newcommand{\T}{\ensuremath{\top}}

\newcommand{\argmin}{\operatornamewithlimits{\arg\,\min}}

\newcommand{\myparagraph}[1]{\smallskip \noindent \textbf{#1}}
\newcommand{\ie}{{i.e.}\xspace}
\newcommand{\eg}{{e.g.}\xspace}

\newcommand{\rebuttal}[1]{\textcolor{black}{#1}}

\newcommand{\cifarairplanefrog}{\textit{airplane vs frog}\xspace}
\newcommand{\cifarbirddog}{\textit{bird vs dog}\xspace}
\newcommand{\cifarairplanetruck}{\textit{airplane vs truck}\xspace}
\newcommand{\imagenettetenchtruck}{\textit{tench vs truck}\xspace}
\newcommand{\imagenetteplayerchurch}{\textit{cassette player vs church}\xspace}
\newcommand{\imagenettetenchparachute}{\textit{tench vs parachute}\xspace}

\definecolor{cyan}{rgb}{0.0, 0.55, 0.55}
\definecolor{green}{rgb}{0.0, 0.5, 0.0}

\newcommand{\sv}[1]{\textcolor{blue}{#1}}

\newcommand{\expnumber}[2]{{#1}\mathrm{e}{#2}}

\renewcommand{\set}[1]{\ensuremath{\mathcal{#1}}}
\renewcommand{\T}{\ensuremath{\top}}

\theoremstyle{thmstyleone}%
\theoremstyle{thmstyletwo}%

\theoremstyle{thmstylethree}%

\def\BibTeX{{\rm B\kern-.05em{\sc i\kern-.025em b}\kern-.08em
    T\kern-.1667em\lower.7ex\hbox{E}\kern-.125emX}}
\usepackage{balance}

\usepackage[numbers,sort&compress]{natbib}

\begin{document}
\title{Backdoor Learning Curves: Explaining Backdoor Poisoning Beyond Influence Functions}

\author{Antonio~Emanuele~Cinà,~\IEEEmembership{Member,~IEEE}, 
        Kathrin~Grosse,
        Sebastiano~Vascon,
        Ambra~Demontis, 
        Battista~Biggio~\IEEEmembership{Fellow,~IEEE},
        Fabio~Roli,~\IEEEmembership{Fellow,~IEEE}, 
        and~Marcello~Pelillo~\IEEEmembership{Fellow,~IEEE}
}

\markboth{Preprint. Paper Accepted at the International Journal of Machine Learning and Cybernetics}%
{Preprint}

\maketitle

\begin{abstract}
Backdoor attacks inject poisoning samples during training, with the goal of forcing a machine learning model to output an attacker-chosen class when presented a specific trigger at test time. Although backdoor attacks have been demonstrated in a variety of settings and against different models, the factors affecting their effectiveness are still not well understood.
In this work, we provide a unifying framework to study the process of backdoor learning under the lens of incremental learning and influence functions.
We show that the effectiveness of backdoor attacks depends on: (i) the complexity of the learning algorithm, controlled by its hyperparameters; (ii) the fraction of backdoor samples injected into the training set; and (iii) the size and visibility of the backdoor trigger. These factors affect how fast a model learns to correlate the presence of the backdoor trigger with the target class. Our analysis unveils the intriguing existence of a region in the hyperparameter space in which the accuracy on clean test samples is still high while backdoor attacks are ineffective, thereby suggesting novel criteria to improve existing defenses.
\end{abstract}


\section{Introduction}\label{sec1}
Machine learning models are vulnerable to backdoor poisoning ~\cite{gu2019badnets,liu2017trojaning,Cina2022Survey, cina2022MLSecurity}. 
These attacks consist of injecting poisoning samples at training time, with the goal of forcing the trained model to output an attacker-chosen class when presented with a specific trigger at test time, while working as expected otherwise. To this end, the poisoning samples typically need not only to embed such a \textit{backdoor trigger} themselves, but also to be labeled as the attacker-chosen class.
As backdoor poisoning preserves model performance on clean test data, it is not straightforward for the victim to realize that the model has been compromised.

Backdoor poisoning has been demonstrated in a plethora of scenarios~\cite{cina2022MLSecurity,Cina2022Survey,cina2022energy}.
In the most common scenario, the user is assumed to download a pre-trained, backdoored model from an untrusted source, to subsequently fine-tune it on their data~\cite{Cina2022Survey}. As backdoors typically remain effective even after this fine-tuning step, they may successfully be exploited by the attacker at test time~\cite{gu2019badnets,liu2017trojaning}.
Alternatively, the attacker is assumed to alter part of the training data collected by the user, either to train the model from scratch or to fine-tune a pre-trained model via transfer learning~\cite{Chen2017TargetedBA,saha2020hidden}. In such cases, the training labels of poisoning samples may be either fully controlled by the attacker~\cite{Chen2017TargetedBA} or subject to validation by the victim~\cite{saha2020hidden}, depending on the considered threat model. 
Backdoor attacks have been shown to be effective in different applications and against a variety of models, including vision~\cite{gu2019badnets} and language models~\cite{chen2020badnl}, graph neural networks~\cite{xi2021graph}, and even reinforcement learning~\cite{kiourti2020trojdrl}.
However, it is still unclear which factors influence the ability of a model to learn a backdoor, i.e., to classify the test samples containing the backdoor trigger as the attacker-chosen class.

In this work, we analyze the process of  \emph{backdoor learning} to identify the main factors affecting the vulnerability of machine learning models against this attack.
To this end, we propose a framework that characterizes the backdoor learning process. The learning process of humans is usually portrayed using learning curves, a graphical representation of the relationship between the hours spent practicing and the proficiency to accomplish a given task. Inspired by this concept, we introduce the notion of \textit{backdoor learning curves} (Section~\ref{sec:backdoor-curves}). 
To generate these curves, we formulate backdoor learning as an incremental learning problem and assess how the loss on the backdoor samples decreases as the target model gradually learns them.
These backdoor learning curves are independent of the threat model as they capture training data and fine-tuning attacks. The slope of this curve, the \textit{backdoor learning slope}, which is connected to the notion of influence functions, quantifies the speed with which the model learns the backdoor samples and hence its vulnerability. 
Additionally, to provide further insights about the backdoor's influence on the learned classifiers, we propose a way to quantify the \textit{backdoor impact on learning parameters}, \ie, how much the parameters of a model deviate from the initial values 
when the model learns a backdoor.

Our experimental analysis (Section~\ref{sec:experiments}) shows that the factors influencing the success of backdoor poisoning are: (i) the fraction of backdoor samples injected into the training data; (ii) the size of the backdoor trigger; and (iii) the complexity of the target model, controlled via its hyperparameters. 
Concerning the latter, our experimental findings reveal a region in the hyperparameter space where models are highly accurate on clean samples while also being robust to backdoor poisoning. This region exists as, to learn a backdoor, the target model is required to increase the complexity of its decision function significantly, and this is not possible if the model is sufficiently regularized. 
This observation will help identify novel criteria to improve existing defenses and inspire new countermeasures.

To summarize, the main contributions of this work are:
\begin{itemize}
    \item We introduce \textit{backdoor learning curves} as a powerful tool to thoroughly characterize the backdoor learning process;
    \item We introduce a metric, named \textit{backdoor learning slope}, to quantify the ability of the classifier to learn backdoors;
    \item We identify three important factors that affect the vulnerability against backdoors;
    \item We unveil a region in the hyperparameter space in which the classifiers are highly accurate and robust against backdoors, which supports novel defensive strategies.
\end{itemize}

We conclude the paper by discussing related work in Section~\ref{sec:relWork}, along with the limitations of our approach and promising future research directions in Section~\ref{sec:concl}.

\section{Backdoor Learning Curves}
\label{sec:backdoor-curves}
In this section, we introduce our framework to characterize backdoor poisoning by means of learning curves and their slope. 
Afterward, we introduce two measures to quantify the backdoor impact on the model's parameters. 
\smallskip

\myparagraph{Notation.} We denote the input data and their labels respectively with $\vct x \in \mathbb{R}^d$ and $y \in \{1,..,c\}$, being $c$ the number of classes. We refer to the untainted, clean training data as $\set D_{\rm tr} =(\vct x_i, y_i)_{i=1}^n$, and to the backdoor samples injected into the training set as $\set P_{\rm tr}=( \hat{\vct x}_j, \hat{y}_j)_{j=1}^m$. We refer to the clean test samples as $\set D_{\rm ts}=( {\vct x}_t, {y}_t)_{t=1}^k$ and to the test samples containing the backdoor trigger as $\set P_{\rm ts}=( \hat{\vct x}_t, \hat{y}_t)_{t=1}^k$. 
\smallskip

\begin{figure*}[t]
\centering

\begin{subfigure}{0.495\textwidth}
\includegraphics[trim=0.1in 0 0.5in 0.35in, clip=true,width=0.45\textwidth]{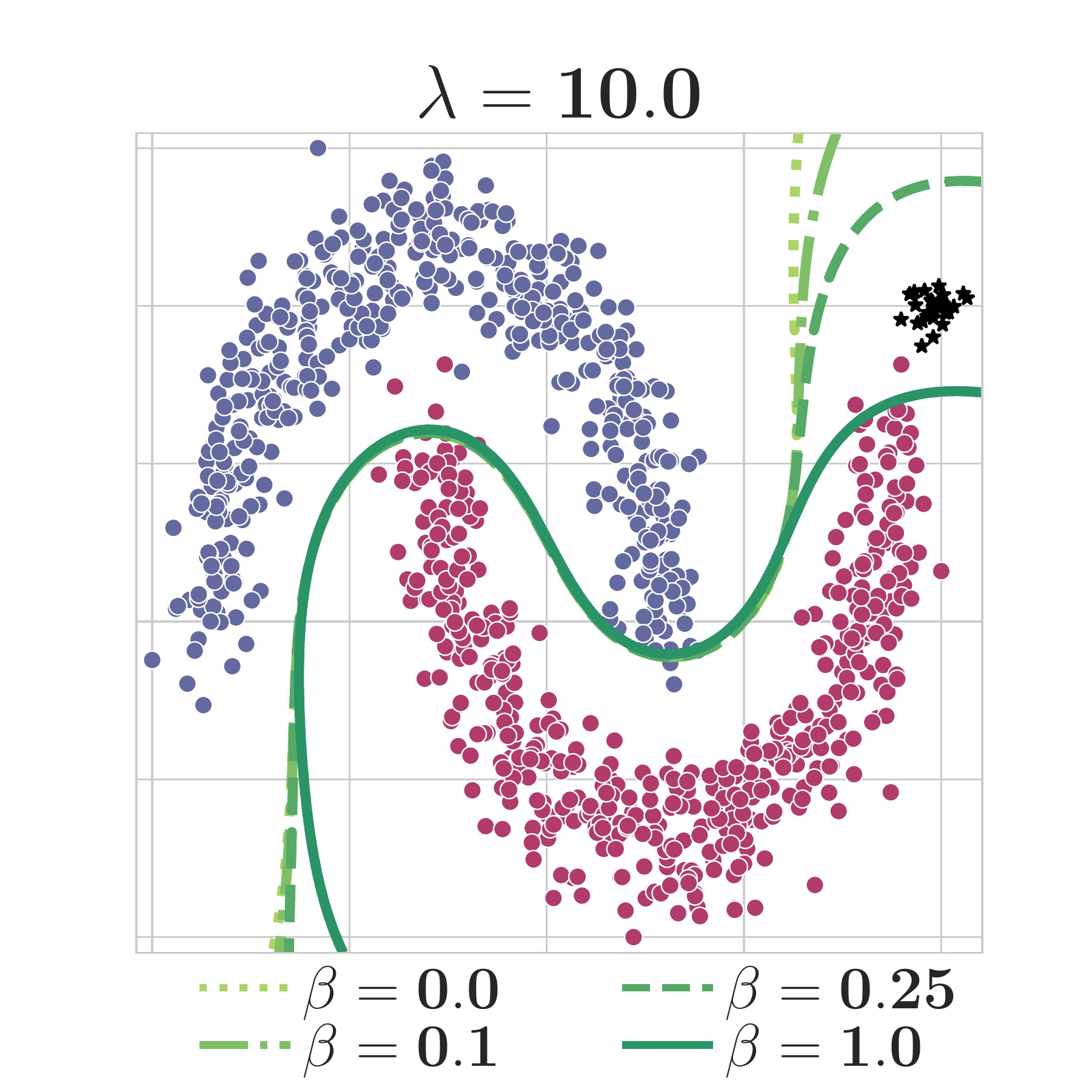}
\includegraphics[trim=0.1in  0 0.5in 0.35in, clip=true,width=0.45\textwidth]{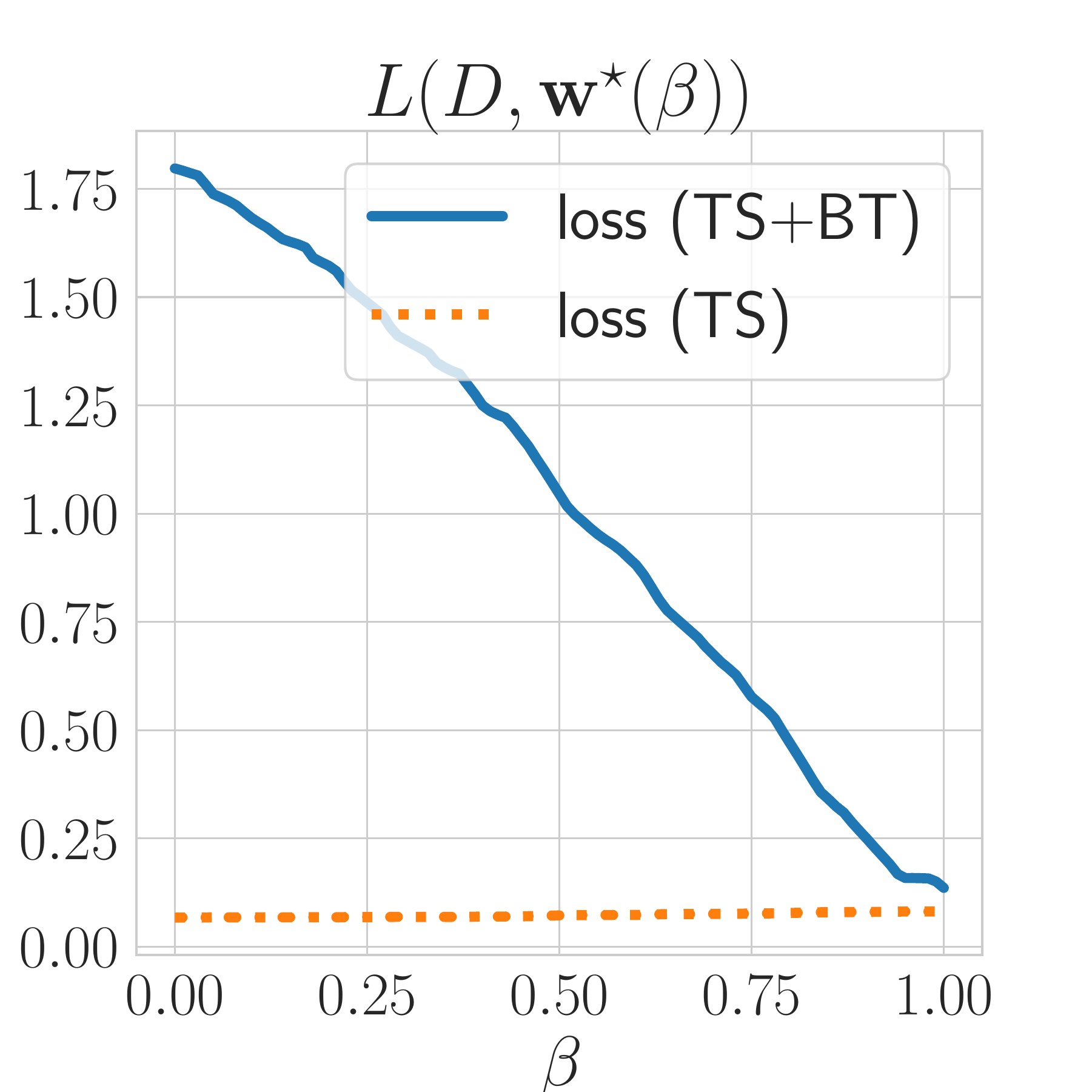}
\caption{\rebuttal{Strong regularization, $\lambda=10~   (C=0.1)$.}}
\end{subfigure}
\begin{subfigure}{0.495\textwidth}
\includegraphics[trim=0.1in  0 0.5in 0.35in, clip=true,width=0.45\textwidth]{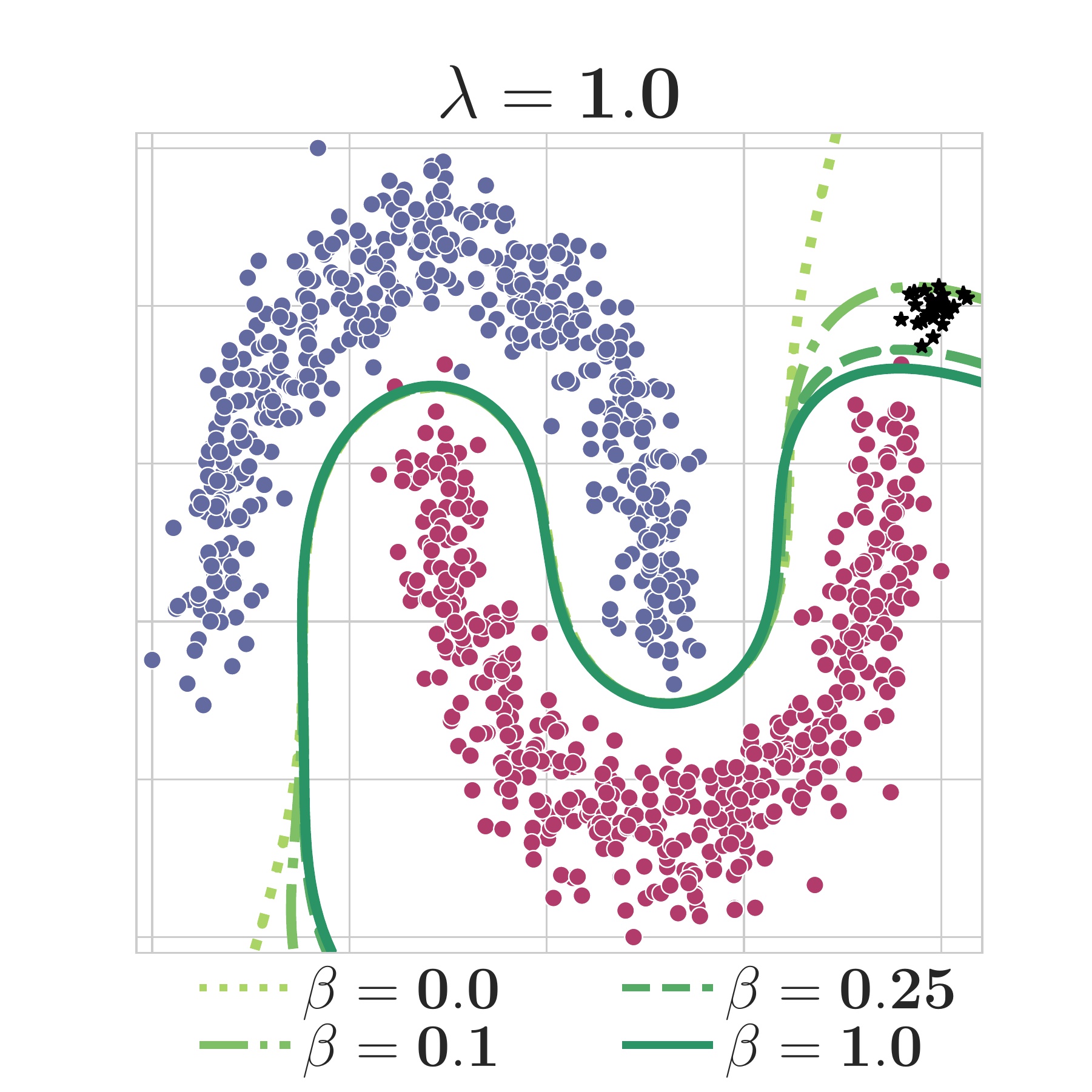}
\includegraphics[trim=0.1in  0 0.5in 0.35in, clip=true,width=0.45\textwidth]{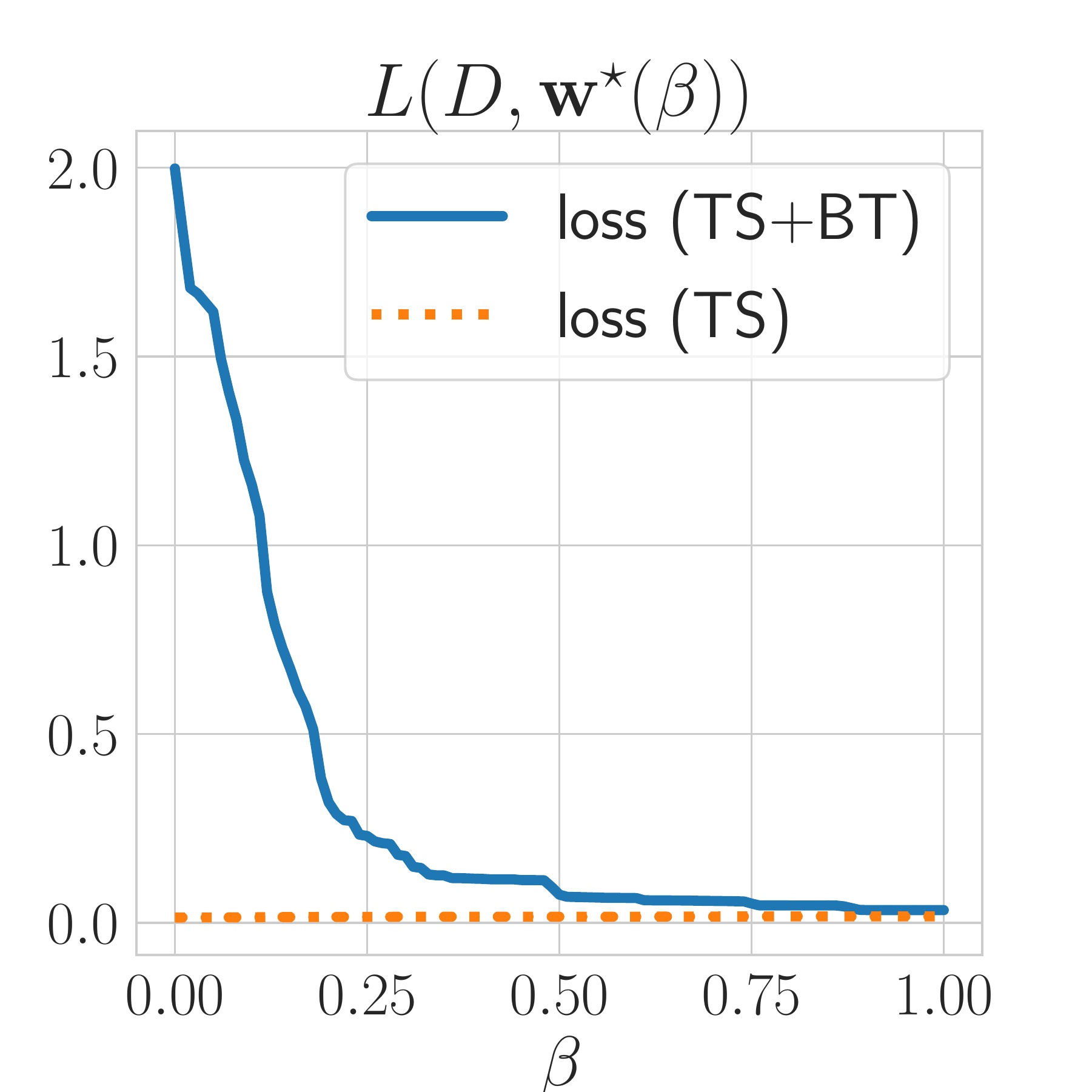} 
\caption{\rebuttal{Weak regularization, $\lambda=1~   (C=1)$.}}
\end{subfigure}
\caption{Backdoor learning curves. Considering an SVM with the RBF kernel ($\gamma = 10$) on a toy dataset in two dimensions, we show the influence of model complexity  \rebuttal{(controlled by the regularization hyperparameter $\lambda=\frac{1}{C}$)} on backdoor learning. For both the strong (\textit{left}) and weak (\textit{right}) regularization settings, we report two plots. The left plot shows the two-dimensional data (dots) and decision surface for different values of $\beta$ (green lines). The right plot shows the backdoor learning curve, i.e. how the loss decreases as $\beta$ ranges from 0 to 1, which amounts to learning the backdoor samples. We plot both the loss on the clean test samples (orange dotted line) and on the test samples with the backdoor trigger (blue line). The slope of these curves represents the speed with which the model learns to classify the backdoor samples (black dots) as blue dots, unveiling that strong regularization slows down such a process.}
\label{fig:backdoor_learning}
\end{figure*}

\myparagraph{Backdoor Learning Curves.} We leverage previous work from incremental learning~\cite{cauwenberghs2001incremental,Hastie2004RegularizationPath} to study how gradually incorporating backdoor samples affects the learned classifier. In mathematical terms, the learning problem can be formalized as:
\begin{align}\label{eq:poisoning_problem_inner}
     \vct w^\star(\beta) \in &\argmin_{\vct w} \set L (\set D_{\rm tr} \cup \set P_{\rm tr}, \vct w)&\\
     &= L (\set D_{\rm tr}, \vct w) + \beta L(\set P_{\rm tr}, \vct w) + \lambda \Omega(\vct w)\nonumber \, ,
\end{align}

where $L$ is the loss attained on a given dataset by the classifier with parameters $\vct w$, and $\set L$ is the loss computed on the training points and the backdoor samples, which also includes a regularization term $\Omega(\vct w)$ (\eg, $\| \vct w\|^2_2 $), weighed by the regularization hyperparameter $\lambda$. To gradually incorporate the backdoor samples $\set P_{\rm tr}$ into the learning process, we introduce the hyperparameter $\beta \in [0,1]$, and increase it from 0 (unpoisoned classifier) to 1 (poisoned classifier). 
As $\beta$ increases, the classifier gradually learns the backdoor by adjusting its parameters; for this reason, we make the dependency of the optimal weights $\vct w^\star$ on $\beta$ explicit as $\vct w^\star(\beta)$.\footnote{Recall that the formulation reported in Eq.~\eqref{eq:poisoning_problem_inner} encompasses many existing learning algorithms, including support vector machines (SVMs), ridge and logistic classifiers. For example, considering either $\beta=0$ or $\beta=1$, the SVM learning problem amounts to minimizing $C \cdot \left (L(\set D_{\rm tr}, \vct w) + \beta L(\set P_{\rm tr}, \vct w)\right)+ \frac{1}{2} \|\vct w\|_2^2$, which is equivalent to our formulation if one sets $L$ to be the hinge loss, $\Omega(\vct w)=\frac{1}{2}\|\vct w\|_2^2$, and $\lambda=\frac{1}{C}$.}

We now define the \textit{backdoor learning curve} as the curve showing the behavior of the classifier loss $L( \set P_{\rm ts}, \vct w^\star(\beta))$ on the test samples with the backdoor trigger as a function of $\beta$. In the following, we abbreviate $L( \set P_{\rm ts}, \vct w^\star(\beta))$ as $L$.
 Intuitively, the faster the backdoor learning curve decreases, the easier the target model is backdoored.
The exact details of how the model is backdoored do not matter for this analysis, e.g. our approach captures for example both the setting where the training data is altered as well as the setting where fine-tuning data is tampered with.
 
We give an example of two such curves under different regularizations in  Figure~\ref{fig:backdoor_learning}. 
The left plots depict a strongly regularized classifier. The corresponding backdoor learning curve (on the right) shows that the classifier achieves low loss and high accuracy on the backdoor samples only after poisoning (when $\beta=1$), i.e. even when the loss on the backdoor samples is considered equally important to the loss on the training samples. The classifier on the right, instead, is less regularized and thus more complex. Consequently, this classifier learns to incorporate the backdoor samples much faster (at low $\beta$), namely when the loss on the backdoor points is taken into account less than the one on the training data. This highlights that this classifier is probably more vulnerable to this attack. \smallskip

\myparagraph{Backdoor Learning Slope.} 
We quantify how fast an untainted classifier can be poisoned by proposing a novel measure, namely the \emph{backdoor learning slope}, that measures the velocity with which the classifier learns to classify the backdoor samples correctly.
This measure allows us to compare the vulnerability of a classifier trained with different hyperparameters or consider different poisoning scenarios (\eg when the attacker can inject a different number of poisoning points or create triggers with different sizes), allowing us to identify factors relevant to backdoor learning. Moreover, as we will show, this measure can be used by the system designer to choose an appropriate combination of hyperparameters for the task at hand. 
To this end, we define the \emph{backdoor learning slope} as the gradient of the backdoor learning curve at $\beta = 0$, capturing the velocity of the curve on learning the backdoor. 
Formally, the \emph{backdoor learning slope} can be formulated as follow:
\begin{equation}
\label{eq:backdoor-learning-slope}
    \diff{L(\set P_{\rm ts}, \vct w^\star(\beta))}{\beta} = \diff{L}{\vct w} \diff{\vct w^\star}{\beta} \, ,
\end{equation}
where the first term is straightforward to compute, and the second term implicitly captures the dependency of the optimal weights on the hyperparameter $\beta$. In other words, it requires us to understand how the optimal classifier parameters change when gradually increasing $\beta$ from 0 to 1, \ie, while incorporating the backdoor samples into the learning process. 

To compute this term, as suggested in previous work in incremental learning~\cite{cauwenberghs2001incremental}, we assume that, while increasing $\beta$, the solution maintains the optimality (Karush-Kuhn-Tucker, KKT) conditions intact. This equilibrium implies that
 $\nabla_\beta \nabla_{\vct w} \set L(\vct w^\star) + \diff{\vct w^\star}{\beta}\nabla^2_{\vct w} \set L(\vct w^\star)  = \vct 0$. Based on this condition, we obtain the derivative of interest,
\begin{equation}
    \label{eq:w-deriv}
    \diff{\vct w^\star}{\beta}  = - (\nabla^2_{\vct w} \set L(\vct w^\star))^{-1} \cdot \nabla_\beta \nabla_{\vct w} \set L(\vct w^\star) \, . 
\end{equation}
Substituting it in Eq.~\ref{eq:backdoor-learning-slope} we obtain the complete gradient:
\begin{equation}
    \diff{L(\set P_{\rm ts}, \vct w^\star(\beta))}{\beta} =  - \nabla_{\vct w} L \cdot (\nabla^2_{\vct w} \set L)^{-1} \cdot \nabla_{\beta} \nabla_{\vct w} \set L \, .
    \label{eq:if_equation}
\end{equation}
The gradient in Eq.~\ref{eq:if_equation} corresponds to the sum of the pairwise influence function values $\set I_{\rm up, loss}(\vct x_{\rm tr}, \vct x_{\rm ts})$ used by Koh et al.~\cite{koh2017understanding}. The authors indeed proposed to measure how relevant a training point is for the predictions of a given test point by computing    $\left.\diff{L}{\beta} \right\vert_{\beta=0} = \sum_t \sum_j \set I_{\rm up, loss}(\hat{\vct x}_t, \hat{\vct x}_j)$. 
To understand how this gradient can be efficiently computed via Hessian-vector products and other approximations,
we refer the reader to~\cite{koh2017understanding} as well as to recent developments in gradient-based bilevel optimization~\cite{pedregosa16-icml,maclaurin15-icml,domke12-aistats}. 
Moreover, we show in Section~\ref{sec:influece-function-example} (Figure~\ref{fig:top_influent_weak}-\ref{fig:top_influent_strong}) an example of the usage of influence functions for weakly and strongly regularized models.

The main difference between the approach by Koh et al.\cite{koh2017understanding} and ours stems from their implicit treatment of regularization and our interest in understanding vulnerability to a subset of backdoor training points, rather than in providing prototype-based explanations. 
However, directly using the gradient of the loss wrt. $\beta$ comes with two disadvantages. First, the slope is inverse to $\beta$, and second, to obtain results comparable across classifiers, we need to rescale the slope. We thus transform the gradient as:
\begin{equation}~\label{eq:backdoor_slope}
    \theta = - \frac{2}{\pi} \arctan{\left ( \left.\diff{L}{\beta} \right\vert_{\beta=0} \right )} \in [-1,1] \, ,
\end{equation}
where we use the negative sign to have positive values correlated with faster backdoor learning (\ie, the loss decreases faster as $\beta$ grows). Computing \nicefrac{2}{$\pi$} of the gradient allows us to rescale the slope to be in the interval between $[-1,1]$.
Hence, a value around $0$ implies that the loss of the backdoor samples does not decrease. In other words, the classifier does not learn the backdoor trigger and is hence very robust.\smallskip

\myparagraph{Backdoor Impact on Learning Parameters.} 
After introducing the previous plot and measure, we can quantify how backdoors are learned by the model. To provide further insights about the backdoor's influence on the learned classifier, we propose to monitor how the classifier's parameters deviate from their initial, unbackdoored values once a backdoor is added. Our approach below captures only convex learners. As shown by Zhang et al.~\cite{Zhang19EqualLayers}, the impact of a network weight in non-convex classifiers' decisions depends on the layer of which it is part. Therefore, measuring the parameter deviation in the non-convex case is challenging, and we leave this unsolved problem for future work.

To capture the backdoor impact on learning parameters in the convex case, we consider the initial weights $\vct w_0 = \vct w^\star(\beta=0)$ and $\vct w_\beta = \vct w^\star(\beta) $ for $\beta>0$, and measure two quantities:
\begin{equation}
 \rho = \| \vct w_\beta \| \in [0, \infty)  , \; {\rm and } \; \nu =  \frac{1}{2} \left(1- \frac{\vct w_0^\T \vct w_\beta}{ \| \vct w_0 \| \| \vct w_\beta \|} \right) \in [0,1] \, .      
\end{equation}
The first measure, $\rho$, quantifies the change of the weights when $\beta$ increases. This quantity is equivalent to the regularization term used for learning. 
The second one, $\nu$, quantifies the change in orientation of the classifier. In a nutshell, we compute the angle between the two vectors and rescale it to be in the interval of $[0,1]$.
Both metrics are defined to grow as $\beta\rightarrow 1$, in other words the backdoored classifier deviates more and more from the original classifier. 
Consequently, in the empirical parameter deviation plots in Section~\ref{sec:experimental_results}, we 
report the value of $\rho(\nu)$ (on the y-axis) as $\beta$ (on the x-axis) varies from 0 to 1, 
to show how the classifier parameters are affected by backdoor learning.


\section{Experiments}\label{sec:experiments}
Employing the previously proposed methodology, we carried out an empirical analysis of linear and nonlinear classifiers.
In this section, we start with the experiments aimed at studying the impact of different factors on backdoor learning. To this end, we employ the backdoor learning curves and the backdoor learning slope to study how the capacity of the model to learn backdoors changes when (a) varying the model's complexity, defined by its hyperparameters, (b) the attacker's strength, defined by the percentage of poisoning samples in the training set and (c) the trigger size and visibility. Our results show that these components significantly determine how fast the backdoor is learned, and consequently, the model's vulnerability. 
Then, leveraging the proposed measures to analyze how the classifier's parameters change during backdoor learning, we provide further insights into the effect of the aforementioned factors on the trained model. 
The results presented in this section will help identify novel criteria to improve existing defenses and inspire new countermeasures. The source code is available on the \href{https://github.com/Cinofix/backdoor_learning_curves}{author's GitHub page}\footnote{\url{https://github.com/Cinofix/backdoor_learning_curves}}.




\subsection{Experimental Setup}~\label{sec:experimental_setup}
Our work investigates which factors influence backdoor vulnerability considering convex learners and neural networks. In the following, we describe our datasets, models, and the backdoor attacks studied in our experiments. \smallskip

\myparagraph{Datasets.}
We carried out our experiments on MNIST~\cite{lecun2010mnist}, CIFAR10~\cite{Krizhevsky09learningmultiple} and Imagenette~\cite{imagenette}\footnote{Imagenette is a subset of 10 classes from Imagenet. We use the 320px version, where the shortest side of each image is resized to that size.}. Supplementary details on the datasets are reported in Appendix~\ref{sec:supplementary_dataset}.

When adopting convex learners, we consider the two-class subproblems as in the work by Saha et al.~\cite{saha2020hidden} and Suya et al.~\cite{suya2021model}. On MNIST, we choose the pairs $7~\rm{vs}~1$, $3~\rm{vs}~0$, and $5~\rm{vs}~2$, as our models exhibited the highest clean accuracy on these pairs. On CIFAR10, analogous to prior work~\cite{saha2020hidden}, we choose \cifarairplanefrog, \cifarbirddog, and \cifarairplanetruck. On Imagenette we randomly choose \imagenettetenchtruck, \imagenetteplayerchurch, and \imagenettetenchparachute. For each two-class subtask, we use $1500$ and $500$ samples as training and test set respectively. In the following section, we focus on the results of one pair on each dataset: $7~\rm{vs}~1$ on MNIST, \cifarairplanefrog ~on CIFAR10, and \imagenettetenchtruck ~on Imagenette. The results of the other pairs (reported in Appendix~\ref{sec:supplementary_experiments}) are analogous.
When testing our framework against neural networks, we train on all ten classes of Imagenette. We use $70\%$ and $30\%$ of the entire dataset for training and testing, respectively.\smallskip

\myparagraph{Models and Training phase.}
To thoroughly analyze how learning a backdoor affects a model, we consider different convex learning algorithms, including linear Support Vector Machines (SVMs), Logistic Regression Classifiers (LCs), Ridge Classifiers (RCs), nonlinear SVMs using the Radial Basis Function (RBF) kernel, and deep neural networks. 
We train the classifiers directly on the pixel values scaled in the range $[0, 1]$ on the MNIST dataset. For CIFAR10 and Imagenette, we instead consider a transfer learning setting frequently adopted in the literature \cite{koh2017understanding,Donahue2014DeCAFAD, Pasquale2015TeachingIT}. Like Saha et al.~\cite{saha2020hidden}, we use the pre-trained model AlexNet~\cite{KrizhevskySH12} as a feature extractor. The convex learners are then trained on top of the feature extractor.
We study these convex learners due to their broad usage in industry~\cite{suya2021model}, derived from their excellent results with smaller dataset~\cite{Vabalas2019MachineLA}, and good interpretability~\cite{tramerSimpleModel21, dacremaProgress2019}. 
In addition, we include in our evaluation pre-trained Resnet18 and Resnet50~\cite{resnet} deep neural networks, sourced from Torchvision~\cite{torchvision2016}, which stand as some of the most extensively employed architectures \cite{Zhang19EqualLayers}. These networks are fine-tuned to classify samples from the Imagenette dataset accurately. \smallskip

\myparagraph{Hyperparameters.}\label{sec:hyperparam-choice}
The choice of hyperparameters has a relevant impact on the learned decision function. For example, some of these parameters control the complexity of the learned function, which may lead to overfitting \cite{Mehta2019AHL}, thereby potentially compromising classification accuracy on test samples. We argue that a high complexity may also lead to higher importance to outlying samples, including backdoors, and thus has a crucial impact on the capacity of the model to learn backdoors. 
To verify our hypothesis, we consider different configurations of the models' hyperparameters. For convex learners, we study two hyperparameters that impact model complexity, \ie, \rebuttal{the regularization hyperparameter $\lambda=\frac{1}{C}$} and the RBF kernel hyperparameter $\gamma$. 
To this end, we take 10 values for $\lambda$ on a uniformly spaced interval on a log scale from $\expnumber{1}{-04}$ to $\expnumber{1}{+02}$. For the Imagenette dataset we extend this interval in  $[\expnumber{1}{-05} , \expnumber{1}{+02}]$. 
Concerning the RBF kernel, we let $\gamma$ take $5$ uniformly spaced values on a log scale in $[\expnumber{5}{-04}, \expnumber{5}{-02}]$ for MNIST, $[\expnumber{1}{-04}, \expnumber{1}{-02}]$ for CIFAR, and $[\expnumber{1}{-05}, \expnumber{1}{-03}]$ for Imagenette.
Furthermore, we take $10$ values of $\lambda$ in the log scale uniformly spaced interval $[\expnumber{1}{-01}, \expnumber{1}{+02}]$ for the RBF kernel. This allowed us to study a combination of $10$ and $50$ hyperparameters for linear classifiers and RBF SVM, respectively.

For deep neural networks, we consider two different numbers of epochs: $10$ and $50$, and increase the number of neurons when using Resnet50 instead of Resnet18. Whereas size intuitively correlates with complexity, previous works, including \cite{Caruana00EarlyStopping}, show that decreasing the number of training epochs reduces the complexity of the trained network as well. Conversely, increasing epochs leads to overfitting on the training dataset, thus, a more complex decision function. Each network is fine-tuned using the SGD optimizer with a learning rate of $0.001$, a momentum of $0.9$, and a batch size of $256$.\smallskip


\myparagraph{Backdoor Attacks.}
We implement the backdoor attacks proposed by Gu et al.~\cite{gu2019badnets} against MNIST and CIFAR10. More concretely, 
we use a random $3 \times 3$ patch as the trigger for MNIST, while on CIFAR10, we increase the size to $8\times 8$ to strengthen the attack~\cite{saha2020hidden}. We add the trigger pattern in the lower right corner of the image~\cite{gu2019badnets}. Samples from MNIST and CIFAR10 with and without trigger can be found in  Figure~\ref{fig:interpretability}. However, in contrast to previous approaches~\cite{gu2019badnets}, we use a separate trigger for each base-class $i$. The reason is that our study encompasses linear models that are unable to associate the same trigger pattern to two different classes. Using different trigger patterns, we enhance the effectiveness of the attack on these linear models.
On the Imagenette dataset, we use the backdoor trigger developed by~\cite{Zhong20backdoor}. This attack injects a patterned perturbation mask into training samples to open the backdoor. A constant value $c_m$ refers to the maximum allowed intensity. 
We apply the backdoor attacks to 10\% of the training data if not stated otherwise, and, as done by Gu et al.~\cite{gu2019badnets}, we force the backdoored model to predict the $i$-th class as class $(i + 1)\% n\_classes$ when the trigger is shown.
We also report additional experiments 
concerning variations in the trigger's size or visibility.

\subsection{Experimental Results}\label{sec:experimental_results}
In the following, we now discuss our experimental results obtained with the datasets, classifiers, and backdoor attacks described above. 
\begin{figure*}[t]
  \centering
  \begin{subfigure}{0.495\textwidth}
\includegraphics[width=1\textwidth]{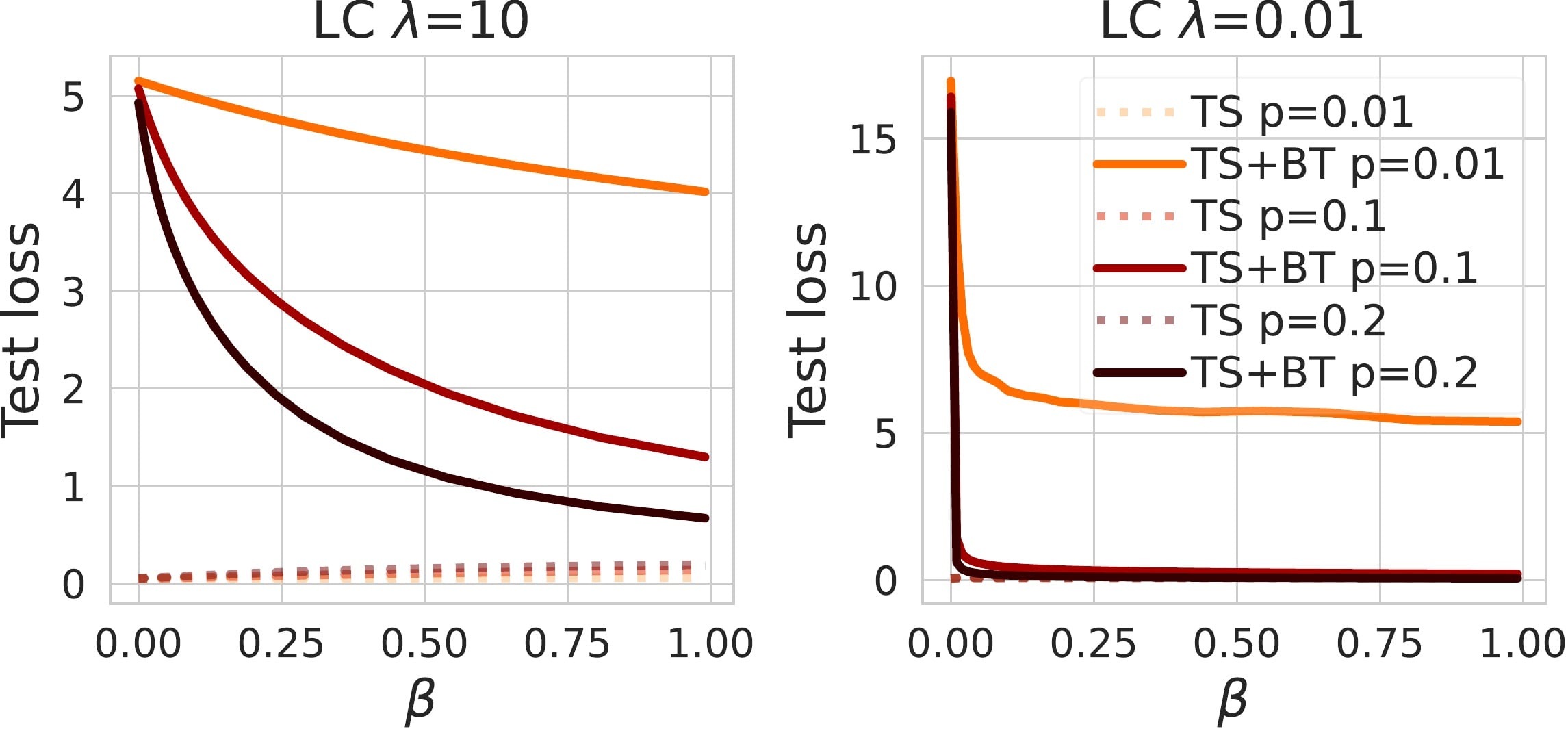}
      \caption{MNIST trigger size $3\times3$.}
      \label{MNIST_trigger_size_curves}
  \end{subfigure}
    \begin{subfigure}{0.495\textwidth}
\includegraphics[width=1\textwidth]{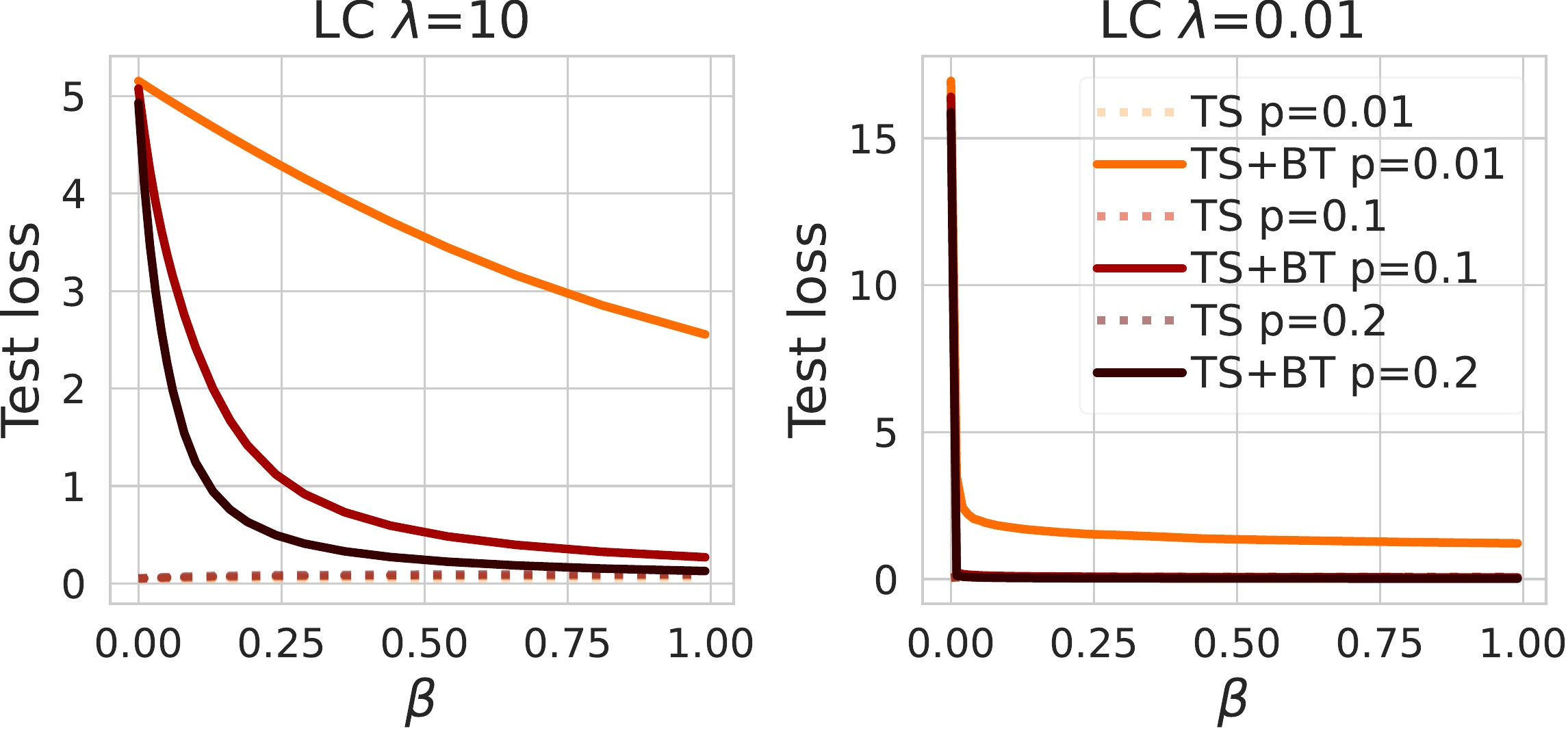}
      \caption{MNIST trigger size $6\times6$.}
      \label{MNIST_trigger_size_curves_double}
  \end{subfigure}

  \begin{subfigure}{0.495\textwidth}
\includegraphics[width=1\textwidth]{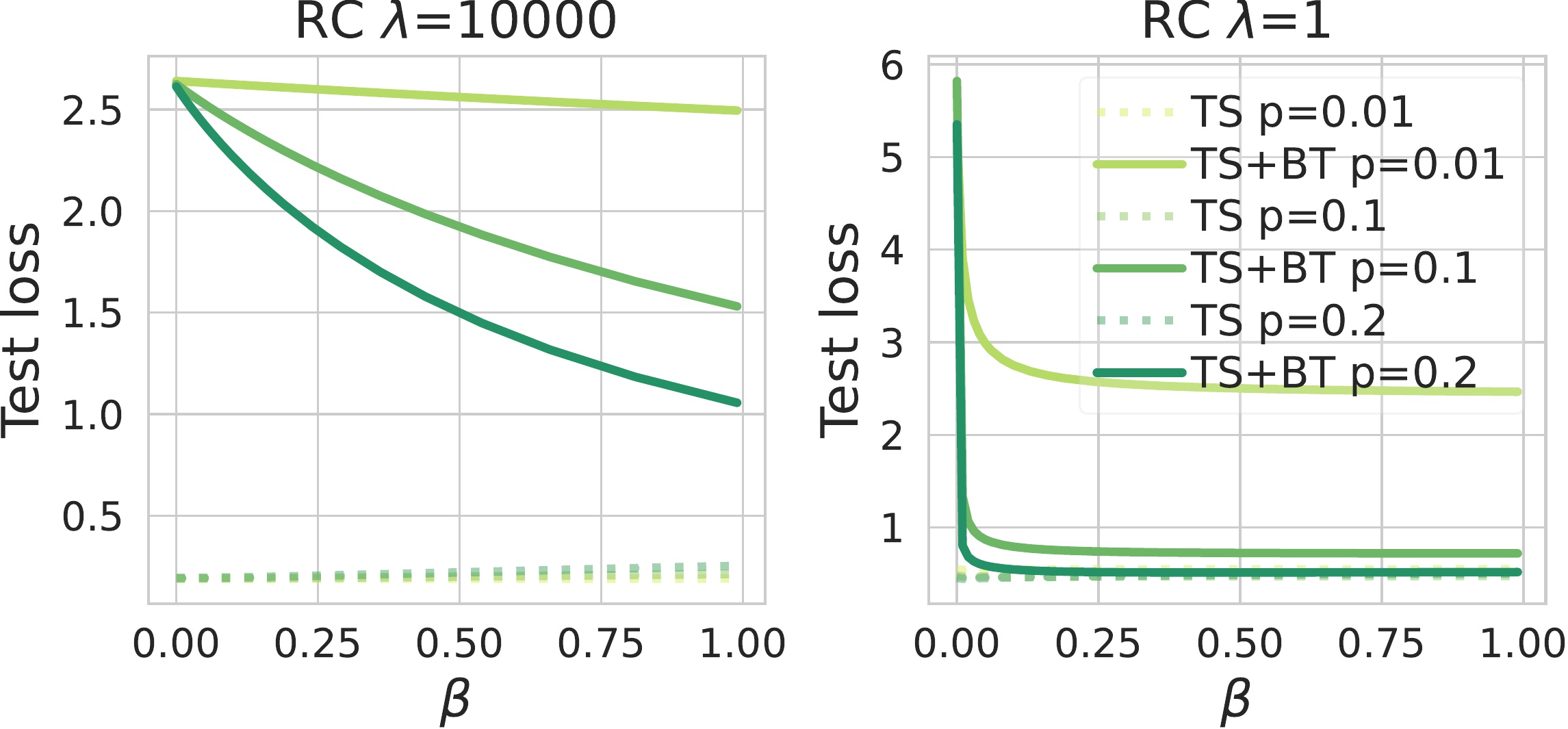}
      \caption{CIFAR10 trigger size $8\times8$.}
      \label{CIFAR_trigger_size_curves}
  \end{subfigure}
    \begin{subfigure}{0.495\textwidth}
\includegraphics[width=1\textwidth]{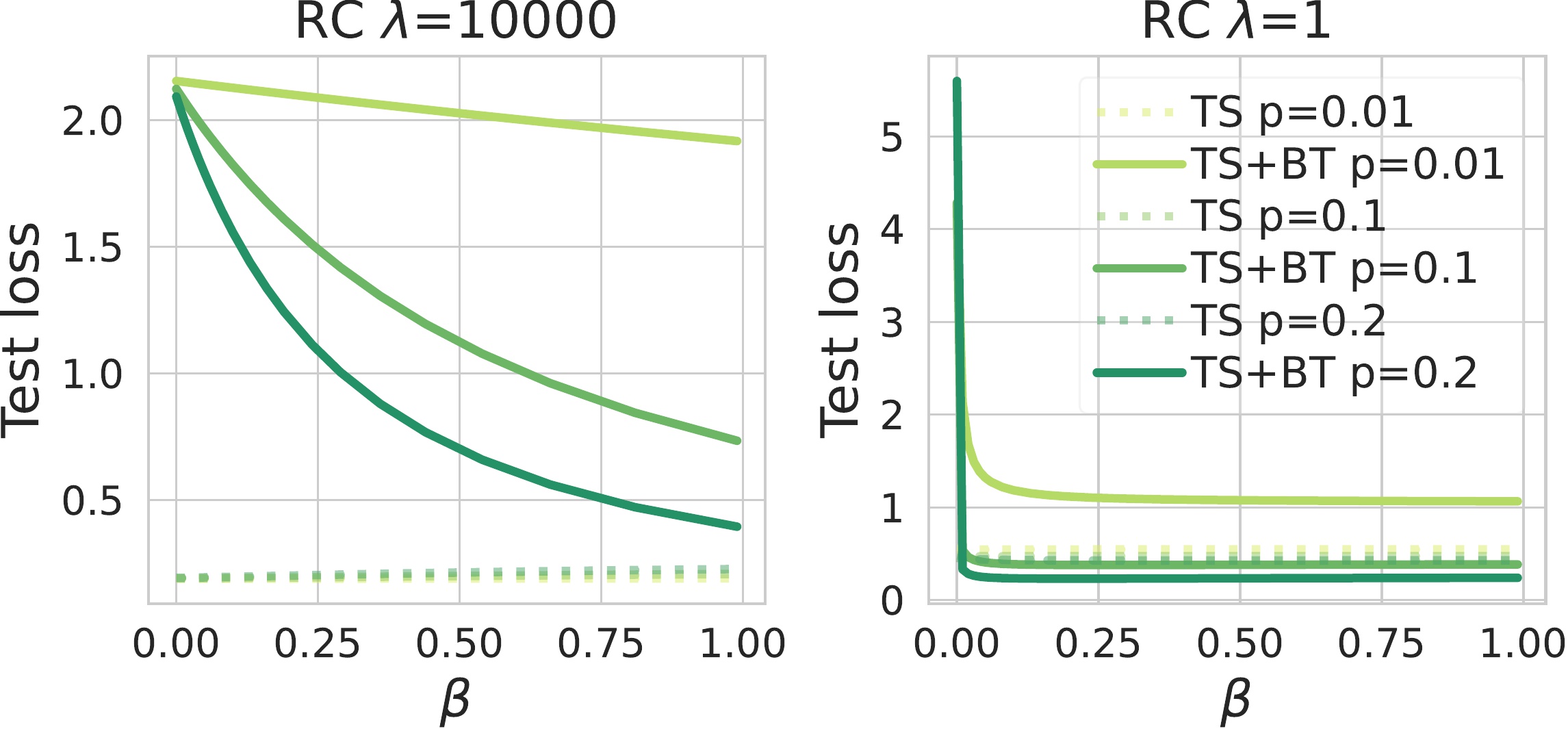}
      \caption{CIFAR10 trigger size $16\times 16$.}
      \label{CIFAR_trigger_size_curves_double}
  \end{subfigure}
  
  \begin{subfigure}{0.495\textwidth}
\includegraphics[width=1\textwidth]{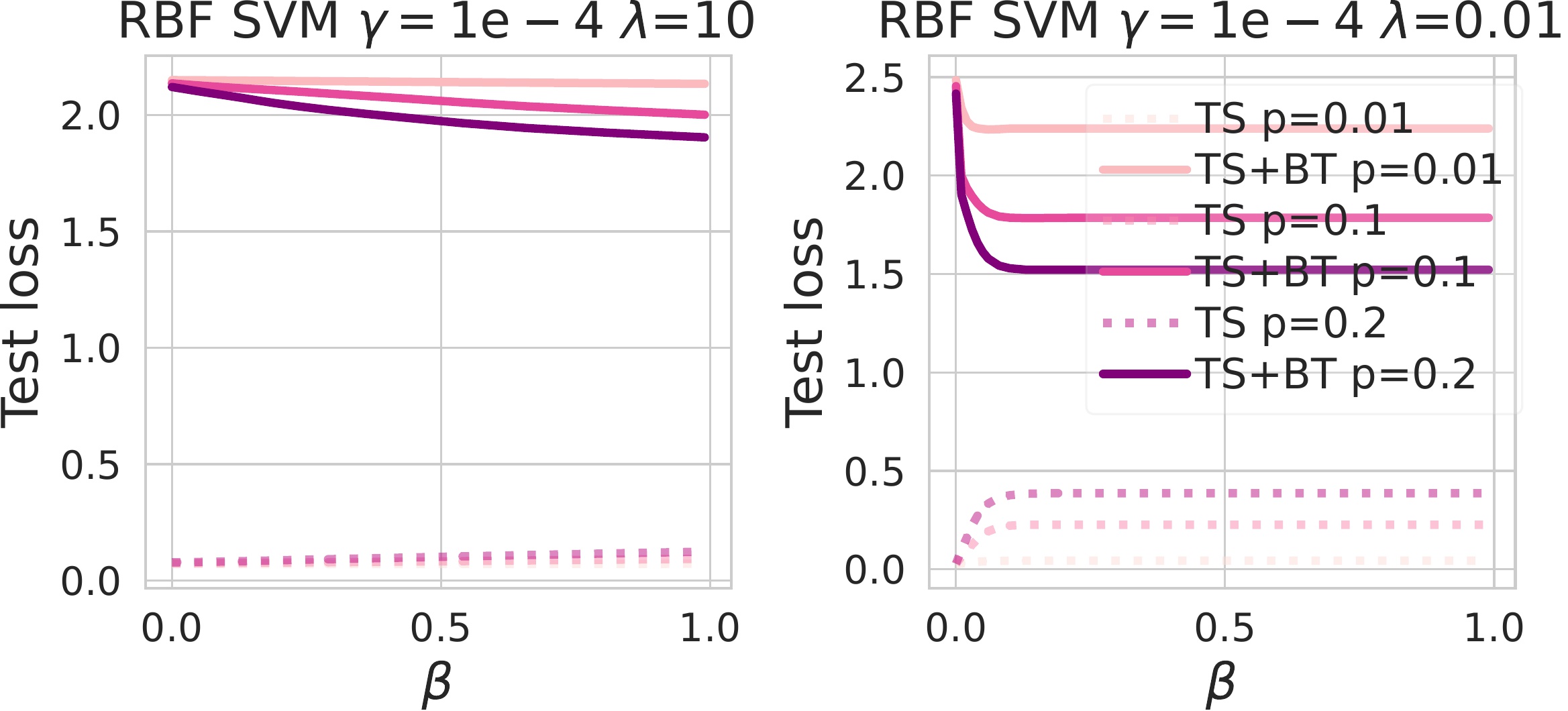}
      \caption{Imagenette trigger visibility $c_m=10$.}
      \label{imagenette_trigger_less_visible}
  \end{subfigure}
    \begin{subfigure}{0.495\textwidth}
\includegraphics[width=1\textwidth]{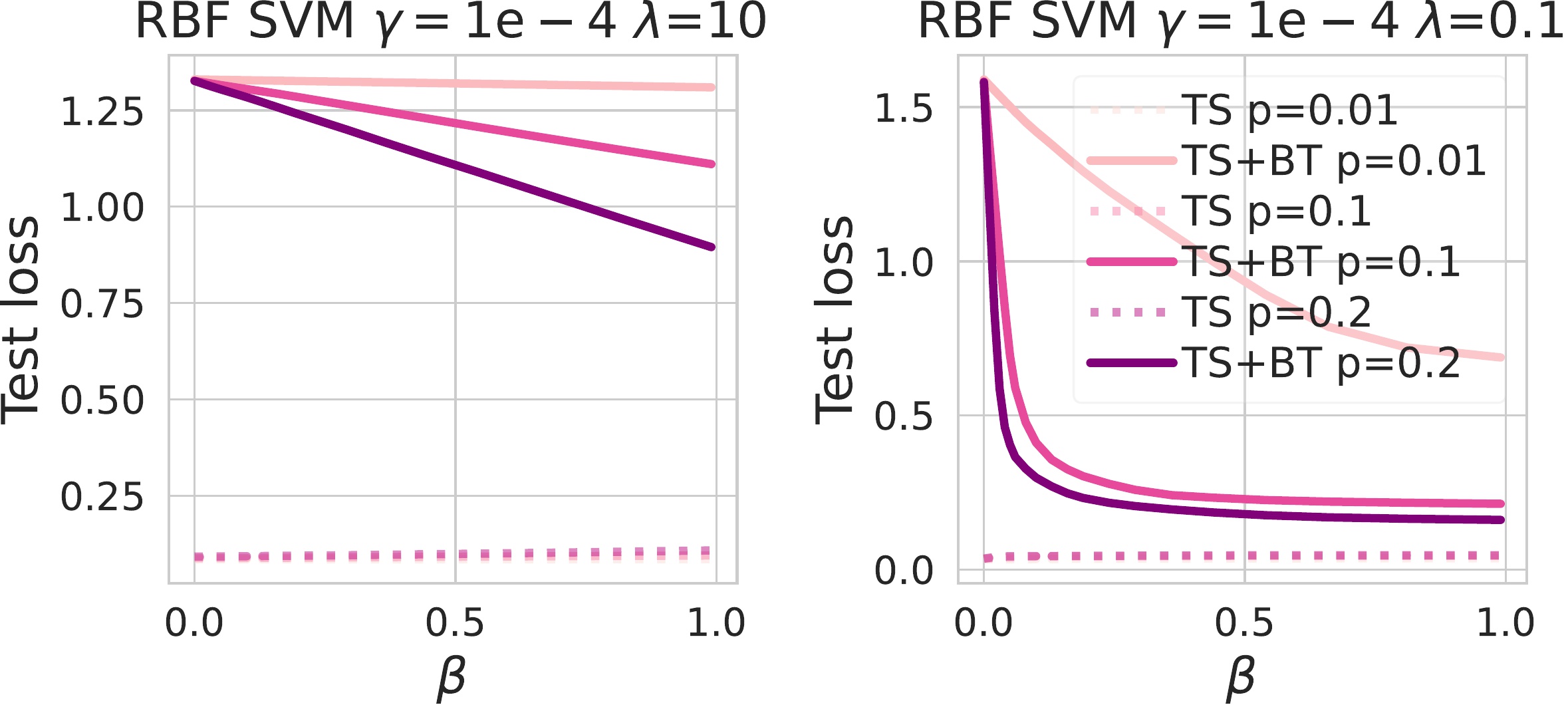}
      \caption{Imagenette trigger visibility $c_m=75$.}
      \label{imagenette_trigger_visible}
  \end{subfigure}
\caption{Backdoor learning curves for: (top row) logistic classifier (LC) on MNIST 7 vs. 1 with $\lambda \in \{10, 0.01\}$ and trigger size $3 \times 3$ \textit{(left)} or $6 \times 6$ \textit{(right)}; (middle row) Ridge classifier on CIFAR10 \cifarairplanefrog with $\lambda \in \{100000, 100\}$ and trigger size $8\times 8$ \textit{(left)} or $16\times 16$ \textit{(right)}; (bottom row) RBF SVM with $\gamma=\expnumber{1}{-04}$ on Imagenette \imagenettetenchtruck with $\lambda \in \{10, 0.1\}$ and trigger visibility $c_m=10$ \textit{(left)} or $c_m=75$ \textit{(right)}. Darker lines represent a higher fraction of poisoning samples $p$ injected into the training set. We report the loss on the \rebuttal{clean test samples (TS)} with a dashed line and on the \rebuttal{test samples with the backdoor trigger (TS+BT)} with a solid line.}
  \label{fig:backdoor_learning_curves_convex}
\end{figure*}

\begin{figure*}[t]
\centering
  \begin{subfigure}{0.495\textwidth}
\includegraphics[width=0.99\textwidth]{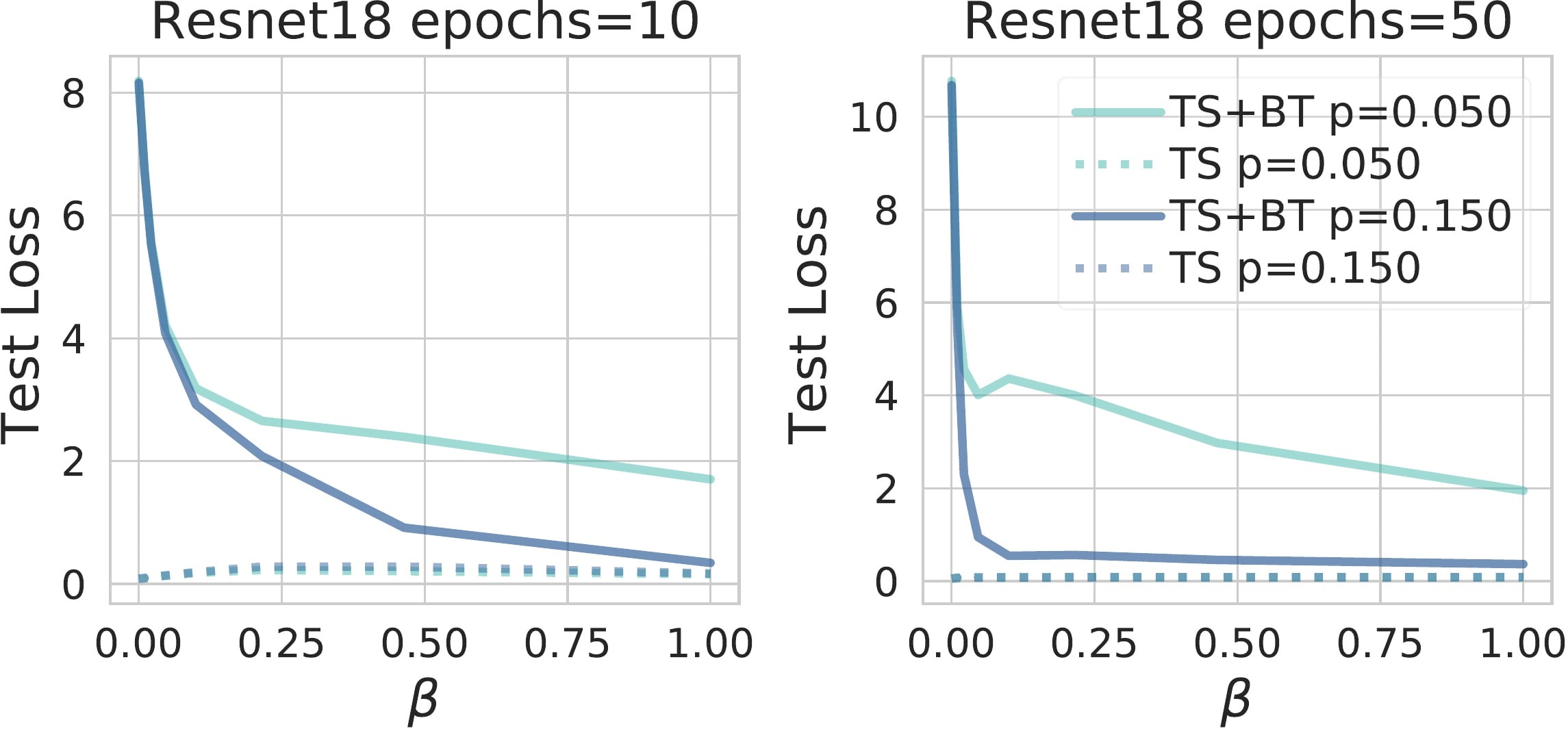}
      \caption{Imagenette with trigger visibility $c_m=10$.}
      \label{imagenette_trigger_less_visible_nn}
  \end{subfigure}
    \begin{subfigure}{0.495\textwidth}
\includegraphics[width=0.99\textwidth]{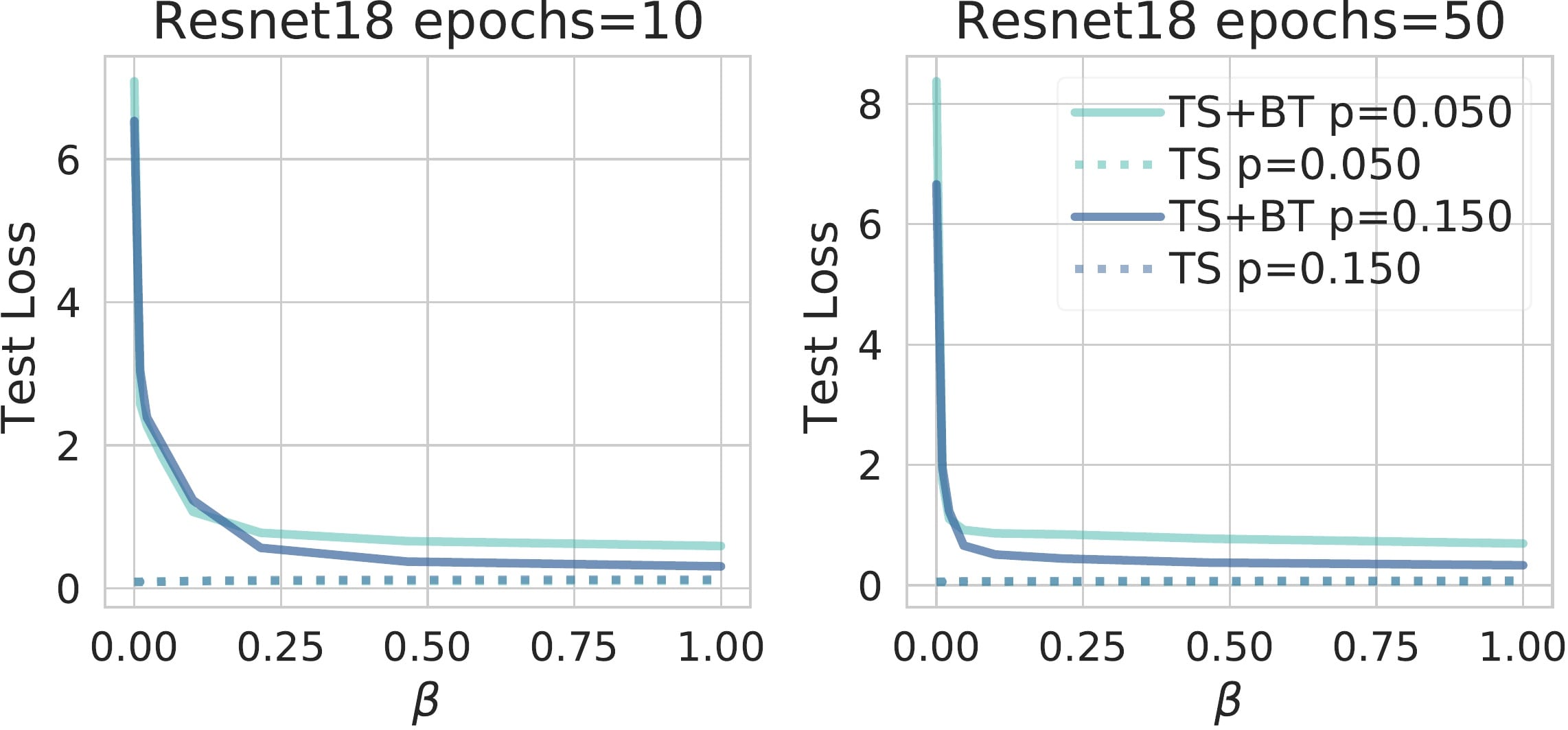}
      \caption{Imagenette with trigger visibility $c_m=75$.}
      \label{imagenette_trigger_visible_nn}
  \end{subfigure}
\caption{Backdoor learning curves for 
Resnet18 trained on the full Imagenette training dataset with $10$ and $50$ epochs. Darker lines represent a higher fraction of poisoning samples $p$ injected into the training set. We report the loss on the clean test samples (TS) with a dashed line and on the test samples with the backdoor trigger (TS+BT) with a solid line.}
\label{fig:backdoor_learning_curves_resnet}
\end{figure*}
\begin{figure*}[h!]
  \centering
    \includegraphics[width=0.495\textwidth]{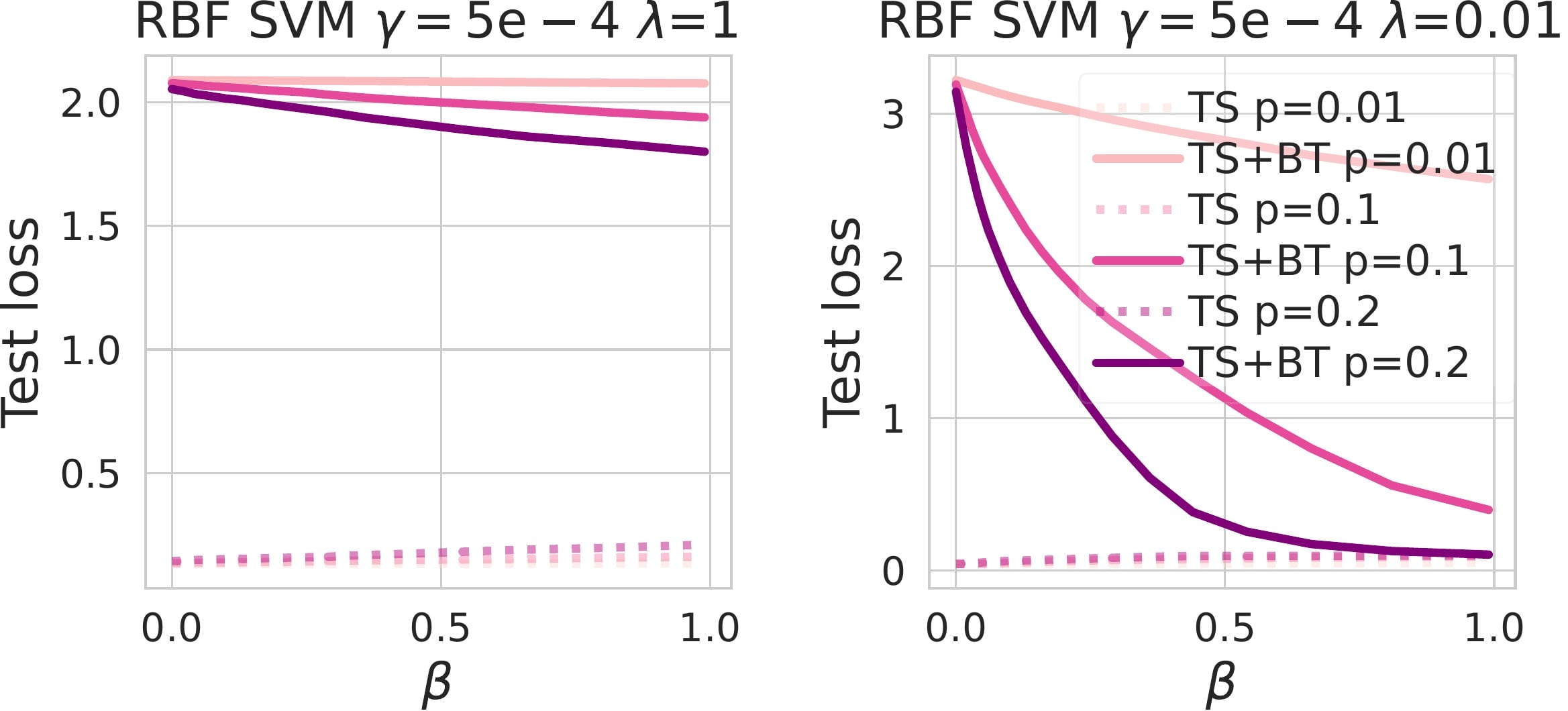}
    \includegraphics[width=0.495\textwidth]{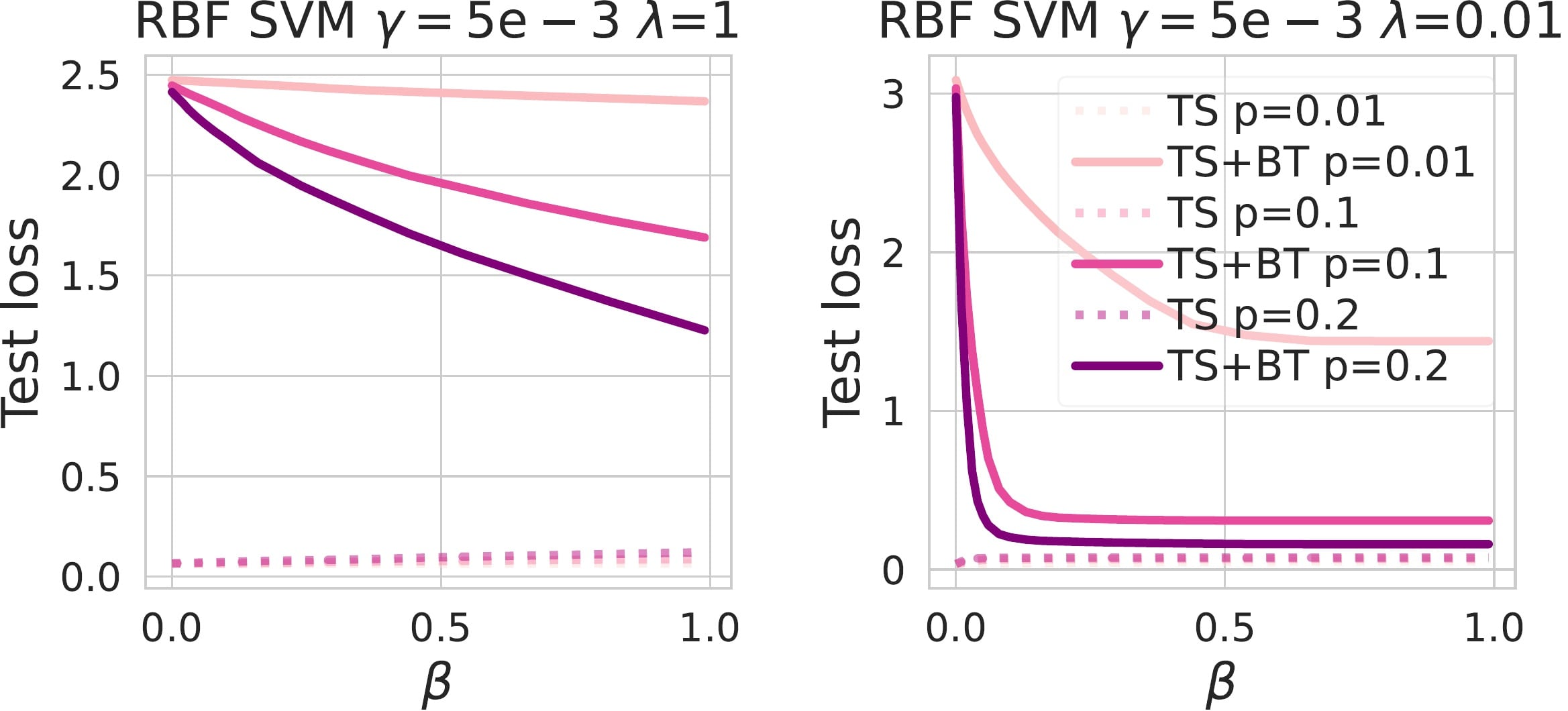}

    \includegraphics[width=0.495\textwidth]{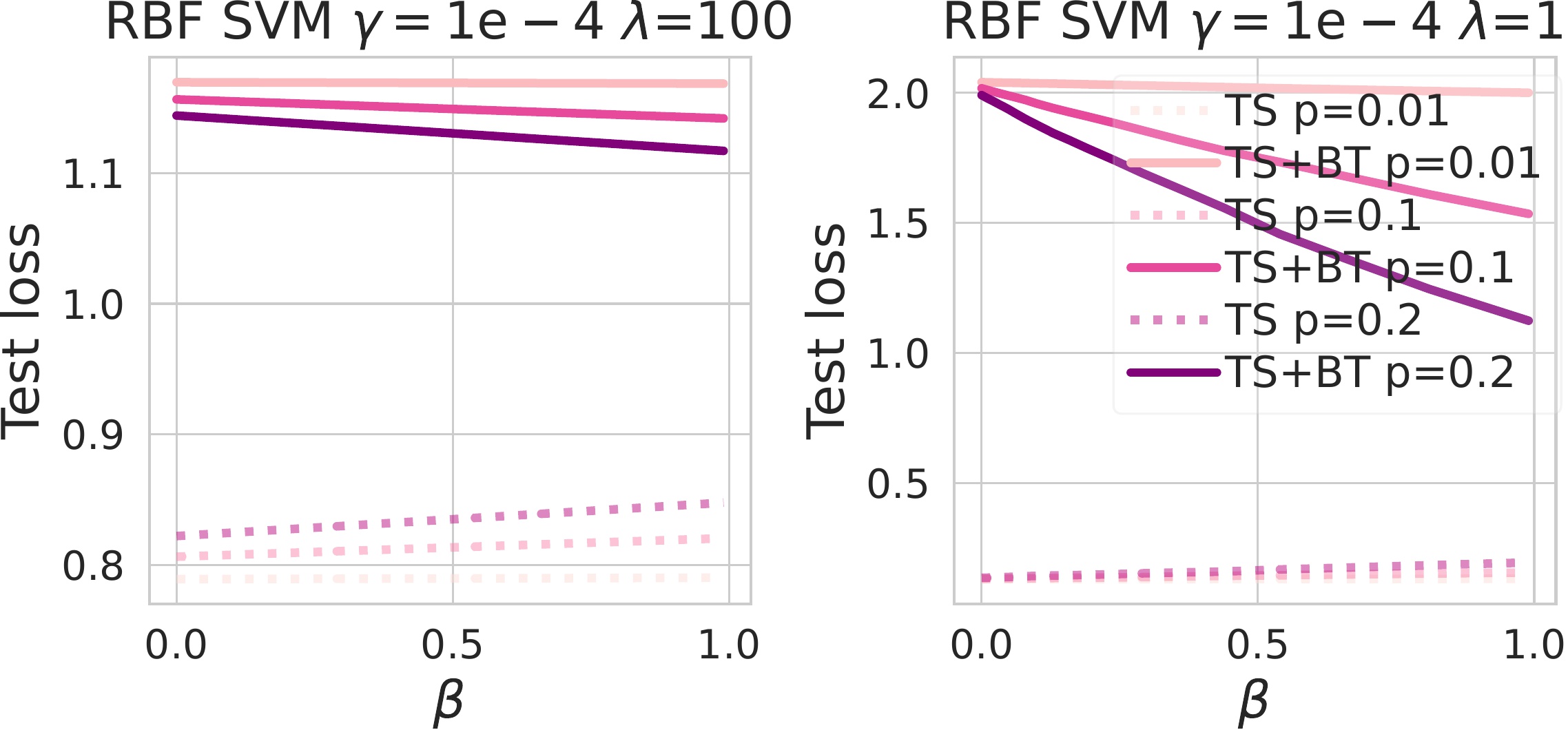}
    \includegraphics[width=0.495\textwidth]{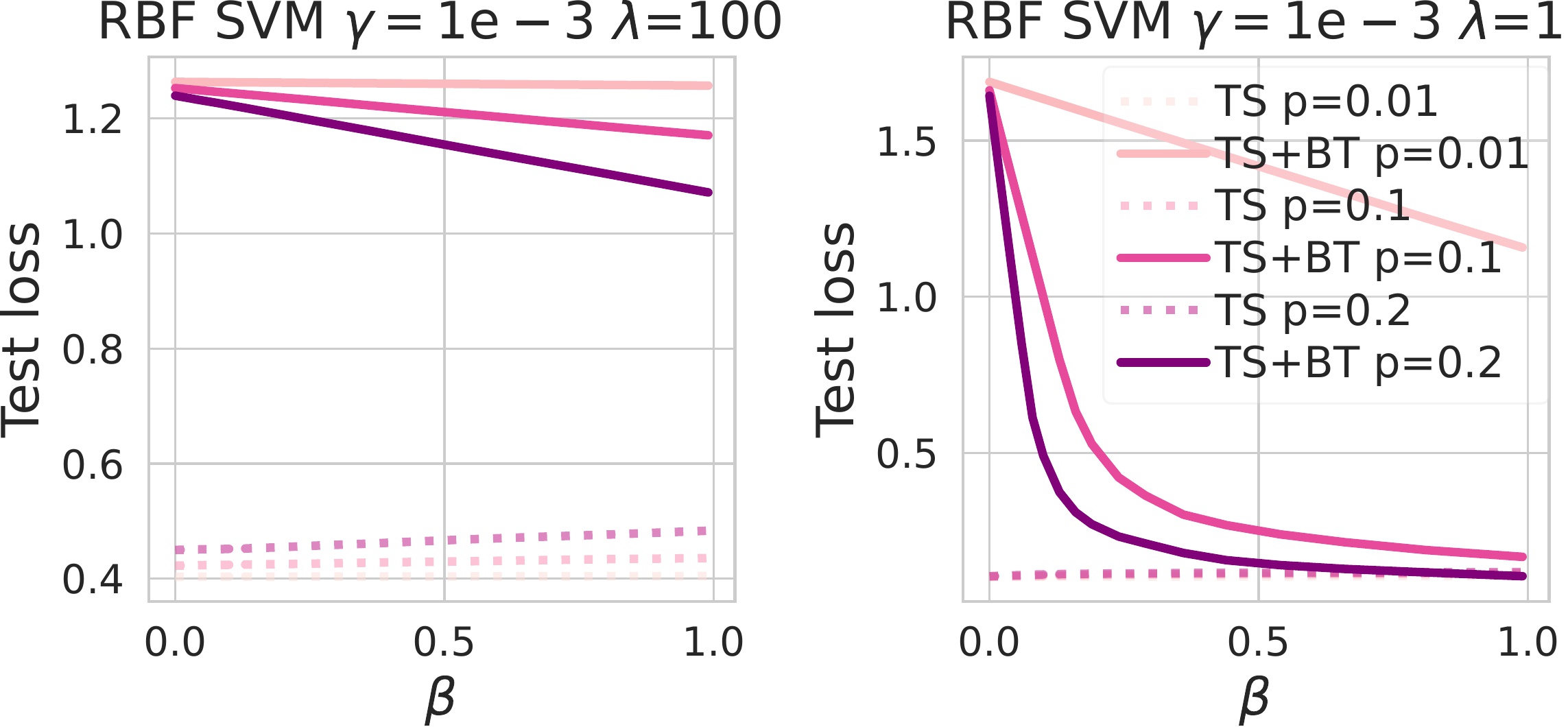}

      \includegraphics[width=0.495\textwidth]{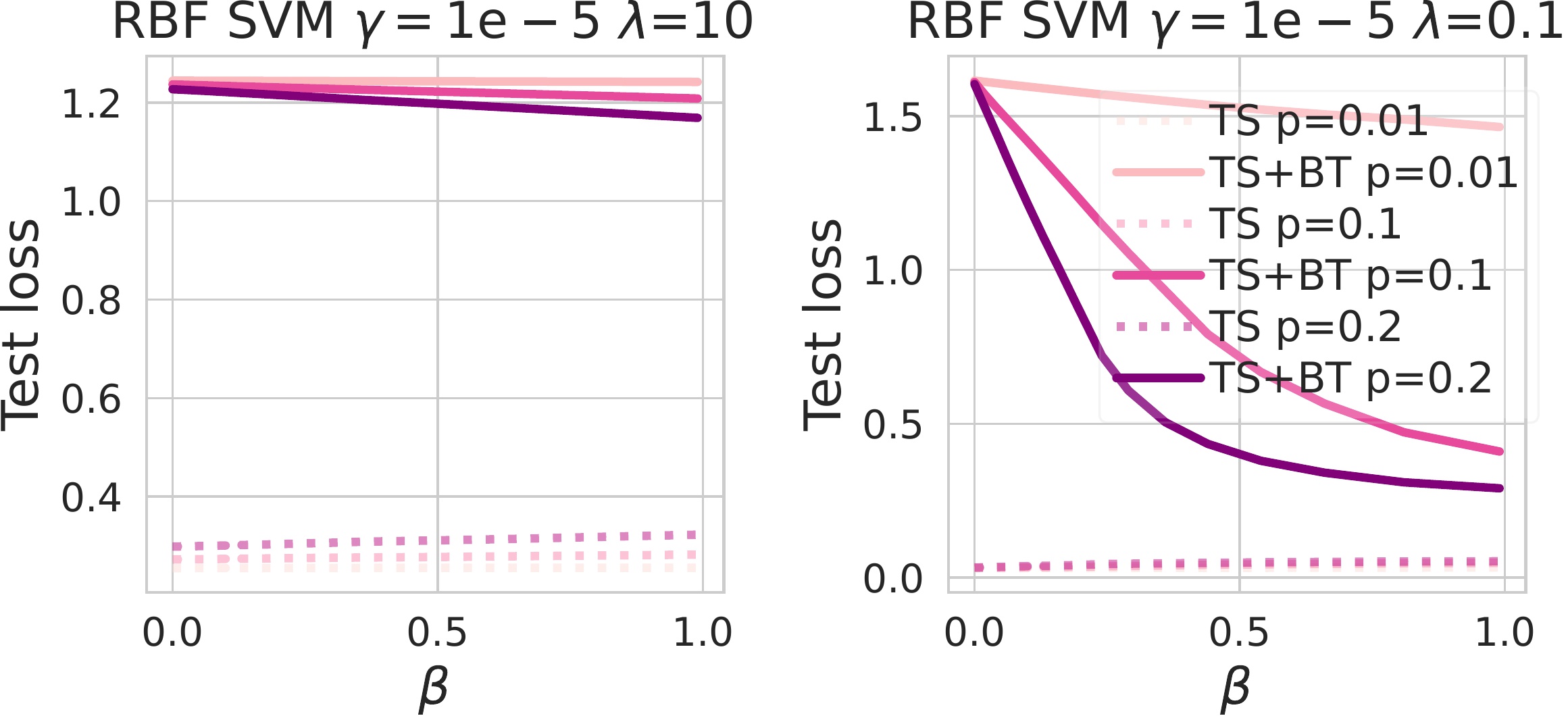}
      \includegraphics[width=0.495\textwidth]{ml_figs-imagenette-incremental-0-6-incremental_backdoor_SVM_RBF_gamma=1e-04_resultsECML.jpeg}

 \caption{Backdoor learning curves for MNIST $7~\rm{vs}~1$ (top row), CIFAR10 \cifarairplanefrog (middle row) and Imagenette \imagenettetenchtruck (bottom row) when changing the kernel parameter $\gamma$ on RBF SVM. Darker lines represent a higher fraction of poisoning samples $p$ injected into the training set. We report the loss on the clean test samples (TS) with a dashed line and on the test samples with the backdoor trigger (TS+BT) with a solid line.
 }
  \label{fig:rbf_learning_curves}
\end{figure*}

\subsubsection{Backdoor Learning Curves} 
Here we present the results obtained using the learning curves that we proposed to study the impact of three different factors on the backdoor learning process: (i) \textit{model complexity}, (ii) the \textit{fraction of backdoor samples injected}, and  (iii)  the \textit{size and visibility of the backdoor trigger}.
We report the impact of these factors on the backdoor learning curves in Figure~\ref{fig:backdoor_learning_curves_convex}  and \ref{fig:backdoor_learning_curves_resnet}. 
More specifically, in Figure~\ref{fig:backdoor_learning_curves_convex} we consider convex classifiers (\ie LC, RC and RBF SVM) trained on two-class subproblems (MNIST, CIFAR10, and Imagenette), whereas in Figure~\ref{fig:backdoor_learning_curves_resnet} we show the results for Resnet18 trained on all the ten classes of Imagenette.\smallskip

To analyze the first factor, we report the results on the same classifiers, changing the hyperparameters that influence their corresponding complexity. In the case of convex learners, we test different values of the regularization coefficient, while for Resnet18, we increase the number of epochs.
To analyze the impact of the second factor, we plot the backdoor learning curves when the attacker injects an increasing percentage of poisoning points $p \in \{0.01, 0.1, 0.2\}$ for convex learners and $p \in \{0.05, 0.15\}$ for Resnet18. 
Finally, to study the third factor, namely the size and visibility of the backdoor trigger, we have created the same backdoor curves doubling the size of the patch triggers for MNIST and CIFAR10, and increasing the trigger's visibility for Imagenette. Even when a high percentage of poisoning points are injected, for flexible enough classifiers, the loss on the clean test samples remains almost constant. 
Instead, the loss on the test set containing the backdoor trigger is highly affected by the factors mentioned above.  
Both a smaller $\lambda$ or a larger number of epochs (low regularization and thus higher complexity), and larger $p$ (a high percentage of poisoning points added) increase the slope of the backdoor learning curve. This means that the classifier learns the backdoor faster. 
When the classifier is sufficiently complex, even a low percentage of low-poisoning points is enough to rapidly induce the classifier to learn the backdoor. On the other hand, this does not apply to highly regularized classifiers, which generally exhibit slower learning of backdoors. Therefore, limiting the classifier's complexity by selecting an appropriate regularization coefficient may mitigate vulnerability to backdoors. 
Furthermore, our results demonstrate that larger trigger sizes lead to faster backdoor learning by classifiers, particularly when they are not regularized. This observation holds when increasing the trigger's visibility, highlighting the well-known trade-off between the attacker's strength and detectability as introduced by Frederickson et al.~\cite{frederickson2018attack}. The attacker can enhance the backdoor's effectiveness by increasing the trigger size or its visibility. However, these adjustments also make it easier for the defender to detect the attack.

Concerning the RBF SVM's robustness to backdoors, we analyzed the backdoors' learning curves for different values of $\gamma$, which determine the RBF kernel's curvature. More precisely, depending on $\gamma$, we have analyzed the backdoor learning curves, and the classifier's parameters change due to backdoor learning. 
We depict the learning curves in Fig.~\ref{fig:rbf_learning_curves}. On both datasets, reducing $\gamma$ results in flatter backdoor learning curves and increased test loss, indicating greater robustness.

\myparagraph{Remarks.} Overall, our experiments show that to learn a backdoor, a classifier has to increase its complexity (if it is not already highly complex). 
However, an increase in complexity is limited when the classifier is highly regularized or when the attack strength is constrained. For this reason, highly regularized classifiers are preferable in terms of backdoor robustness. 
We present supplementary results for other classifiers in \autoref{sec:supplementary_experiments}, validating the trends above described.

\begin{figure*}[t]
  \centering
\includegraphics[width=0.242\textwidth]{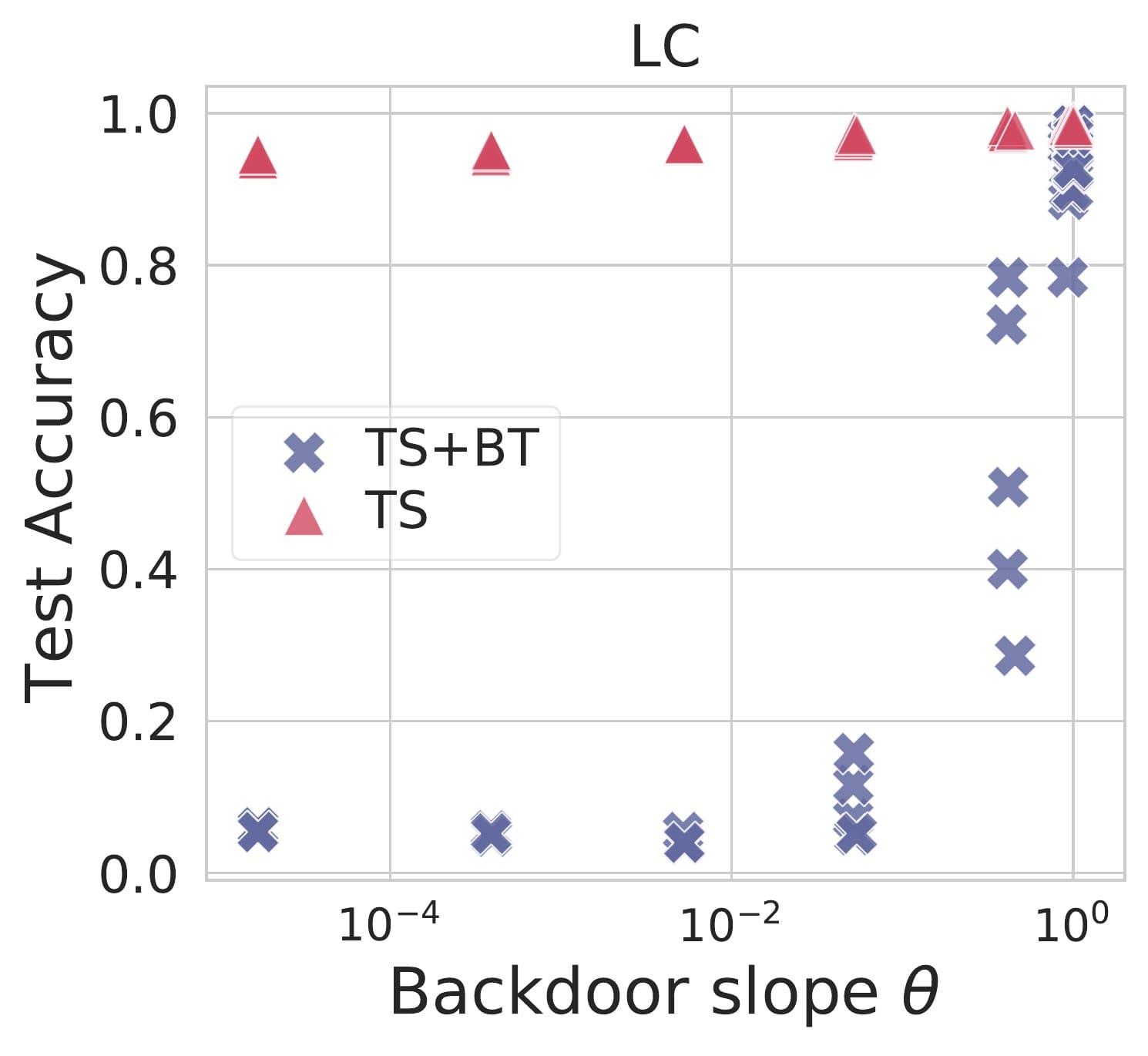}
\includegraphics[width=0.242\textwidth]{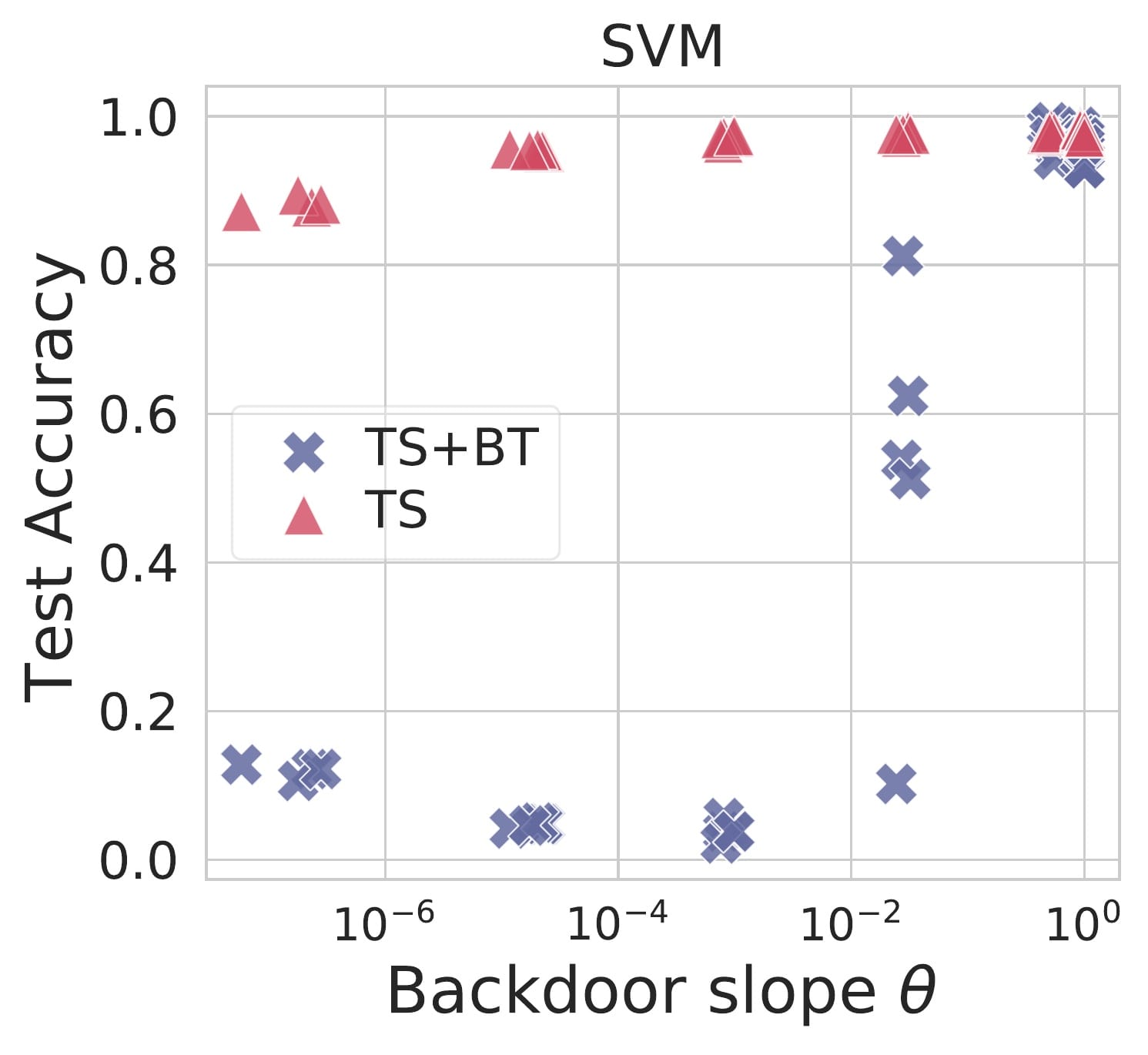}
\includegraphics[width=0.242\textwidth]{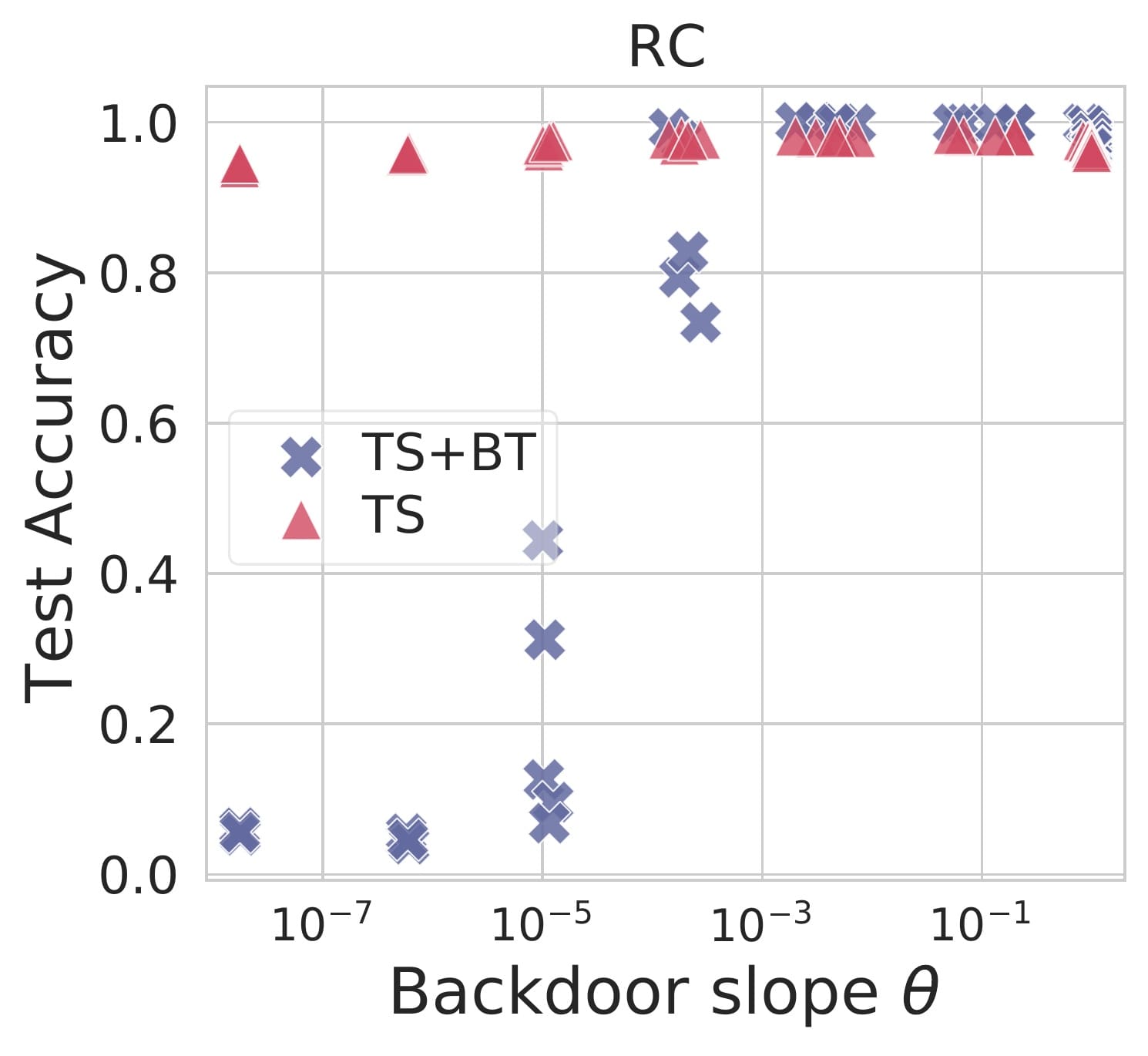}
\includegraphics[width=0.242\textwidth]{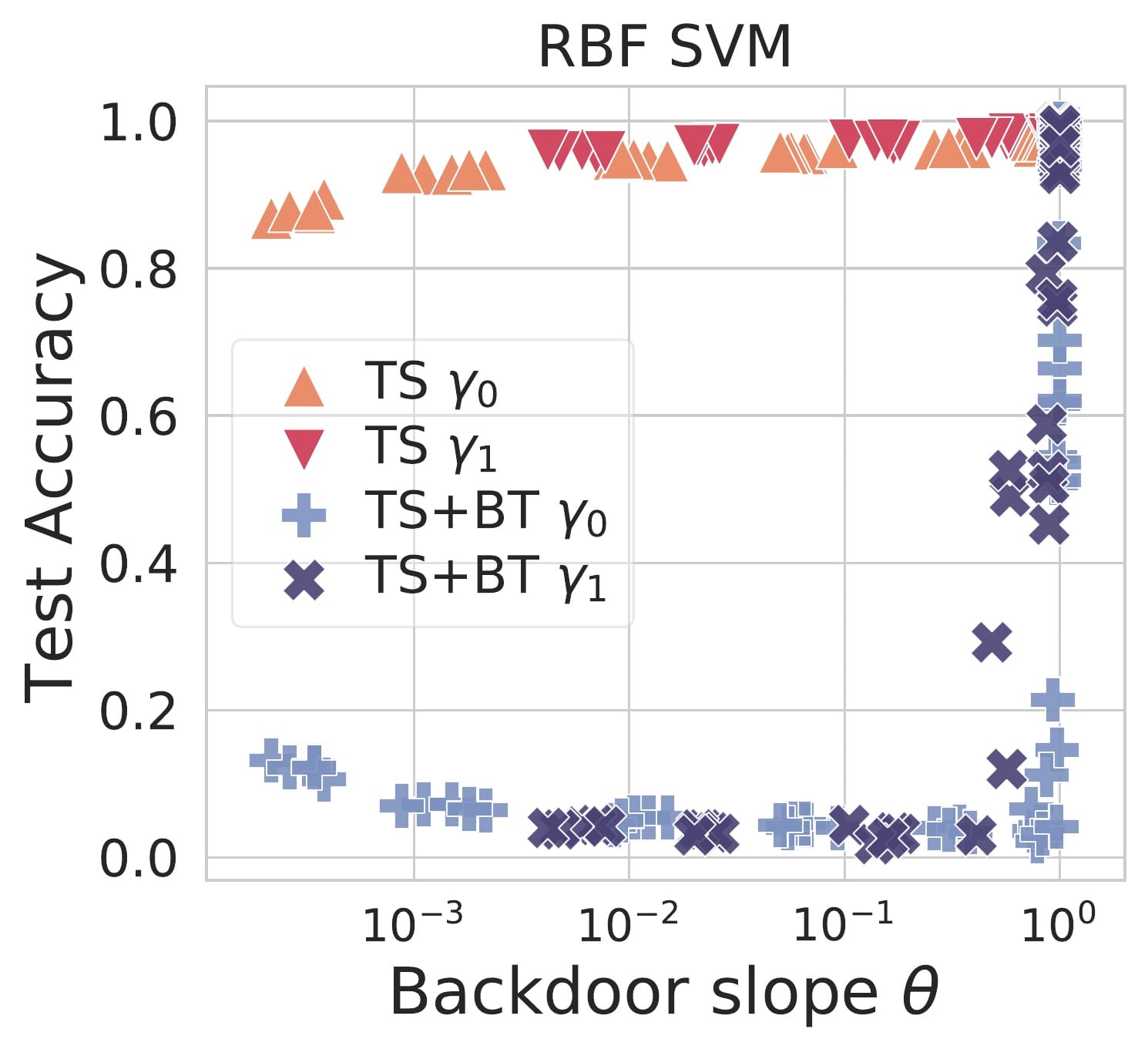}

\includegraphics[width=0.242\textwidth]{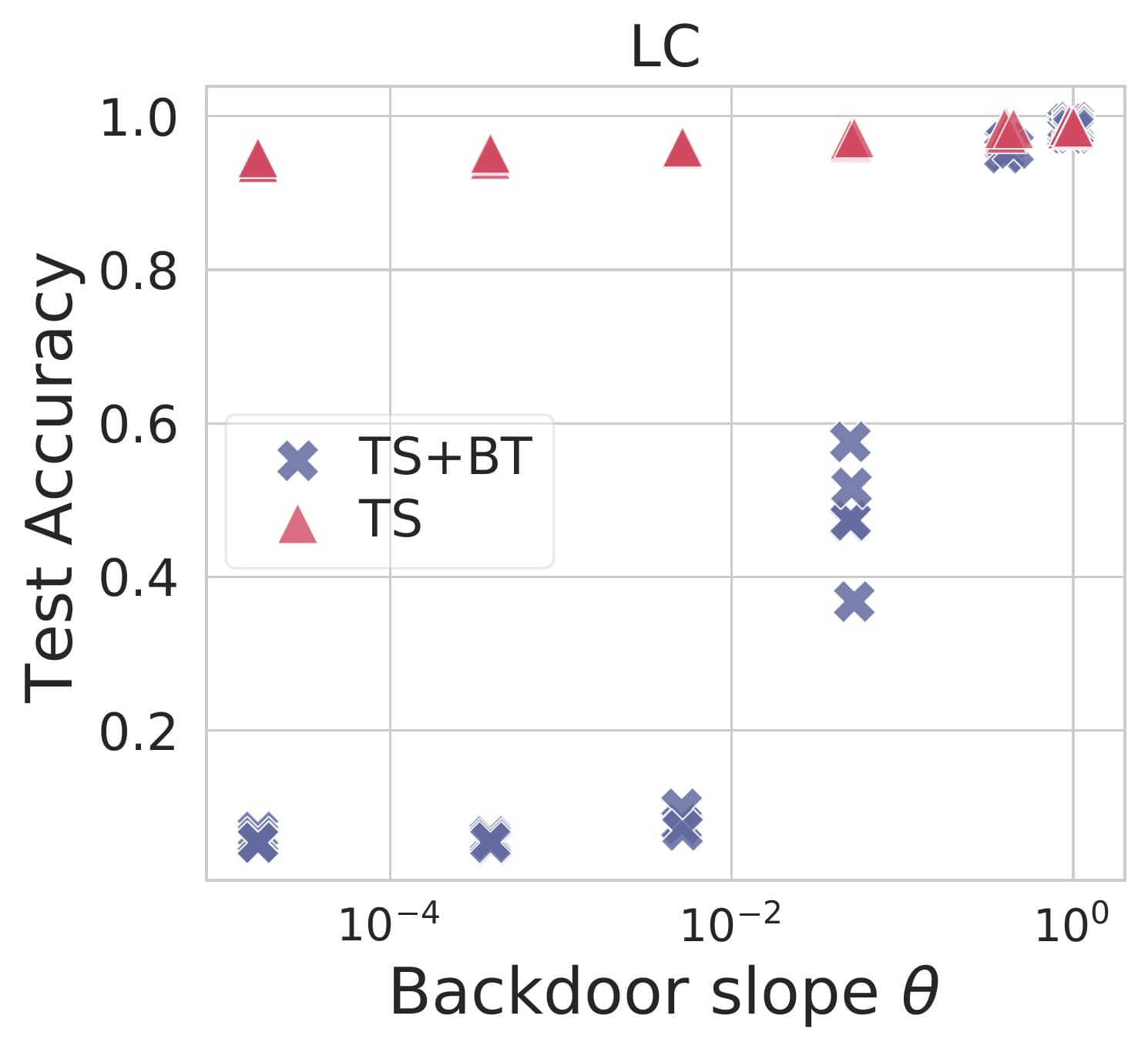}
\includegraphics[width=0.242\textwidth]{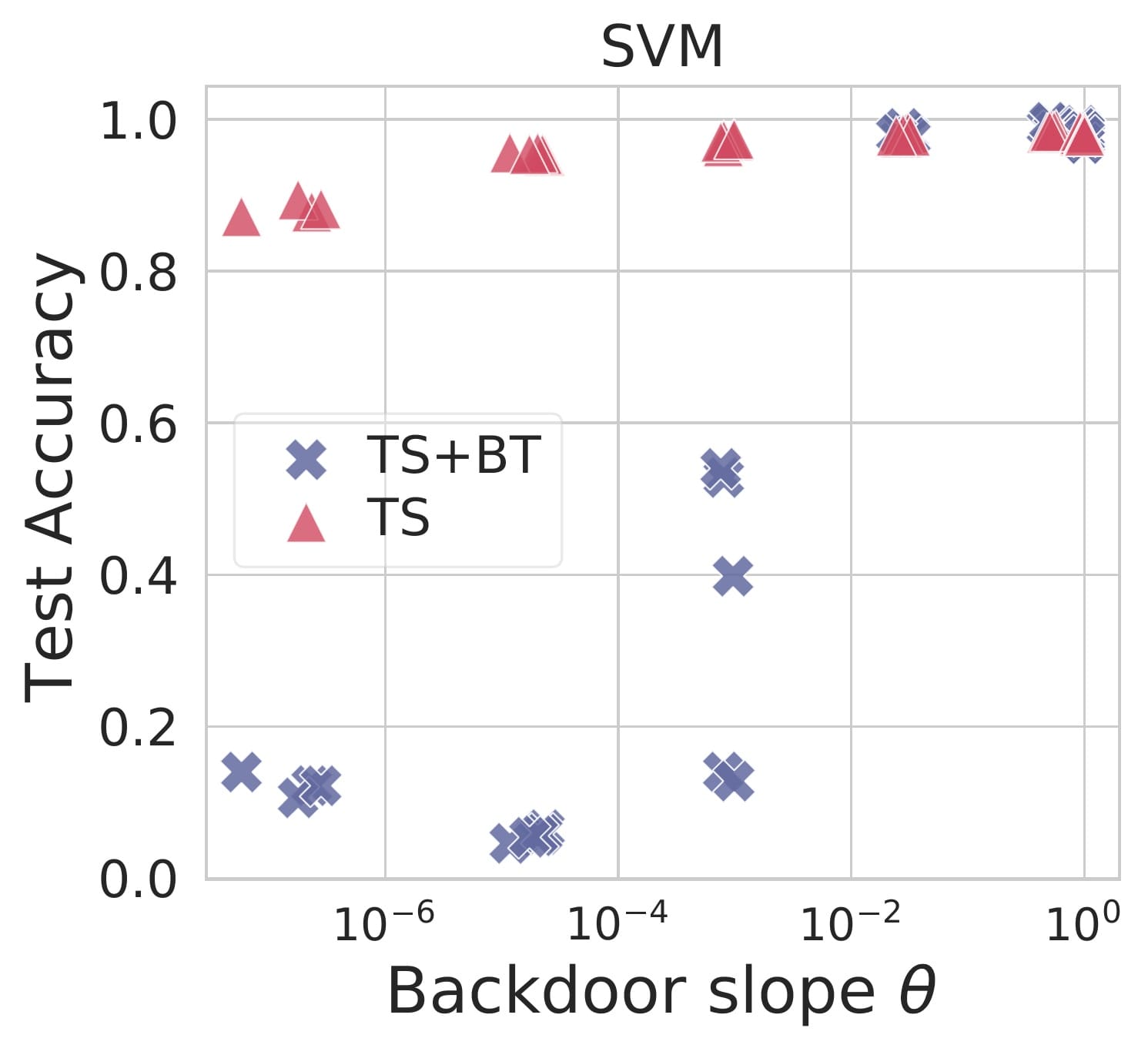}
\includegraphics[width=0.242\textwidth]{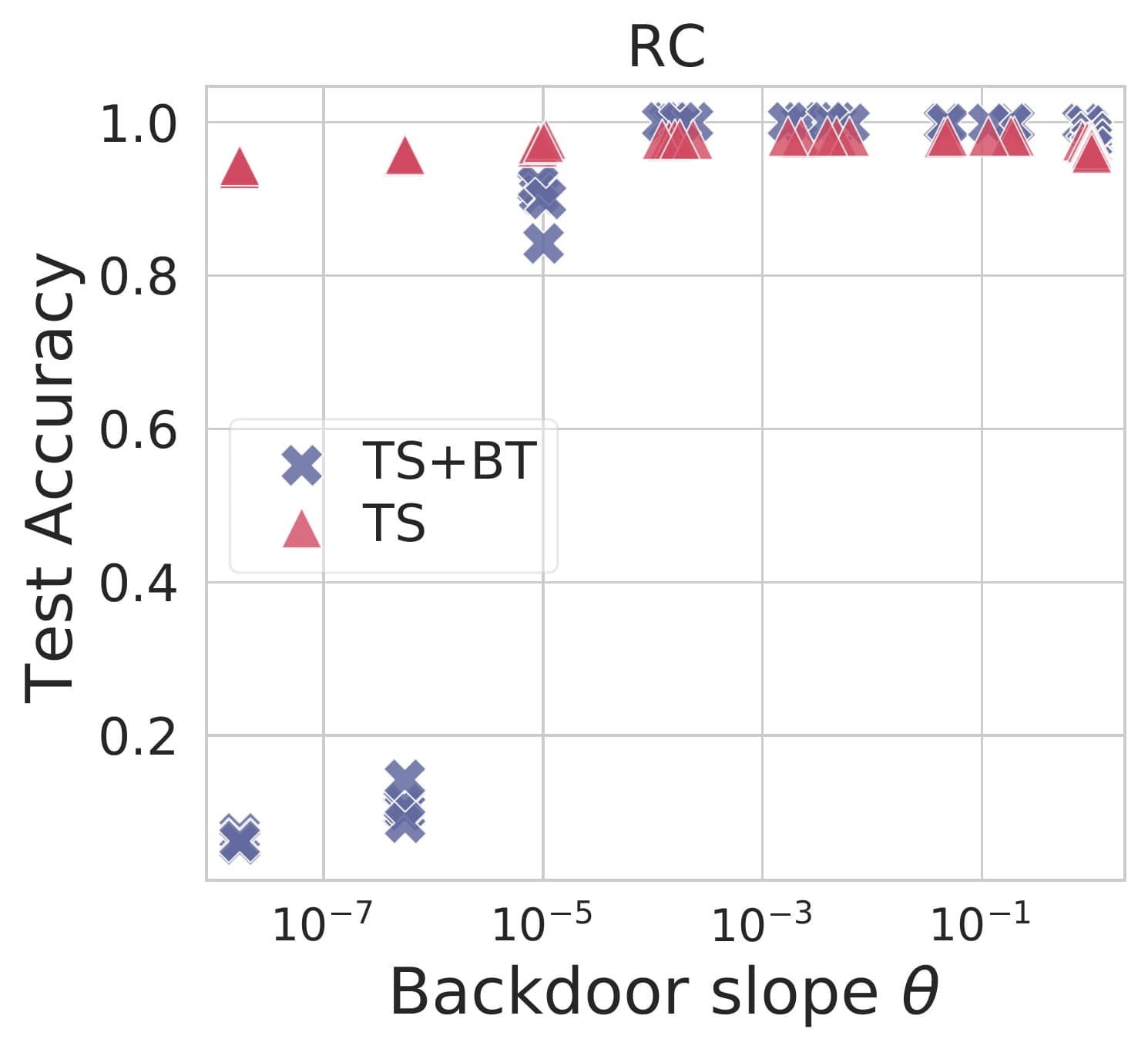}
\includegraphics[width=0.242\textwidth]{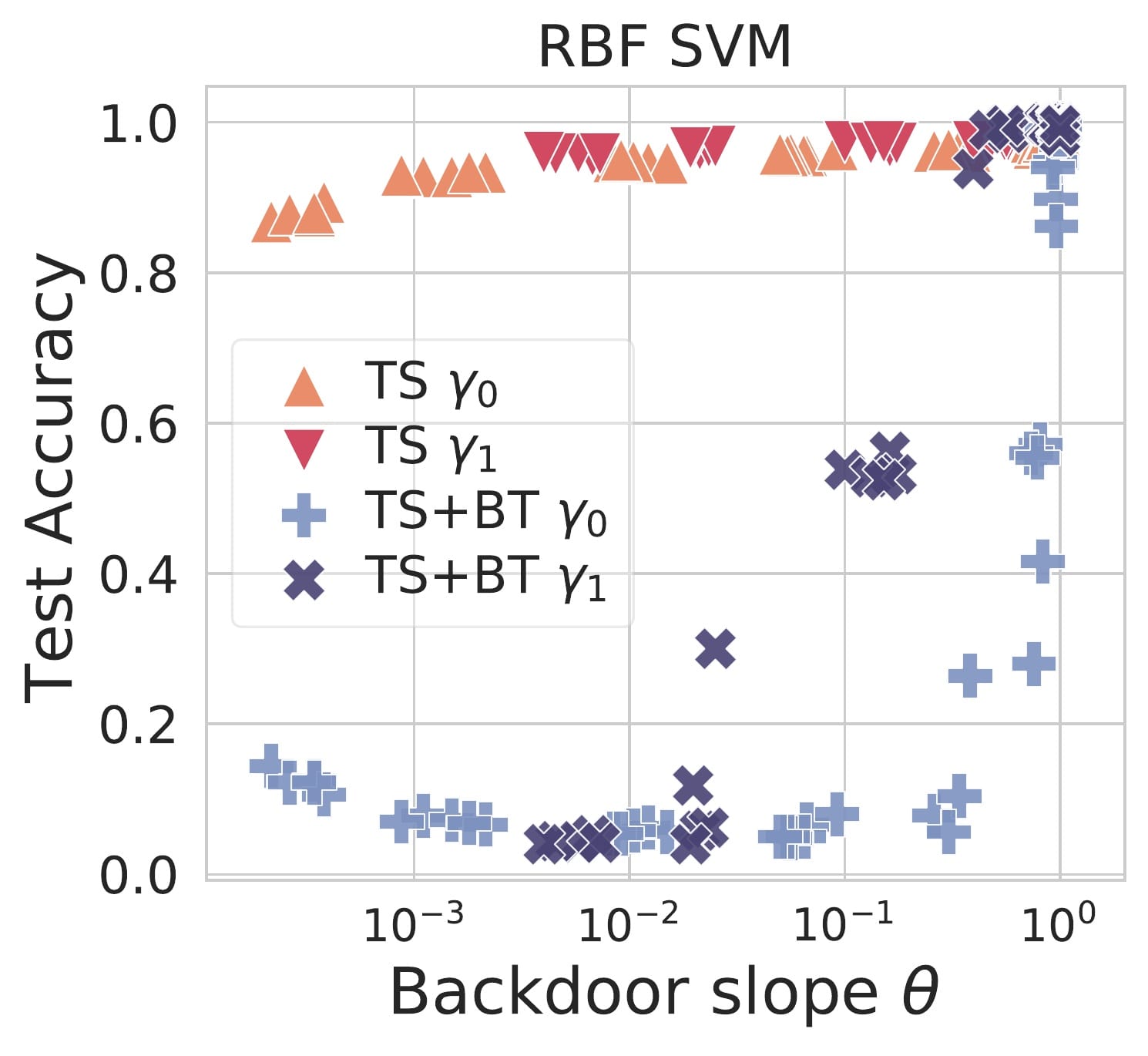}

 \caption{Backdoor slope $\theta$ vs clean accuracy (red) and backdoor effectiveness (blue) on MNIST 7 vs. 1 with backdoor trigger size $3\times 3$ (top row) and $6 \times 6$ (bottom row). We measure the classification accuracy on the \rebuttal{untainted test samples (TS)}, and on the same samples \rebuttal{after injecting the backdoor trigger (TS+BT)}. We chose the $\gamma$ parameter for the RBF kernel as $\gamma_0=\expnumber{5}{-04}$ (orange triangle for clean data, light blue plus for data with trigger) and $\gamma_1=\expnumber{5}{-03}$ (red inverted triangle for clean data, dark blue x for data with trigger).}
  \label{fig:resultsSlopeMNIST}
\end{figure*}
\begin{figure*}[t]
  \centering
\includegraphics[width=0.242\textwidth]{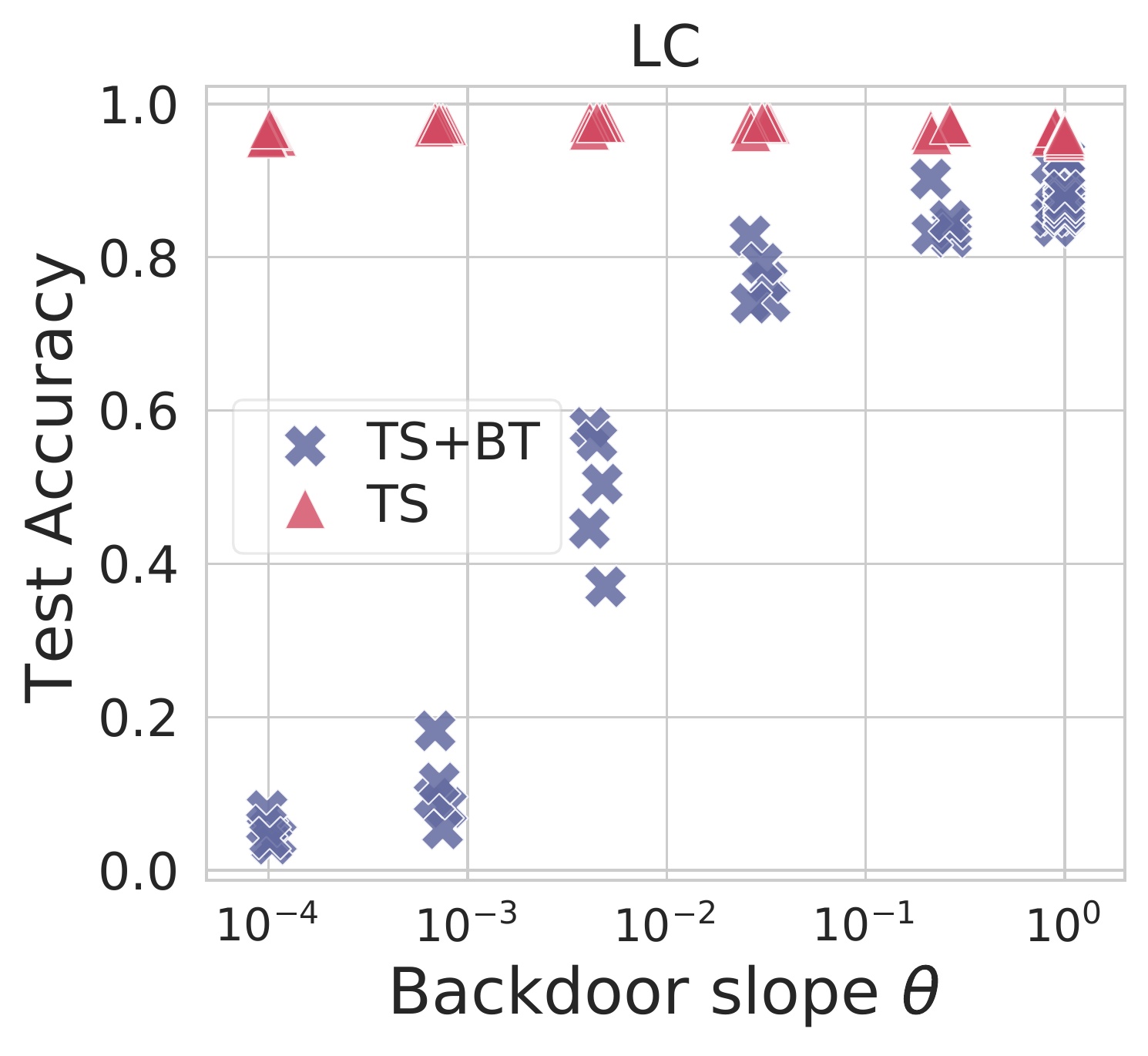}
\includegraphics[width=0.242\textwidth]{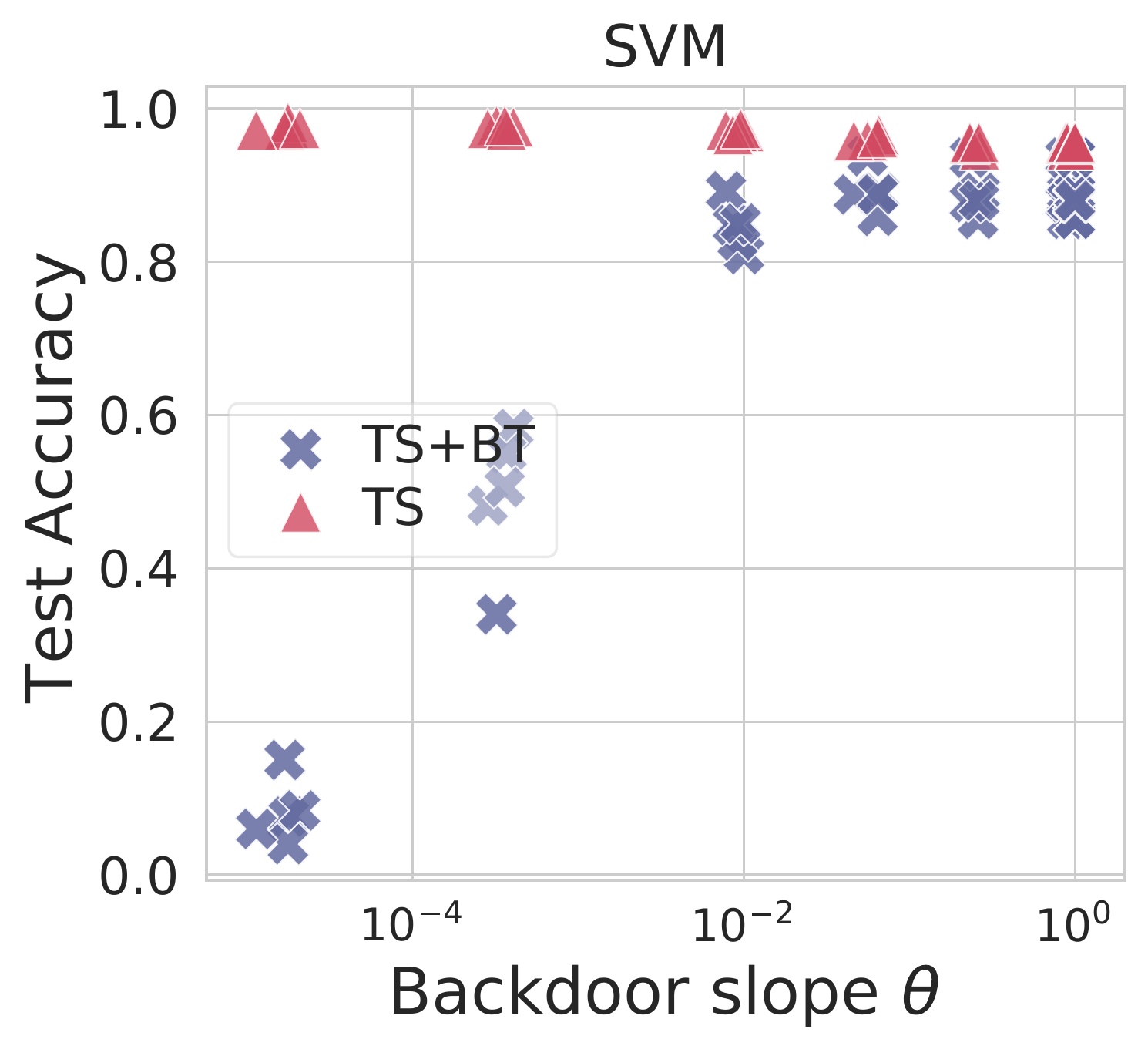}
\includegraphics[width=0.242\textwidth]{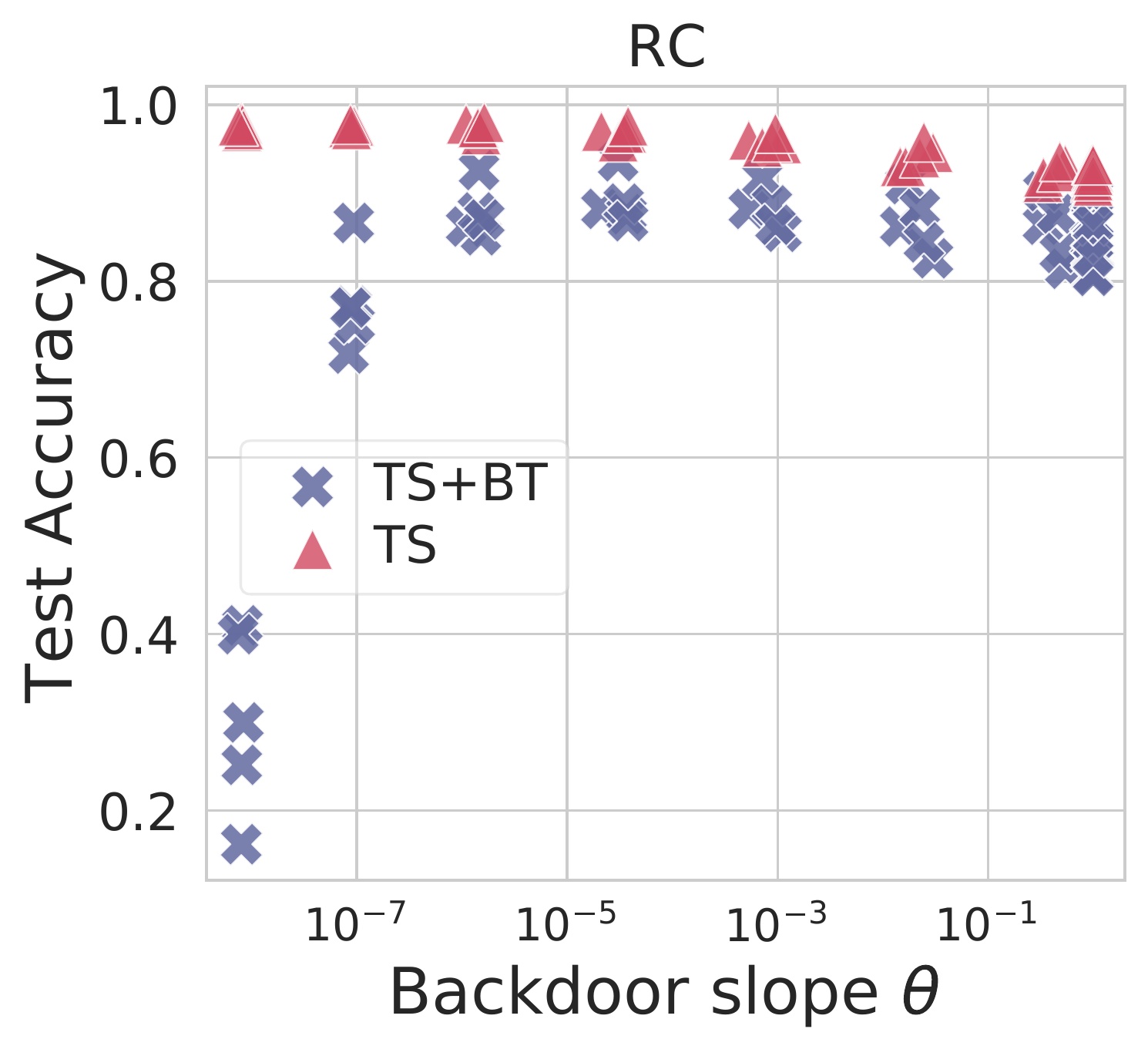}
\includegraphics[width=0.242\textwidth]{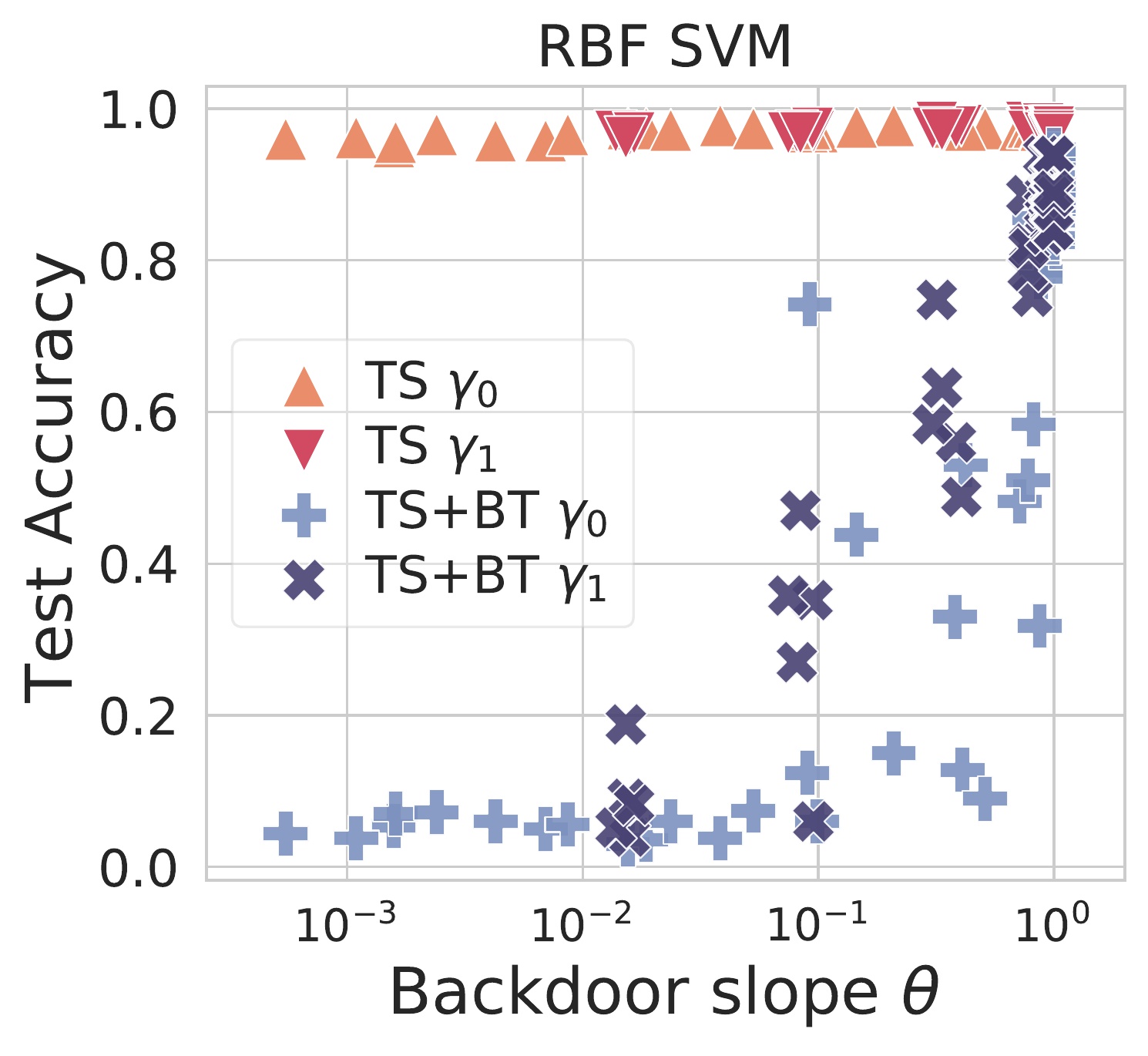}

\includegraphics[width=0.242\textwidth]{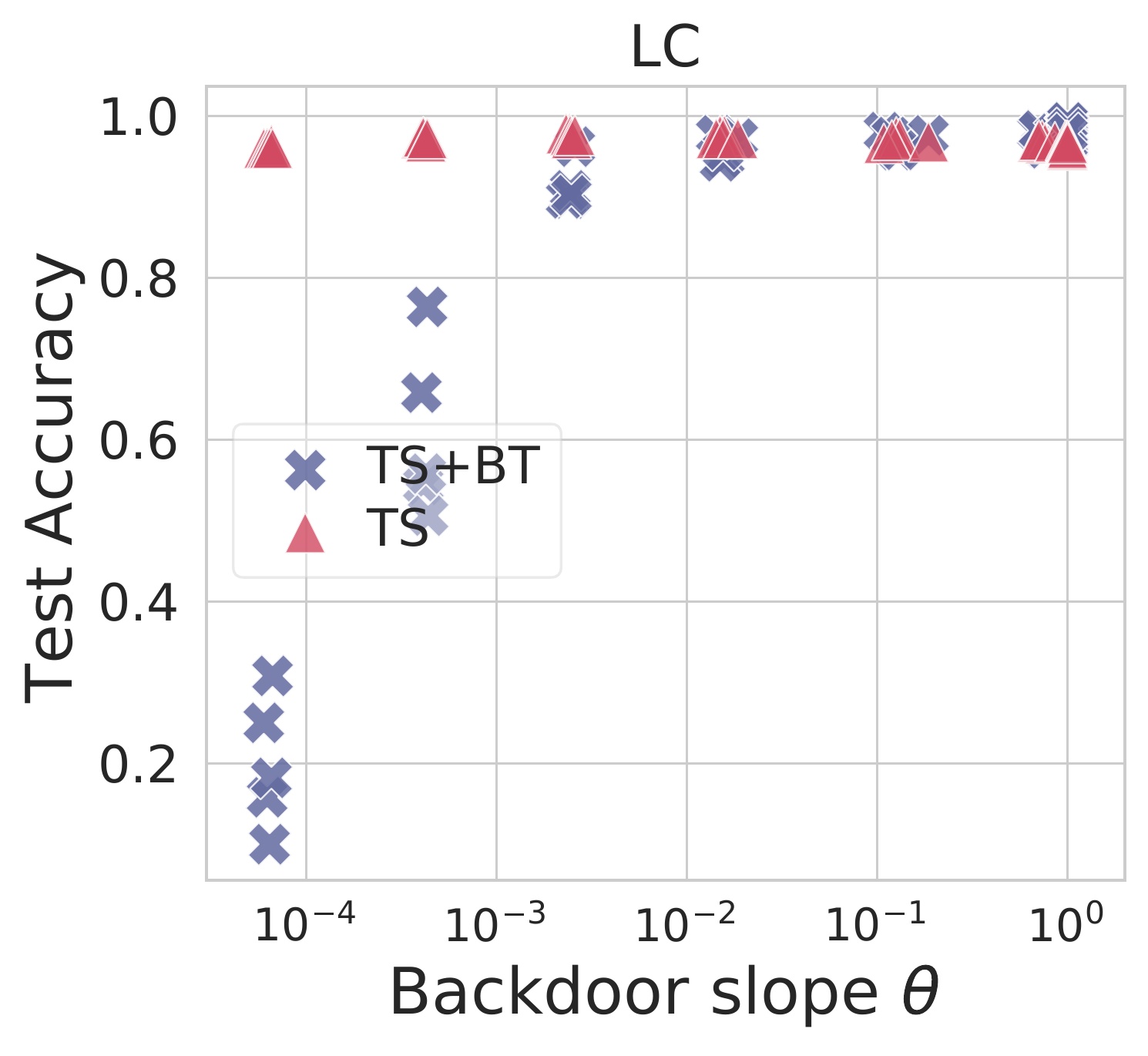}
\includegraphics[width=0.242\textwidth]{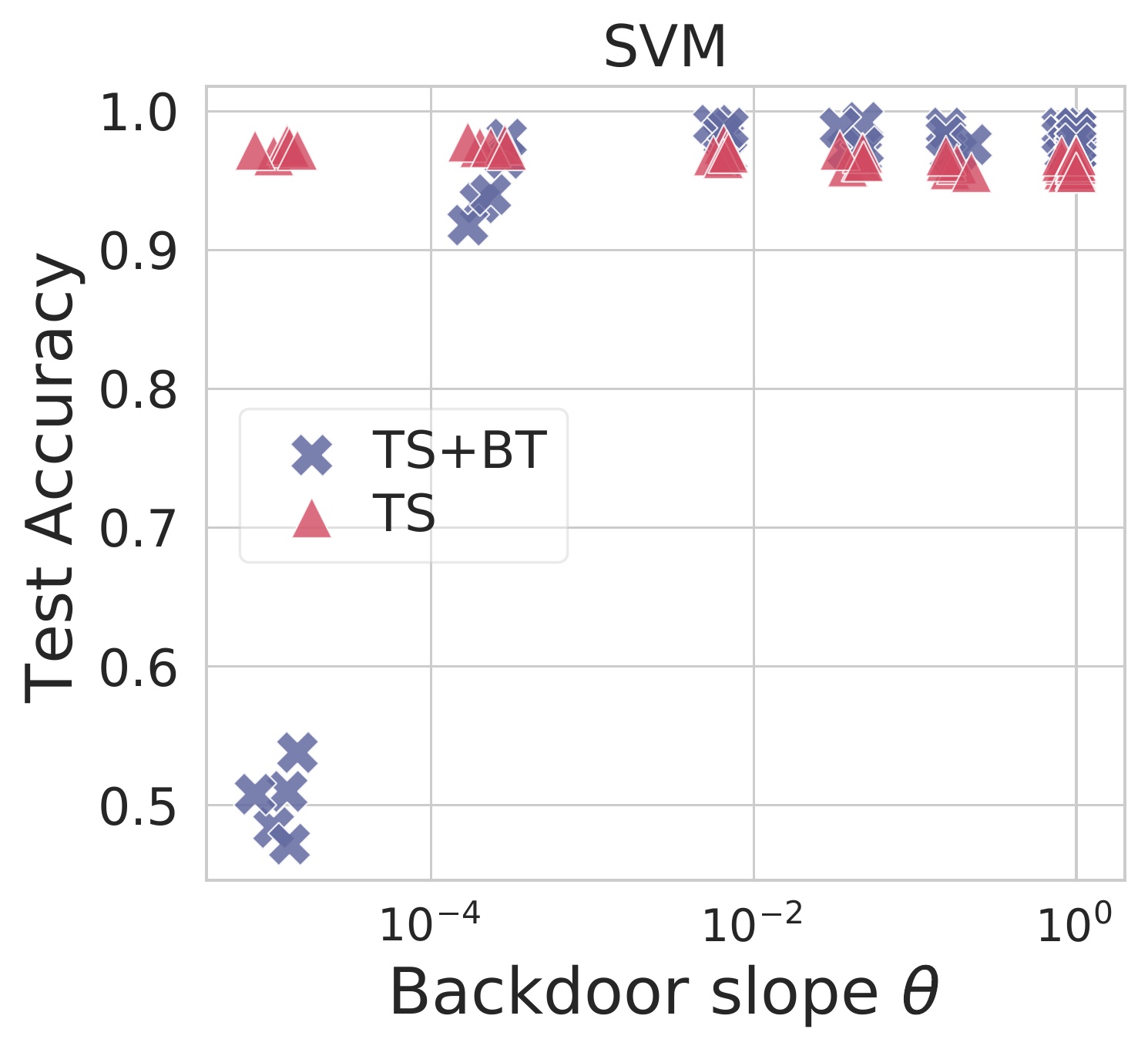}
\includegraphics[width=0.242\textwidth]{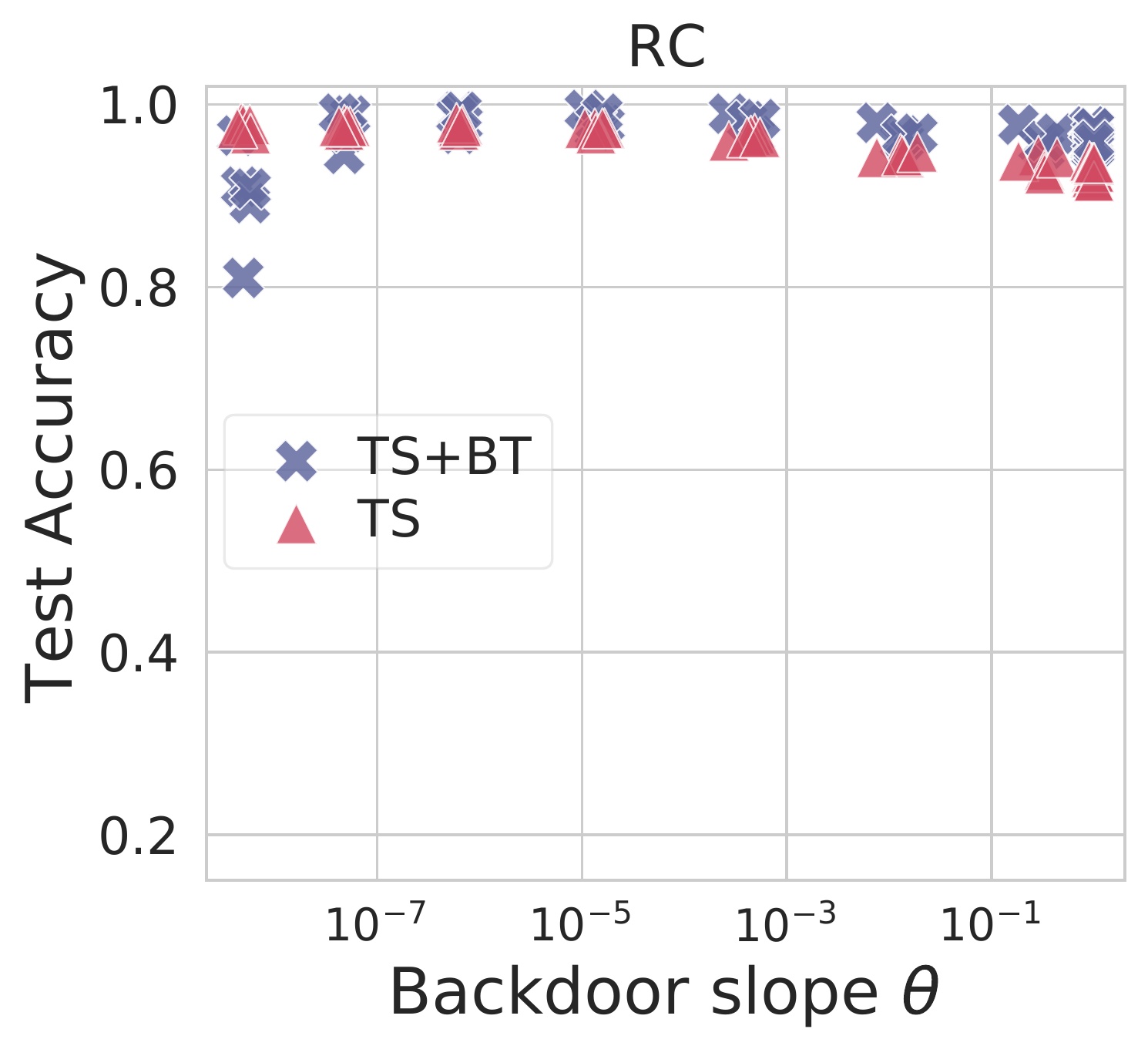}
\includegraphics[width=0.242\textwidth]{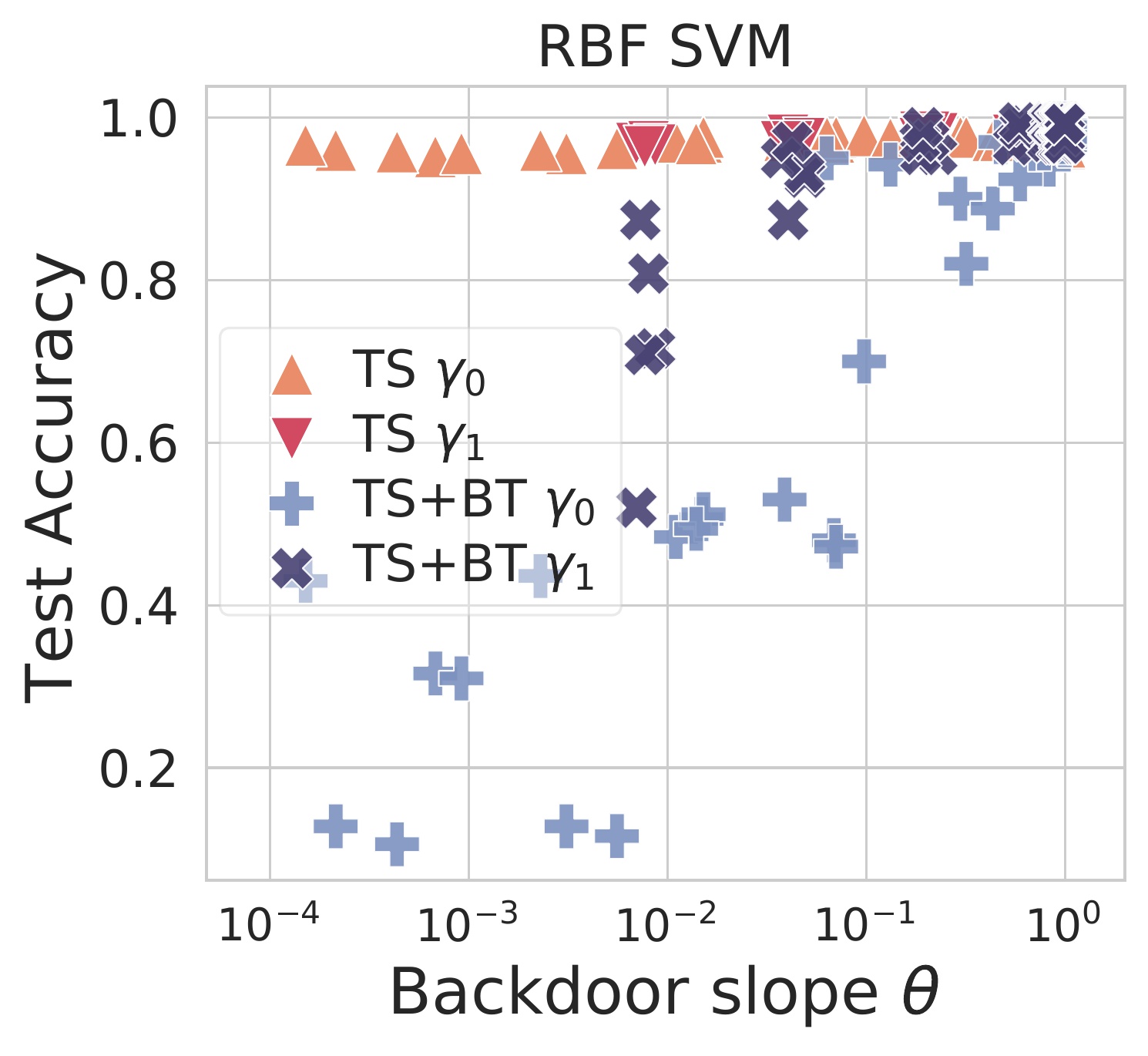}

 \caption{Backdoor slope $\theta$ vs clean accuracy (red) and backdoor effectiveness (blue) on CIFAR10 \cifarairplanefrog with backdoor trigger size $8 \times 8$ (top row), and $16 \times 16$ (bottom row). We measure the classification accuracy on the \rebuttal{untainted test samples (TS)}, and on the same samples \rebuttal{after injecting the backdoor trigger (TS+BT)}. We chose the $\gamma$ parameter for the RBF kernel as $\gamma_0=\expnumber{1}{-04}$ (orange triangle for clean data, light blue plus for data with trigger) and $\gamma_1=\expnumber{1}{-03}$ (red inverted triangle for clean data, dark blue x for data with trigger).}
  \label{fig:resultsSlopeCIFAR}
\end{figure*}
\begin{figure*}[h!]
  \centering
  
\includegraphics[width=0.242\textwidth]{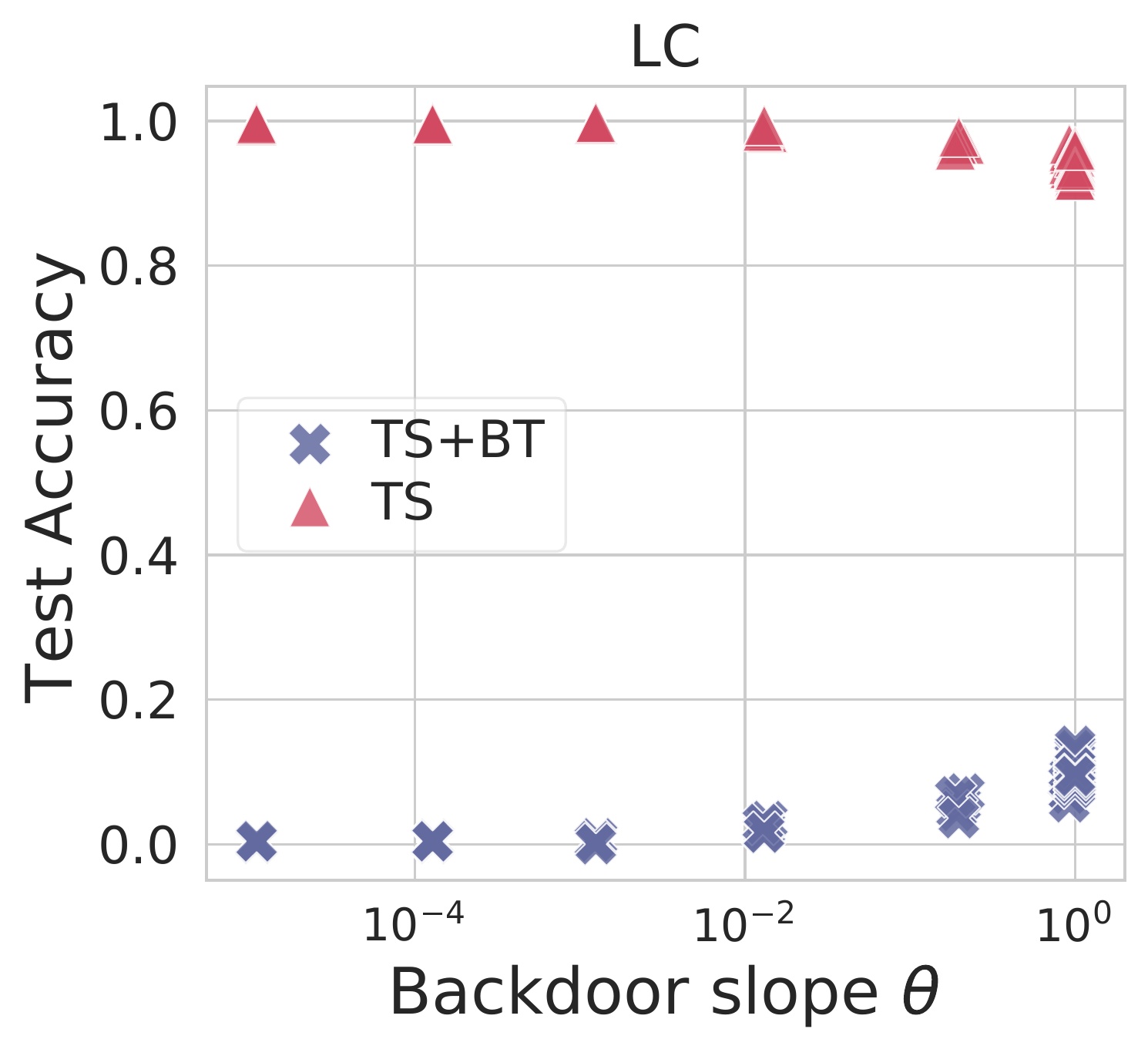}
\includegraphics[width=0.242\textwidth]{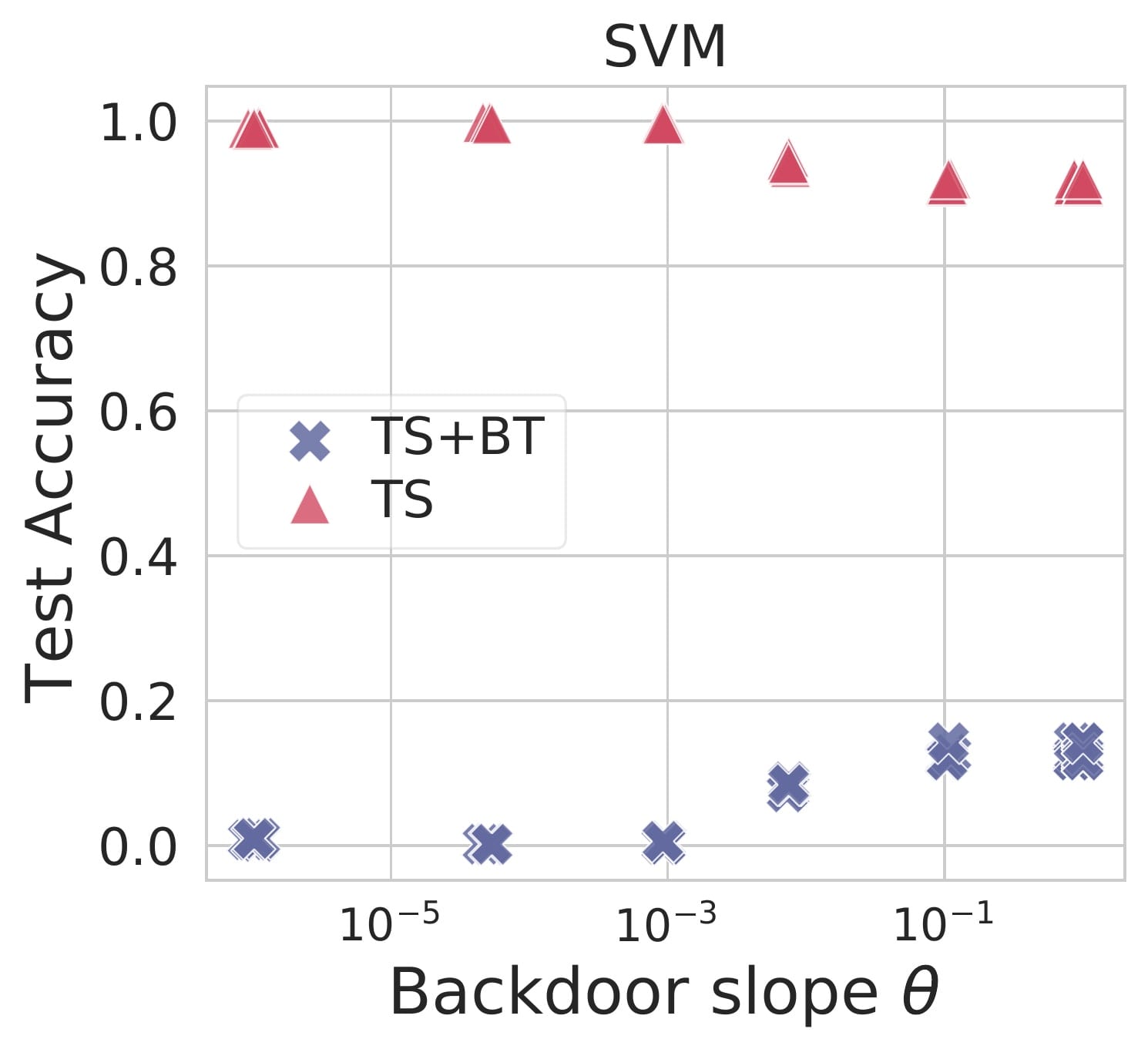}
\includegraphics[width=0.242\textwidth]{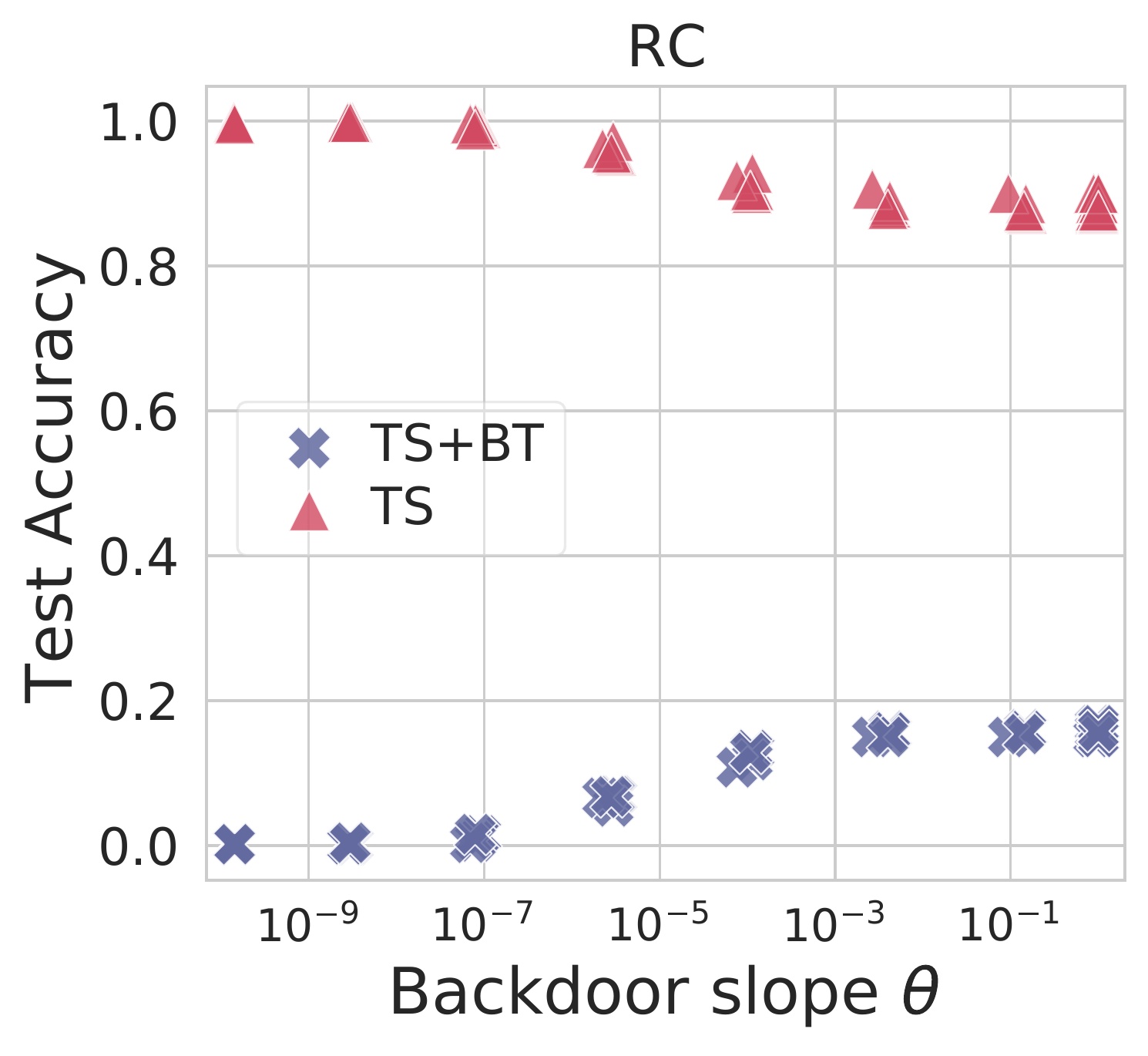}
\includegraphics[width=0.242\textwidth]{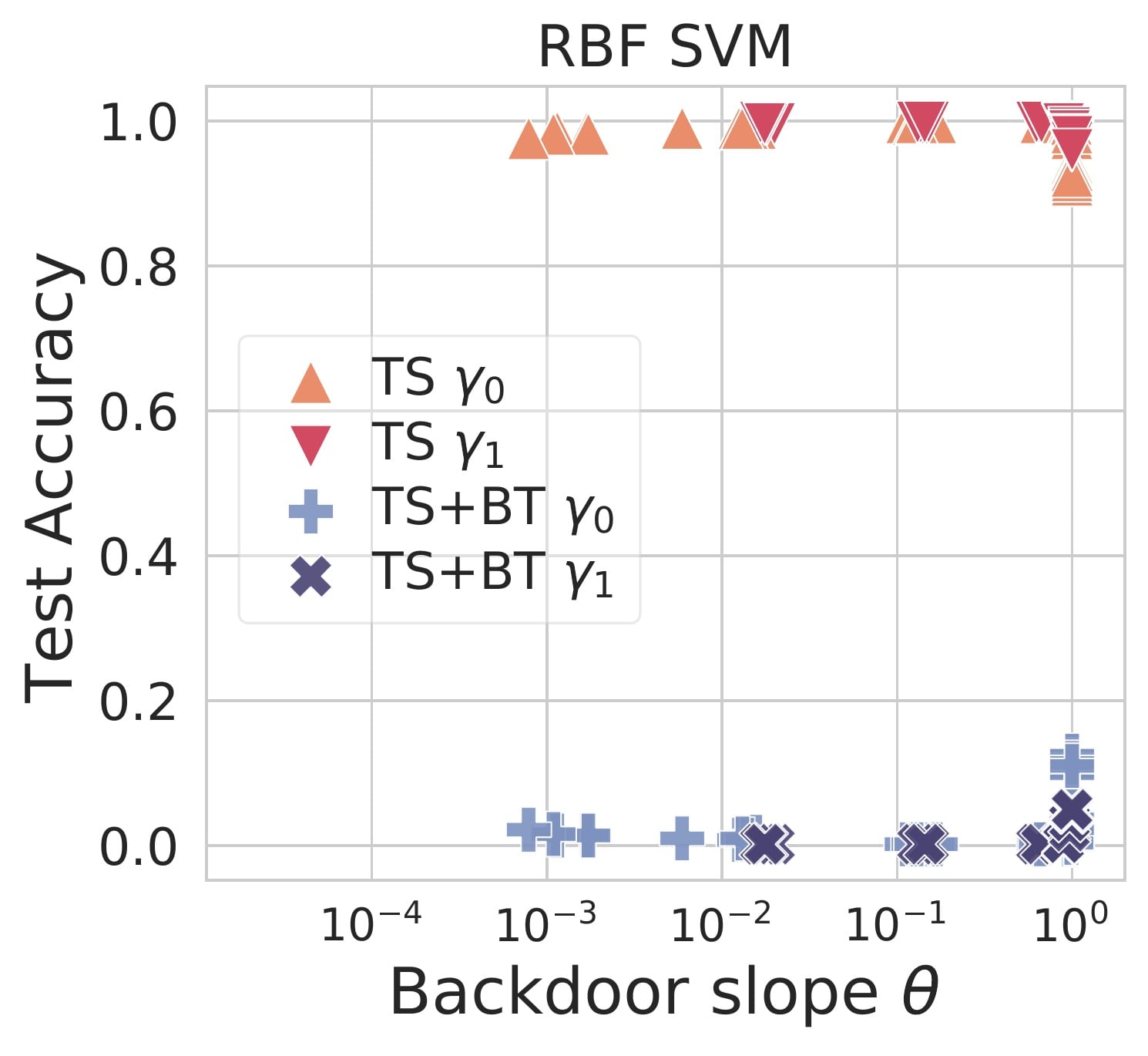}

\includegraphics[width=0.242\textwidth]{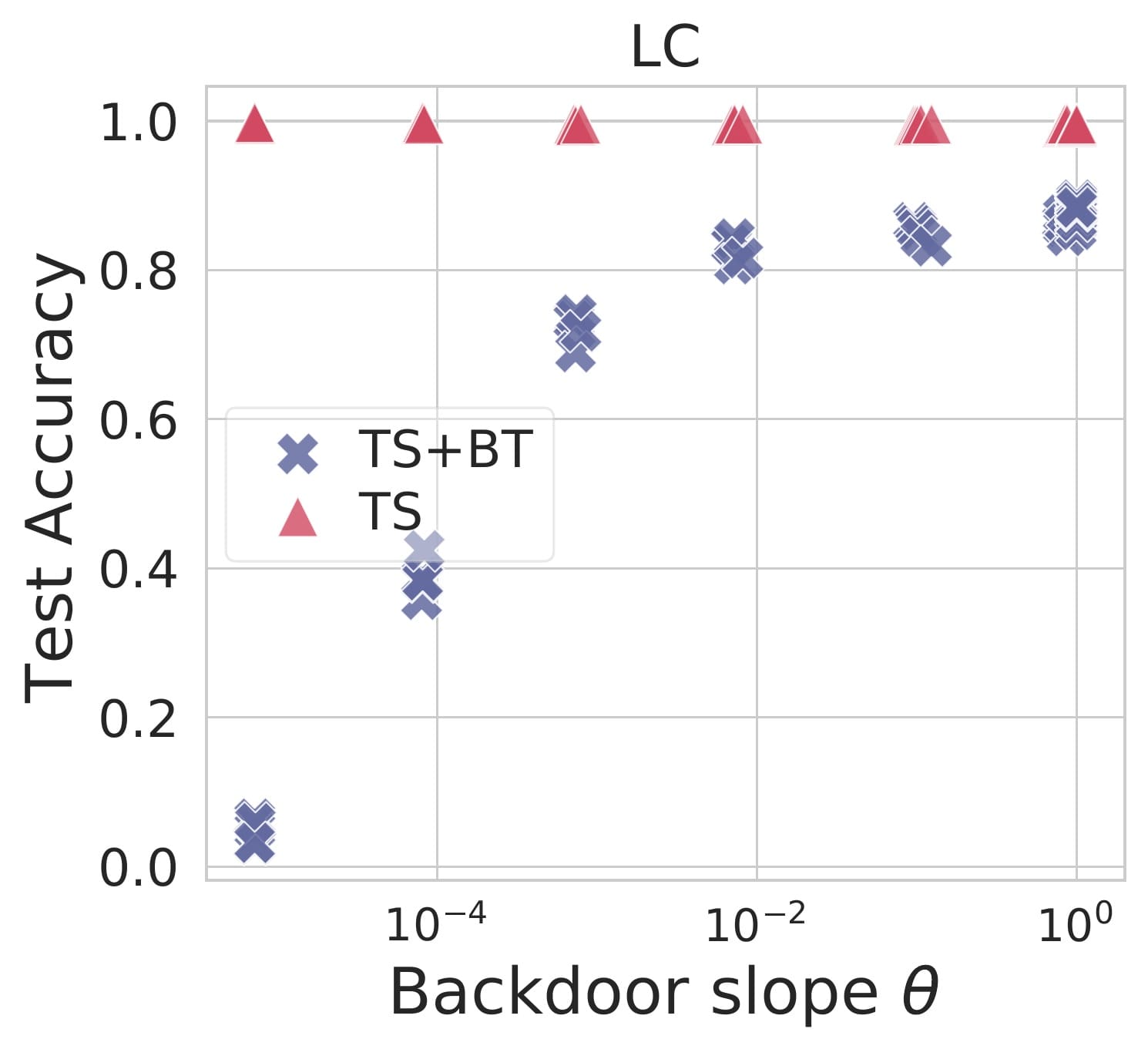}
\includegraphics[width=0.242\textwidth]{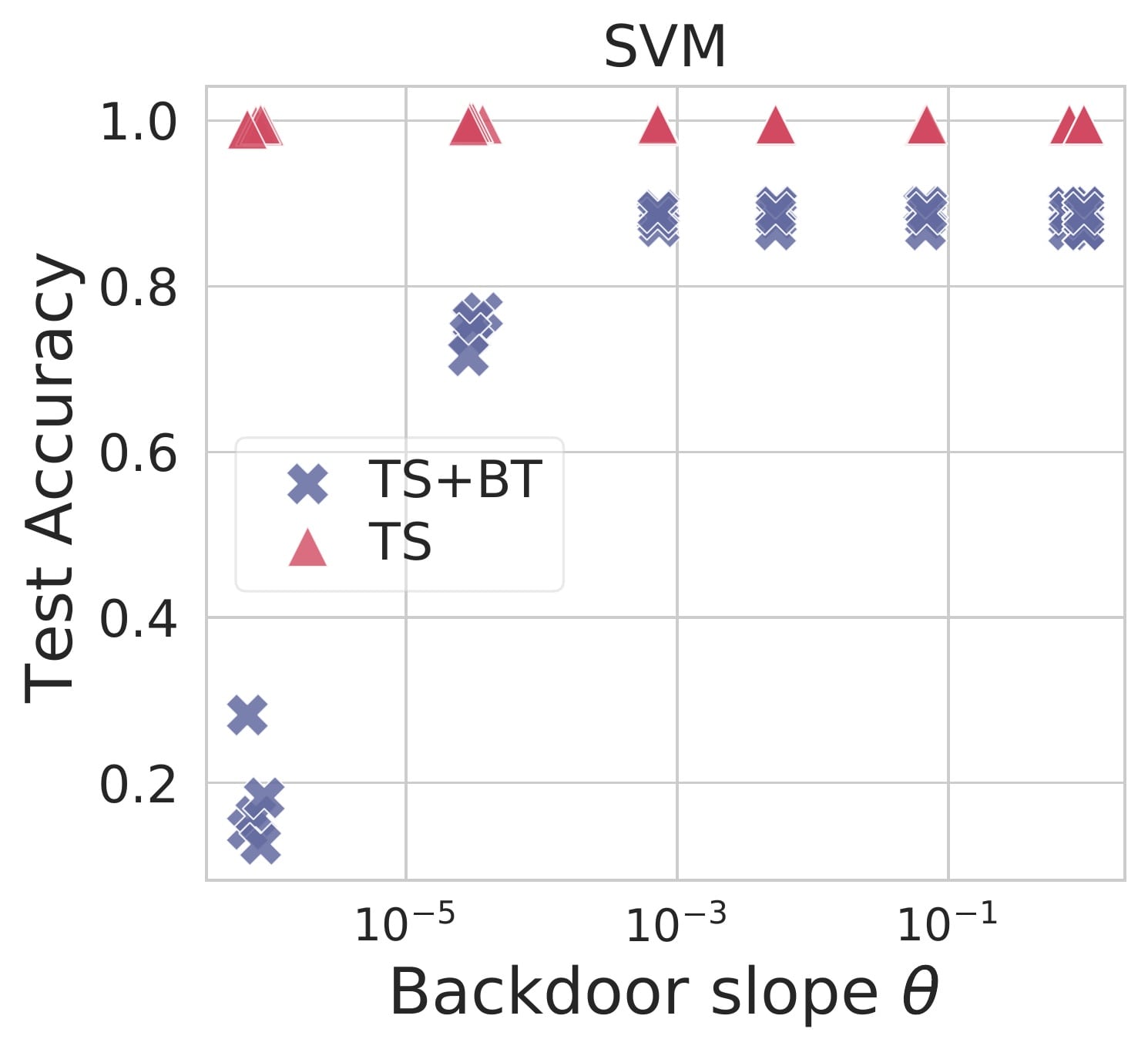}
\includegraphics[width=0.242\textwidth]{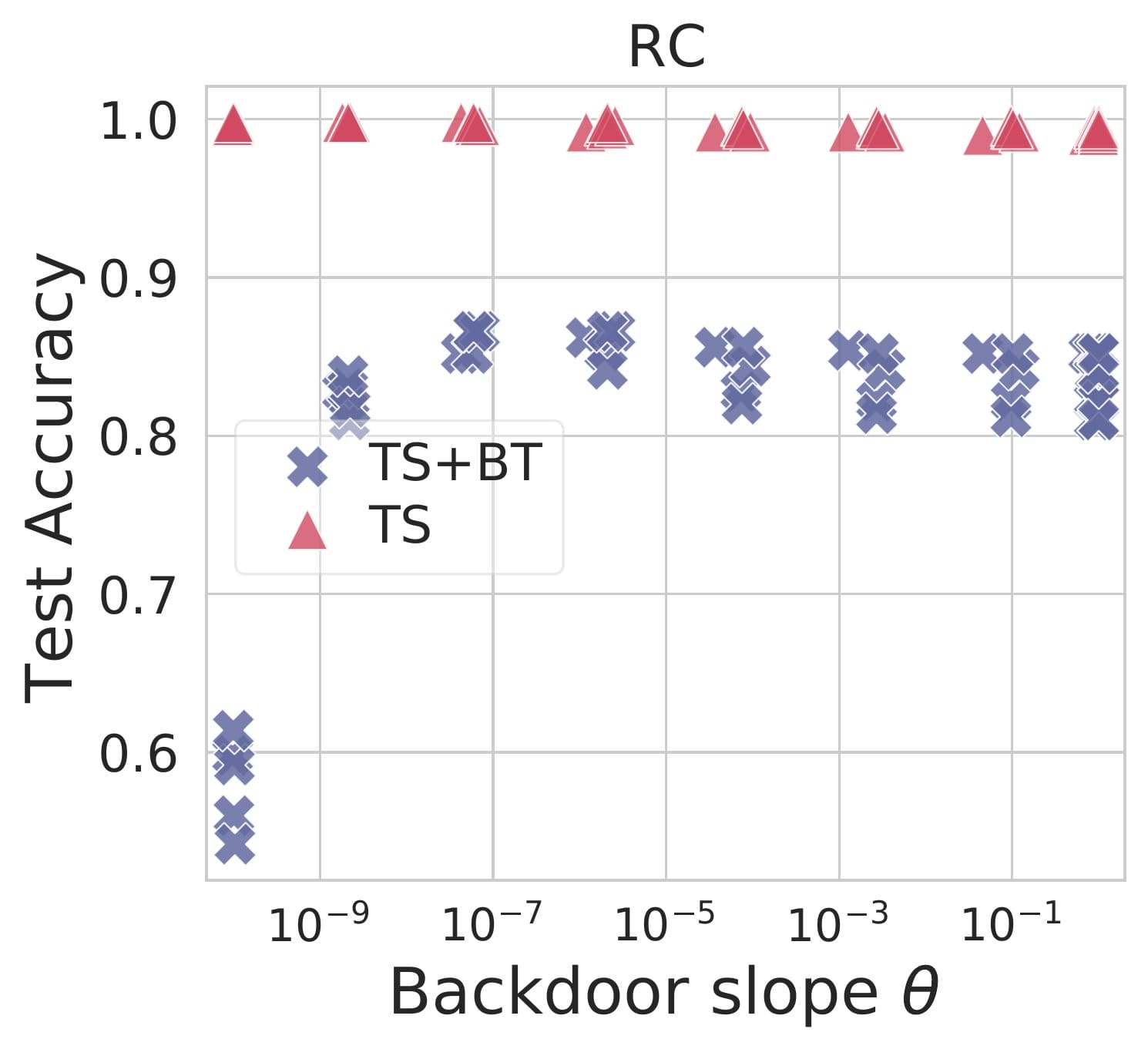}
\includegraphics[width=0.242\textwidth]{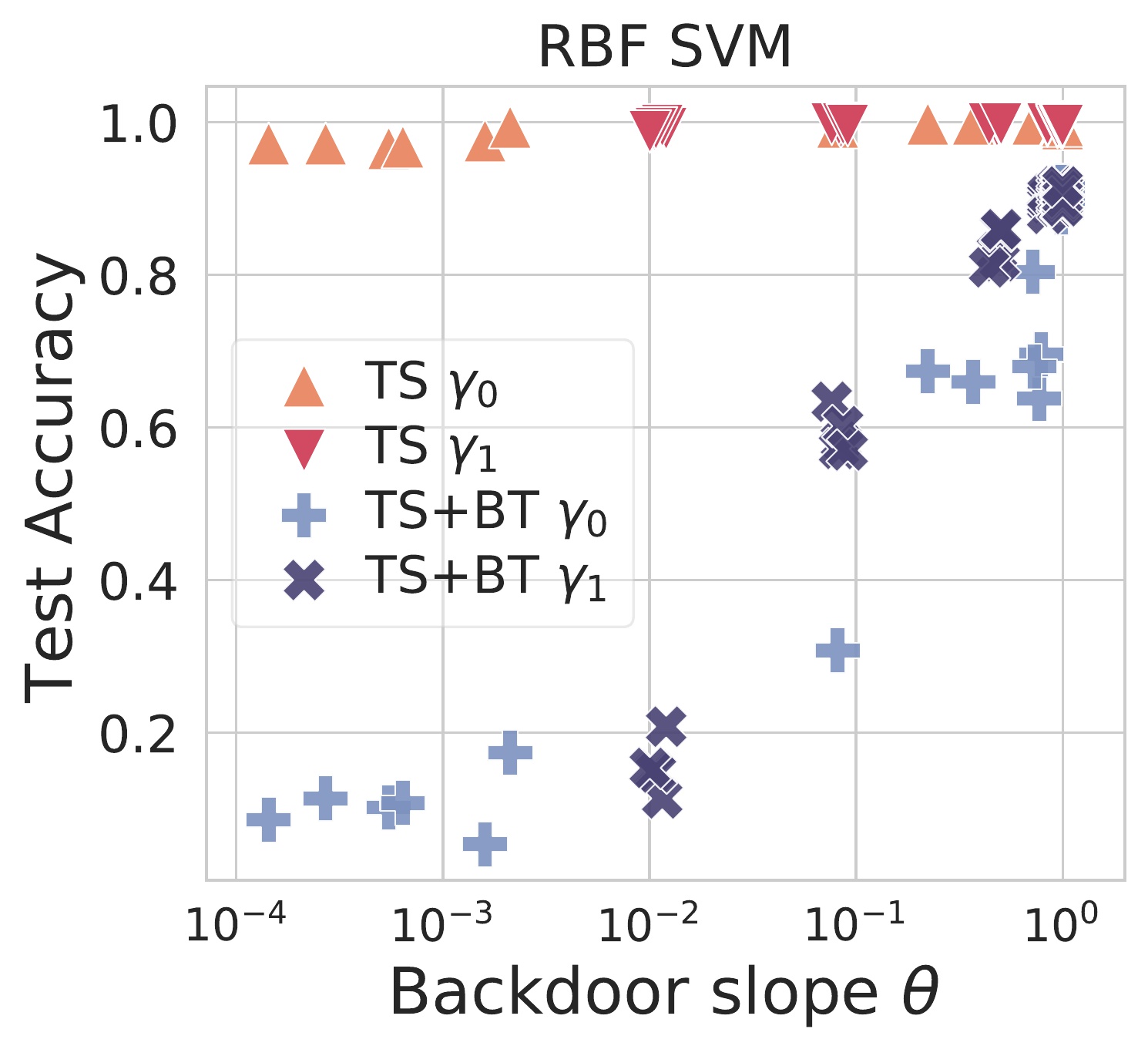}

 \caption{Backdoor slope $\theta$ vs clean accuracy (red) and backdoor effectiveness (blue) on Imagenette \imagenettetenchtruck with trigger visibility $c_m=10$, \ie almost imperceptible, (top row) and $c_m=75$ (bottom row). We measure the classification accuracy on the \rebuttal{untainted test samples (TS)}, and on the same samples \rebuttal{after injecting the backdoor trigger (TS+BT)}.  We chose the $\gamma$ parameter for the RBF kernel as $\gamma_0=\expnumber{1}{-05}$ (orange triangle for clean data, light blue plus for data with trigger) and $\gamma_1=\expnumber{1}{-04}$ (red inverted triangle for clean data, dark blue x for data with trigger).}
  \label{fig:resultsSlopeImagenette}
\end{figure*}

\subsubsection{Backdoor Slope} 
From the previous results, we have seen that reducing complexity through regularization increases robustness against backdoors. For a deeper understanding of model complexity on backdoor learning, we leverage the proposed backdoor slope. 
In our experiments for convex learners, we fix the fraction of injected poisoning points to $0.1$, as by Gu et al.~\cite{gu2019badnets}, and we report a dot for each combination of $\lambda$ and $\gamma$ as specified in Section~\ref{sec:hyperparam-choice}.  Figures~\ref{fig:resultsSlopeMNIST}-\ref{fig:resultsSlopeImagenette} show the relationship between the backdoor slope and the backdoor effectiveness, measured as the percentage of samples with the trigger that mislead the classifier, respectively for MNIST, CIFAR10 and Imagenette. 
\rebuttal{We report the accuracy on the clean test dataset (TS) and the test dataset with the backdoor trigger (TS+BT)}. For the RBF SVM, we report the accuracy for two different $\gamma$ values. 

Interestingly, our plots show a region where the accuracy of the classifiers on benign samples is high, yet the classifier exhibits low accuracy on samples with triggers.
For linear classifiers, this region equals low-regularized classifiers. In the case of the RBF SVM, the best trade-off is achieved with high $\lambda$ (strong regularization) and small $\gamma$, which also constrain SVM's complexity. 
Our results thus indicate that in these cases, the classifier is not flexible enough to learn the backdoor in addition to the clean test samples. Conversely, as long as the classifier has enough flexibility, it can learn the backdoor without sacrificing clean test accuracy. In a nutshell, by choosing the hyperparameters appropriately, we can obtain a classifier able to learn the original task but not the backdoor. However, there is a trade-off between the accuracy of the original task and the robustness of backdoor classification.  
In Figures~\ref{fig:resultsSlopeMNIST}-\ref{fig:resultsSlopeImagenette}, we extend the comparison between the backdoor learning slope and the attack effectiveness, considering a stronger attack that exploits larger or more visible triggers. With these attacks, the trade-off region is diminished, resulting in fewer viable configurations of hyperparameters that yield a robust model.  
This result aligns with our earlier findings based on backdoor learning curves: as the attack strength increases, the learning curve descends more steeply, exhibiting a higher backdoor slope.
Our results suggest that system designers should thus prioritize maximizing regularization in models, ensuring the tradeoff with accuracy remains acceptable, to deploy a more robust and effective machine learning model against potential backdoor attacks.
\begin{figure*}[t]
  \centering
    \includegraphics[width=1\textwidth]{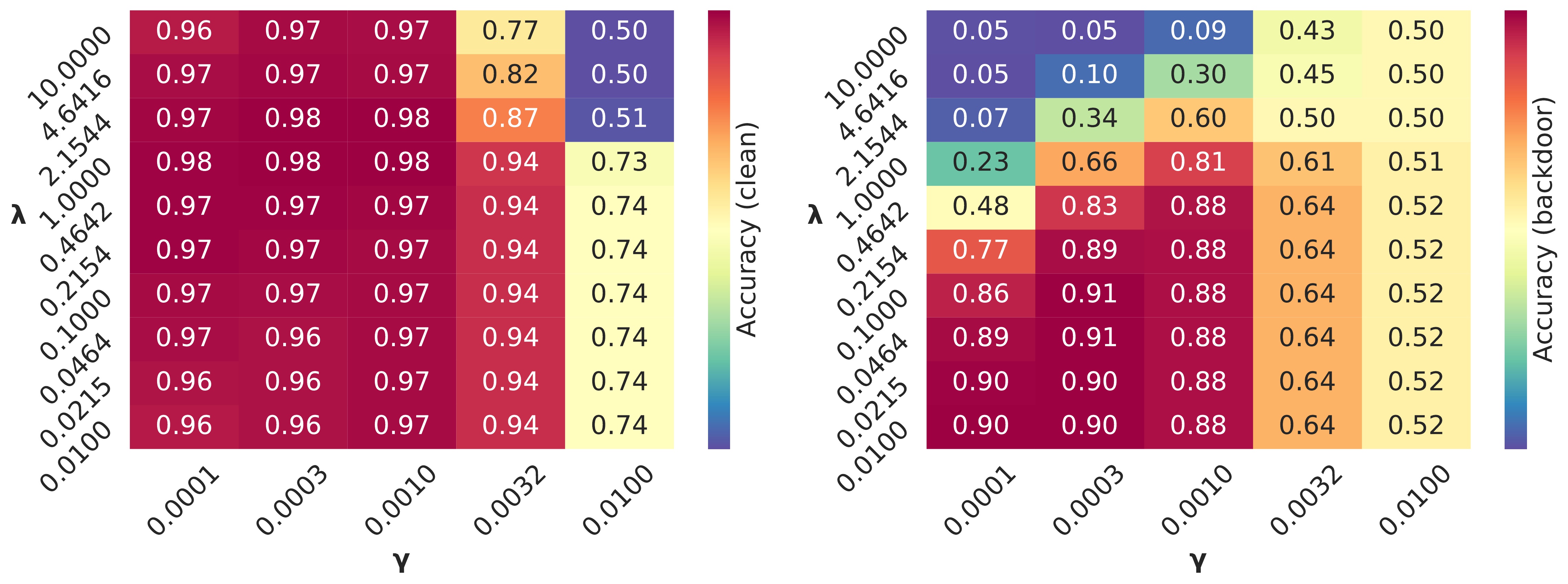}
    \includegraphics[width=1\textwidth]{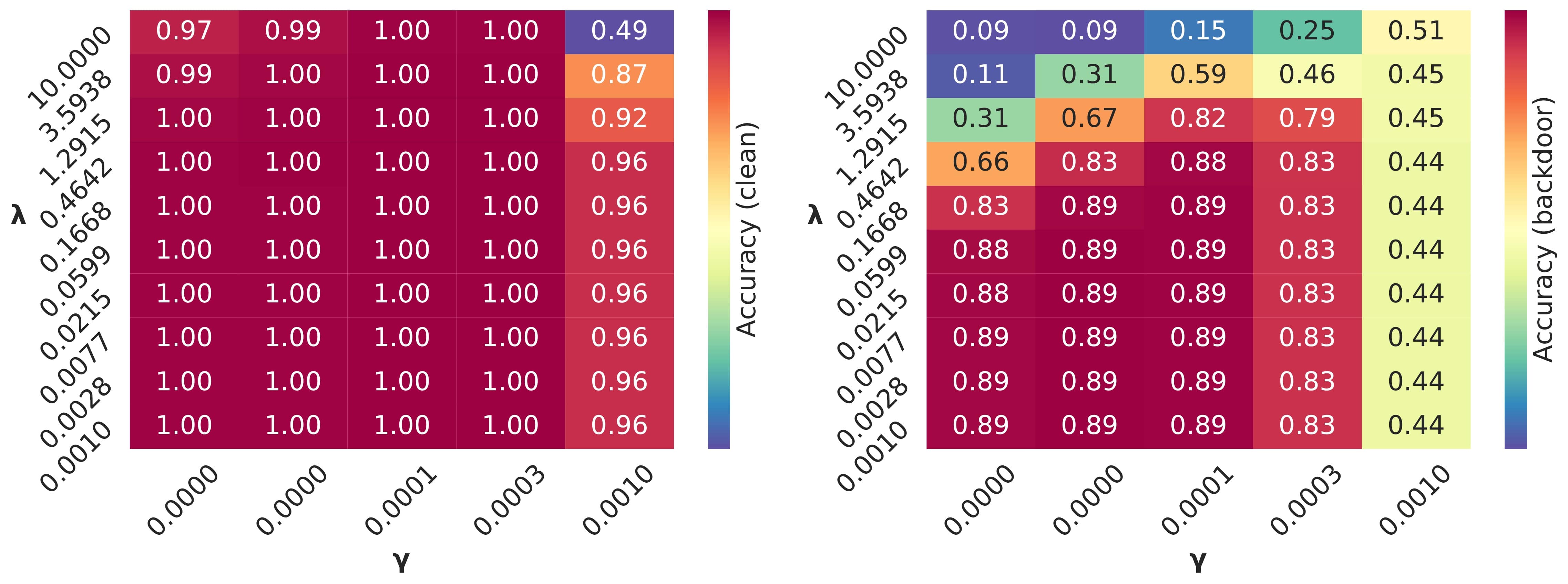}
 \caption{Influence of $\gamma$ (x-axis) and $\lambda$ (y-axis) on the backdoor effectiveness (left) and clean accuracy (right) for CIFAR10 \cifarairplanefrog (top row) and Imagenette \imagenettetenchtruck (bottom row). The Backdoor is mounted with trigger size $8 \times 8$ for CIFAR10 and visibility $c_m=75$ for Imagenette. The plots show that there are hyperparameter configurations for which clean accuracy is high (red regions on the left plots), while the accuracy on the backdoored points is low (blue regions on the right plots).}
  \label{fig:influence_of_gamma_rbf}
\end{figure*}

As a final check to assess the reliability of the backdoor learning slope, we plot in Figure~\ref{fig:influence_of_gamma_rbf} clean and backdoor accuracy when training an RBF SVM with different hyperparameter configurations. We train, for each configuration, the classifier on the poisoned dataset. On the top row, we show the results for CIFAR10 \cifarairplanefrog, and on the bottom row the results for Imagenette \imagenettetenchtruck. We followed the same backdoor setting for the backdoor slope in Figure~\ref{fig:resultsSlopeMNIST}-\ref{fig:resultsSlopeImagenette}, \ie trigger size $8 \times 8$ for MNIST and trigger visibility $c_m=75$ for Imagenette. 
Analogous to our previous findings, there exists a trade-off region where the clean accuracy is high (red), while the backdoor accuracy is low (blue), suggesting a higher robustness against backdoors. Analogous to the backdoor slope, the best region is obtained with reduced complexity, thus regularizing it or reducing $\gamma$. Consequently, we conclude that smaller values of $\gamma$ can effectively guide the model toward a more robust configuration against backdoor attacks. Nevertheless, it also emerges that the most decisive factor influencing model robustness remains the regularization term.

\begin{table*}[t]
\caption{Backdoor learning slope for Resnet18 and Resnet50 when increasing the percentage of backdoor poisoning $p$, the number of epochs \textit{(\#Epochs)}, and parameter $h$ for estimate in Eq.~\ref{eq:estimate_influence}. We also report the corresponding backdoor effectiveness \textit{(Accuracy TS+BT)} and clean accuracy \textit{(Accuracy TS)}, measures respectively as the percentage correctly classified test samples with and without the backdoor trigger.}
\centering
\begin{tabular}{ccccccccc}
\toprule

Model & $p$ & \#Epochs & 
\begin{tabular}[c]{@{}c@{}}Slope\\h=0.01\end{tabular} & 
\begin{tabular}[c]{@{}c@{}}Slope\\h=0.1\end{tabular} & \begin{tabular}[c]{@{}c@{}}Slope\\h=0.2\end{tabular} & \begin{tabular}[c]{@{}c@{}}Slope\\h=1\end{tabular} & \begin{tabular}[c]{@{}c@{}}Accuracy\\TS+BT\end{tabular} & \begin{tabular}[c]{@{}c@{}}Accuracy\\TS\end{tabular} \\
\midrule
Resnet18 & 0.05 & 10 & 0.9955 & 0.9872 & 0.9752 & 0.9026 & 0.4163 & 0.9588 \\
Resnet50 & 0.05 & 10 & 0.9965 & 0.9895 & 0.9785 & 0.9169 & 0.7197 & 0.9781 \\
\addlinespace
Resnet18 & 0.05 & 50 & 0.9986 & 0.9900 & 0.9797 & 0.9281 & 0.5256 & 0.9737 \\ 
Resnet50 & 0.05 & 50 & 0.9992 & 0.9936 & 0.9849 & 0.9377 & 0.8067 & 0.9881 \\
\addlinespace
\hdashline
\addlinespace
Resnet18 & 0.15 & 10 & 0.9955 & 0.9878 & 0.9774 & 0.9189 & 0.8804 & 0.9568 \\
Resnet50 & 0.15 & 10 & 0.9966 & 0.9902 & 0.9943 & 0.9231 & 0.9440 & 0.9826 \\
\addlinespace
Resnet18 & 0.15 & 50 & 0.9987 & 0.9937 & 0.9864 & 0.9384 & 0.8893 & 0.9720 \\ 
Resnet50 & 0.15 & 50 & 0.9992 & 0.9939 & 0.9971 & 0.9403 & 0.9509 & 0.9890 \\ 

\bottomrule

\end{tabular}%
\label{tab:nn-slope}
\end{table*}

While this measure works well on convex learners, its roots in influence functions prevent a direct application on deep neural networks. As pointed out in~\cite{basu2020influence} the analytical gradient in Eq.~\ref{eq:backdoor_slope} at $\beta = 0$ is unstable for deep neural networks. To overcome this deficiency, we estimate it with finite difference approximation, obtaining:
\begin{equation}
    \left.\diff{L}{\beta} \right\vert_{\beta=0}  = \frac{L(\set P_{\rm ts}, \vct w^\star(h)) - L(\set P_{\rm ts}, \vct w^\star(0))}{h} \, .
    \label{eq:estimate_influence}
\end{equation}
We report the the results for Resnet18 and Resnet50 in Table~\ref{tab:nn-slope} where we used $h = \{0.01, 0.1, 0.2, 1\}$. For each combination of poisoning percentage and number of epochs, we report the estimate of the backdoor learning slope when choosing different $h$ values. The closer $h$ is to $0$, the closer to $1$ is the backdoor slope of the neural network. This result is consistent with Figure~\ref{fig:backdoor_learning_curves_resnet}, where the backdoor learning curves drop similarly fast, suggesting a high vulnerability of the model in the presence of backdoor samples. 
A subtle difference is that when increasing $h$, there is more evidence for higher vulnerability of neural networks trained with more epochs or when increasing the percentage of poisoning points.

\myparagraph{Remarks on Model Selection / Hyperparameter Tuning.} The observed degradation of accuracy on the backdoored samples (blue dots) at lower slopes in Figures~\ref{fig:resultsSlopeMNIST}-\ref{fig:resultsSlopeImagenette} suggests that models with a minimal backdoor learning slope, induced by a strong regularization, retain the ability to remain robust. Quite surprisingly, this phenomenon does not have a substantial impact on the classification accuracy of the pristine samples (red triangles), highlighting an advantageous trade-off for defending against backdoor attacks.
In other words, there is a wide region in the hyperparameter space in which the model still keeps a very high accuracy on clean data, but it is essentially unable to learn the backdoor samples. This trade-off enables the defender to find a sweet spot in which the model can be sufficiently robust to backdoor attacks. 
In practice, this means that well-regularized models can be made resilient to backdoor attacks with a negligible impact on classification accuracy, by simply performing an appropriate choice of the hyperparameters, including the regularization term $\lambda$, number of epochs or neurons. In particular, such hyperparameters can be tuned to regularize the learning process as much as possible, while retaining an acceptable classification accuracy for the task at hand. 
We conclude this section by pointing out that this finding is all but trivial, as also highlighted in recent work: Bagdasaryan and Shmatikov~\cite{Bagdasaryan2023HyperparameterSI} claim that using well-regularized models is the only \textit{effective} and \textit{applicable} defense for systems that are deployed and maintained in practical machine-learning operations (MLOps) pipelines, as any other defensive technique will require significant and costly changes to the deployment pipeline infrastructures, inducing a high technical debt on future system maintenance. \smallskip

\subsubsection{Empirical Parameter Deviation Plots}
After having investigated which factors influence backdoor effectiveness, we shift our focus to examining how the model's weights change during the training process when the dataset is tainted with backdoor samples. We aim to determine whether there is an increase in complexity or not.

\begin{figure*}[t]
  \centering
  \includegraphics[width=0.235\textwidth]{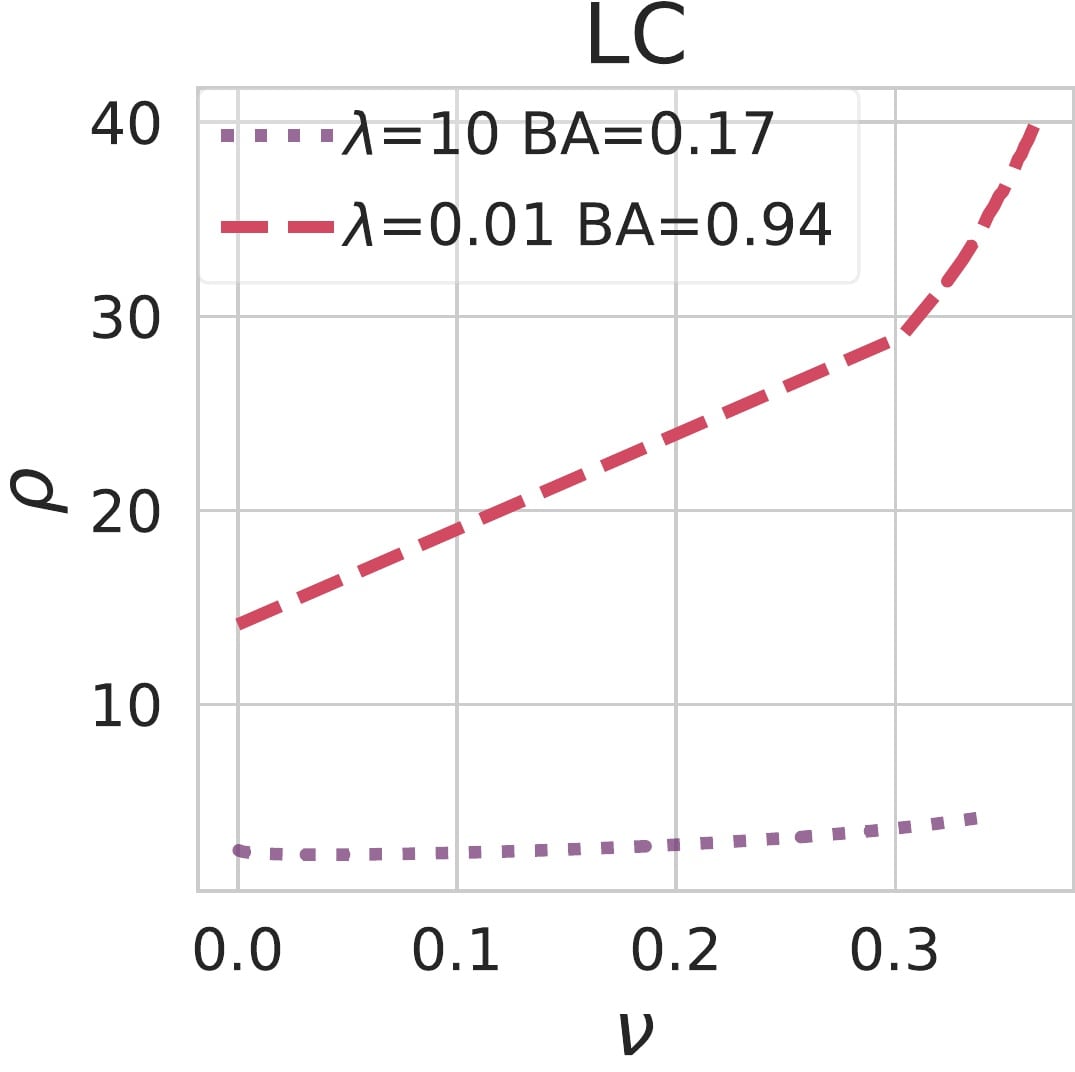}
  \includegraphics[width=0.24\textwidth]{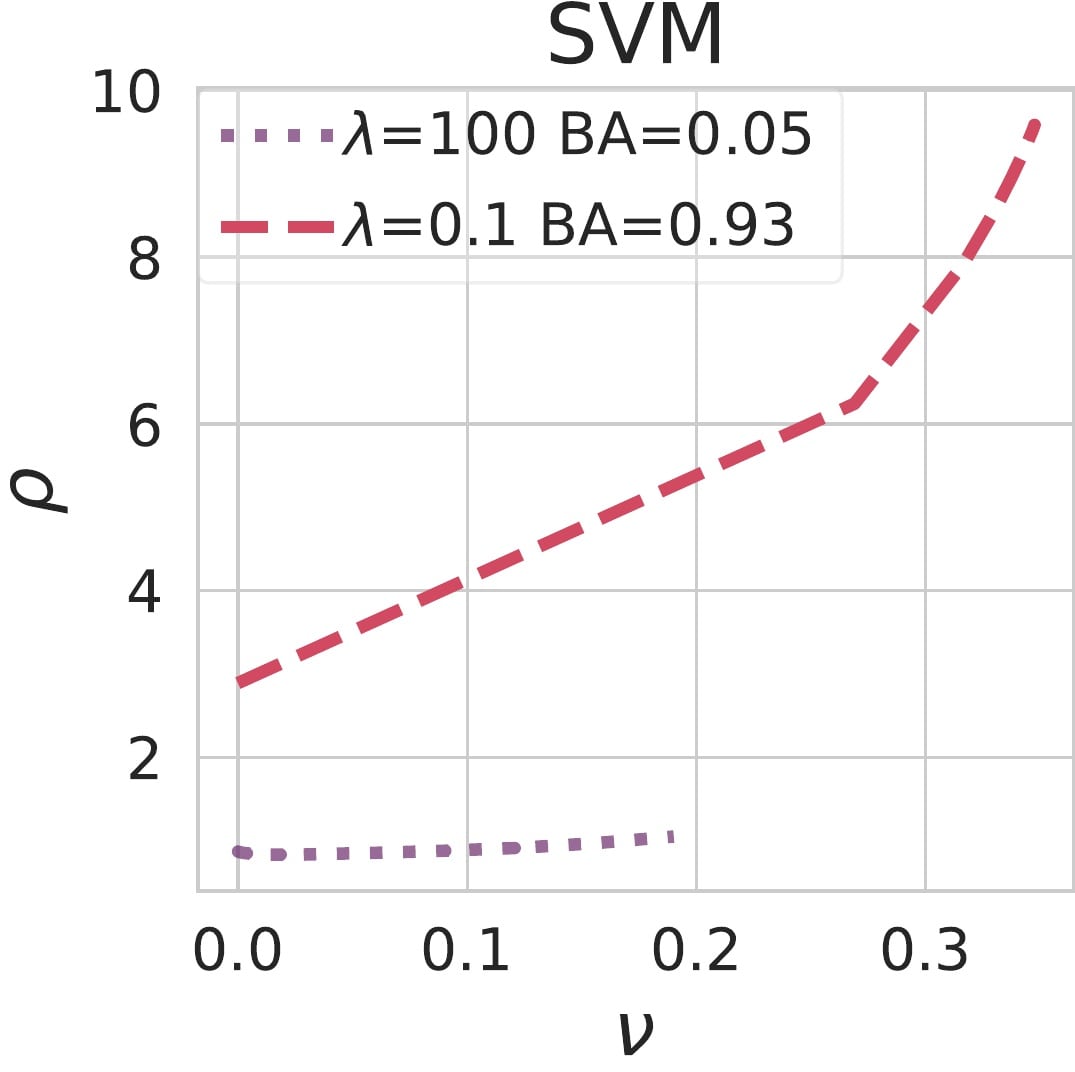}
  \includegraphics[width=0.24\textwidth]{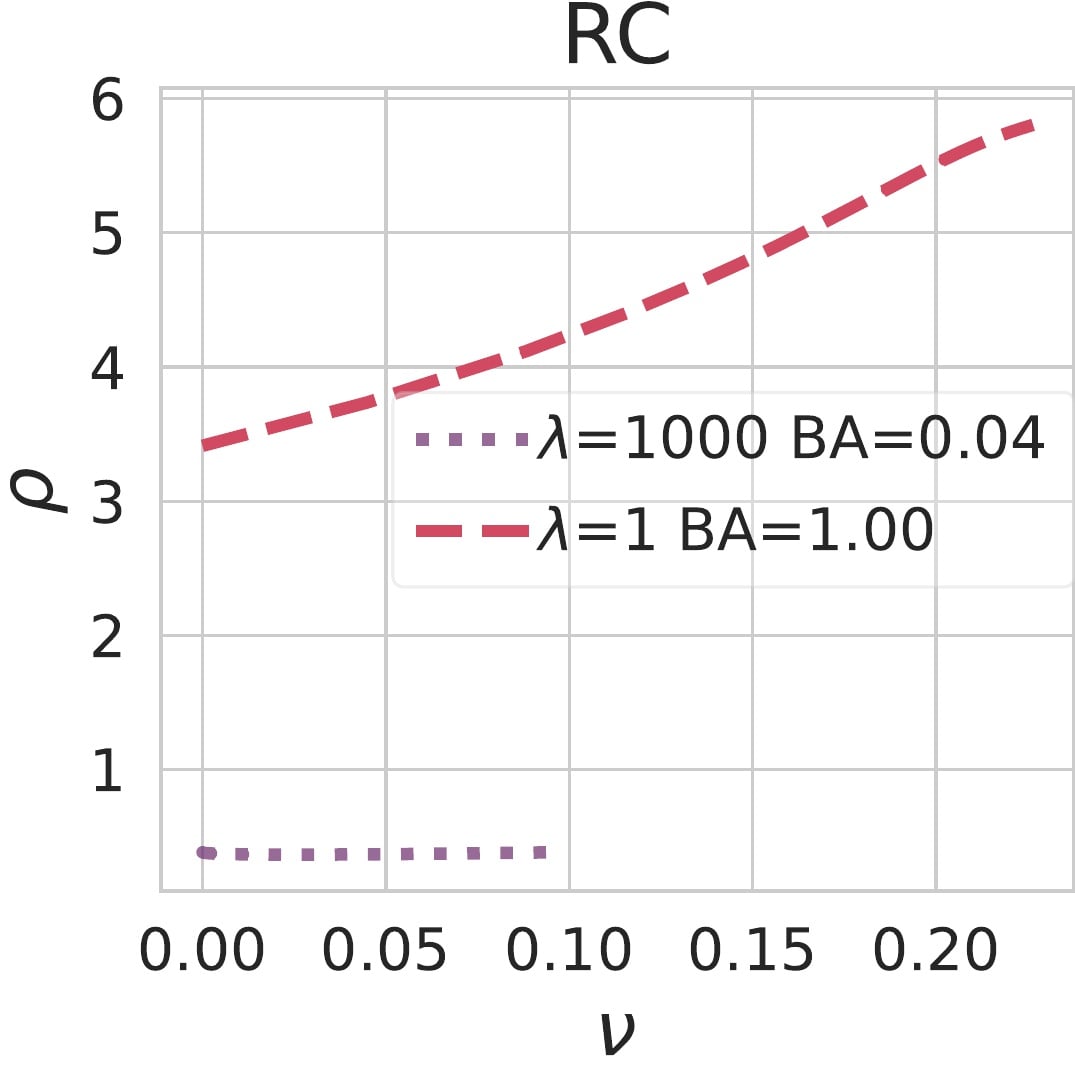}
  \includegraphics[width=0.24\textwidth]{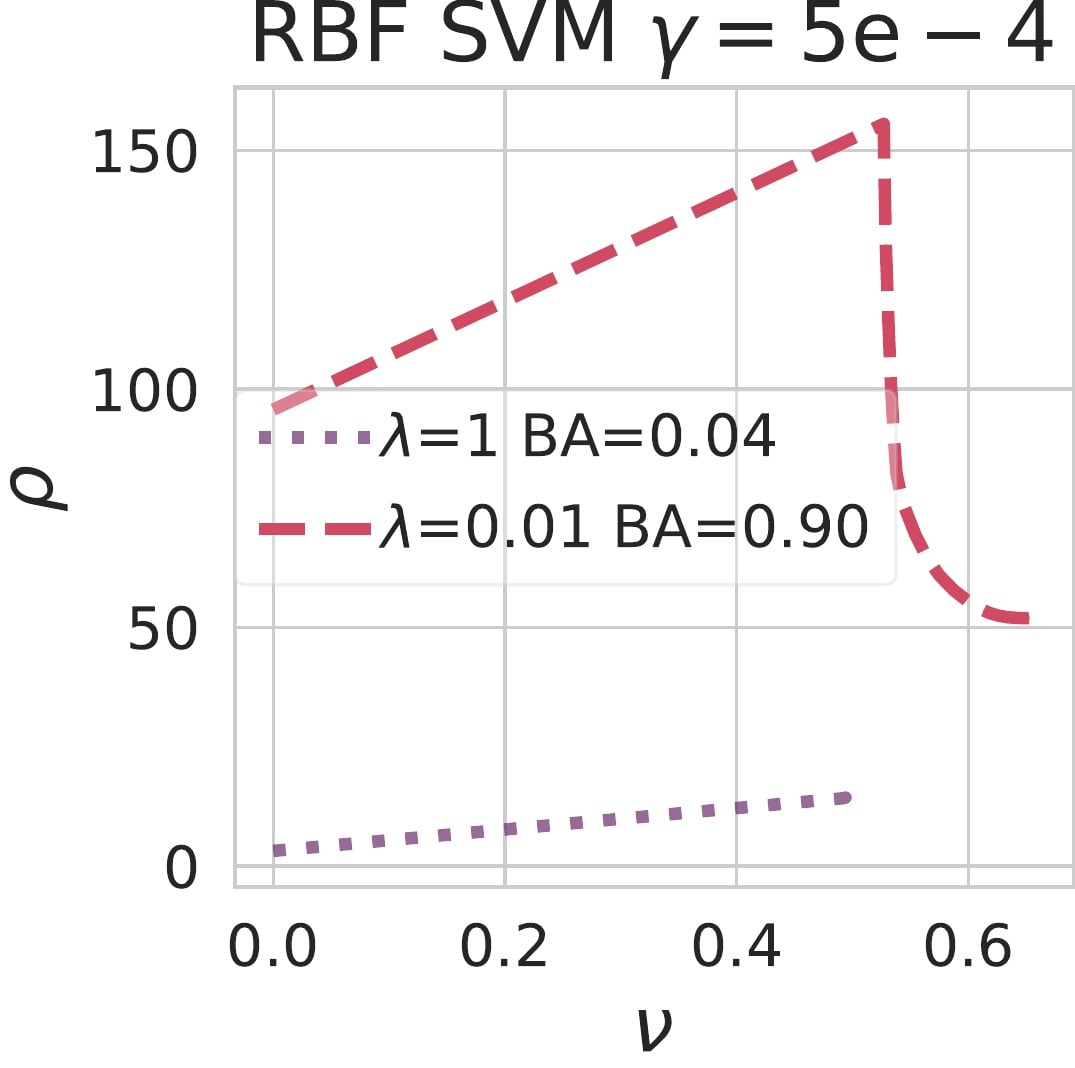}
  \hfill \break
  \includegraphics[width=0.235\textwidth]{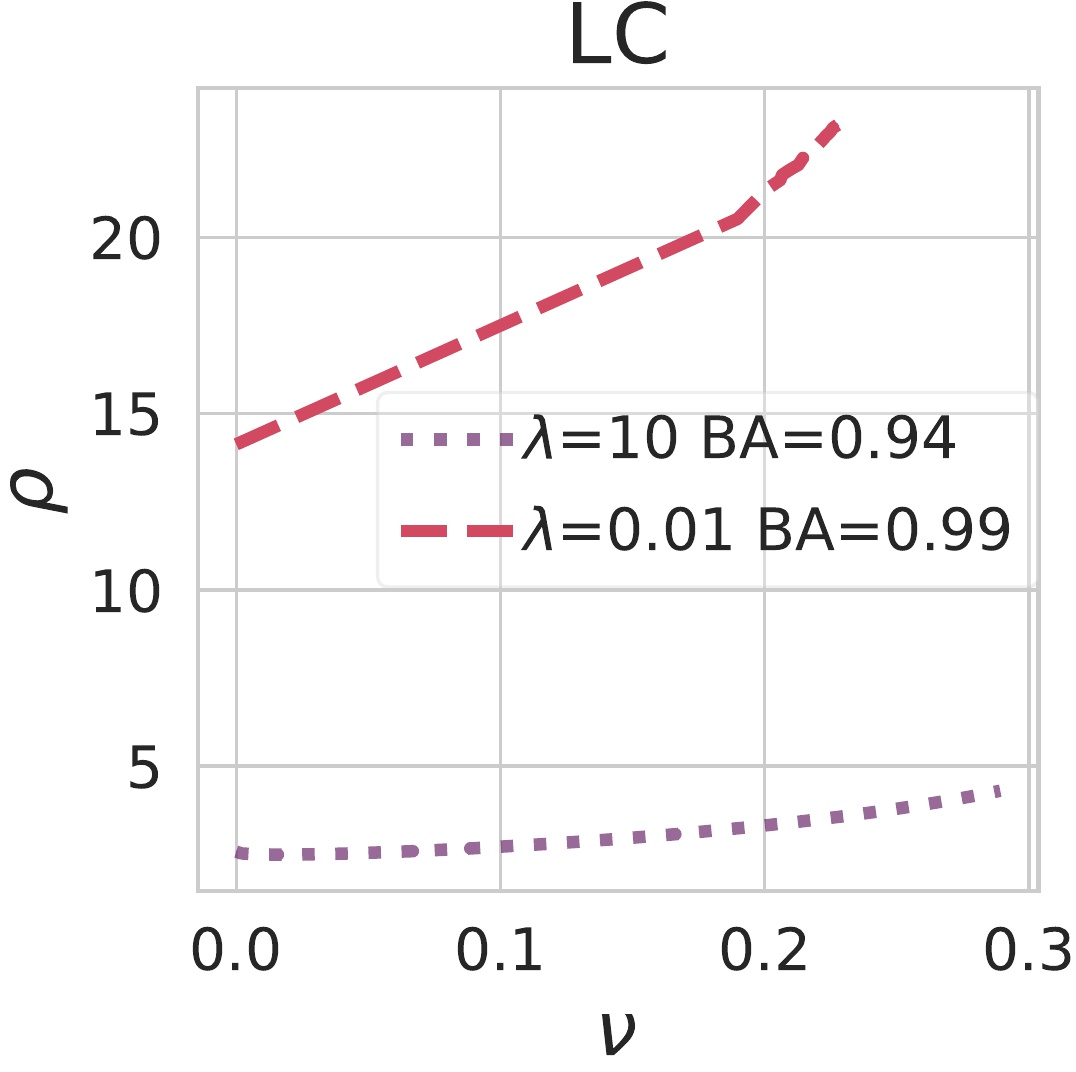}
  \includegraphics[width=0.235\textwidth]{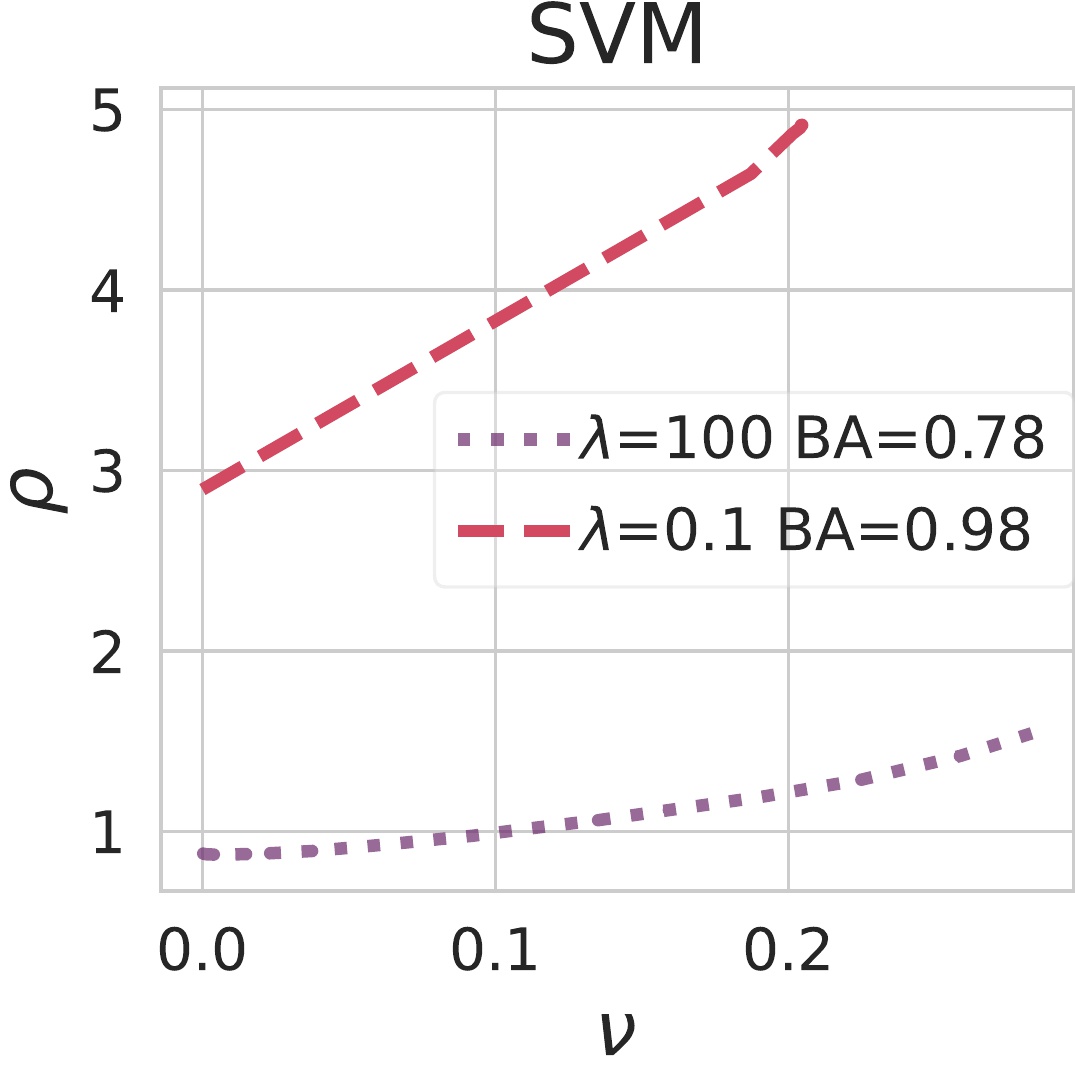}
  \includegraphics[width=0.235\textwidth]{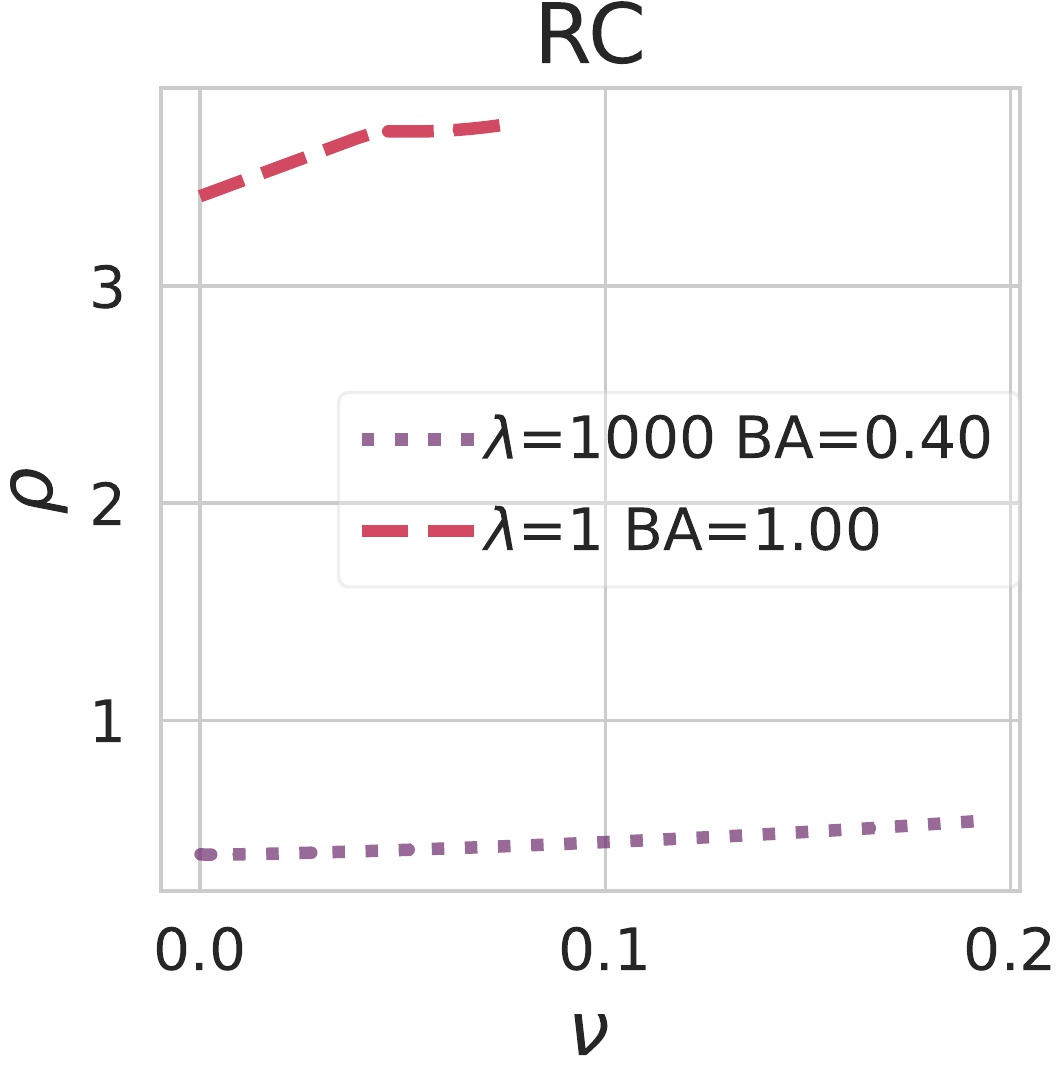}
  \includegraphics[width=0.23\textwidth]{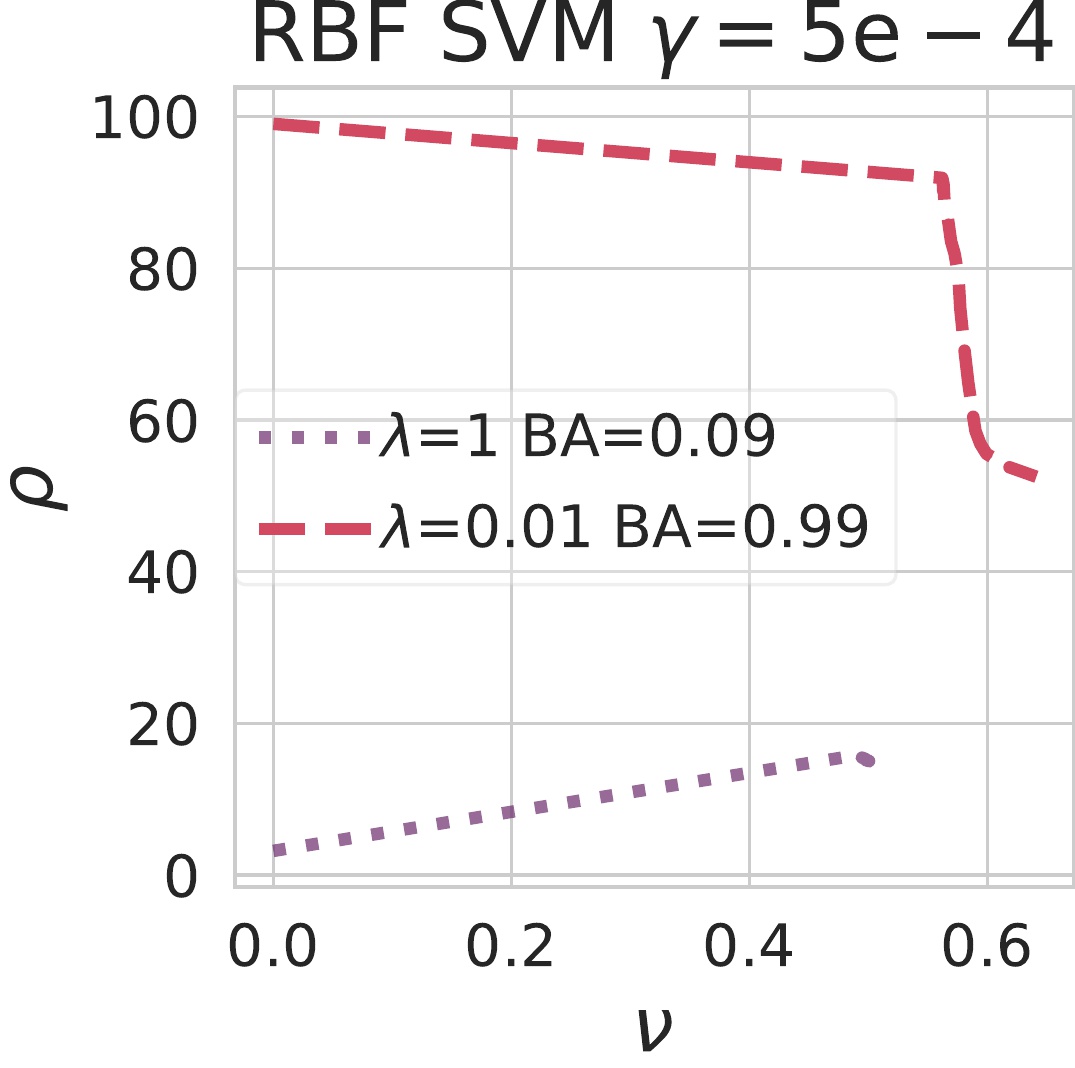}
  \caption{Backdoor weight deviation for the logistic classifier (LC), support vector machine (SVM), the ridge classifier (RC) and SVM with RBF kernel on MNIST $7~\rm{vs.}~1$ poisoned with backdoor trigger \cite{gu2019badnets}. We report the results for trigger size $3 \times 3$ (top row) and $6 \times 6$ (bottom row). We specify the regularization parameter $\lambda$ and backdoor accuracy (BA) for each setting in the legend of each plot.}
  \label{fig:backdoorParametersDeviationMNIST}
\end{figure*}

\begin{figure*}[h!]
  \centering
  \includegraphics[width=0.24\textwidth]{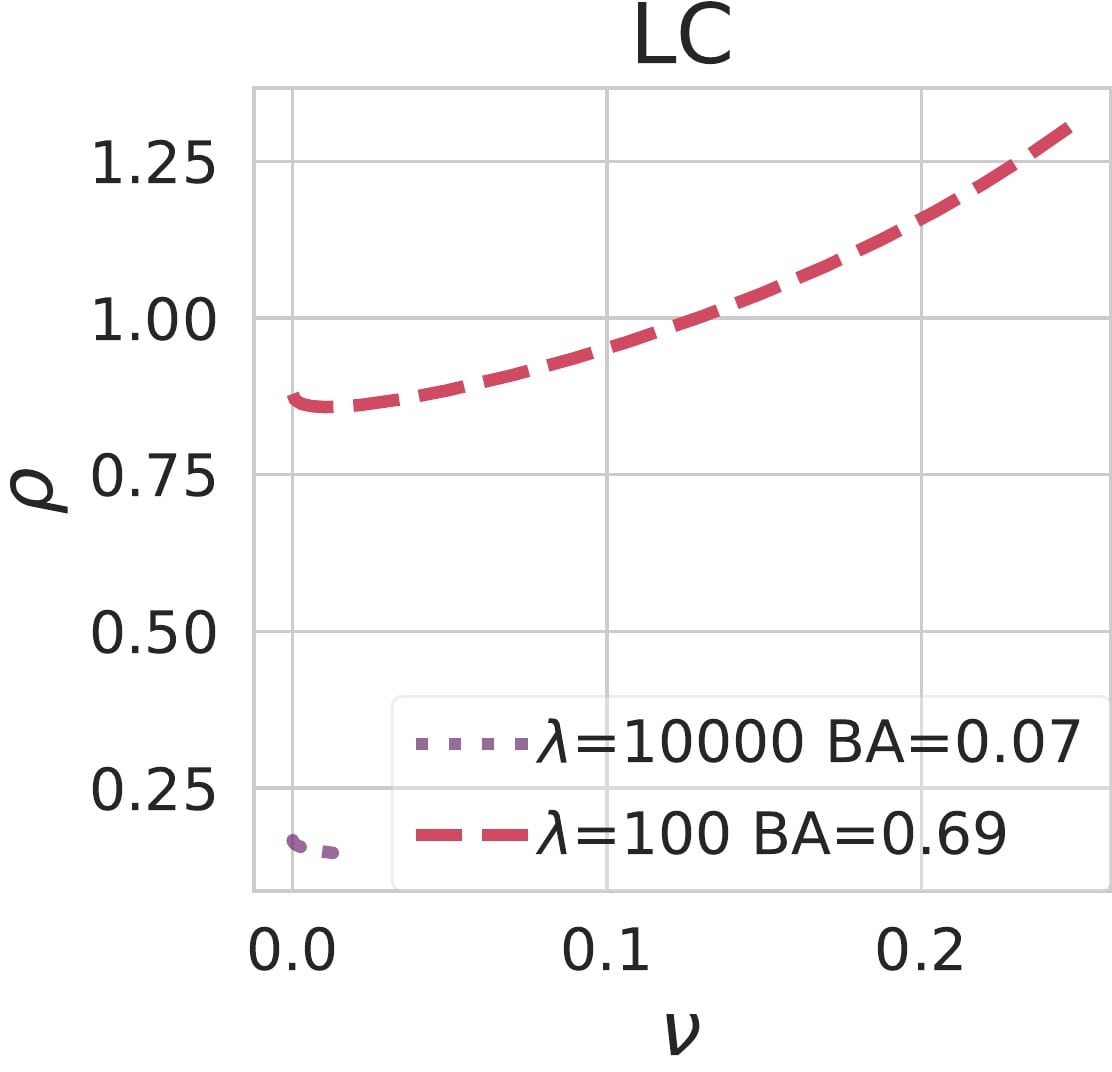}
  \includegraphics[width=0.235\textwidth]{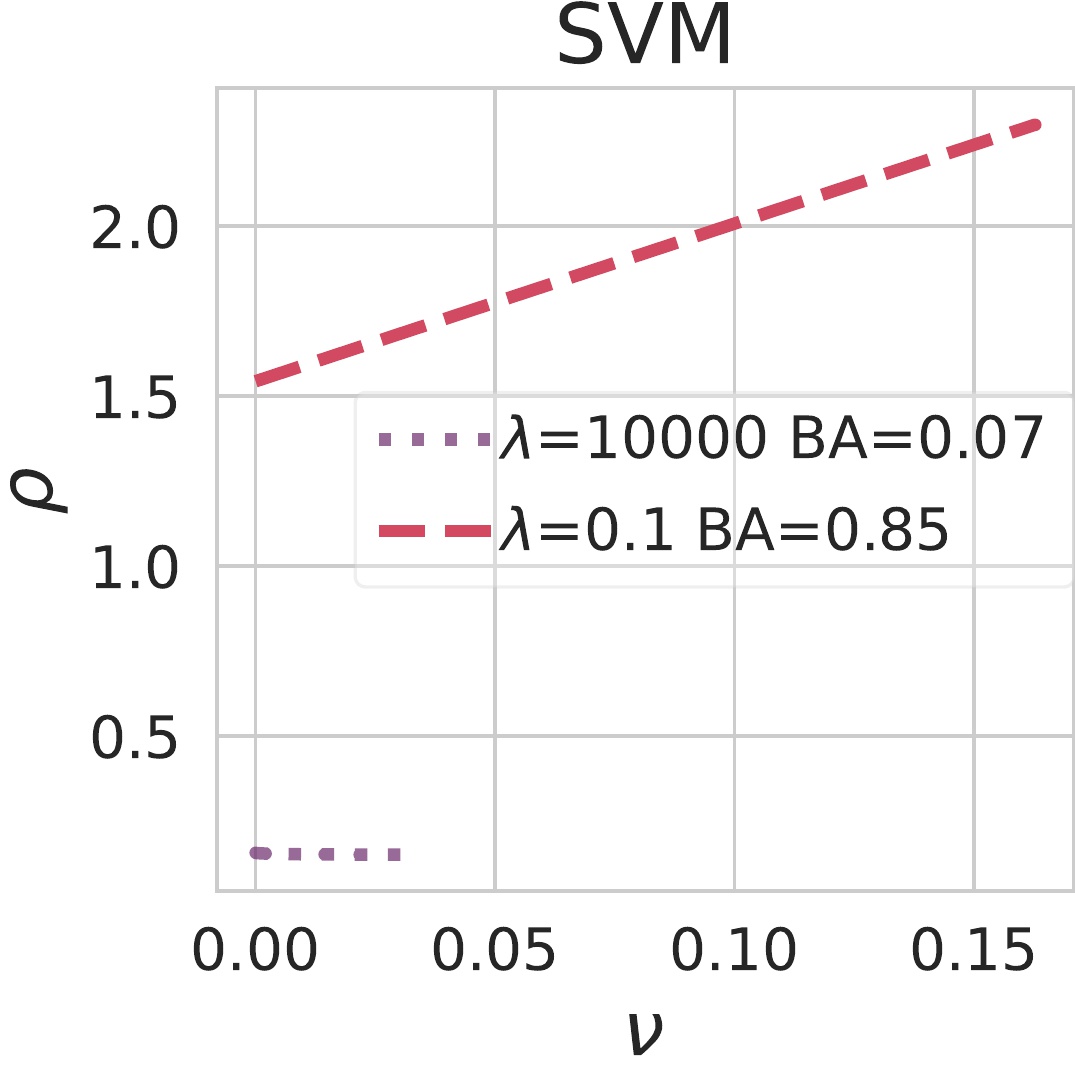}
  \includegraphics[width=0.235\textwidth]{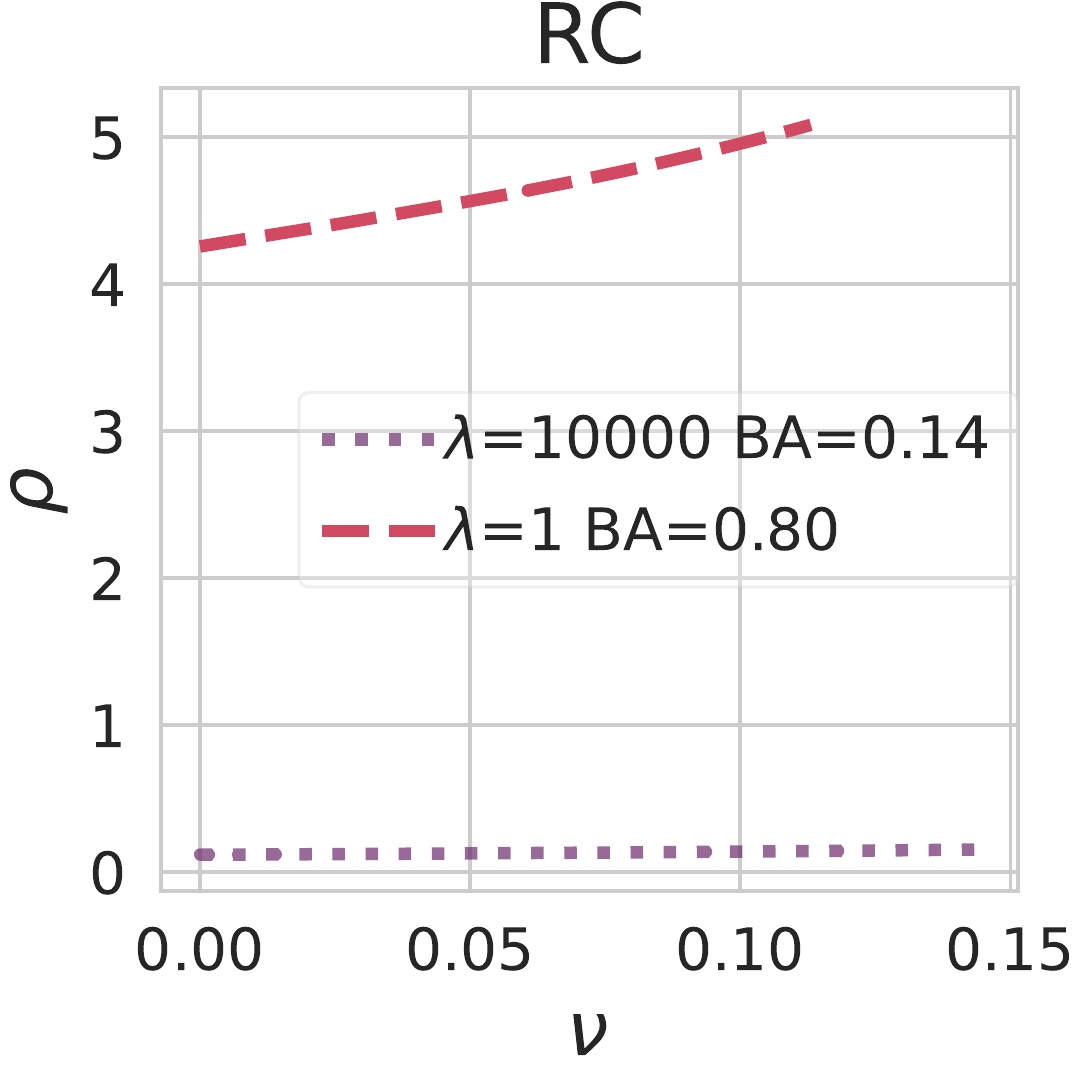}
  \includegraphics[width=0.23\textwidth]{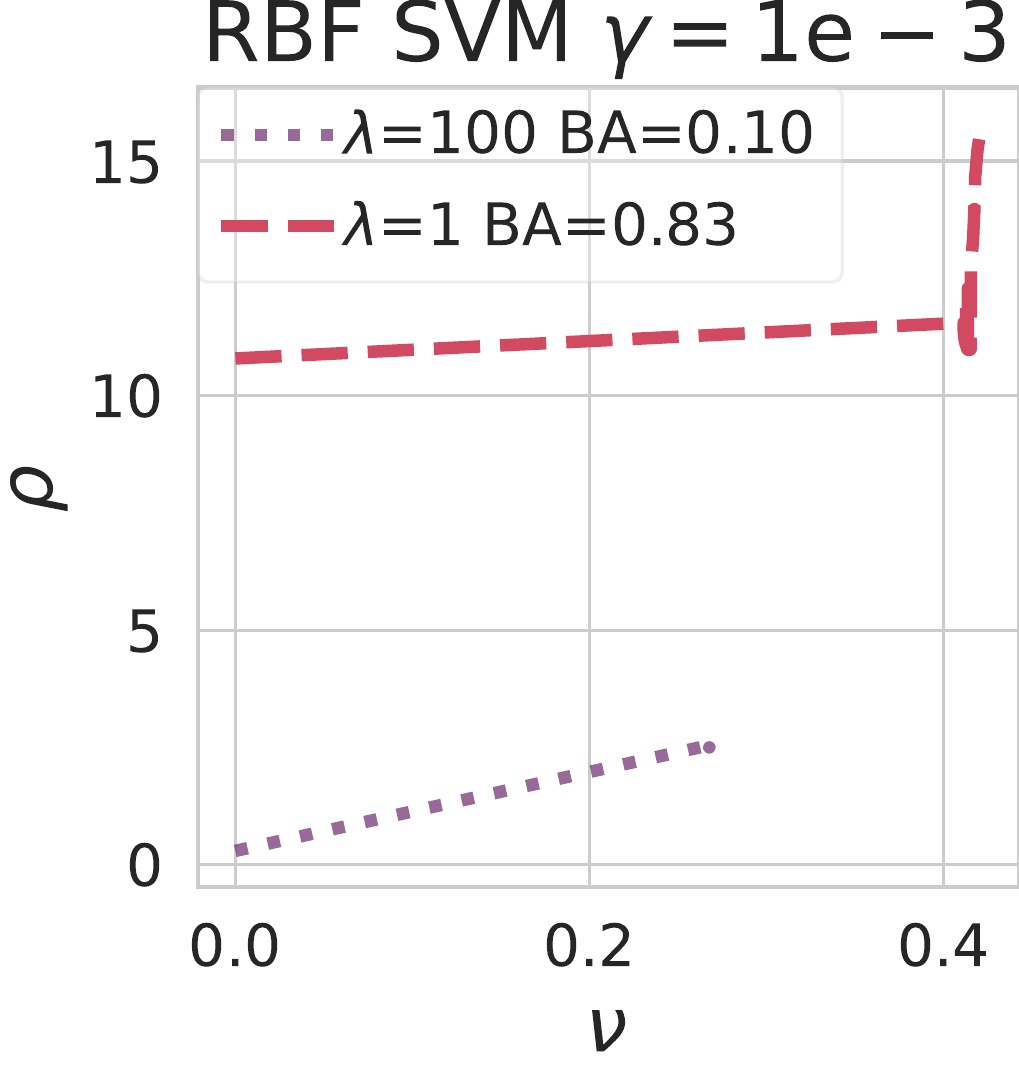}
  \hfill \break
  \includegraphics[width=0.24\textwidth]{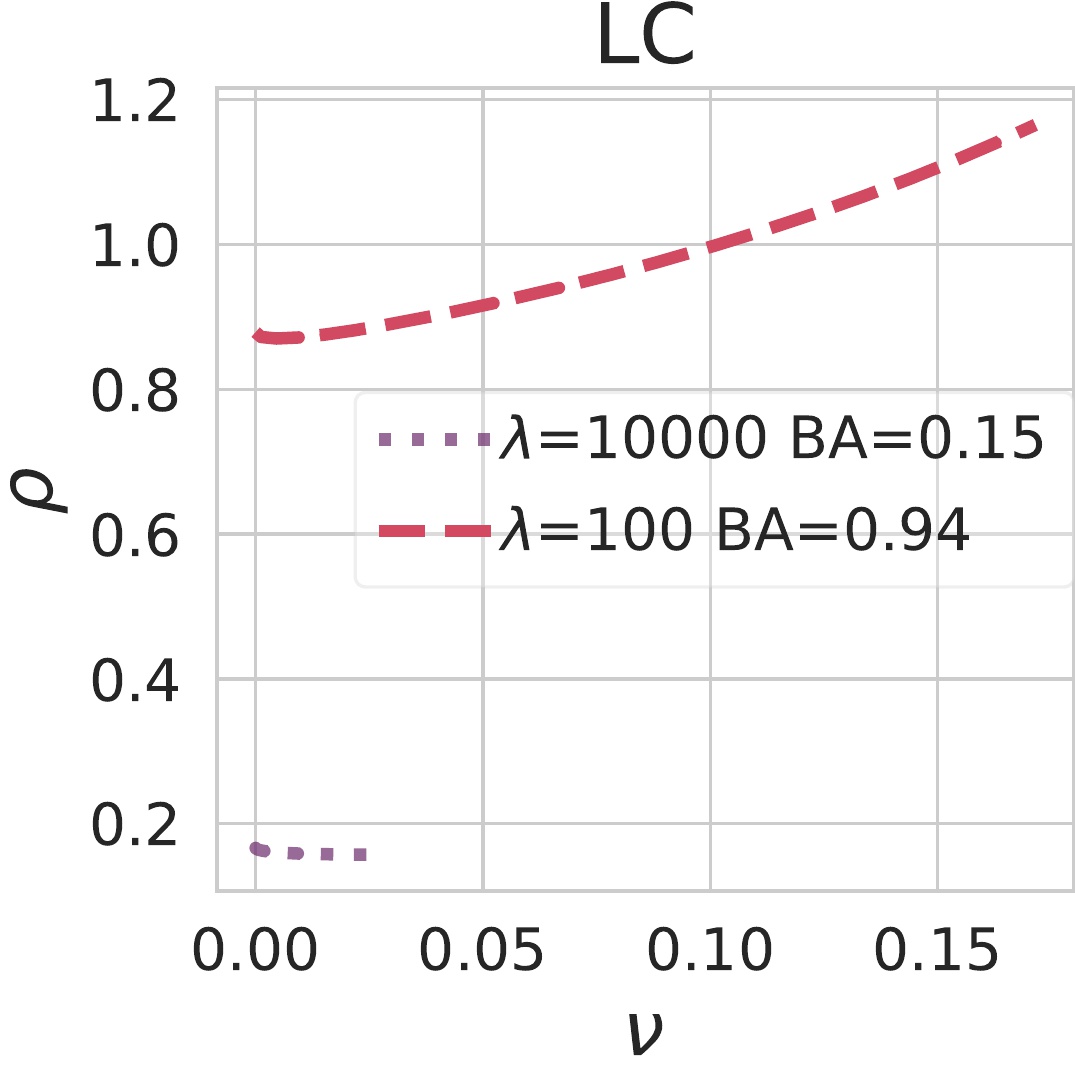}
  \includegraphics[width=0.235\textwidth]{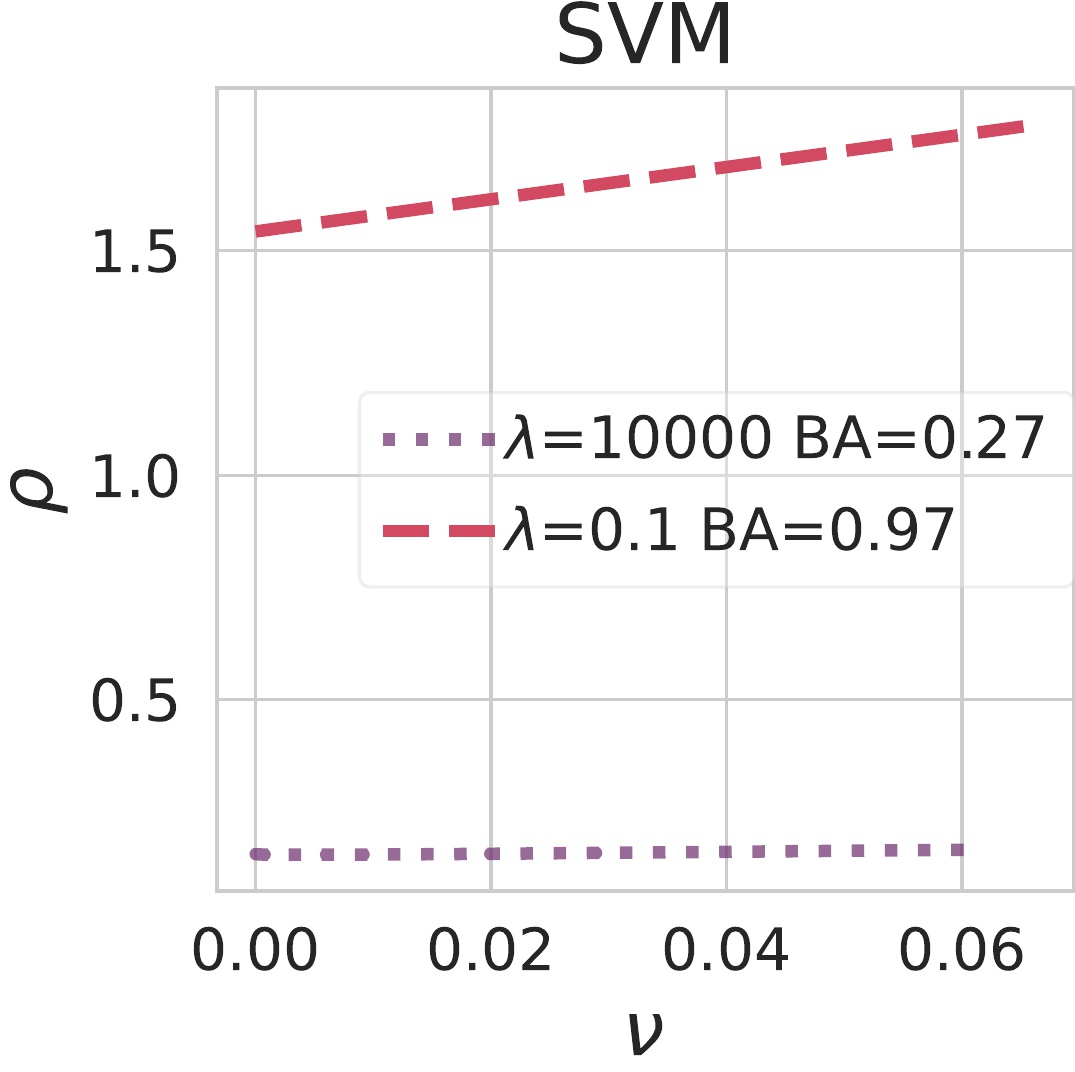}
  \includegraphics[width=0.235\textwidth]{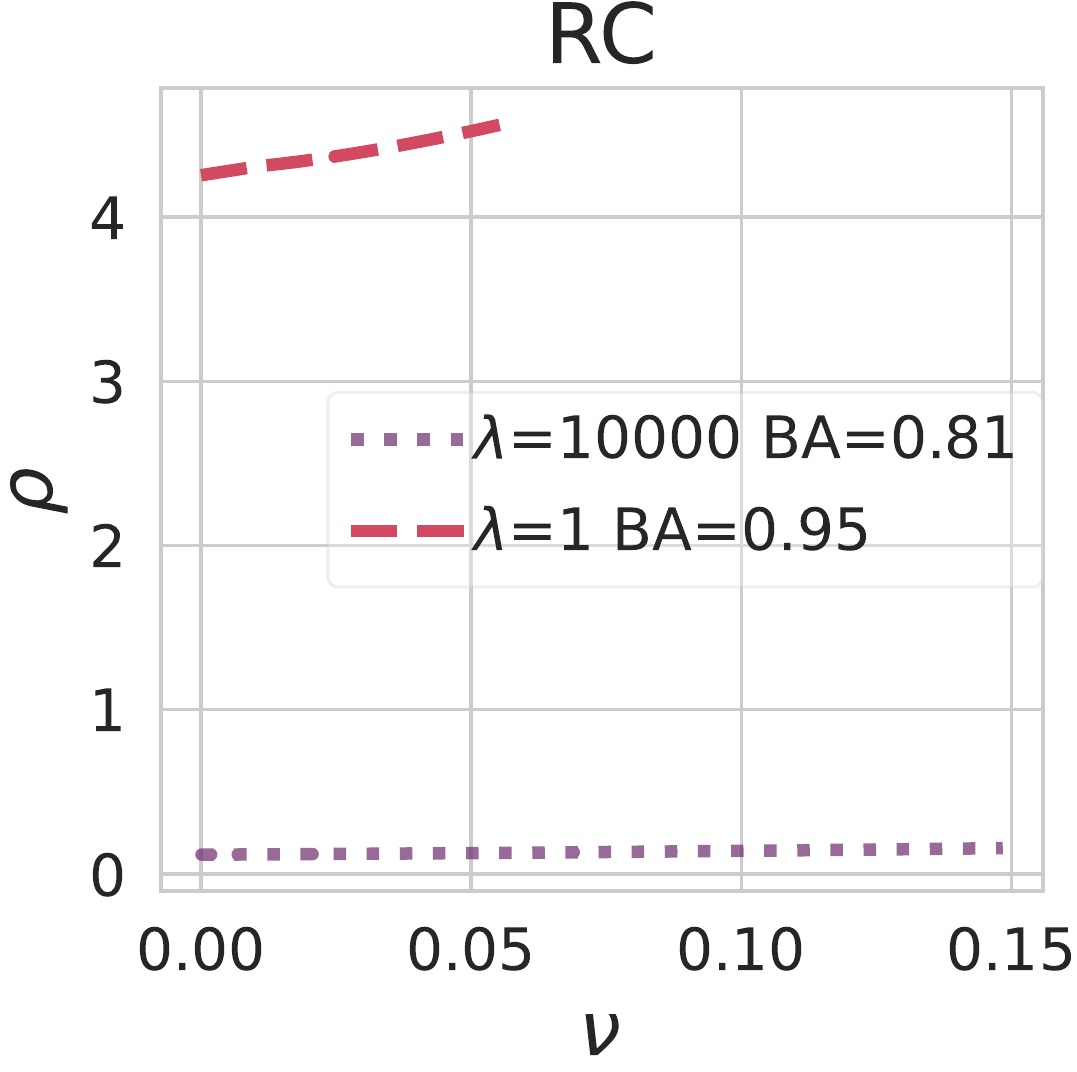}
  \includegraphics[width=0.235\textwidth]{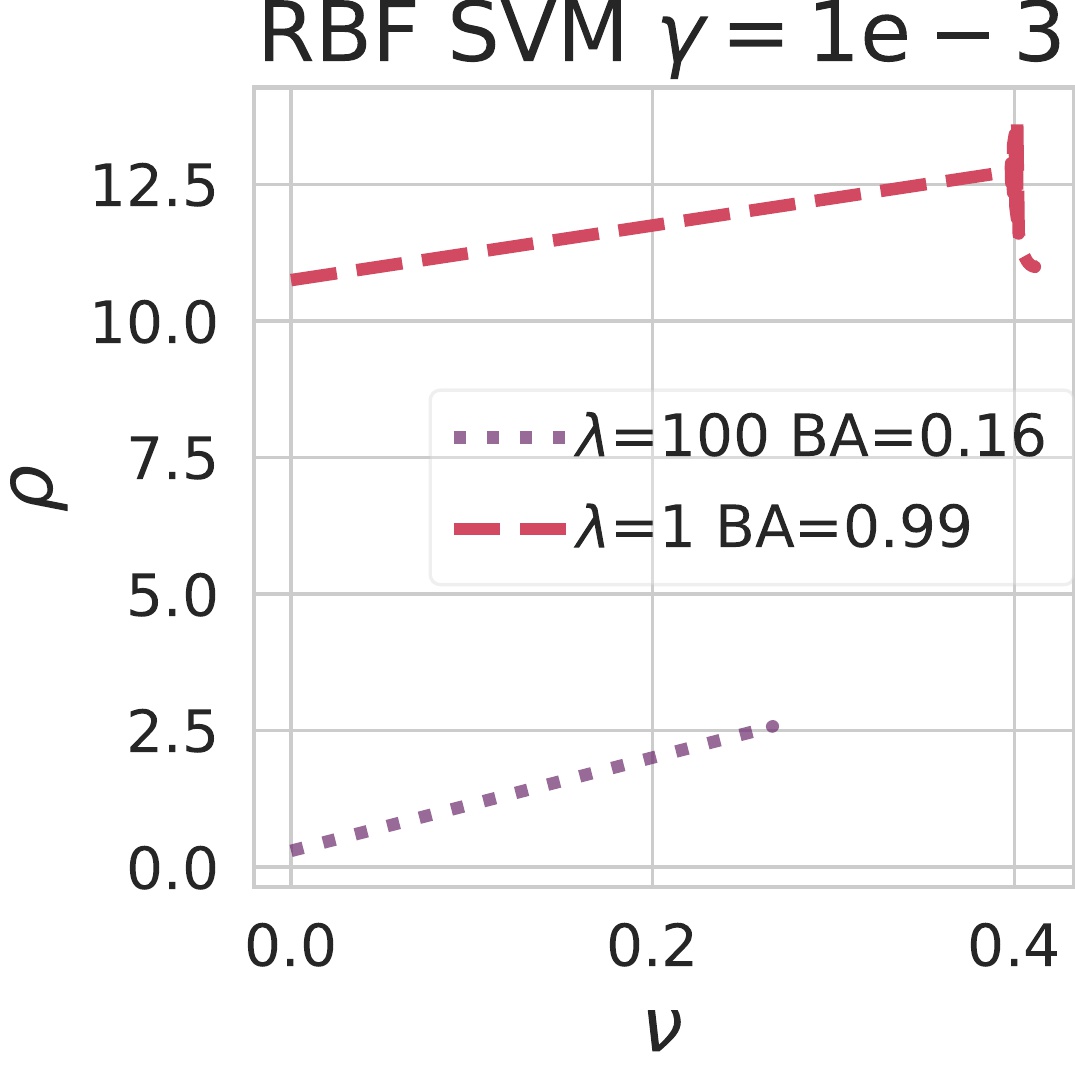}
  \hfill 
  \caption{Backdoor weight deviation for the logistic classifier (LC), support vector machine (SVM), the ridge classifier (RC) and SVM with RBF kernel on CIFAR10 \cifarairplanefrog poisoned with backdoor trigger \cite{gu2019badnets}. We report the results for trigger size $8 \times 8$ (top row) and $16\times16$ (bottom row). We specify the regularization parameter $\lambda$ and backdoor accuracy (BA) for each setting in the legend of each plot.}
  \label{fig:backdoorParametersDeviationCIFAR}
\end{figure*}

\begin{figure*}[h!]
  \centering
\includegraphics[width=0.23\textwidth]{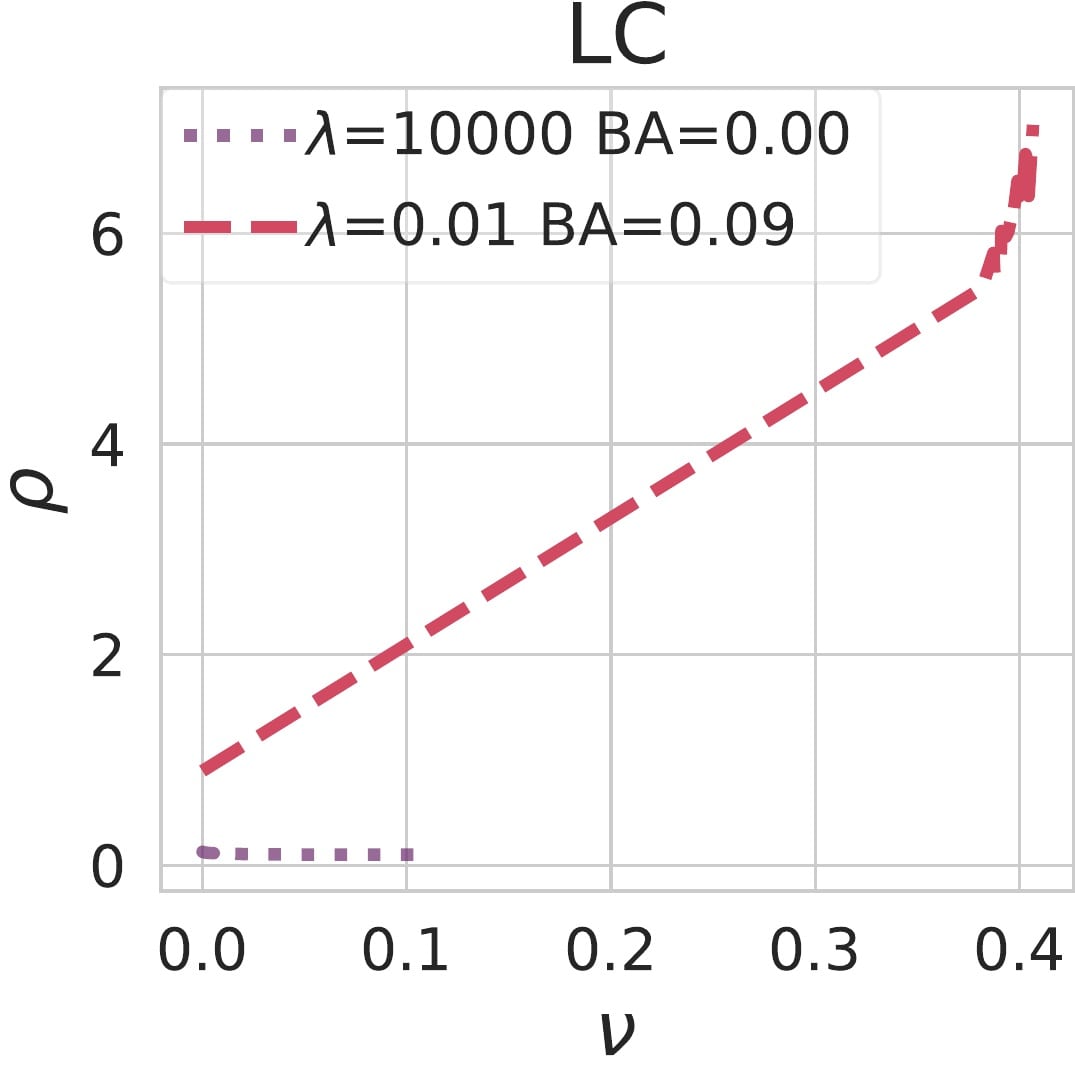}
  \includegraphics[width=0.238\textwidth]{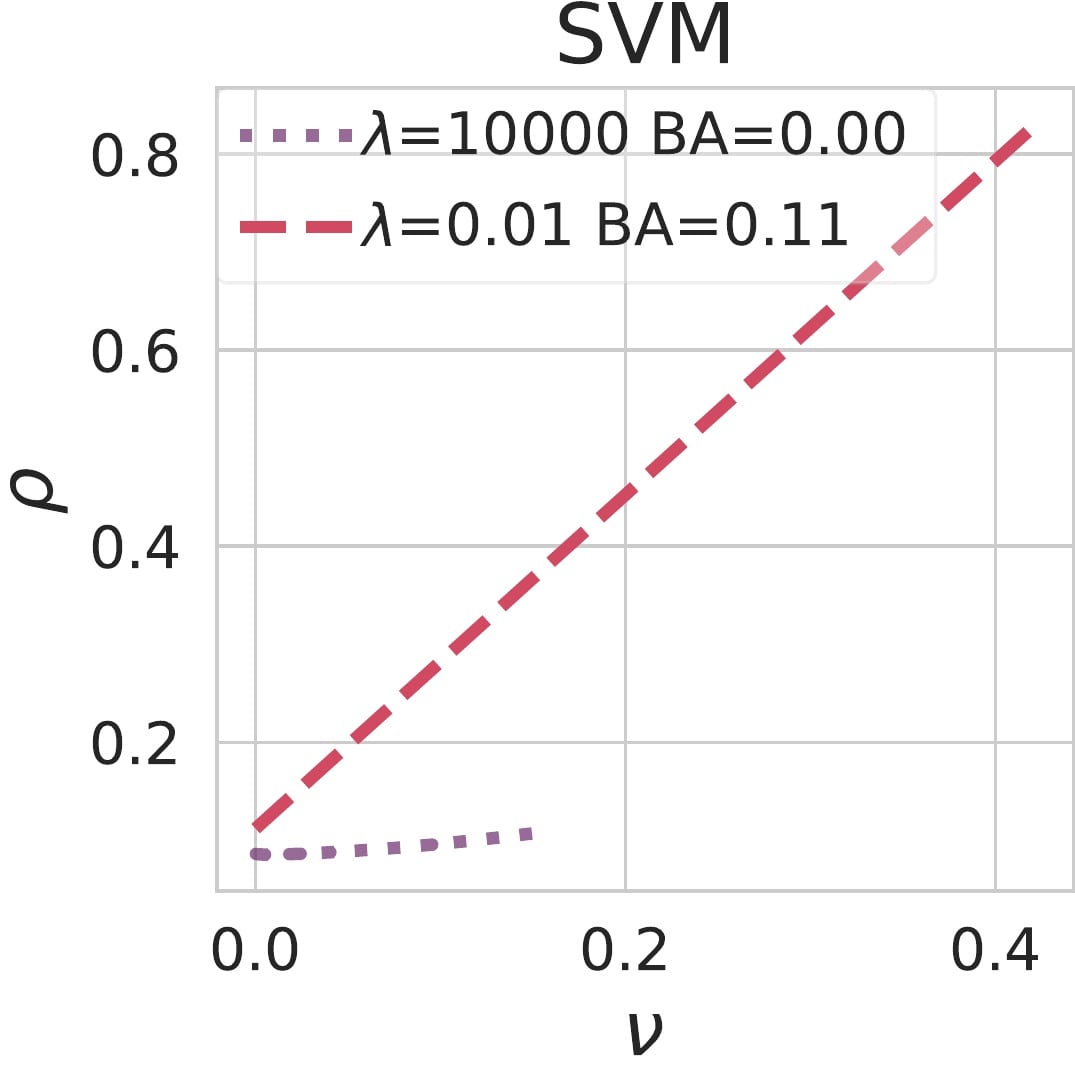}
  \includegraphics[width=0.238\textwidth]{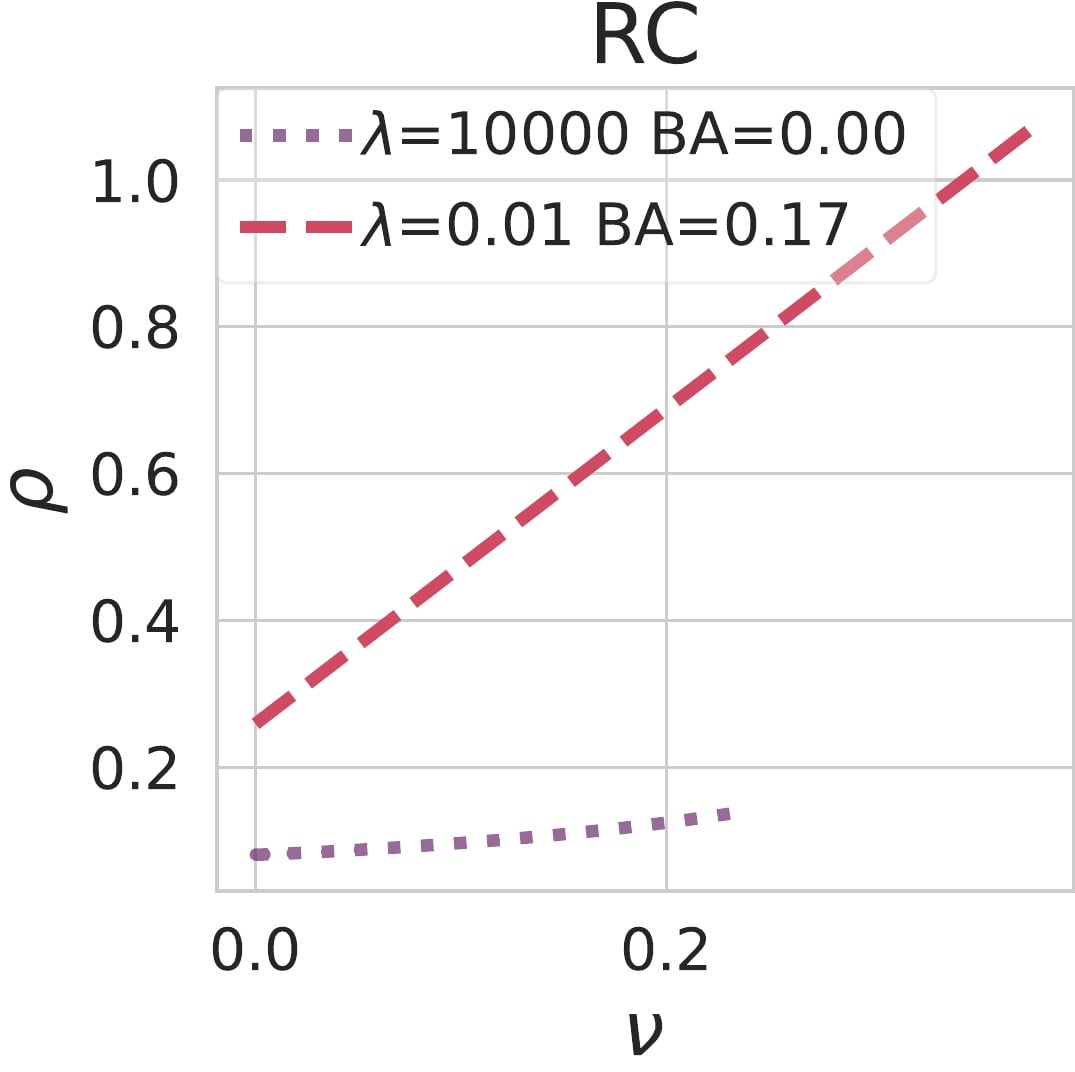}
  \includegraphics[width=0.238\textwidth]{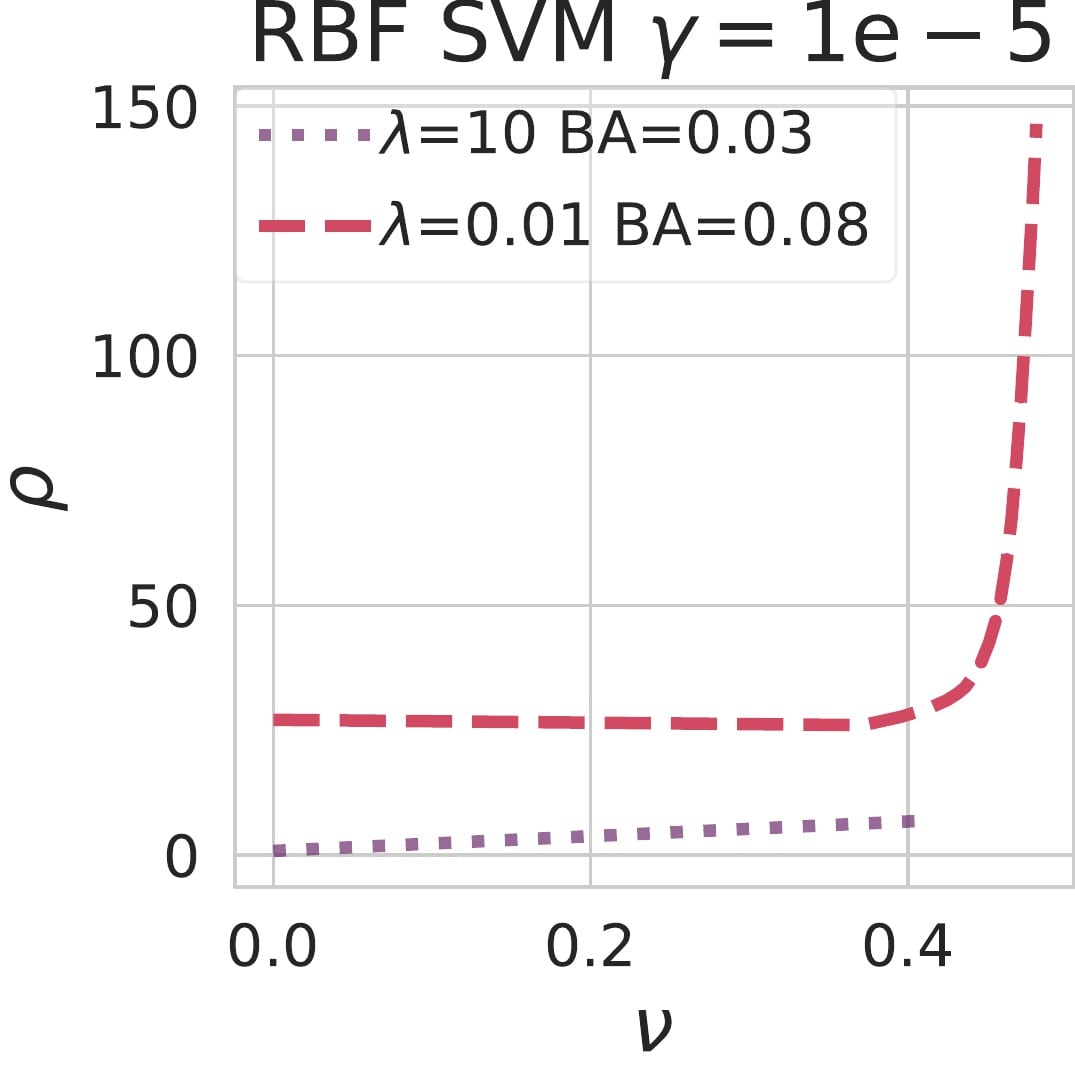}
    \hfill \break
\includegraphics[width=0.228\textwidth]{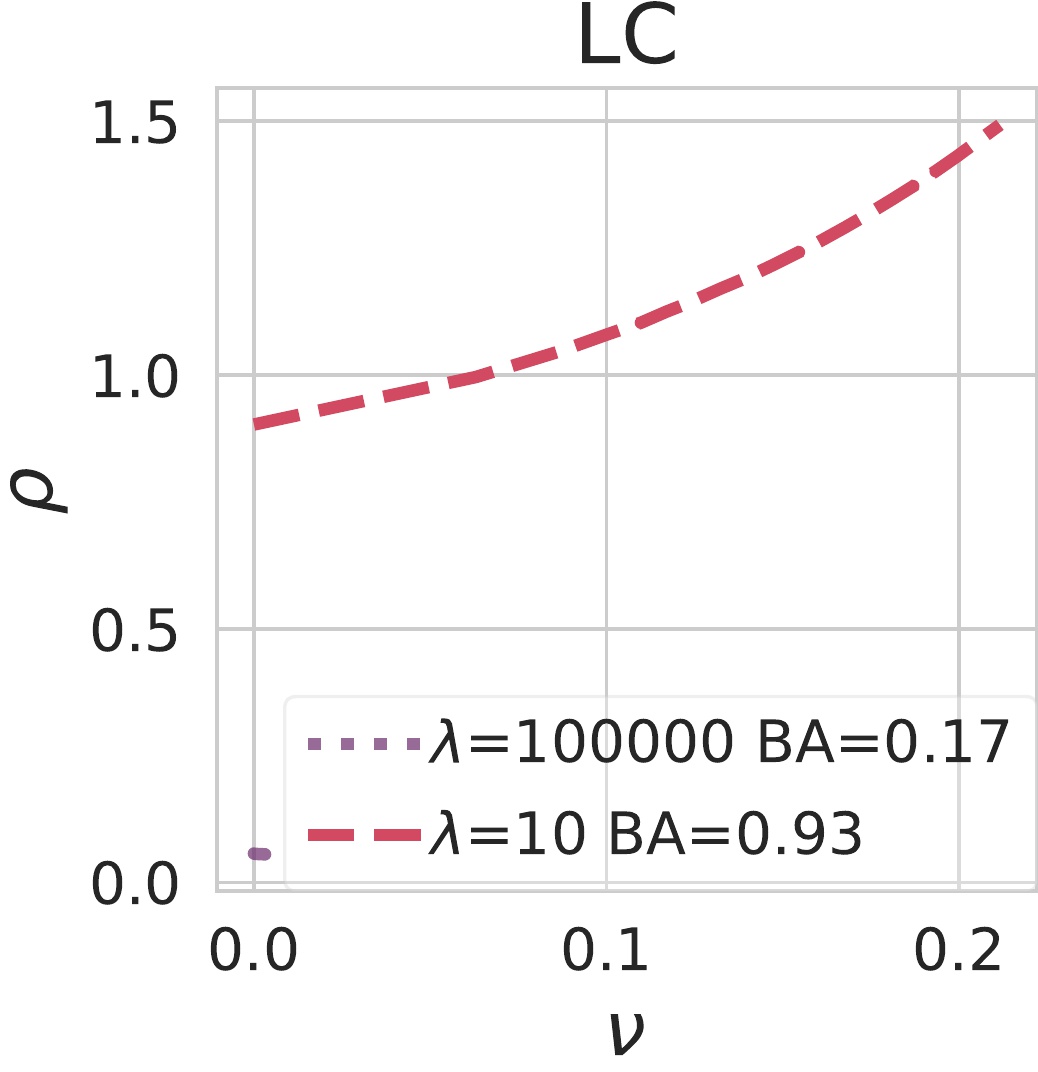}
  \includegraphics[width=0.238\textwidth]{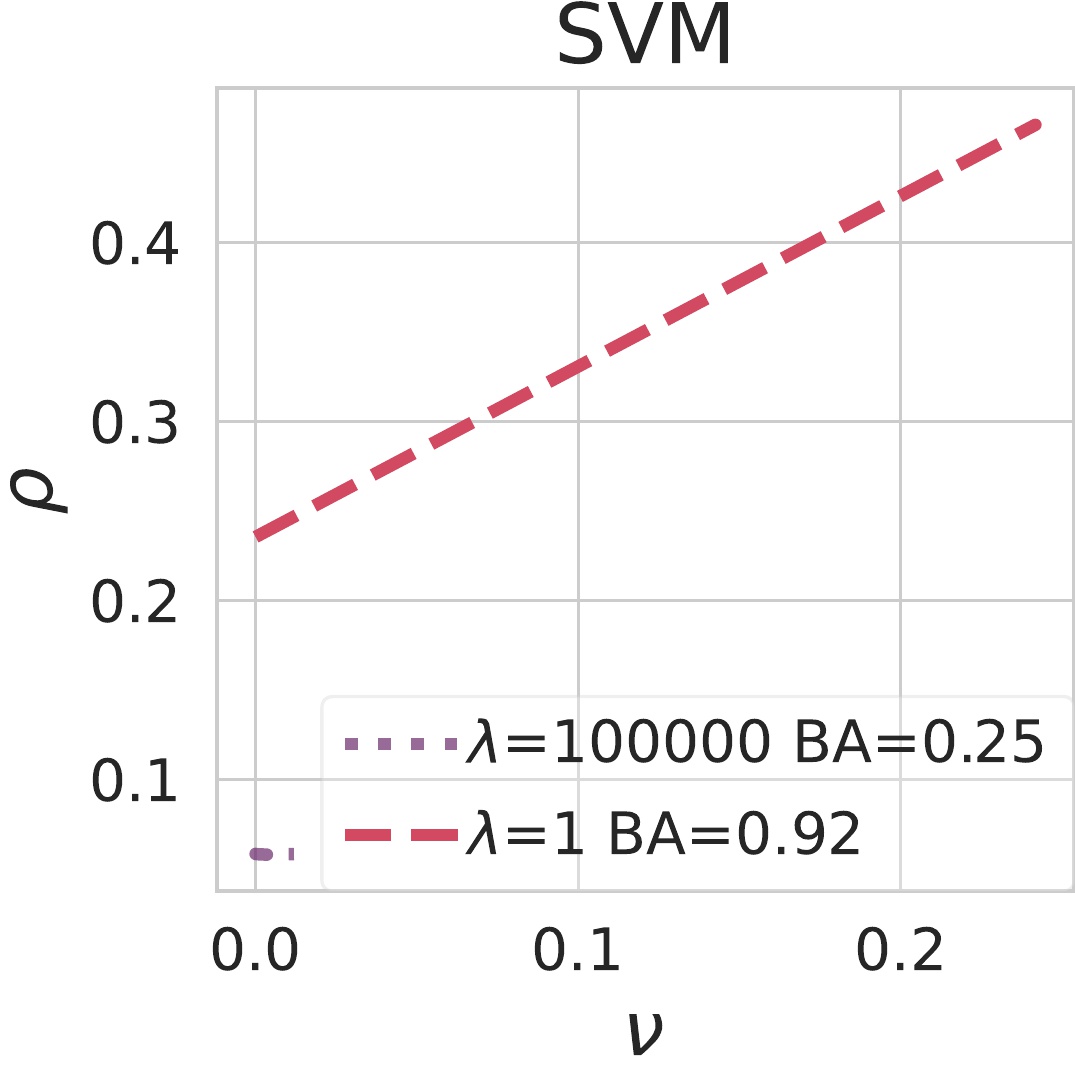}
  \includegraphics[width=0.238\textwidth]{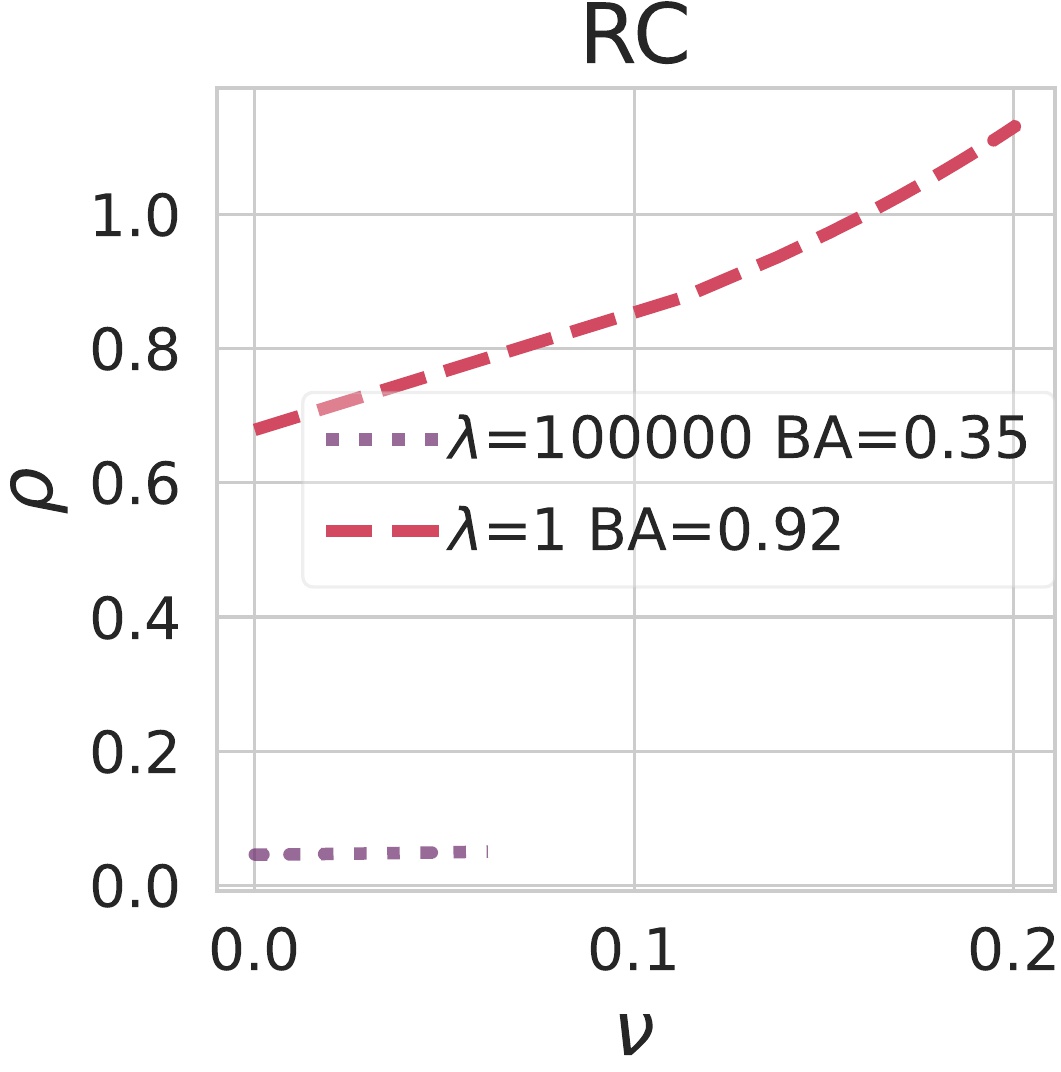}
  \includegraphics[width=0.238\textwidth]{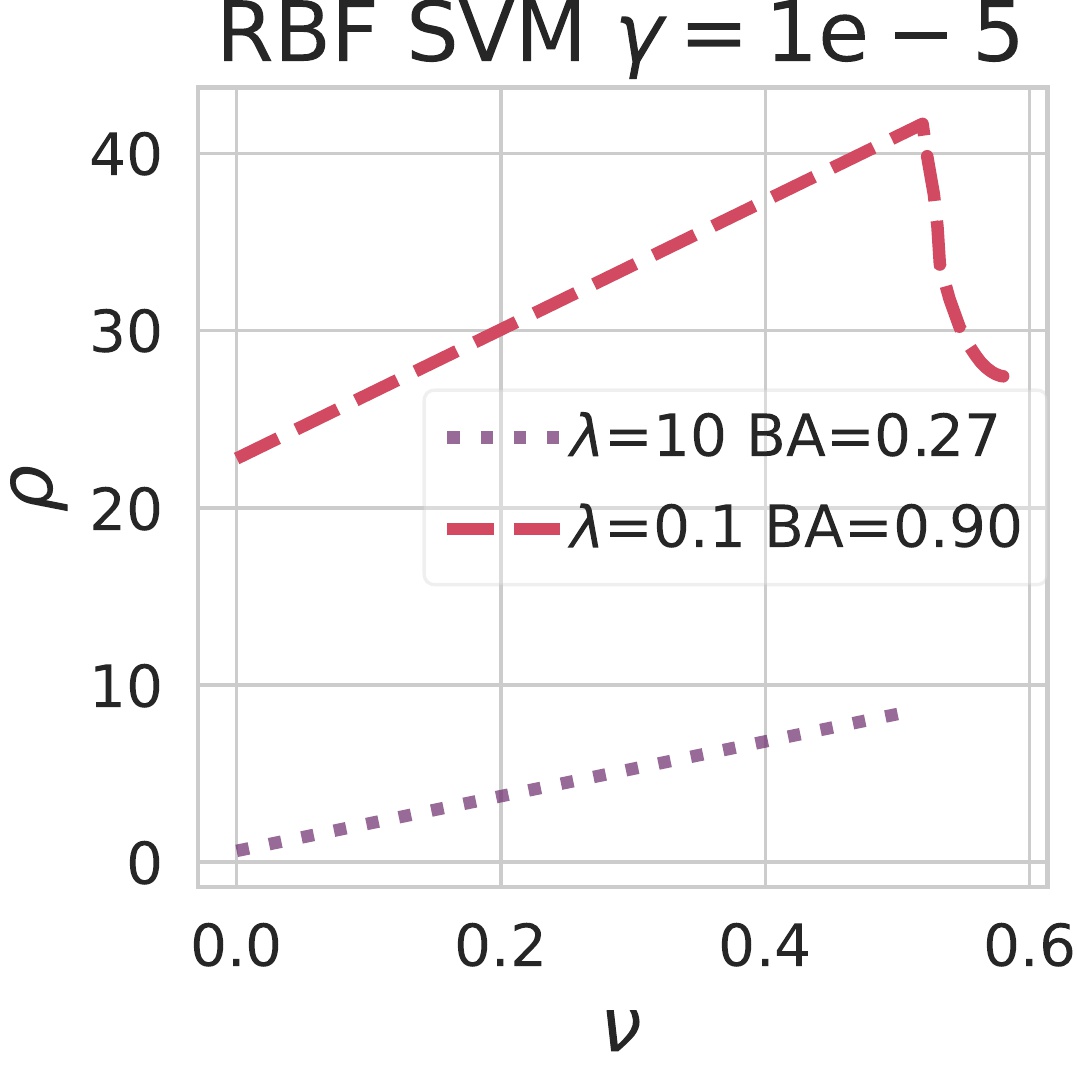}
    \hfill
  \caption{Backdoor weight deviation for the logistic classifier (LC), support vector machine (SVM), the ridge classifier (RC) and SVM with RBF kernel on Imagenette \imagenettetenchtruck poisoned with backdoor trigger \cite{Zhong20backdoor}. We report the results for visibility $c_m=10$ (top row) and $c_m=75$ (bottom row). We specify the regularization parameter $\lambda$ and backdoor accuracy (BA) for each setting in the legend of each plot.}
  \label{fig:backdoorParametersDeviationImagenette}
\end{figure*}

We use our two measures proposed in Section~\ref{sec:backdoor-curves}, $\rho$ and $\nu$
to analyze the parameter change. The former, $\rho$, monitors the change of the weights, for example, whether they increase or decrease. The latter, $\nu$, measures the change in orientation or angle of the classifier.
We plot both measures with different regularization parameters, trigger size, or visibility with a fraction of poisoning points to $p=0.1$ in Figure~\ref{fig:backdoorParametersDeviationMNIST}-\ref{fig:backdoorParametersDeviationImagenette}. Within each plot, we also report the backdoor accuracy (BA) representing the model's performance on backdoor samples at the end of training. 

On linear classifiers, $\rho(\vct w)$ increases during the backdoor learning process. This equals an increase in the weights' values, suggesting that the classifiers become more complex while learning the backdoor.
However, when investigating the RBF SVM, the results are slightly different. Indeed, when increasing $\gamma$ and decreasing $\lambda$, 
the classifier becomes flexible and complex enough to learn the backdoor without increasing its complexity. On the other hand, when decreasing $\gamma$, the model is constrained to behave similarly to a linear classifier. In this way, analogously to linear classifiers, the model needs to increase its complexity to learn the backdoor.
When increasing the trigger size or visibility the results are similar, thus confirming the previous analysis. However, as a result of increasing the attacker's strength\sv{,} the backdoor accuracy turns out to be higher.
\smallskip


\subsubsection{Explaining Backdoor Predictions}
\begin{figure*}[t]
  \centering
  \includegraphics[width=0.14\textwidth]{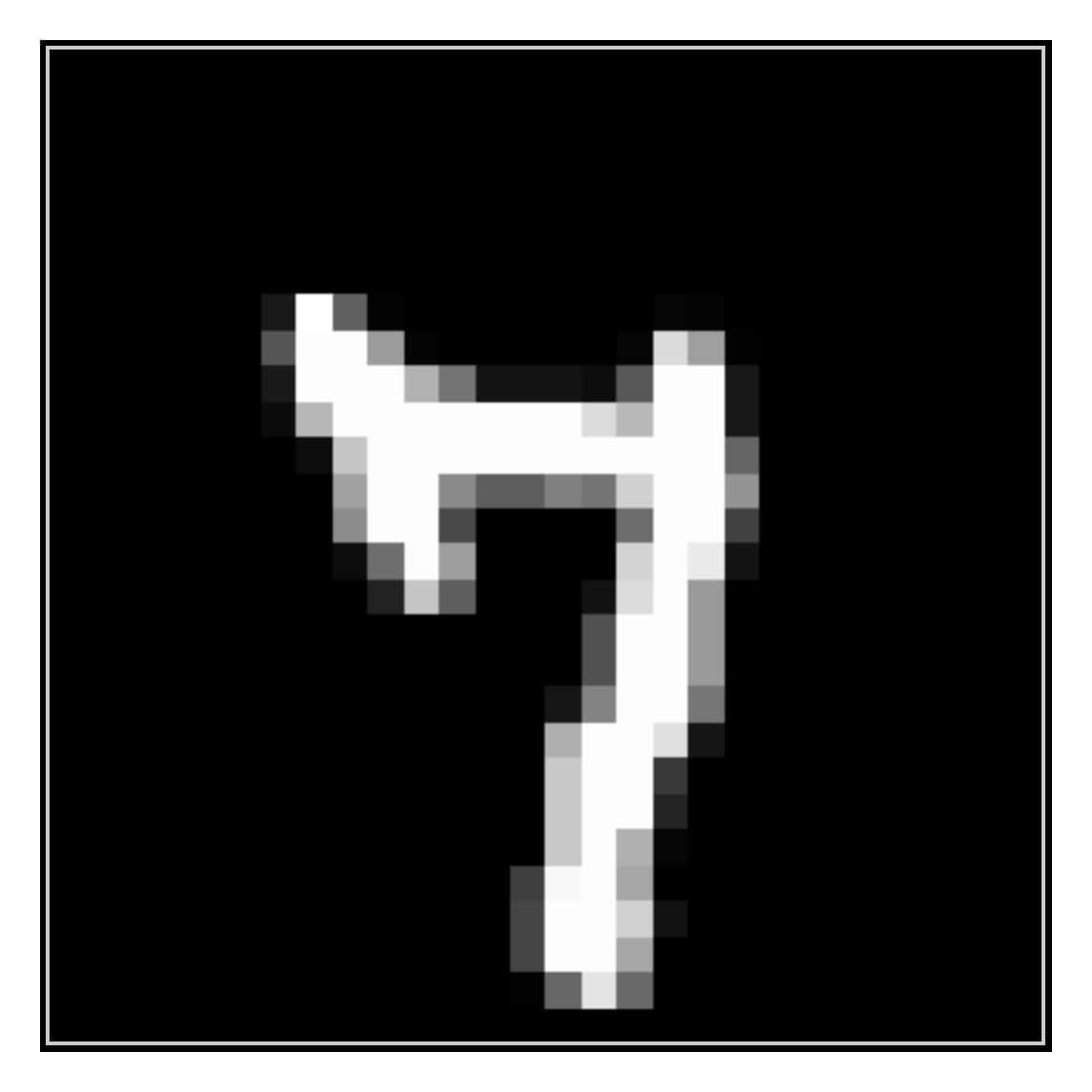}
  \includegraphics[width=0.17\textwidth]{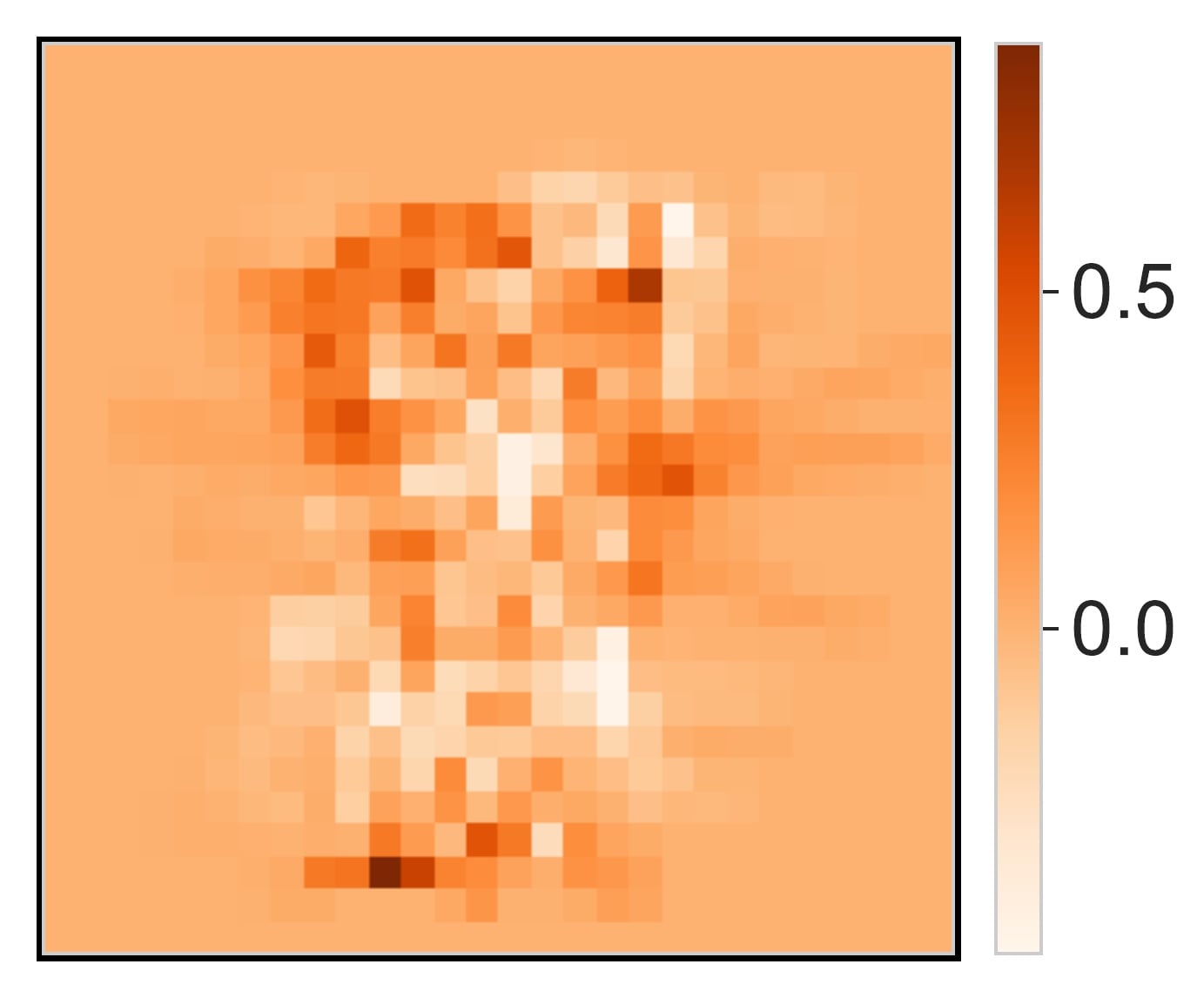}
  \hspace{-0.4em}
  \includegraphics[width=0.172\textwidth]{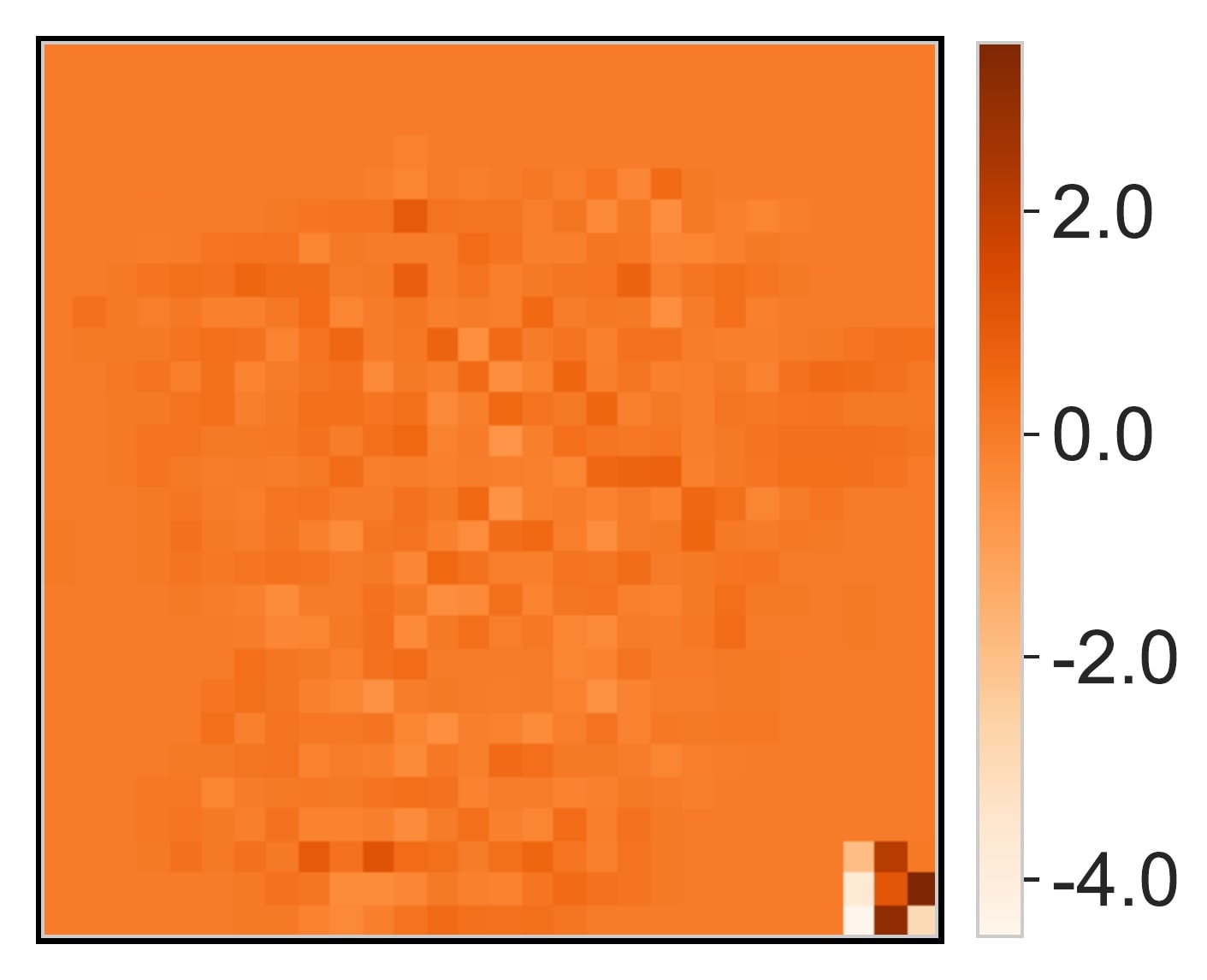}
  \includegraphics[width=0.14\textwidth]{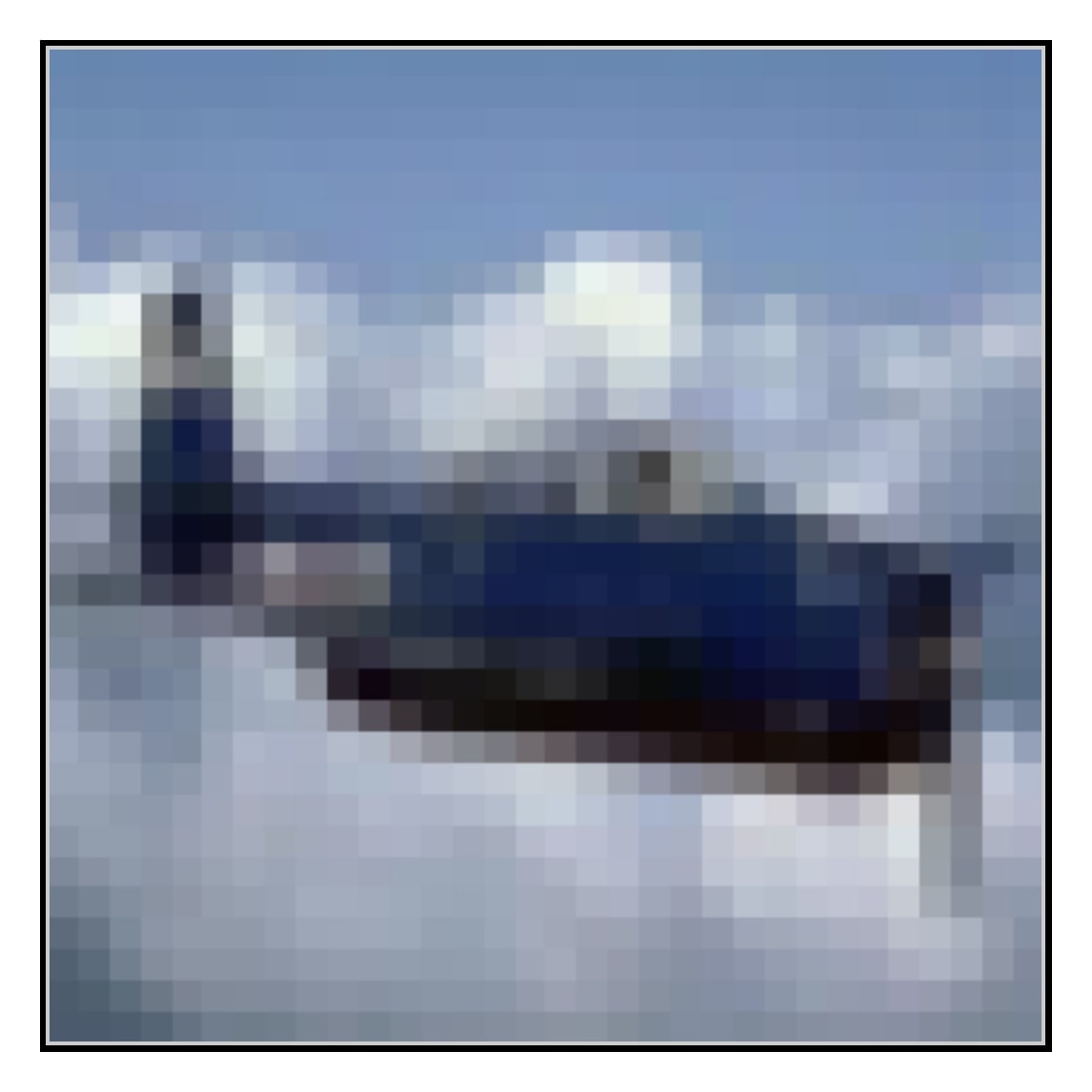}
  \includegraphics[width=0.17\textwidth]{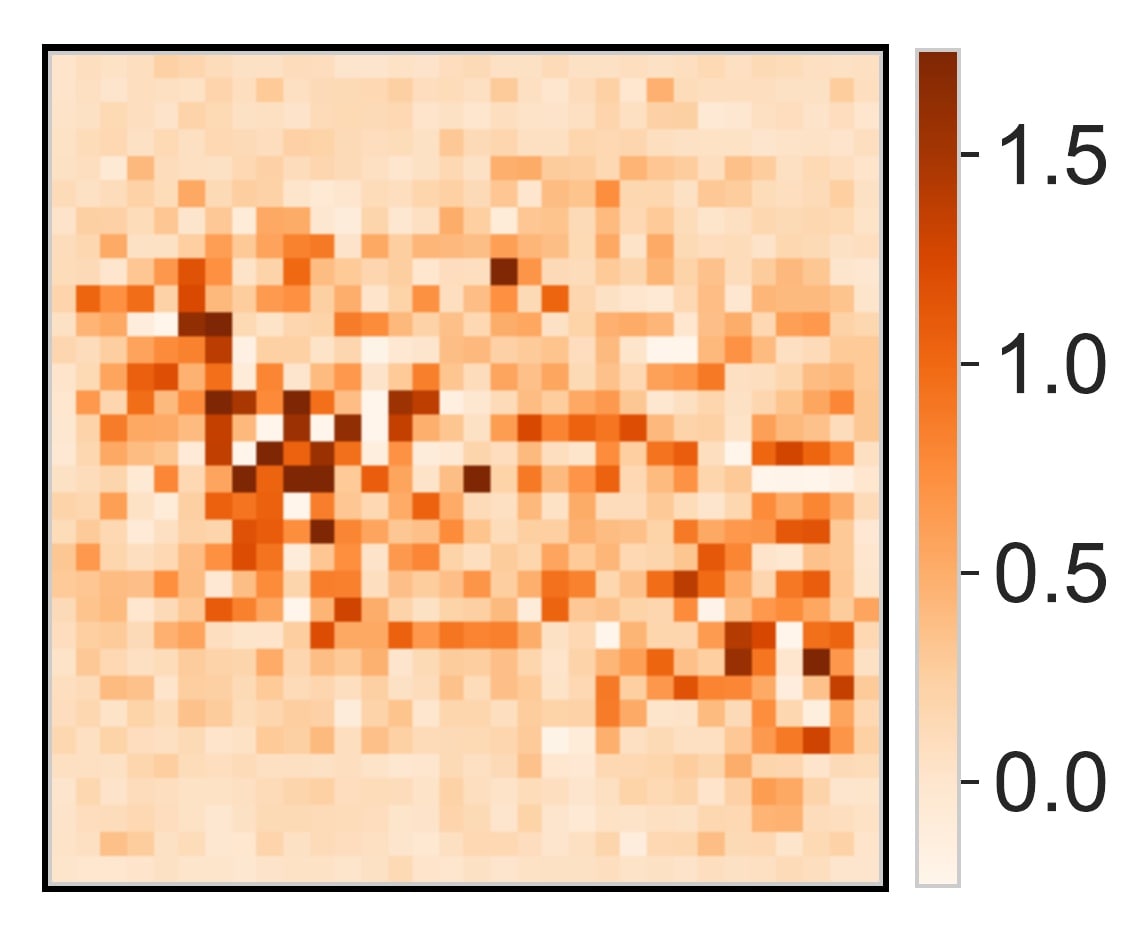}
  \includegraphics[width=0.172\textwidth]{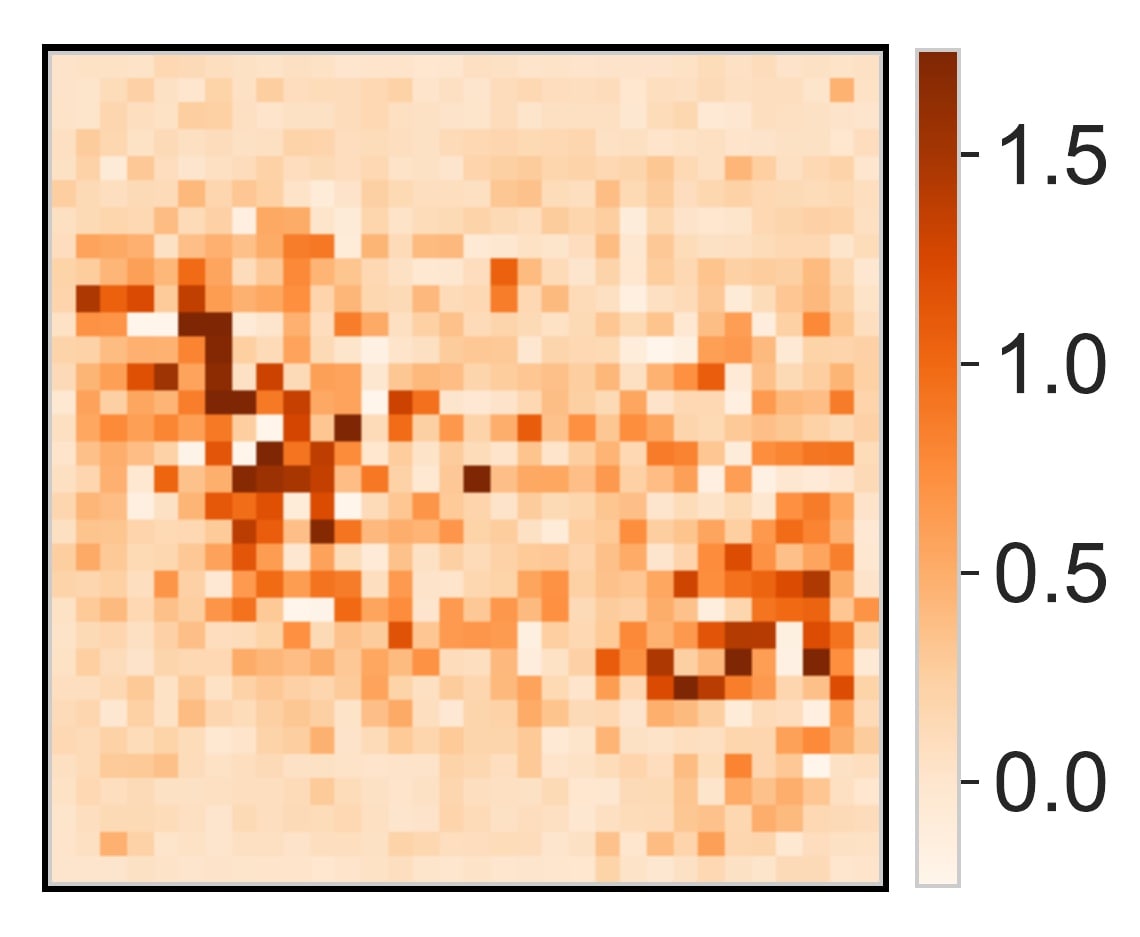}

  \includegraphics[width=0.14\textwidth]{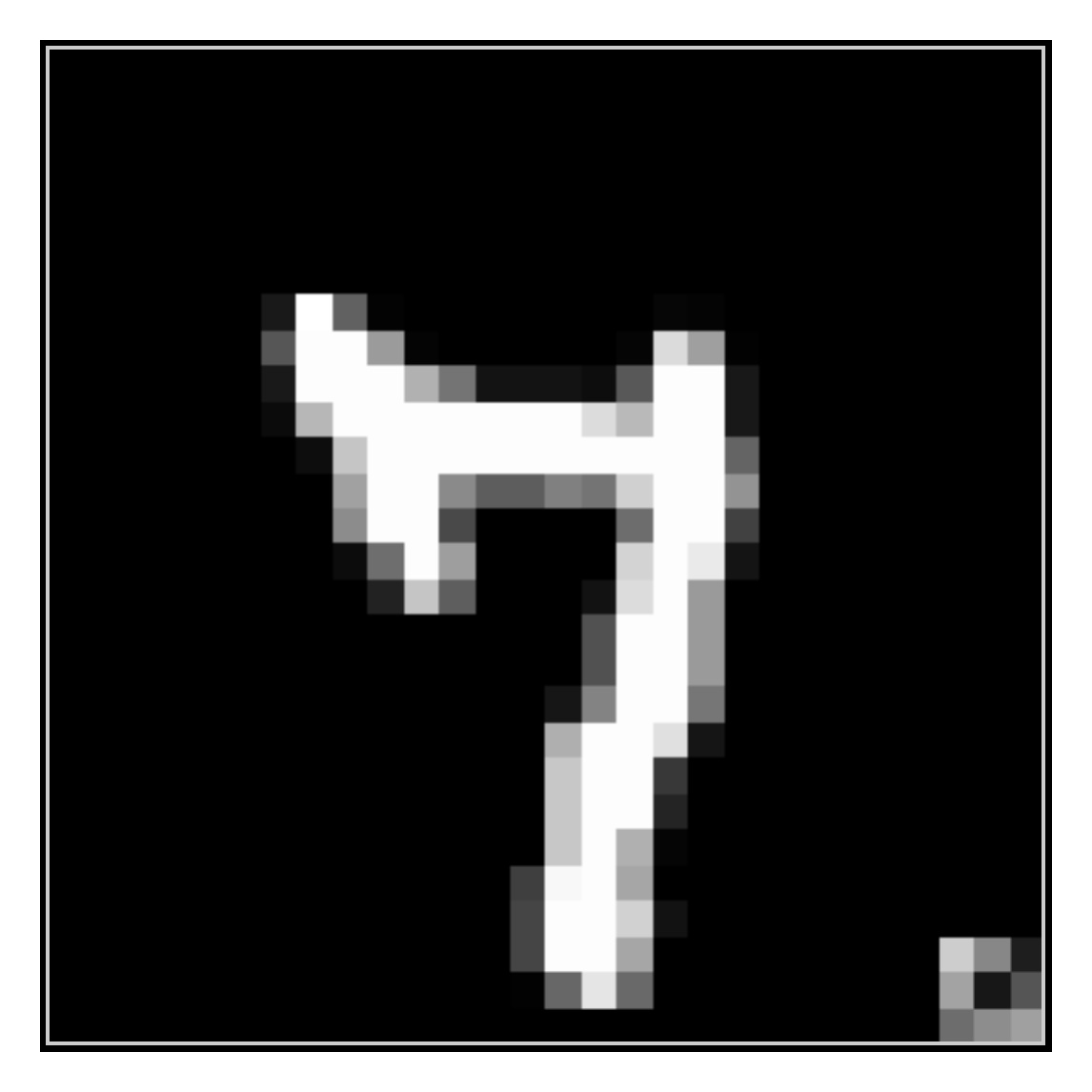}
  \includegraphics[width=0.17\textwidth]{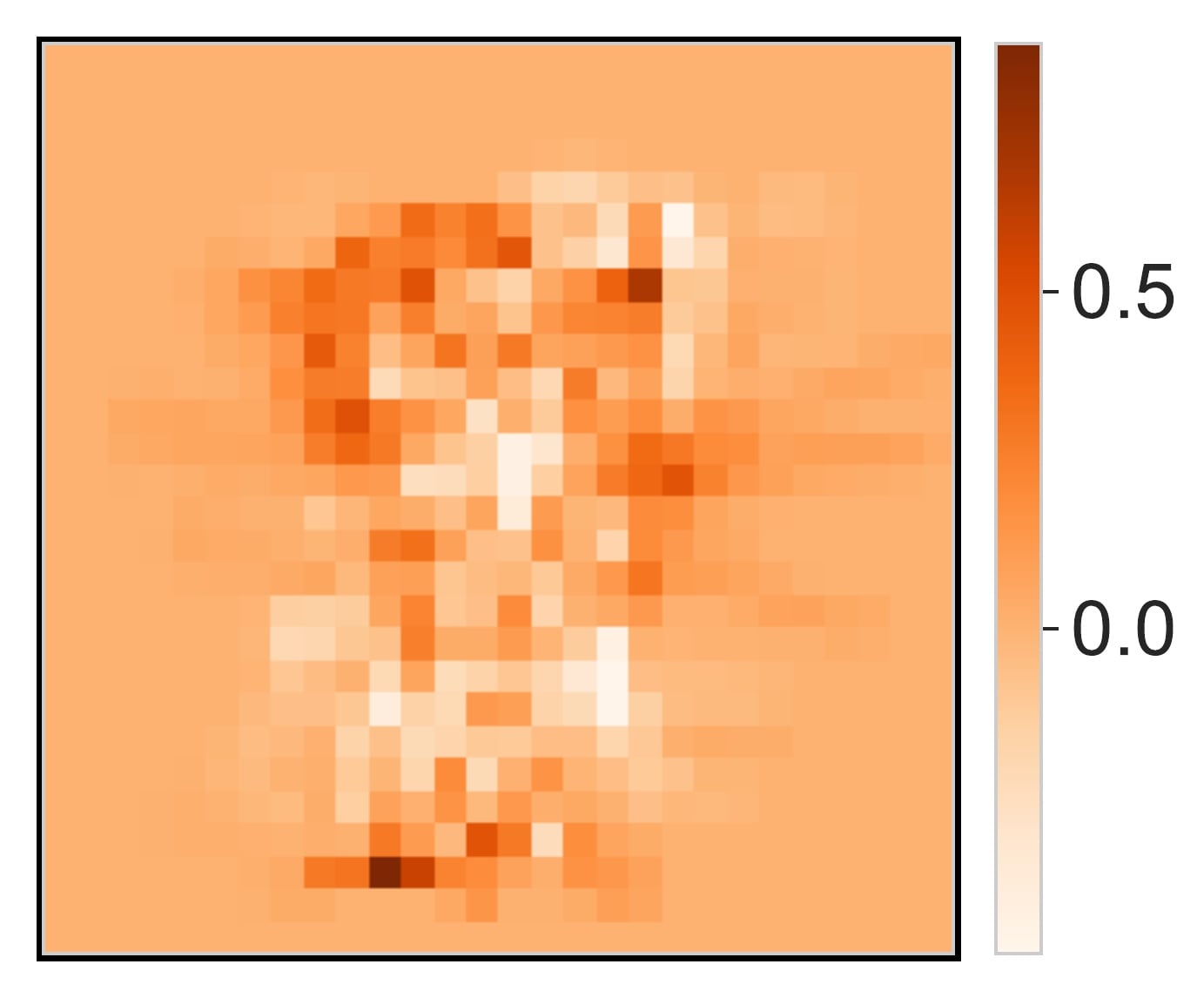}
  \hspace{-0.4em}
  \includegraphics[width=0.172\textwidth]{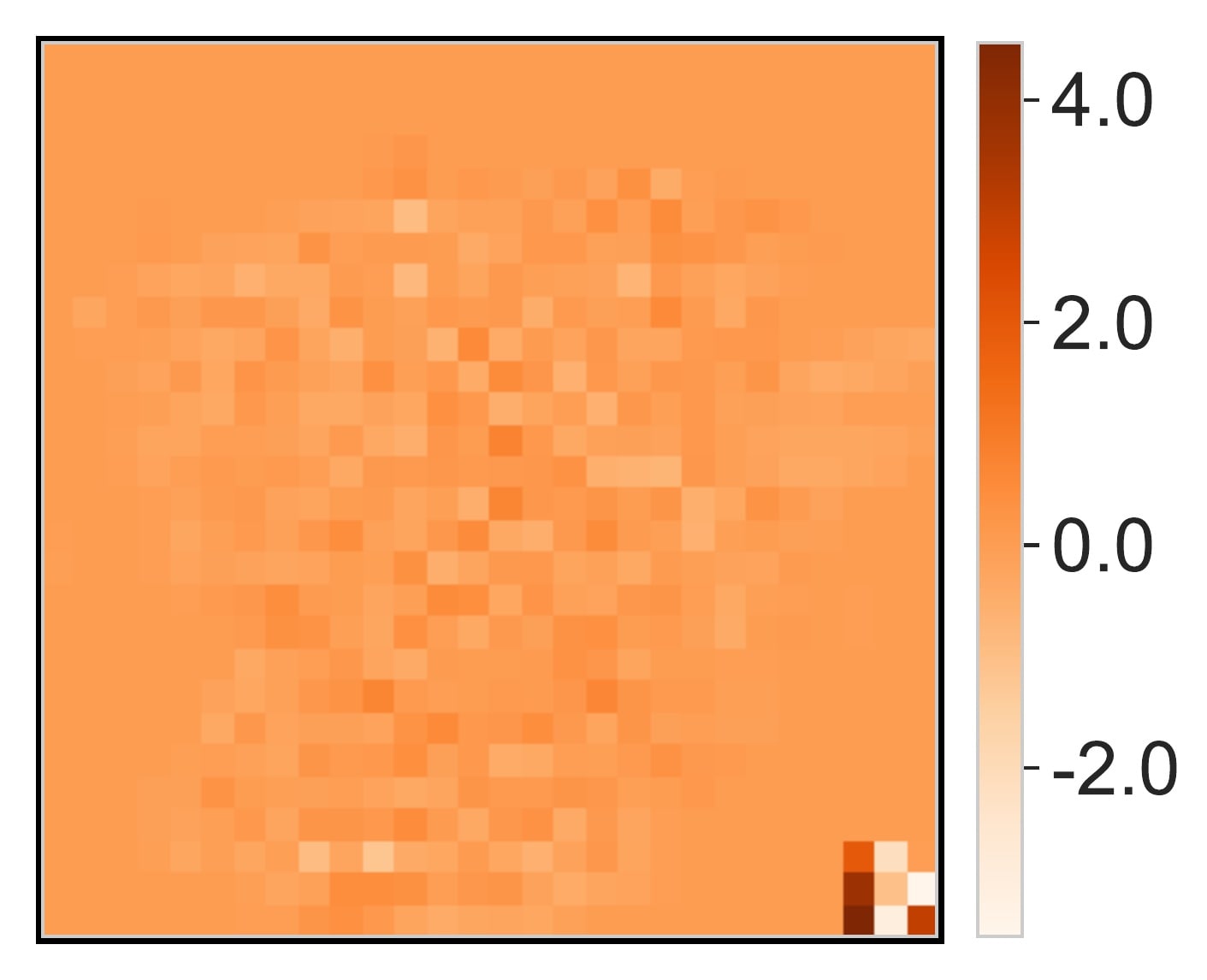}
  \includegraphics[width=0.135\textwidth]{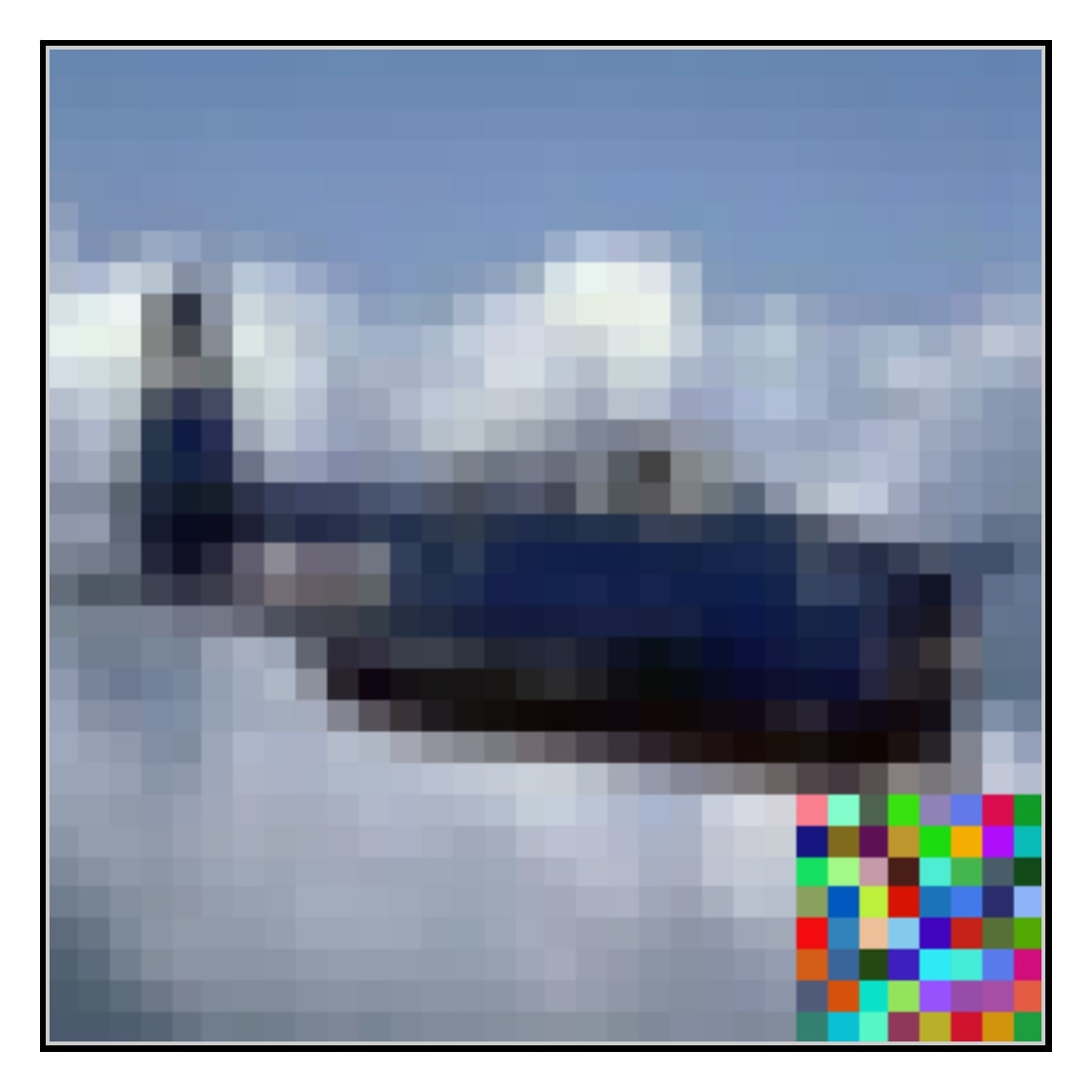}
  \includegraphics[width=0.172\textwidth]{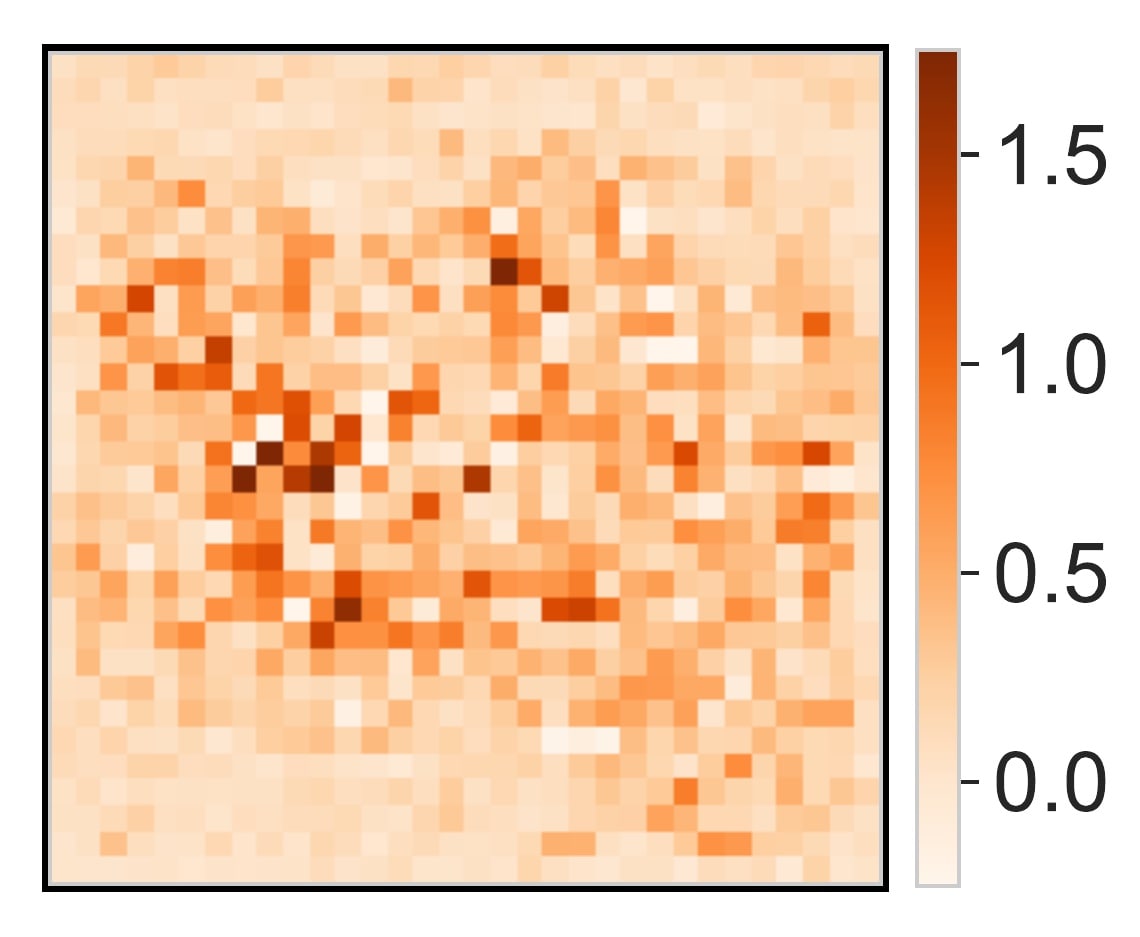}
  \includegraphics[width=0.172\textwidth]{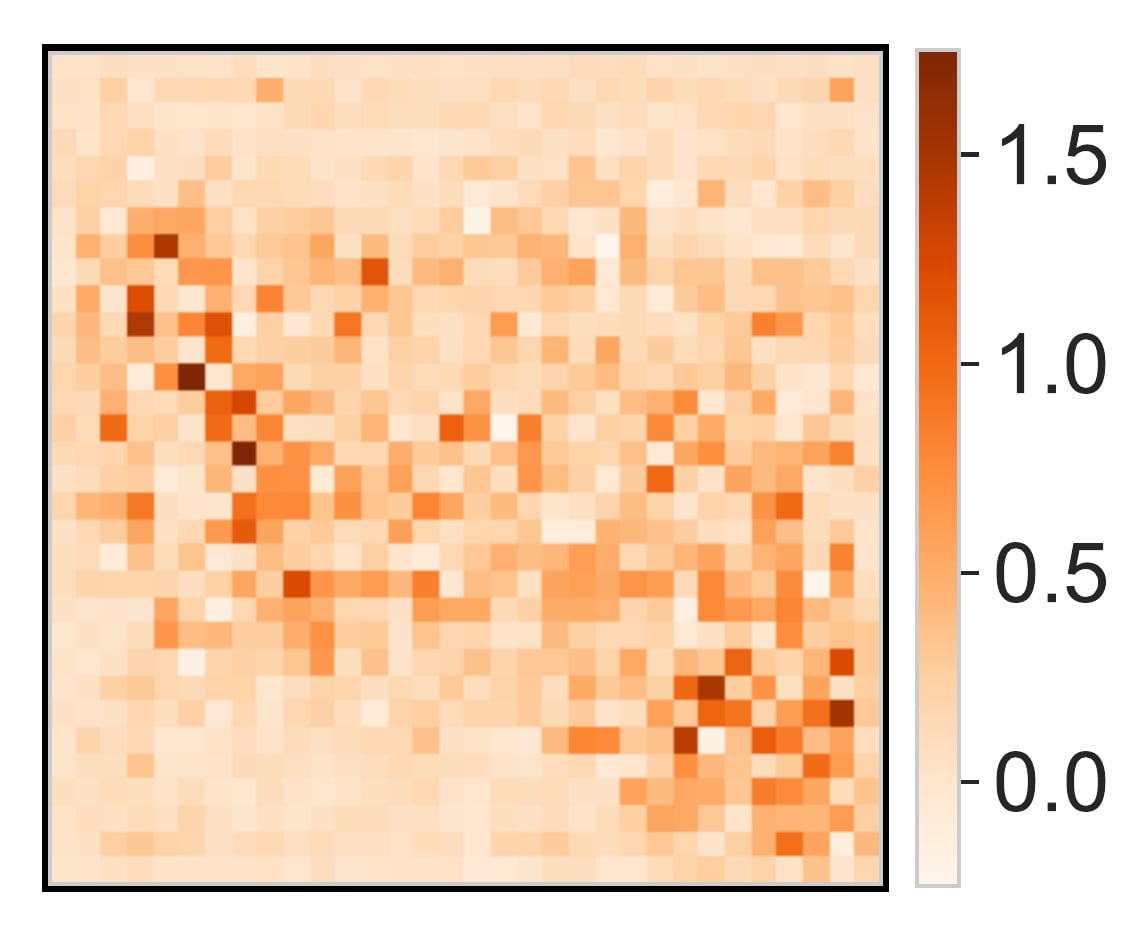}

    \caption{
    Input gradients of untainted and poisoned SVMs on pristine (top row) and backdoored (bottom row) test samples. Each row shows two sets of three images. Each set contains an example from MNIST \textit{7 vs 1} or CIFAR10~\cifarairplanefrog (left), along with the corresponding input gradient of the untainted SVM (middle), and of the poisoned SVM (right). For CIFAR10, we consider the maximum gradient of each pixel among the three channels.}
    
  \label{fig:interpretability}
\end{figure*}

In the following, we give a graphical interpretation of the poisoned convex-classifier's decision function, expressed by its internal weights, for which interpretation of their results is easier \cite{tramerSimpleModel21, dacremaProgress2019}. We consider the results for a backdoor trigger \cite{gu2019badnets} in a specific position, as its influence on the classifier decision is easier to see. Conversely, the backdoor trigger by for example Zhong et al.~\cite{Zhong20backdoor} spans the entire image, and therefore its influence is harder to spot from the interpretability plots.
In particular, given a sample $x$ we aim to compute and show the gradient of the classifier's decision function with respect to $x$. We use an SVM with regularization $\lambda = \expnumber{1}{-02}$ for MNIST $7~\rm{vs}~1$ and CIFAR10 \cifarairplanefrog, and report the results in Figure~\ref{fig:interpretability}.
For MNIST, we consider the digit $7$ with the trigger, showcasing the gradient of the clean classifier's decision function. We present the results of the gradient from the clean and poisoned classifiers corresponding to the clean and backdoored inputs. Since we train a linear classifier on the input space, the derivative coincides with the classifier's weights. Intriguingly, the classifier's weights increase in magnitude and now exhibit high values in the bottom right corner, where the trigger is located. 
From CIFAR10, we show a poisoned airplane. We report the gradient mask obtained by considering the maximum value for each channel, both for the clean and backdoored classifier.
Also, in this case, the backdoored model shows higher values in the bottom right region, corresponding to the trigger location.
This means that the analyzed classifiers assign high importance to the trigger to discriminate the class of the input points.

Summarizing, the plots in Figure~\ref{fig:interpretability} further confirm our findings regarding the change of the internal parameters during the backdoor learning process. 
In particular, we have seen that less regularized classifiers need to increase their weights and thus complexity to learn the backdoor. Conversely, when the flexibility of the classifier increases then it can learn the backdoor easier without significantly altering its complexity.

\subsubsection{Visualizing Influential Training Data Points}\label{sec:influece-function-example}
\begin{figure*}[h!]
  \centering
  \includegraphics[width=1\textwidth]{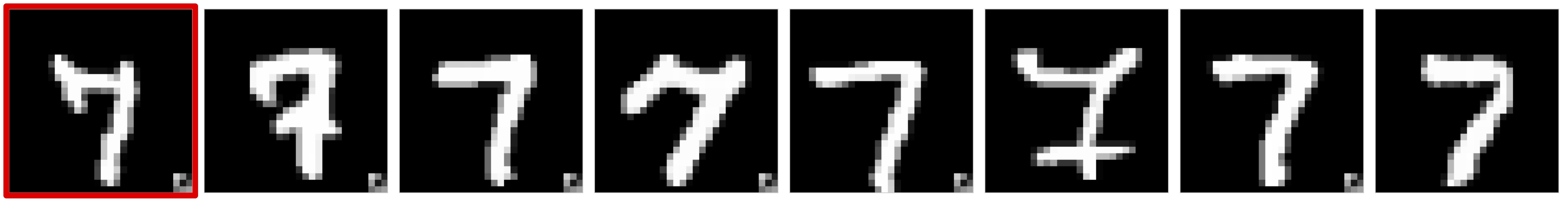}
    \includegraphics[width=1\textwidth]{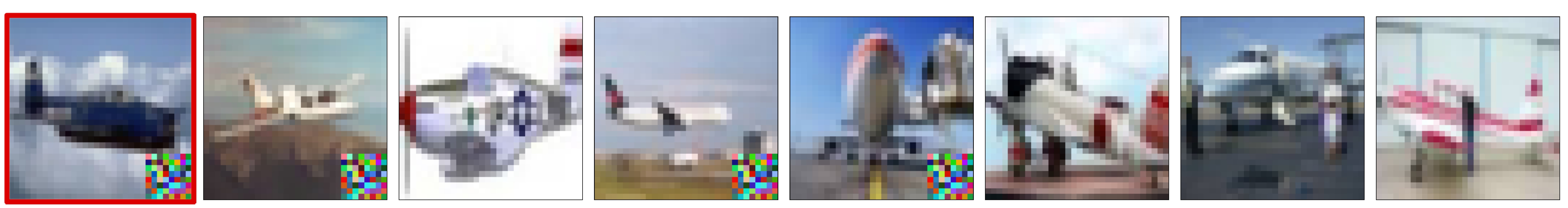}
    \includegraphics[width=1\textwidth]{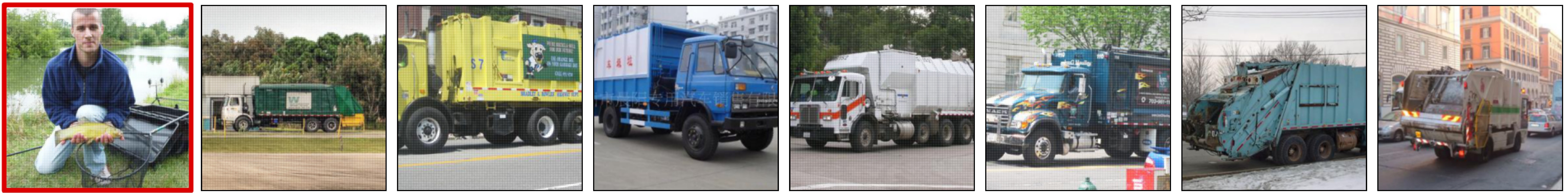}

    \caption{Influential training points for a high-complexity classifier. Considering an SVM with $\lambda=0.01$ trained on MNIST, and with $\lambda=0.1$ trained on CIFAR10, and Imagenette, we show the top $7$ most influential training samples on the prediction of the samples with the red border.}
  \label{fig:top_influent_weak}
\end{figure*}

\begin{figure*}[h!]
  \centering
  \includegraphics[width=1\textwidth]{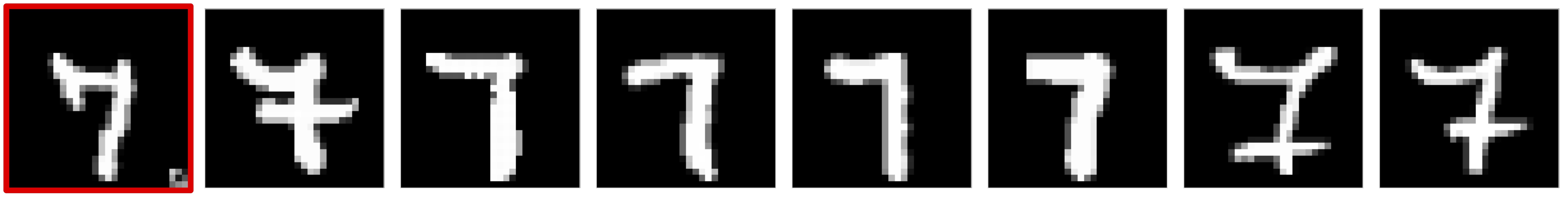}
    \includegraphics[width=1\textwidth]{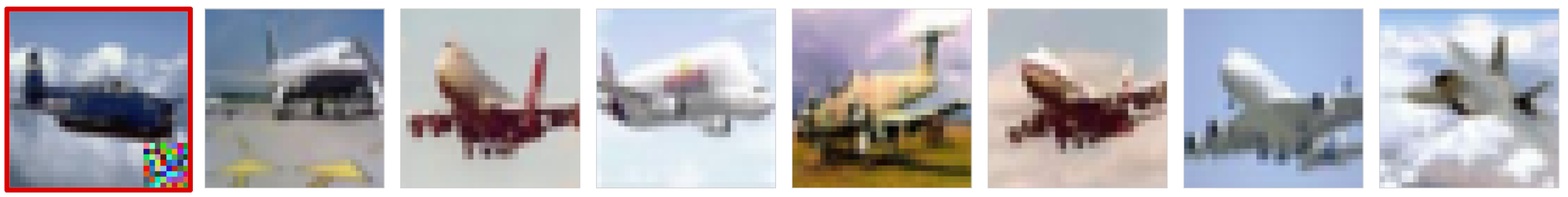}
    \includegraphics[width=1\textwidth]{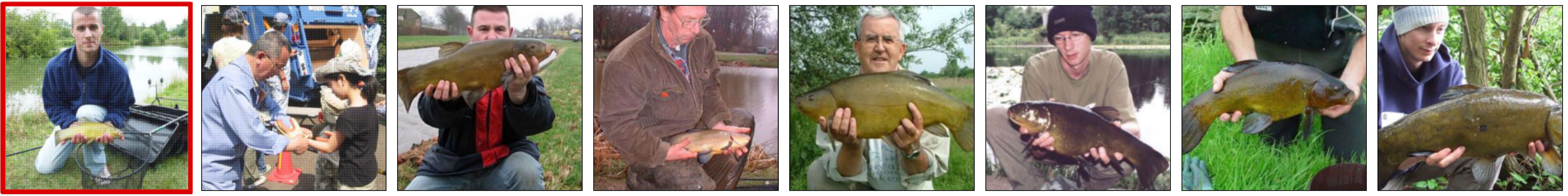}
    \caption{Influential training points for low-complexity classifiers. Considering an SVM with $\lambda=1e-3$ trained on MNIST, and with $\lambda = 1e-5$ trained on CIFAR10, and Imagenette, we show the top $7$ most influential training samples on the prediction of the samples with the red border.}
  \label{fig:top_influent_strong}
\end{figure*}

Influence functions are used in the context of ML to identify the training points more responsible for a given prediction \cite{koh2017understanding}. In Section~\ref{sec:backdoor-curves} we have seen how they represent the basis of our backdoor learning slope measure. In this section, we employ them to show their outcomes and provide further insight into the relationship between complexity and backdoor effectiveness. To this end, as in Section \ref{sec:experimental_setup}, we poison $10\%$ of the training dataset.
According to previous experiments, we employed the backdoor trigger in \cite{gu2019badnets} for MNIST and CIFAR10 with trigger size $3 \times 3$ and $6 \times 6$ respectively, while for Imagenette we employed the trigger in Zhong et al.~\cite{Zhong20backdoor} with higher visibility (\ie $c_m=75$).
In Figure~\ref{fig:top_influent_weak} and ~\ref{fig:top_influent_strong}, considering respectively a high- and a low-complexity classifier, we report the seven most influential training samples 
on the classification of a randomly chosen test point. 
For high-complexity classifiers, many of these training samples contain the trigger. In contrast, this is not the case for low-complexity classifiers. 
These results suggest that low-complexity classifiers rely less on the samples containing the backdoor trigger in their predictions.

\section{Related Work}\label{sec:relWork}
We first review the literature about backdoor poisoning attacks and defenses. 
Afterward, we focus on defenses that increase the robustness against backdoors by reducing the model's complexity. 
We conclude the section by discussing the relationship between our proposed framework and influence functions.\smallskip

\myparagraph{Backdoor Poisoning.}
Although backdoors were introduced recently~\cite{gu2019badnets,Cina2022Survey,Chen2017TargetedBA}, a plethora of backdoor attacks and defenses have been published. For a more detailed overview, we refer the reader to surveys in this area~\cite{Cina2022Survey,goldblum2020data,cina2022MLSecurity}.
Despite the quickly-growing literature about this topic, the majority of the previous works~\cite{suya2021model,frederickson2018attack,cano21reg,cina2021hammer} study different types of poisoning attacks, \ie, not backdoors. In contrast, only a few works have studied factors that influence the success of this attack. Baluta et al.~\cite{baluta2019quantitative} and \cite{Lin2022MeasuringTE} studied the relationship between backdoor effectiveness and the percentage of backdoored samples. Salem et al.~\cite{salem2020Dynamic} experimentally investigated the relationship between the backdoor effectiveness and the trigger size. Similarly, Severi et al.~\cite{Severi2021ExplanationGuidedBP} have analyzed the correlation between the backdoor success and the attacker's strength on malware classifiers. Schwarzschild et el.~\cite{Schwarzschild2020JustHT} evaluated the performance of backdoor attacks when scaling the dataset size while fixing the poison budget. Finally, Li et al.~\cite{Li2020RethinkingTT} demonstrated that the backdoor performance is sensitive to the location of the trigger on the attacked image. We instead do not limit our study to neural networks but also study other models. Furthermore, we also investigate other relevant factors, e.g., regularization and visibility, and their interaction at once. \smallskip

\myparagraph{Complexity and Backdoor Defenses.}
In this work, we have analyzed the relationship between backdoor effectiveness and different factors, including complexity, controlled via regularization and the RBF kernel's hyperparameter.
In this study, we have demonstrated that reducing complexity by choosing appropriate hyperparameter values improves robustness against backdoors.  Our findings align with the insights presented in Frnay et al.~\cite{Frnay2014ClassificationIT}, who suggested that overfitting avoidance techniques like, e.g., regularization, can offer partial mitigation against random label noise~\cite{Teng2000EvaluatingNC,Teng2001ACO}. Expanding upon their discourse, we apply and extend this consideration to the context of backdoor attacks, wherein the noise is intentionally and strategically introduced to deceive the machine learning model. 
Some of the defenses proposed against backdoors use different techniques to reduce complexity. These techniques include pruning~\cite{liu2018fine,bajcsy2021baseline}, data augmentation~\cite{zeng2020deepsweep, borgnia2021strong} and gradient shaping~\cite{hong2020effectiveness}. However, from these works, it remains unclear why reducing complexity alleviates the threat of backdoor poisoning. To the best of our knowledge, our work is the first to investigate this aspect.\smallskip

\myparagraph{Relation to Influence Functions.} %
Influence functions originated in robust statistics~\cite{cook1980characterizations} and were later used as a tool to measure the influence of specific training points on the classification output~\cite{christmann2004robustness, koh2017understanding}.
In our work, we clarify that influence functions naturally descend from the incremental learning formulation in Equation \ref{eq:poisoning_problem_inner}, showing that they quantify the velocity with which the classifier will learn new points.
As seen in Section~\ref{sec:backdoor-curves},  they correspond to the partial derivative of the learning curve at the point $\beta = 0$. Moreover, we leveraged them by proposing a measure, namely the backdoor slope, which quantifies the ability of a classifier to learn backdoors. This measure allowed us to study the factors that impact backdoor effectiveness.

Several defense approaches confirm that the influence functions, or gradients during training, are indeed related to backdoor learning. 
For example, some defenses are directly based on the gradient~\cite{geiping2021doesn}, based on gradient differences~\cite{li2021anti,yo2022CreatedEqual}, or based on differential privacy that noises the gradients during training~\cite{hong2020effectiveness,borgnia2021dp,du2019diffprivacy}.

\section{Conclusions, Limitations and Future Work}\label{sec:concl}

In this paper, we presented a framework to analyze the factors influencing the effectiveness of backdoor poisoning. We carried out experiments on convex learners, also used in transfer-learning scenarios, and neural networks. As in previous work~\cite{koh2017understanding, saha2020hidden}, we focus our analysis on two-class classification problems for convex learners, and on multiclass classification when considering neural networks.

Our analysis shows that the effectiveness of backdoor attacks inherently depends on (i) the complexity of the target model, (ii) the fraction of backdoor samples in the training set, and (iii) the size and visibility of the backdoor trigger.
By analyzing the influence of the first factor on backdoor learning, we are the first to unveil a region in the hyperparameter space where the accuracy on clean test samples remains high while the accuracy on backdoor samples is low.  Specifically, we discovered that the target model needs to significantly increase the complexity of its decision function to learn backdoors, which is only possible when the model is not regularized enough. Conversely, when raising the model's regularization, we can keep high performance on clean samples and be unaffected by potential backdoor attacks. However, increasing the attacker's strength, i.e., the last two factors, makes the attack more effective, shrinking this region and thus exposing the model to greater vulnerability. We, therefore, conclude that a prudent strategy to preserve robustness against potential poisoning attacks is to regularize as much as possible during the hyperparameter optimization phase, thereby reducing the backdoor learning slope while ensuring that the trade-off with accuracy remains acceptable.

The study of more factors, like, for example, the dimensionality of the data, is straightforward using the proposed framework but left for future work. Our current results already provide important insights and provide a starting point to derive guidelines for designing models that are more robust against backdoor poisoning.

\section*{Declarations}

\myparagraph{Funding.} This work has been partially supported by Spoke 10 ``Logistics and Freight'' within the Italian PNRR National Centre for Sustainable Mobility (MOST), CUP I53C22000720001; the projects SERICS (PE00000014) and FAIR (PE00000013) under the NRRP MUR program funded by the EU - NGEU; the PRIN 2017 project RexLearn
(grant no. 2017TWNMH2), funded by the Italian MUR; and by BMK, BMDW, and the Province of Upper Austria in the frame of the COMET Programme managed by FFG in the COMET Module S3AI.

\myparagraph{Conflict of interest.} The authors declare that they have no competing
interests.

\myparagraph{Data availability.} The data MNIST, CIFAR10, and Imagenette used in this article can be obtained from the following links: MNIST~\url{http://yann.lecun.com/exdb/mnist/}, CIFAR10~\url{https://www.cs.toronto.edu/~kriz/cifar.html}, and Imagenette~\url{https://github.com/fastai/imagenette}.

\myparagraph{Code availability.} The codebase is publicly available at \url{https://github.com/Cinofix/backdoor_learning_curves}.

\bibliographystyle{IEEEtranN}
\bibliography{sn-bibliography-short}

\clearpage
\appendix
\subsection{Datasets}\label{sec:supplementary_dataset} 
The MNIST dataset~\cite{lecun2010mnist} contains $70,000$ observations representing $28 \times 28$ grayscale images of handwritten digits from $0$ to $9$. The CIFAR10 dataset~\cite{Krizhevsky09learningmultiple} contains $60,000$ colour images of size $32 \times 32$ pixels divided in 10 classes, each with $6,000$ observations. Finally, the Imagenette dataset~\cite{imagenette} is a subset of 10 classes (i.e., tench, English springer, cassette player, chain saw, church, French horn, garbage truck, gas pump, golf ball, parachute) from Imagenet. We use the 320px version, where the shortest side of each image is resized to that size.

\subsection{Additional Experimental Results}\label{sec:supplementary_experiments} 
\noindent In the paper, we have shown the backdoor learning curves only for some classifiers. Here, we report them for all the classifiers considered in this work. As we will discuss later in this section, these results confirm the ones obtained in the paper. 
In particular, here we consider:
\begin{itemize}
    \item Support Vector Machine (SVM) with $\lambda \in \{100, 0.1\}$ for MNIST, $\lambda \in \{10000, 0.1\}$ for CIFAR10, and $\lambda \in \{100000, 1\}$ for Imagenette.
    \item Ridge Classifier (RC) with $\lambda \in \{1000, 1\}$ for MNIST, $\lambda \in \{10000, 1\}$ for CIFAR10, and $\lambda \in \{100000, 1\}$ for Imagenette.
    \item Logistic Classifier (LC) with $\lambda \in \{10, 0.01\}$ for MNIST,  $\lambda \in \{10000, 100\}$ for CIFAR10, and $\lambda \in \{100000, 10\}$ for Imagenette.
    \item SVM with an RBF kernel, where $\lambda \in \{1, 0.01\}$ and $\gamma=\expnumber{5}{-04}$ for MNIST,  $\lambda \in \{100, 1\}$ and $\gamma=\expnumber{1}{-03}$ for CIFAR10, and $\lambda \in \{10, 0.1\}$ and $\gamma=\expnumber{1}{-05}$ for Imagenette.
\end{itemize}
Moreover, we compare the results obtained on the class pairs considered in the paper ($7~\rm{vs}~1$ on MNIST, \cifarairplanefrog~on CIFAR10 and Imagenette \imagenettetenchtruck) with the ones obtained on different pairs.\smallskip

\begin{figure*}[t]
\centering
\includegraphics[width=0.495\textwidth]{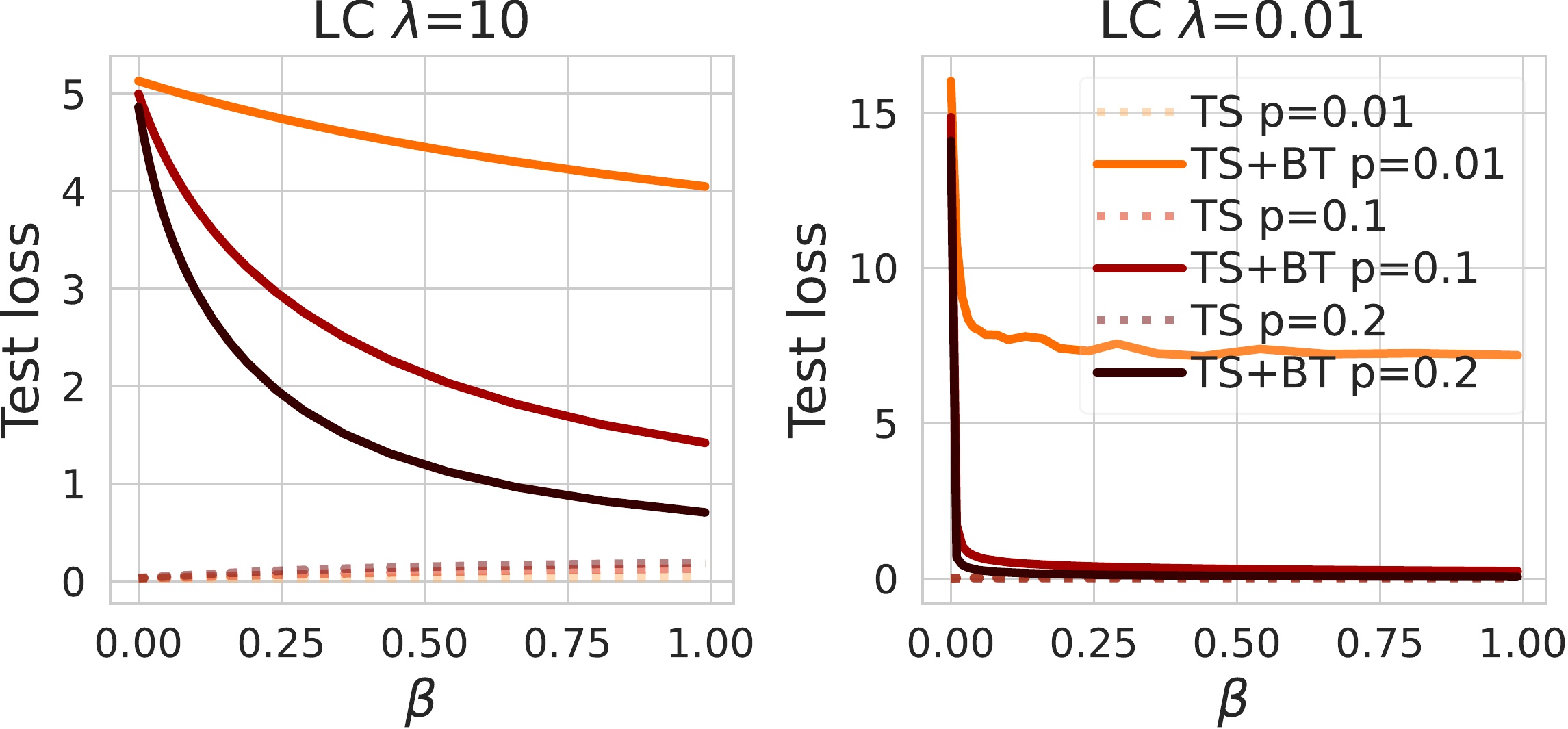}
\includegraphics[width=0.495\textwidth]{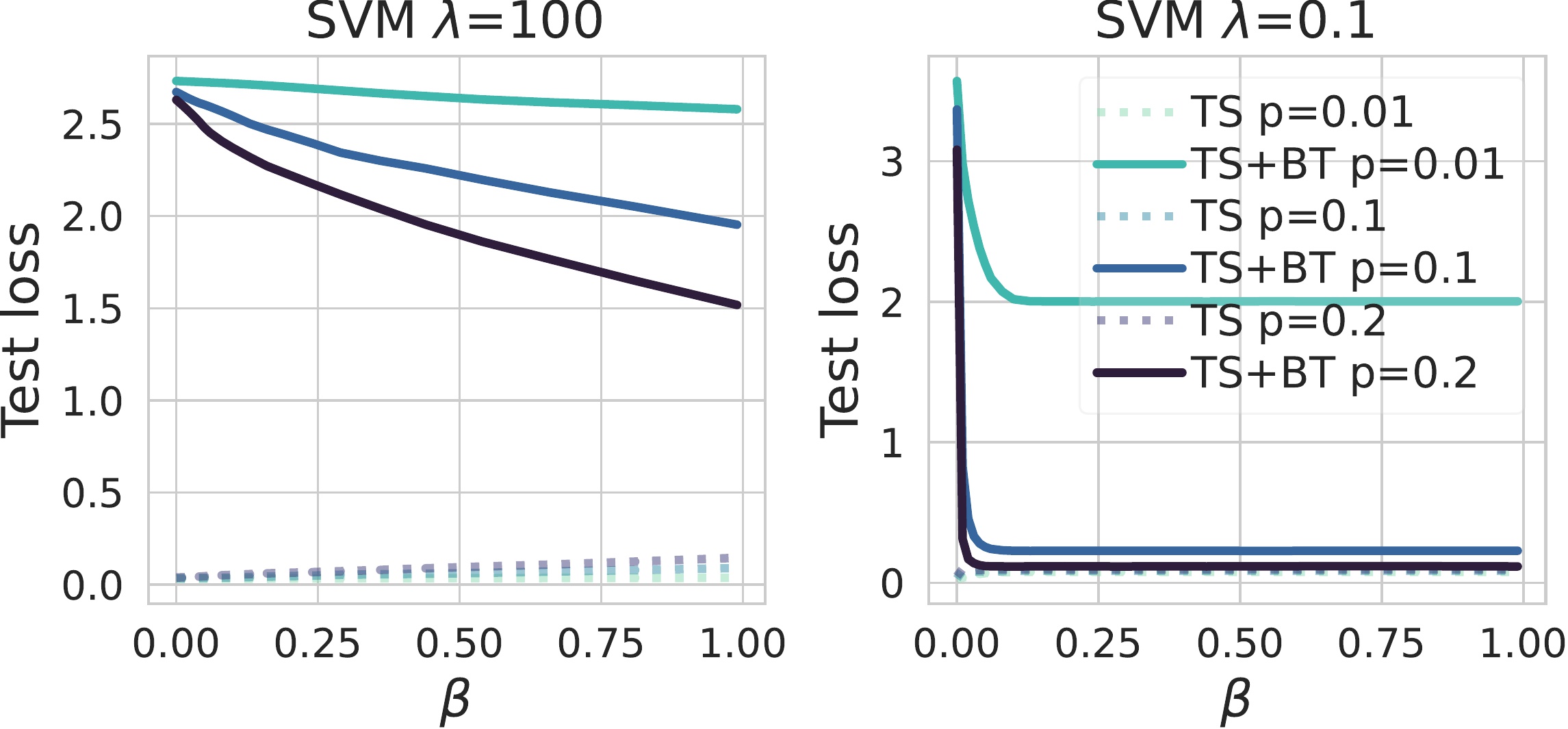}
\includegraphics[width=0.495\textwidth]{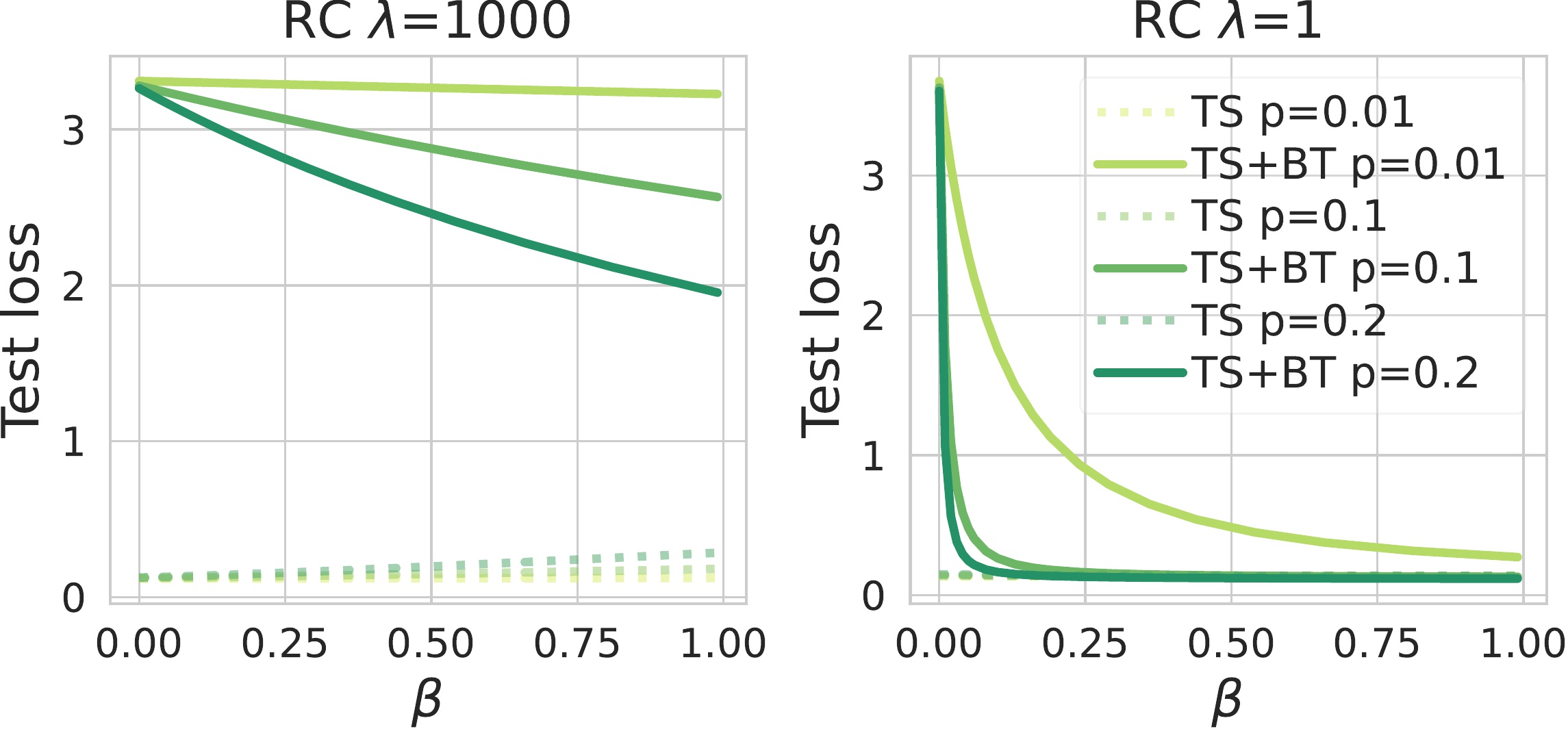}
\includegraphics[width=0.495\textwidth]{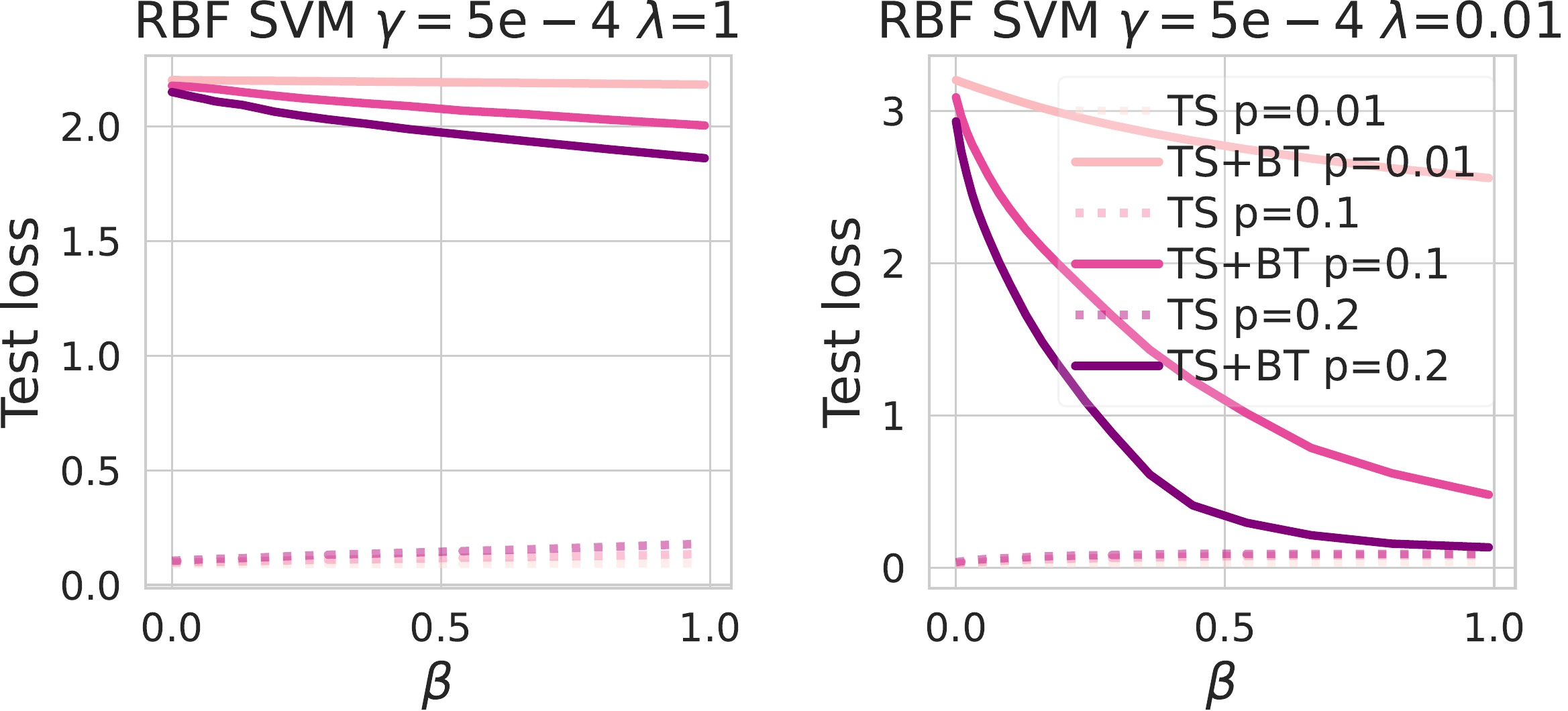}
\caption{Backdoor learning curves for different classifiers trained on MNIST 3-0. Darker lines represent a higher fraction of poisoning samples $p$ injected into the training set. We report the loss on the clean test samples (TS) with a dashed line and on the test samples with the backdoor trigger (TS+BT) with a solid line.}\label{fig:appendix_backdoor_learning_curves_mnist30}
\end{figure*}

\begin{figure*}[h!]
\centering
\includegraphics[width=0.495\textwidth]{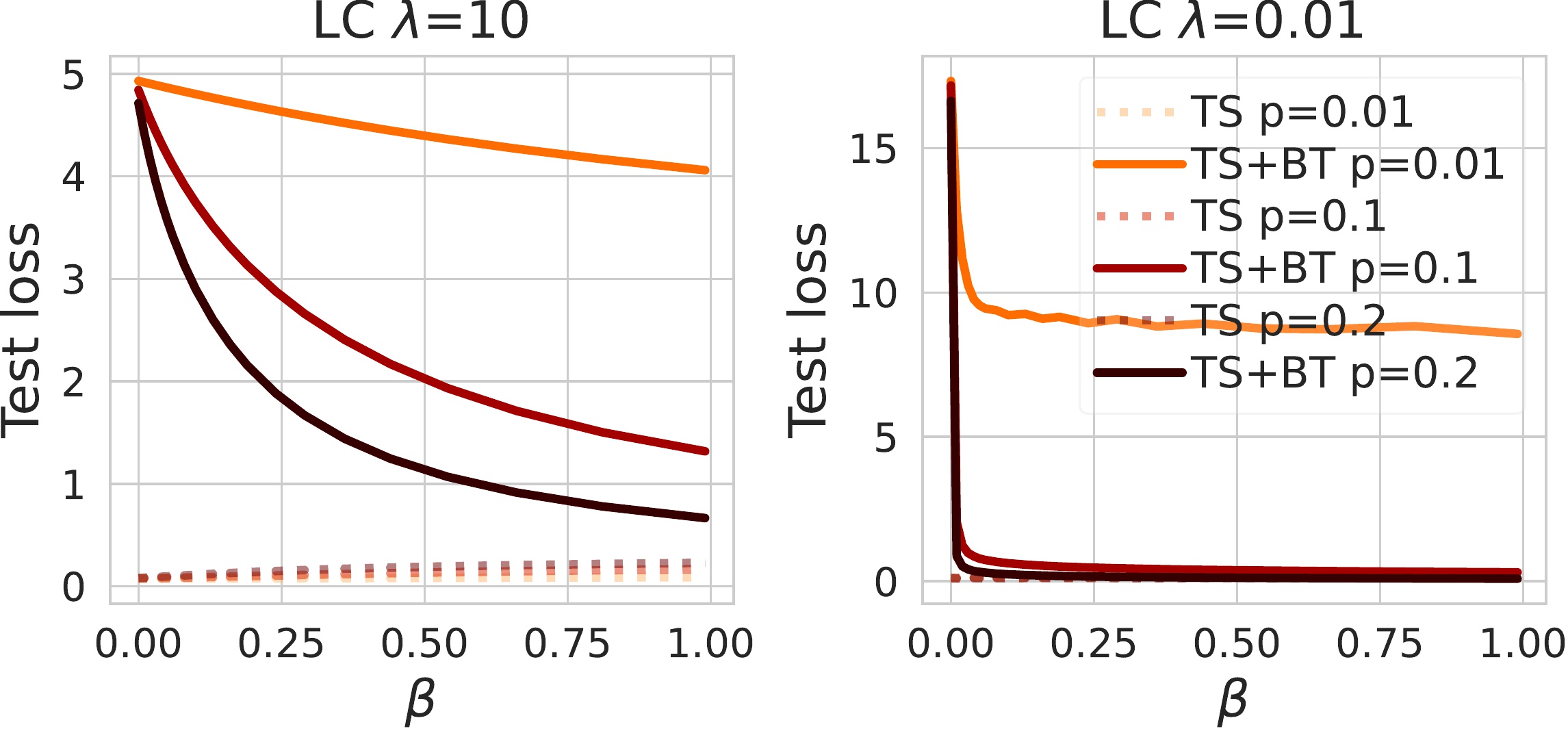}
\includegraphics[width=0.495\textwidth]{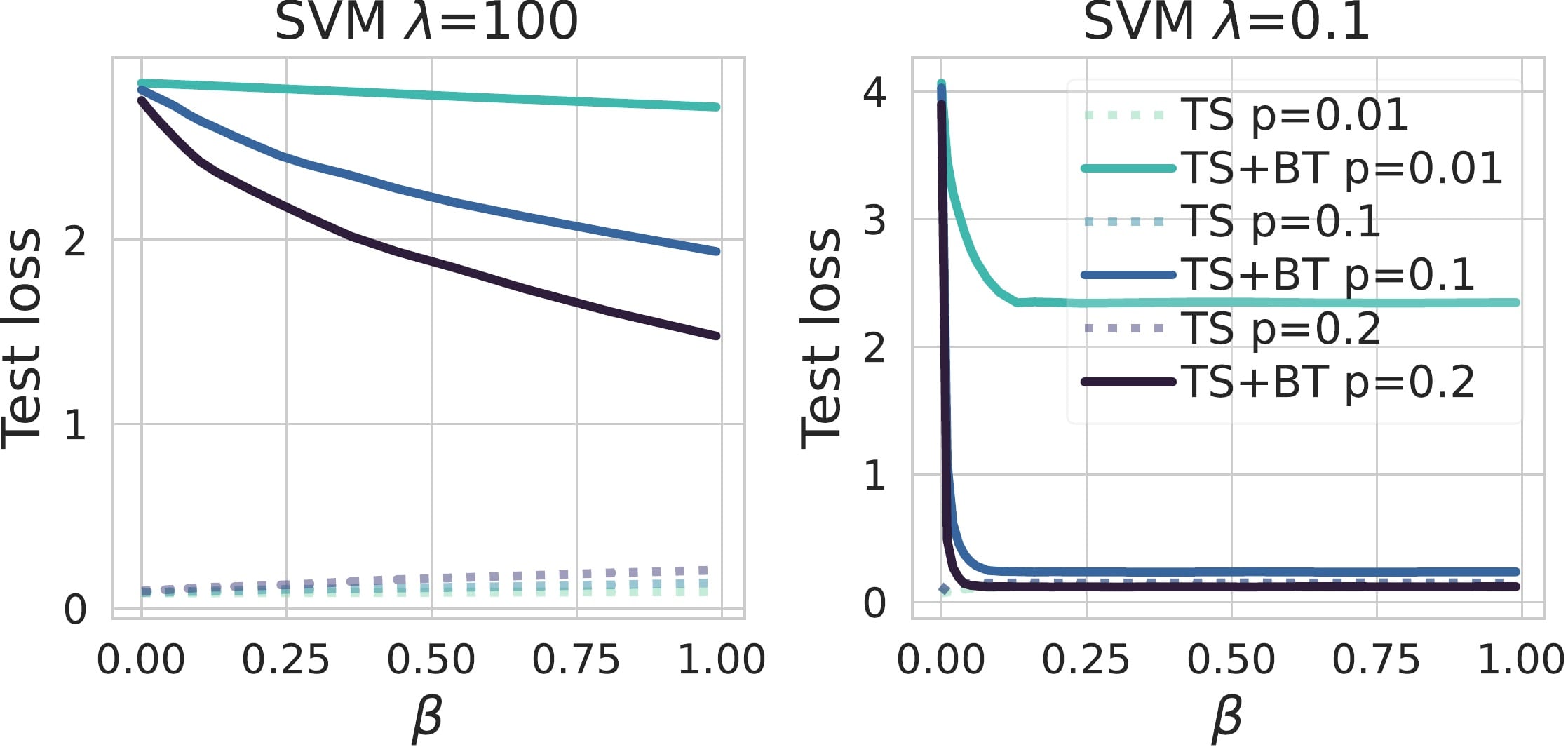}
\includegraphics[width=0.495\textwidth]{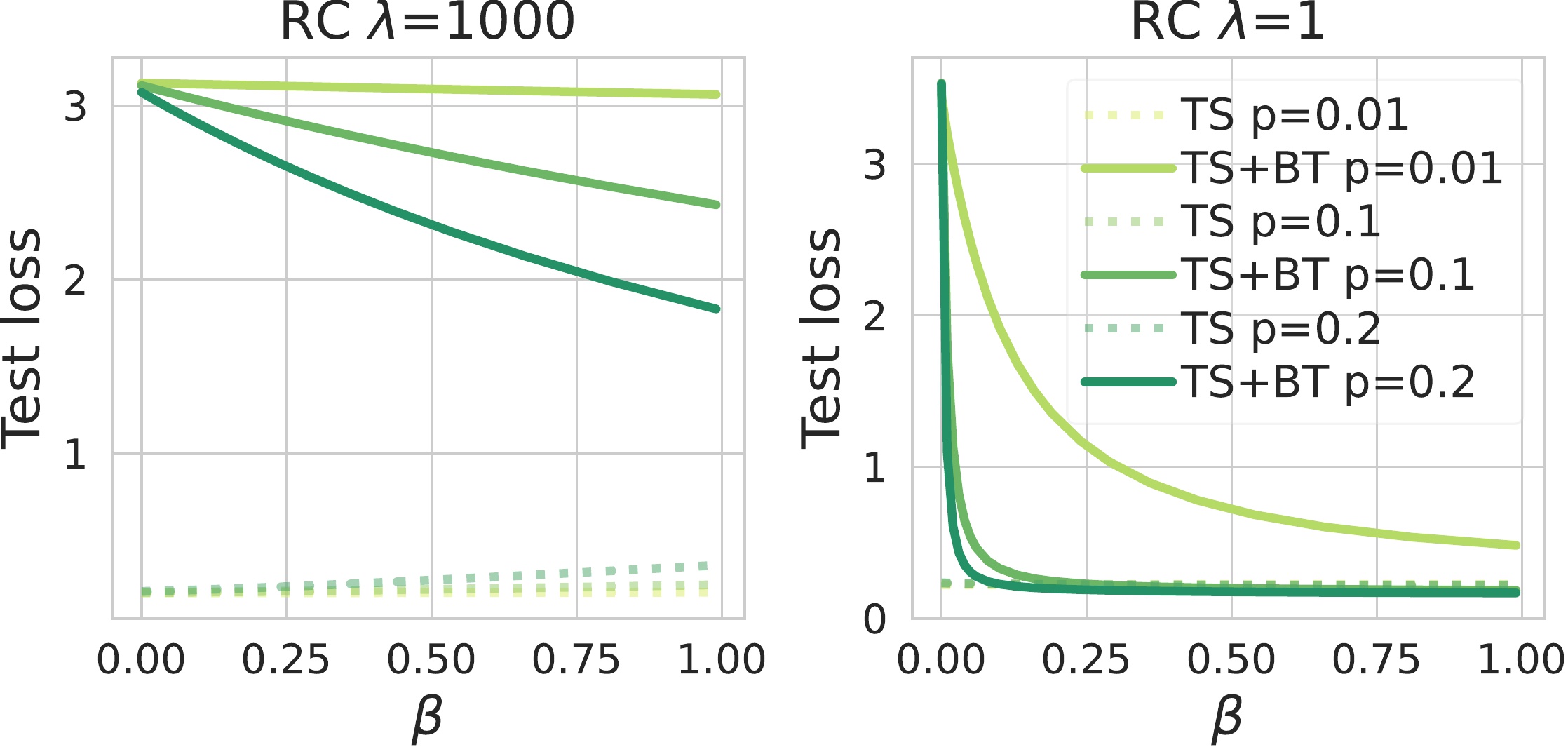}
\includegraphics[width=0.495\textwidth]{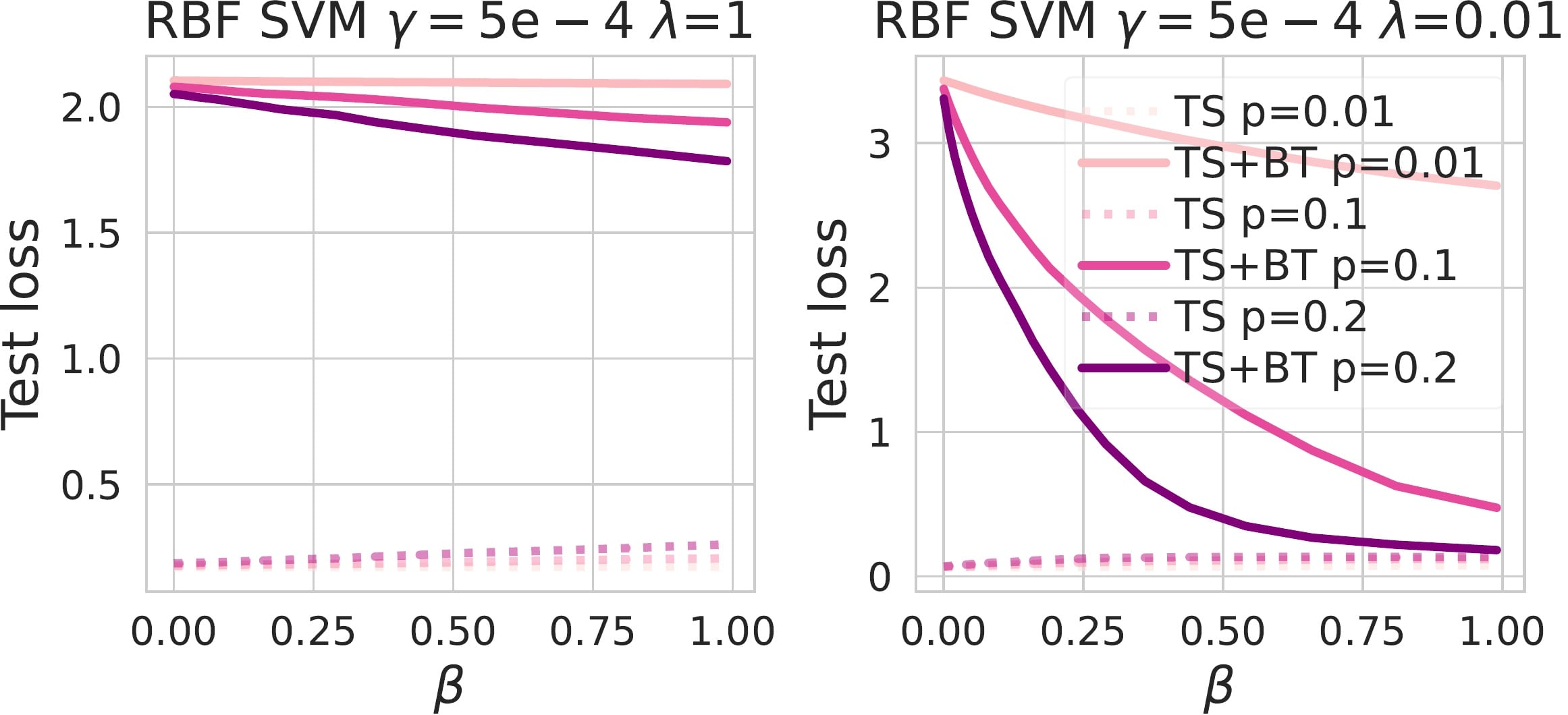}
\caption{Backdoor learning curves for different classifiers trained on MNIST 5-2.
See the caption of Figure~\ref{fig:appendix_backdoor_learning_curves_mnist30} for further details.}\label{fig:appendix_backdoor_learning_curves_mnist52}
\end{figure*}

\begin{figure*}[h!]
\centering
\includegraphics[width=0.495\textwidth]{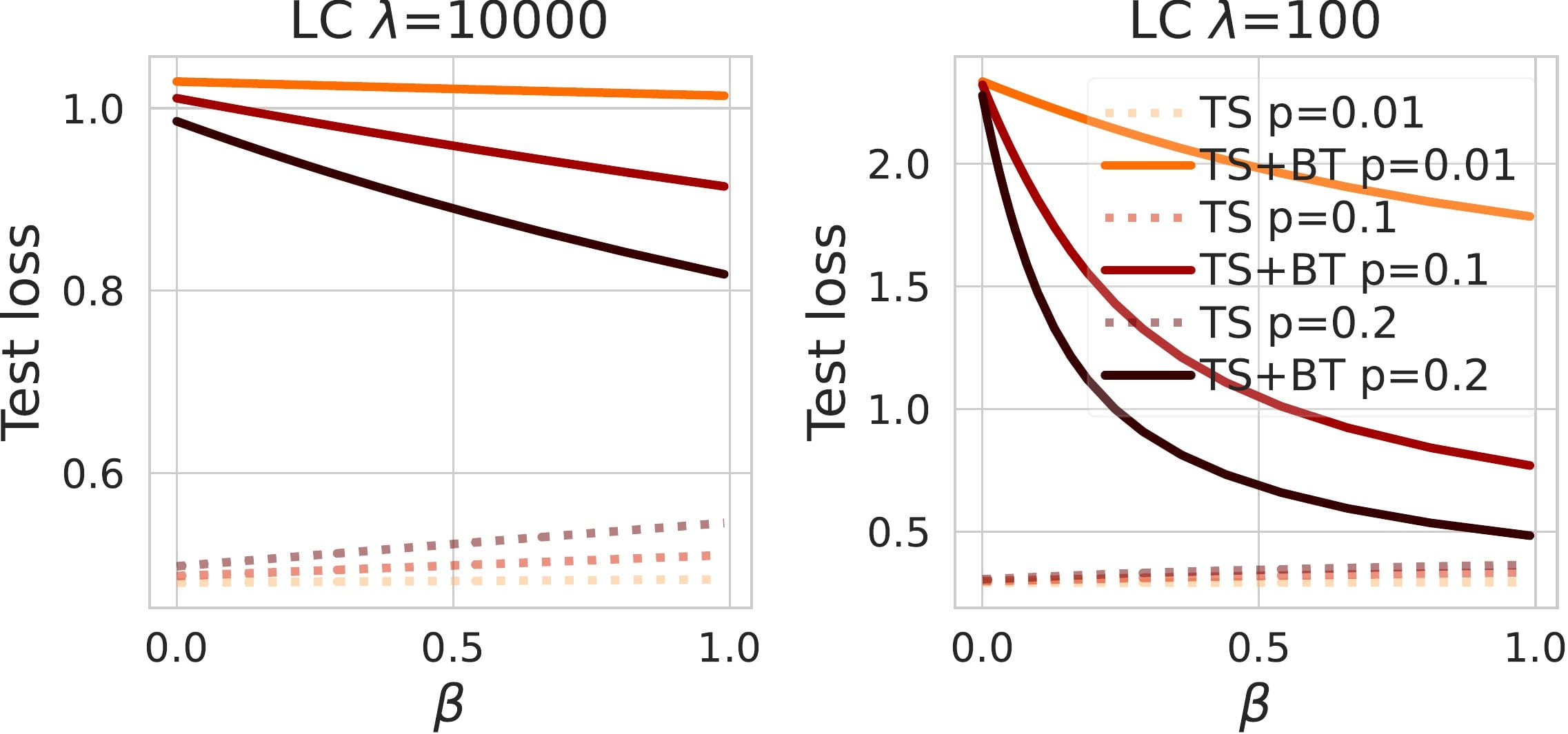}
\includegraphics[width=0.495\textwidth]{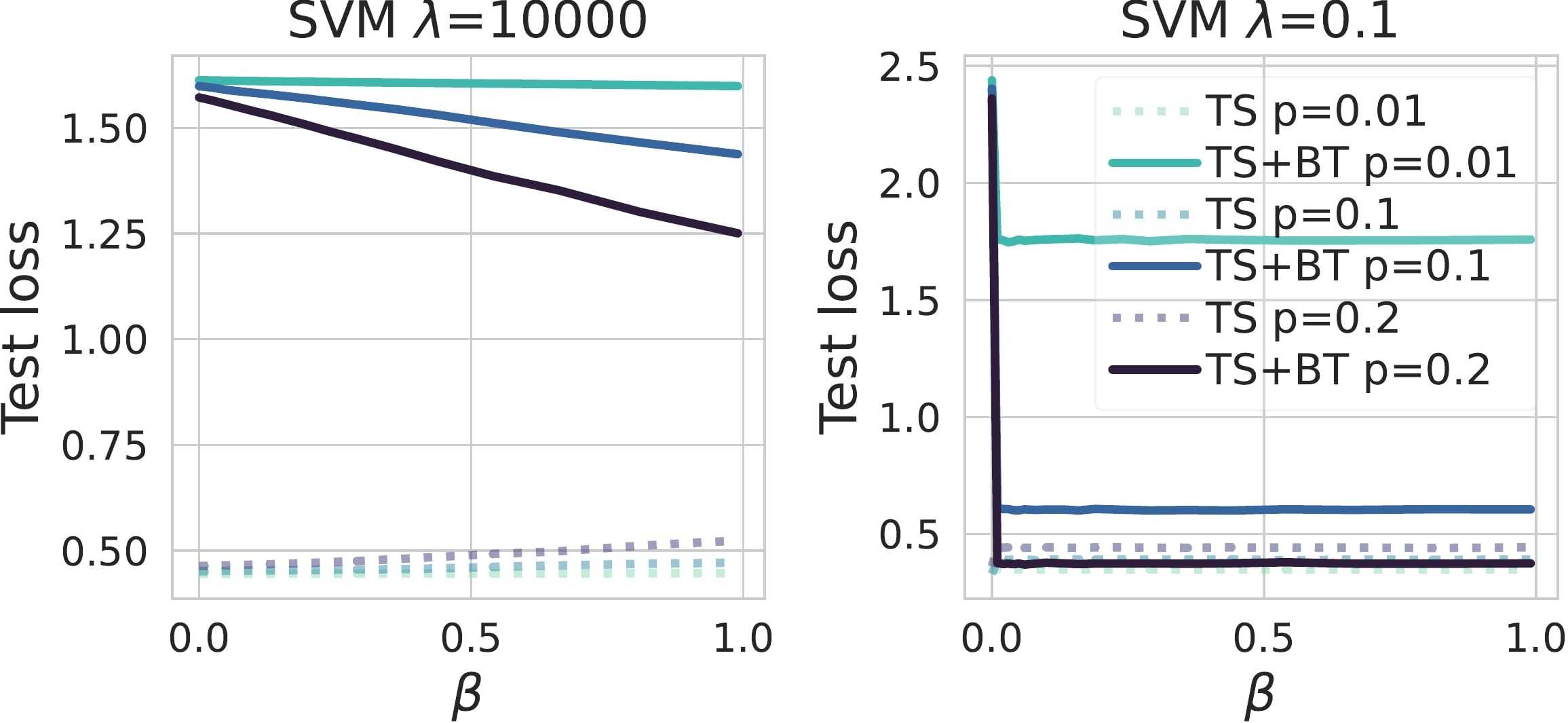}
\includegraphics[width=0.495\textwidth]{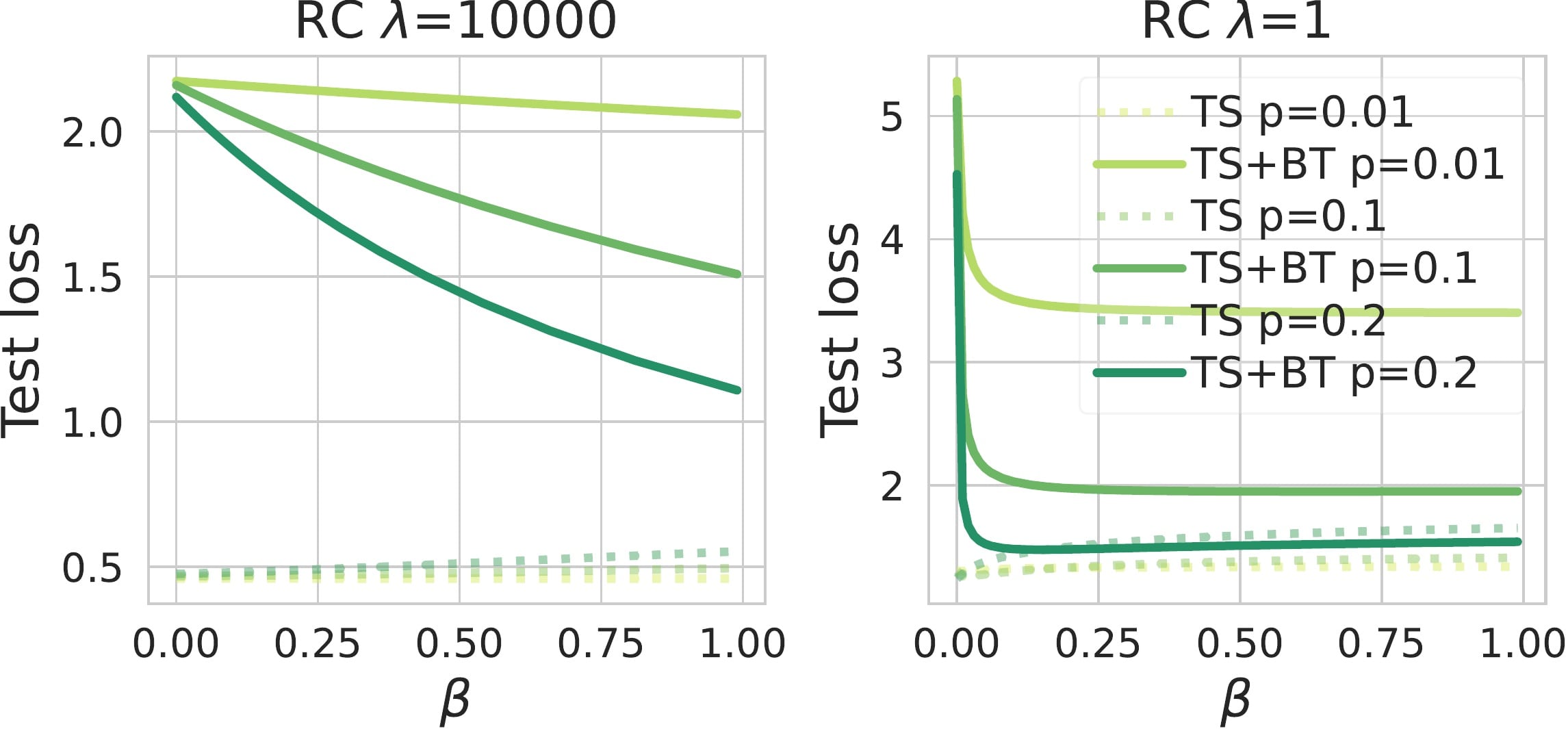}
\includegraphics[width=0.495\textwidth]{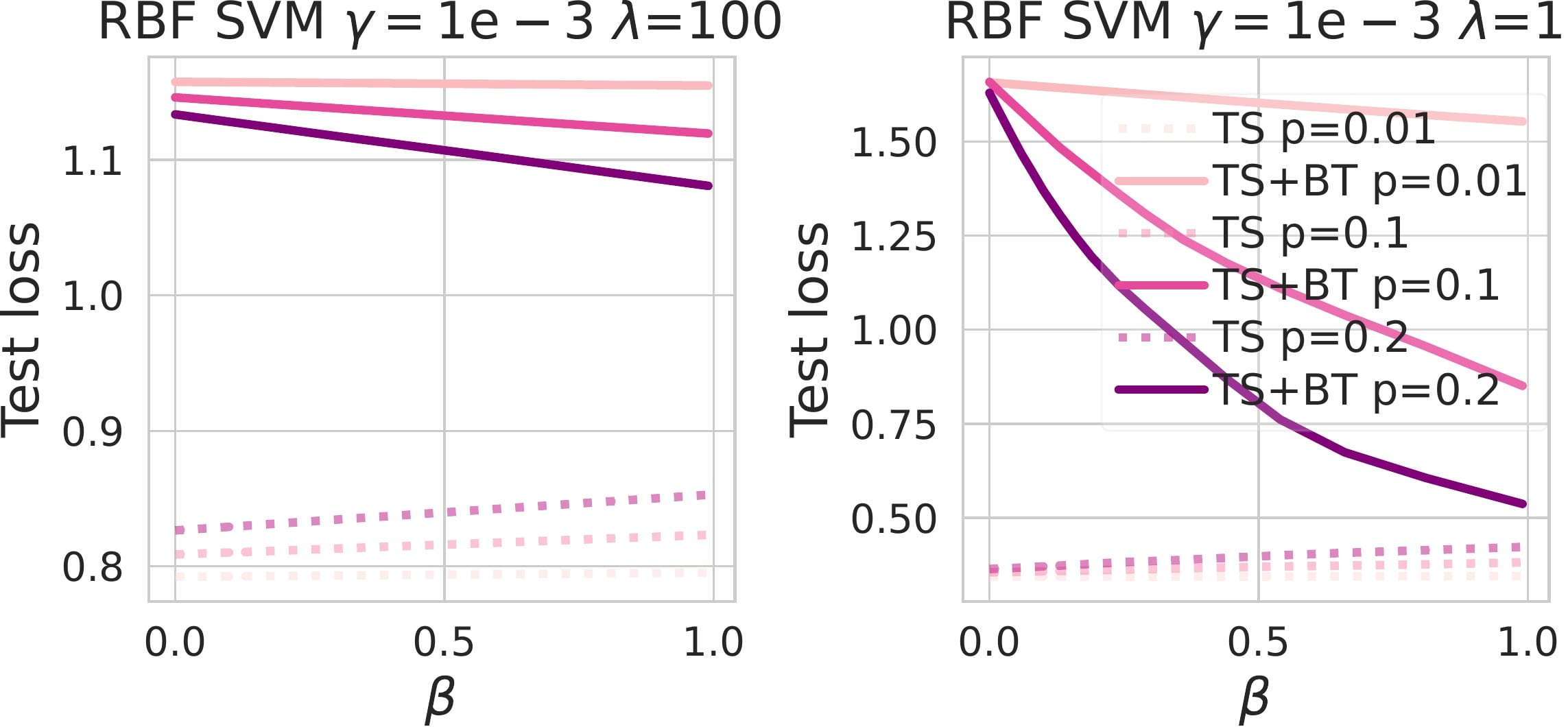}
\caption{Backdoor learning curves for different classifiers trained on  CIFAR10 \cifarbirddog. See the caption of Figure~\ref{fig:appendix_backdoor_learning_curves_mnist30} for further details.
}\label{fig:appendix_backdoor_learning_curves_cifar25}
\end{figure*}

\begin{figure*}[h!]
\centering
\includegraphics[width=0.495\textwidth]{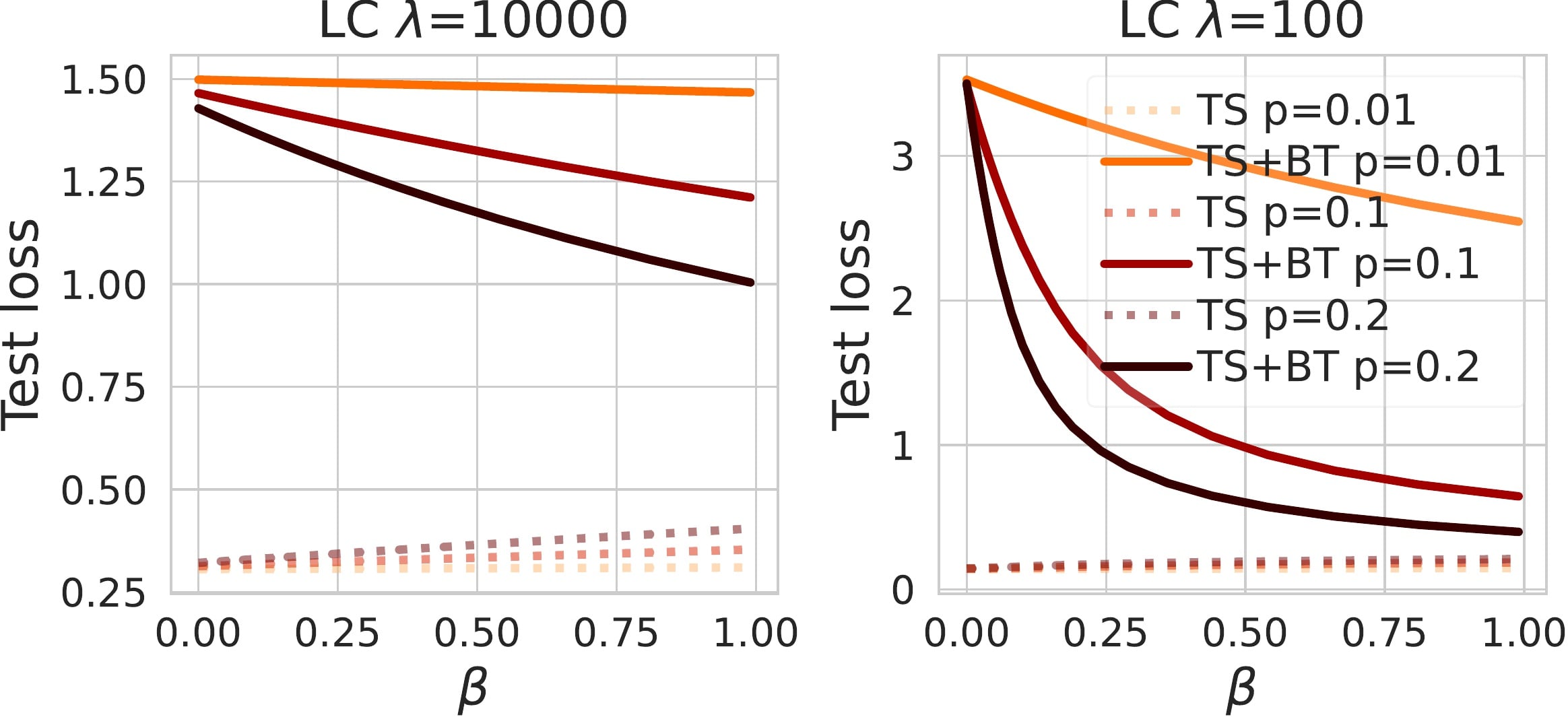}
\includegraphics[width=0.495\textwidth]{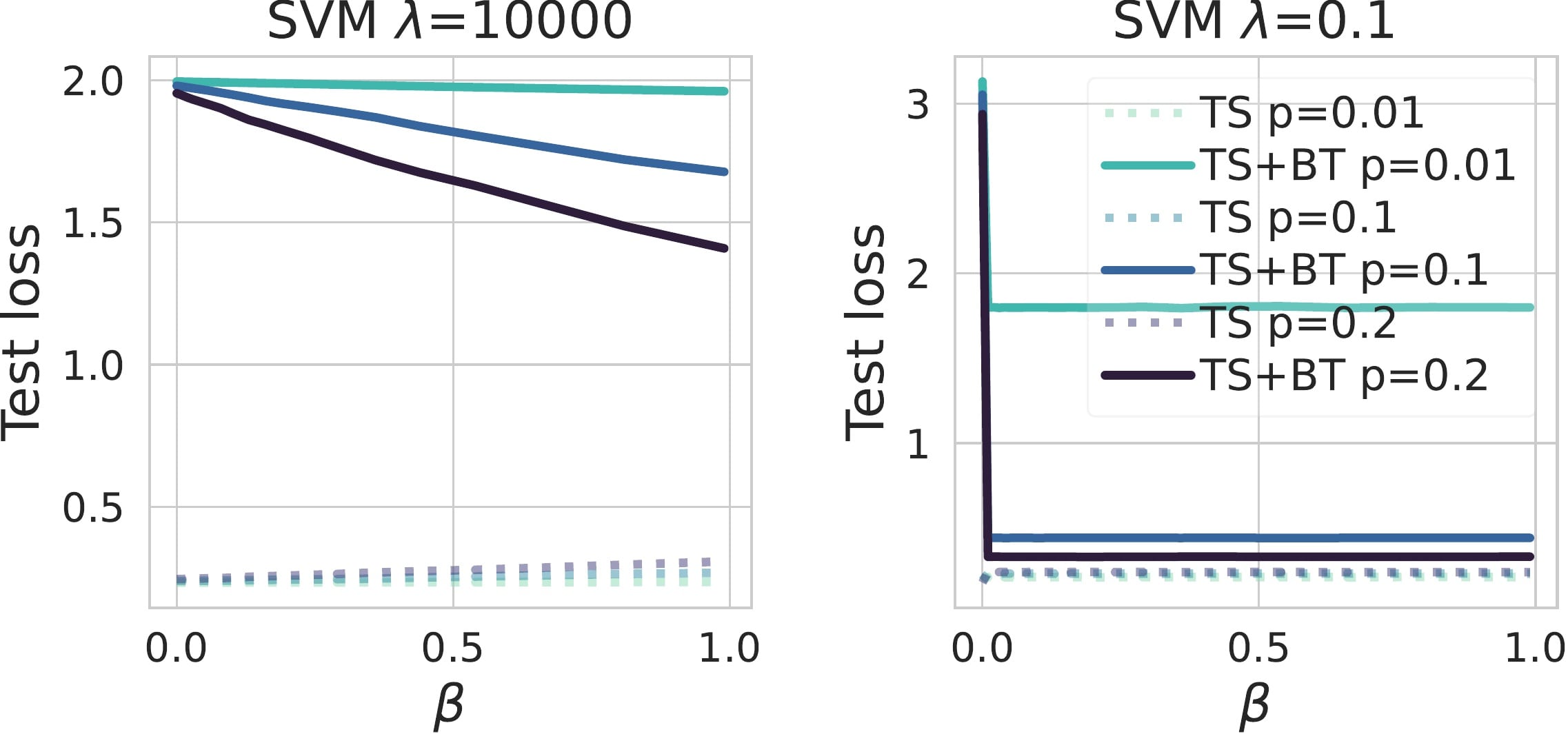}
\includegraphics[width=0.495\textwidth]{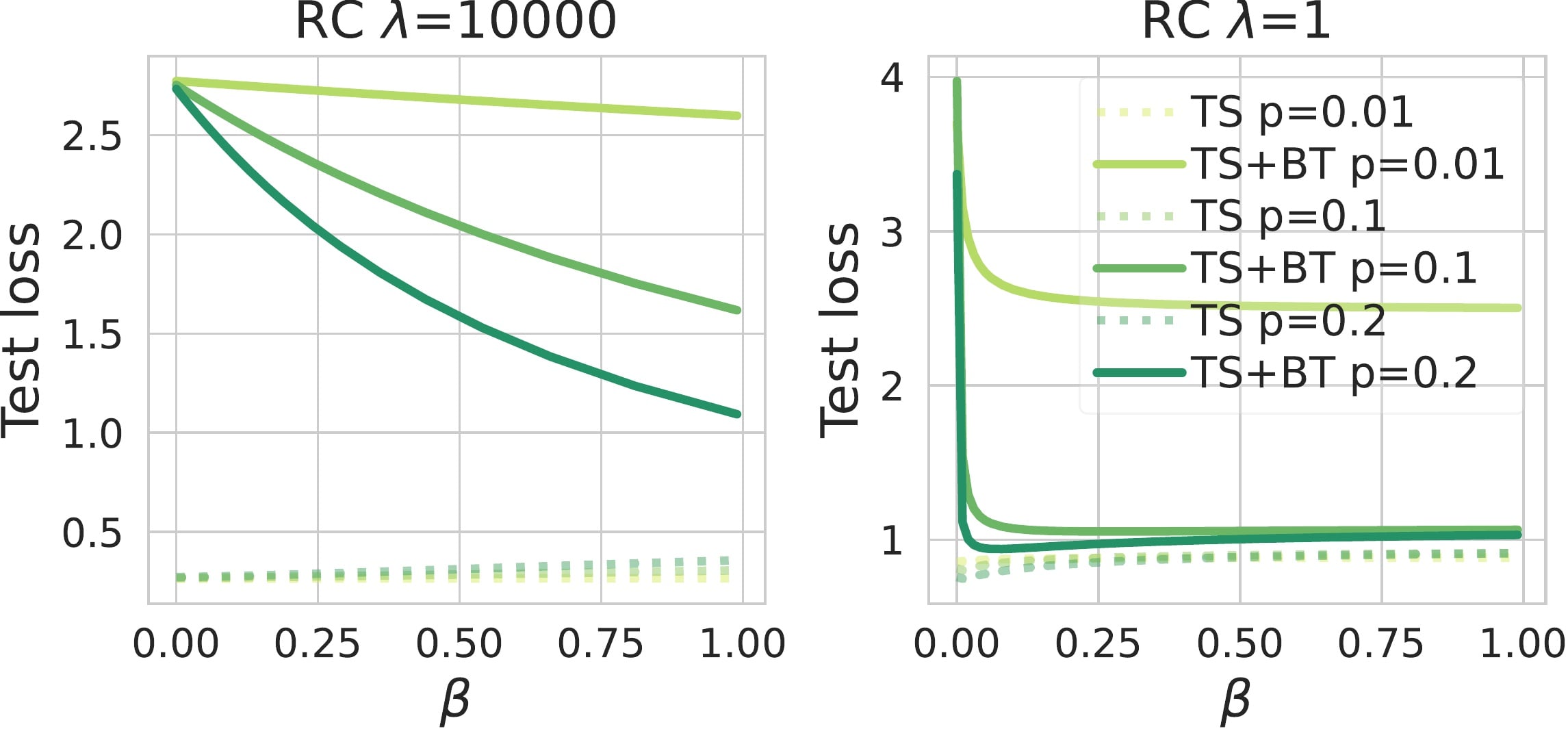}
\includegraphics[width=0.495\textwidth]{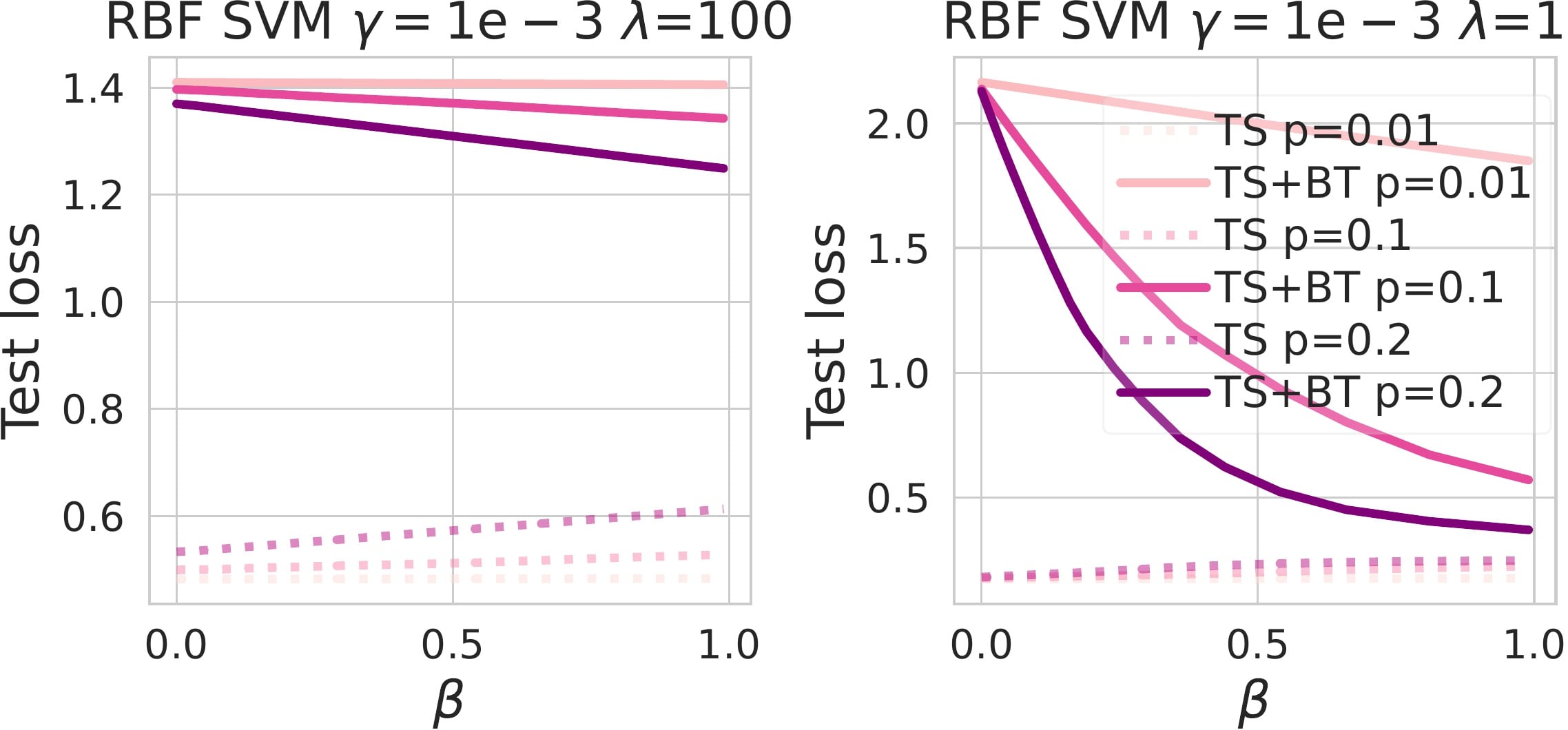}
\caption{Backdoor learning curves for different classifiers trained on  CIFAR10 \cifarairplanetruck. See the caption of Figure~\ref{fig:appendix_backdoor_learning_curves_mnist30} for further details.}\label{fig:appendix_backdoor_learning_curves_cifar90}
\end{figure*}

\begin{figure*}[h!]
\centering
\includegraphics[width=0.495\textwidth]{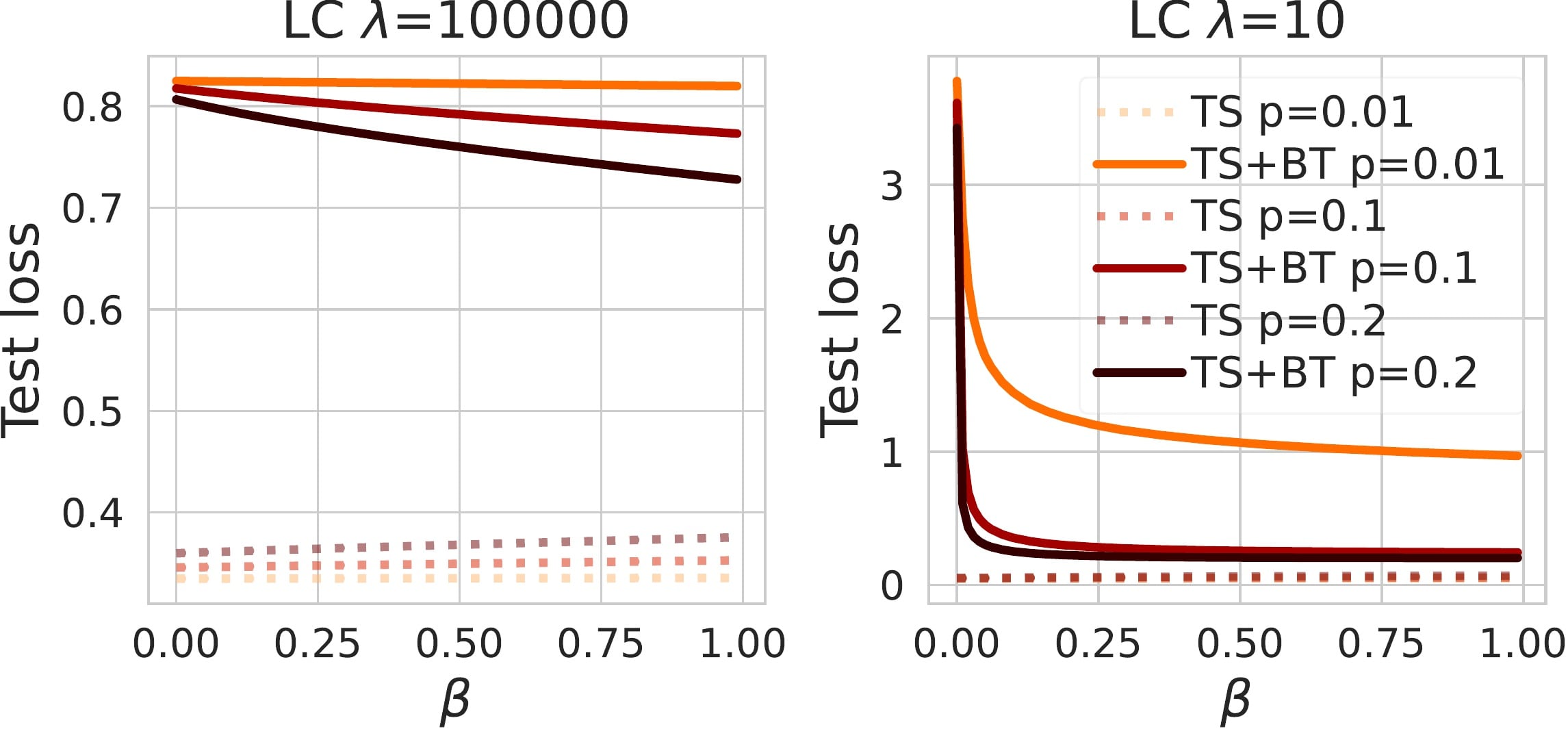}
\includegraphics[width=0.495\textwidth]{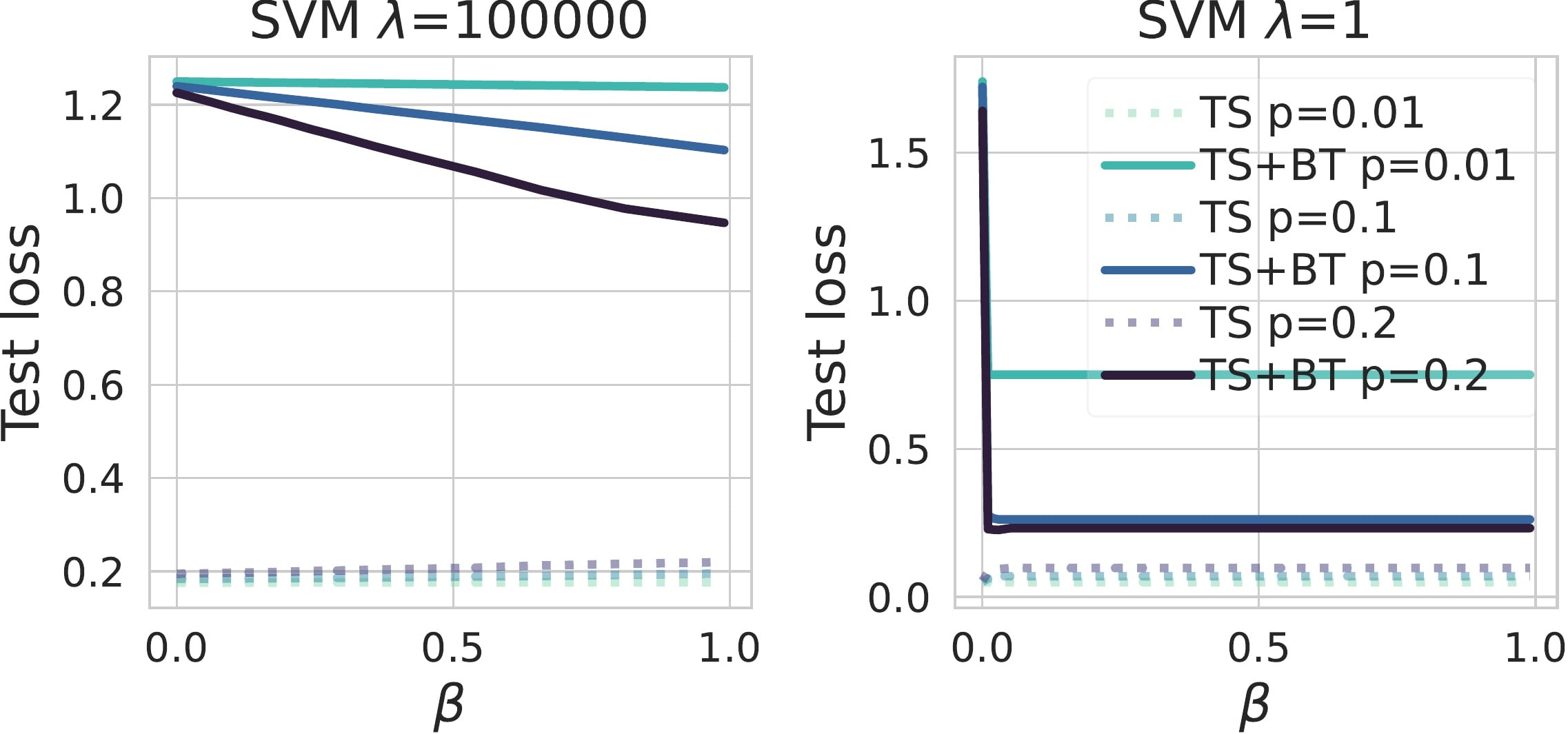}
\includegraphics[width=0.495\textwidth]{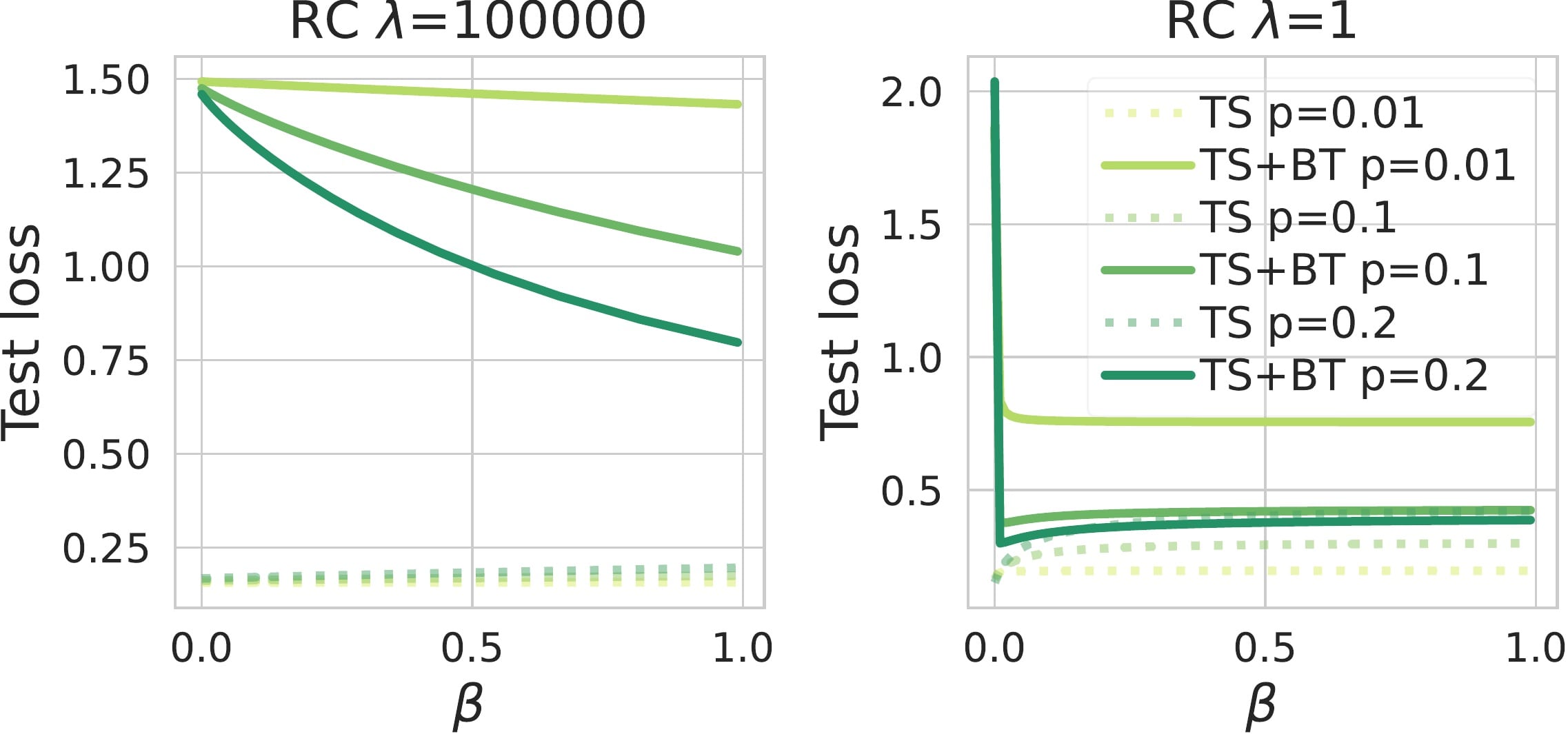}
\includegraphics[width=0.495\textwidth]{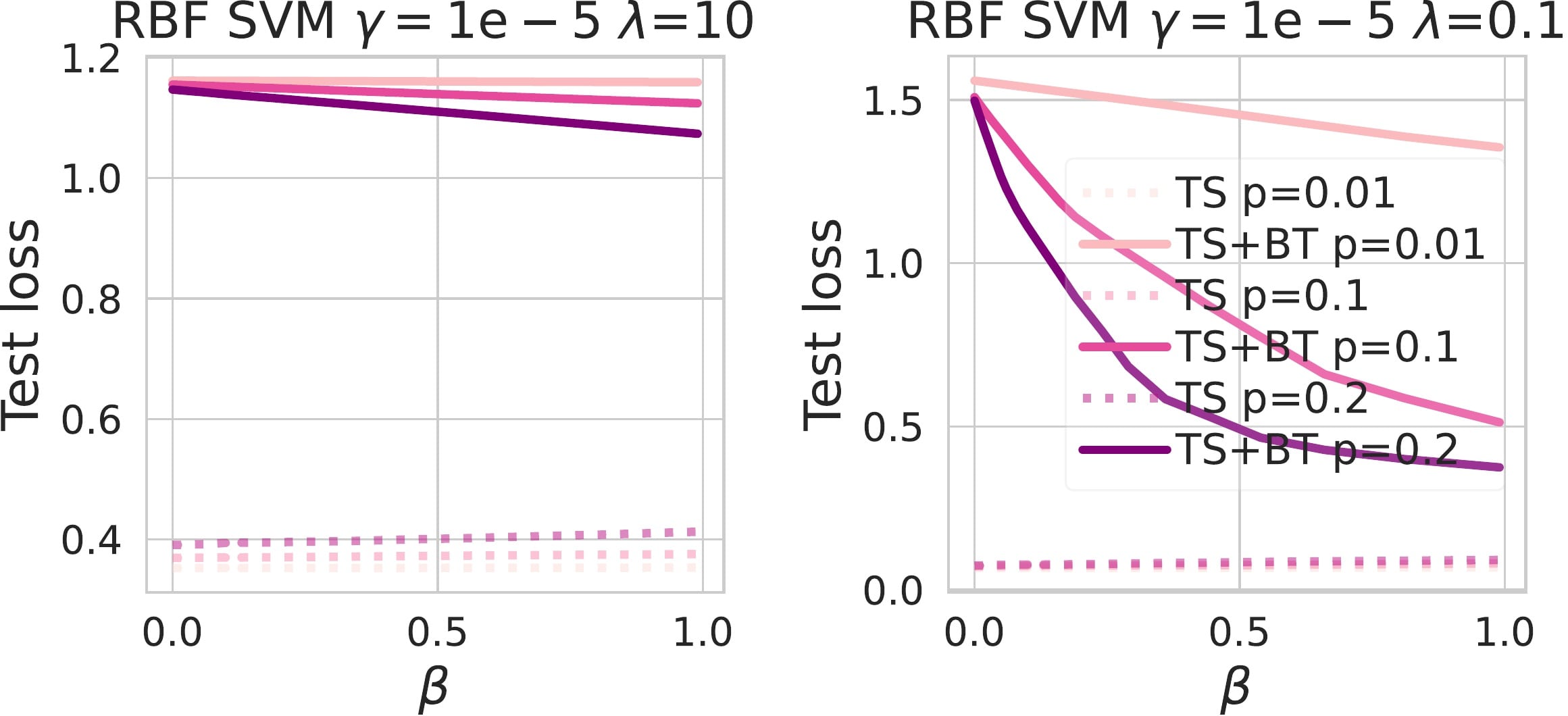}
\caption{Backdoor learning curves for different classifiers trained on Imagenette \imagenetteplayerchurch. See the caption of Figure~\ref{fig:appendix_backdoor_learning_curves_mnist30} for further details.
}
\label{fig:appendix_backdoor_learning_curves_imagenette09}
\end{figure*}

\begin{figure*}[h!]
\centering
\includegraphics[width=0.495\textwidth]{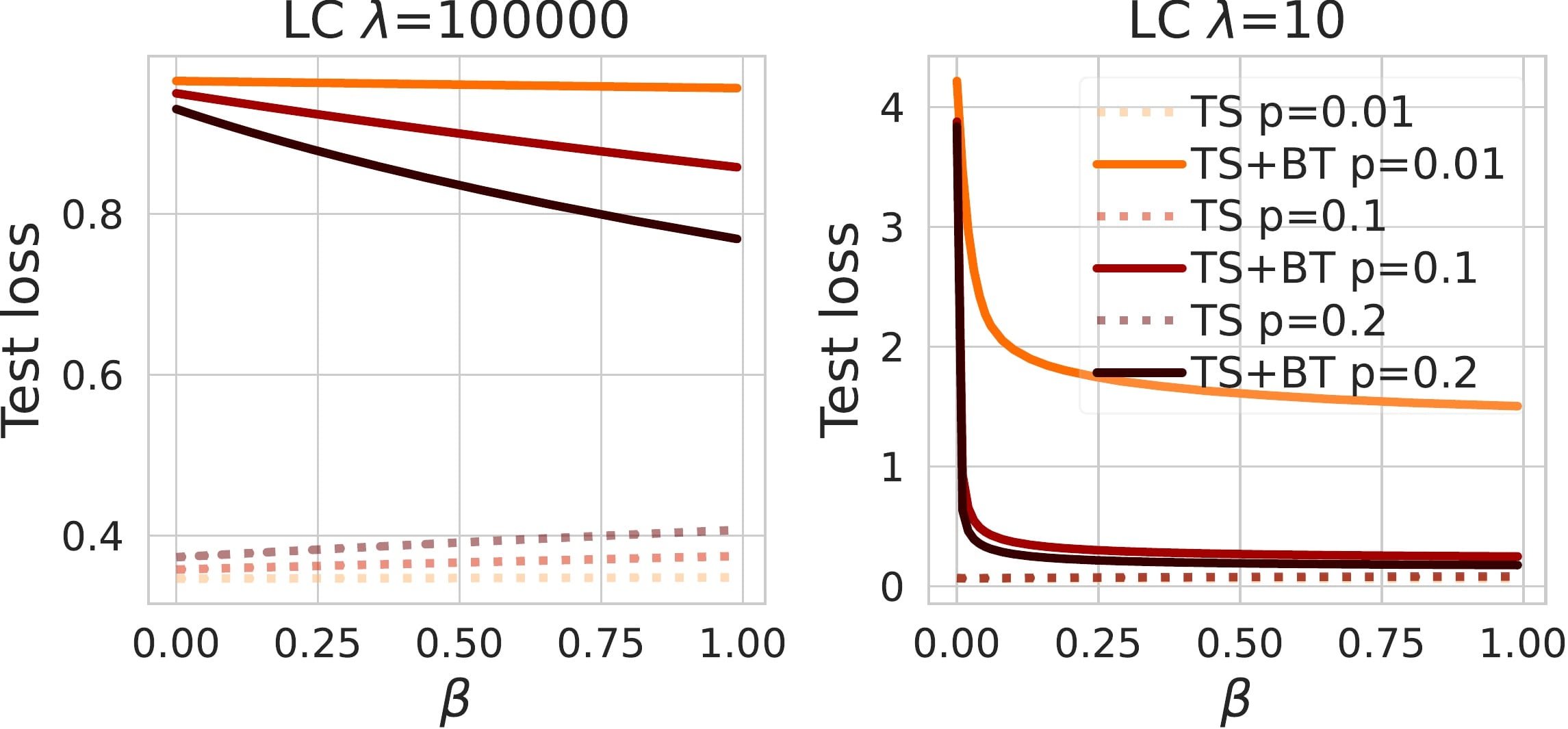}
\includegraphics[width=0.495\textwidth]{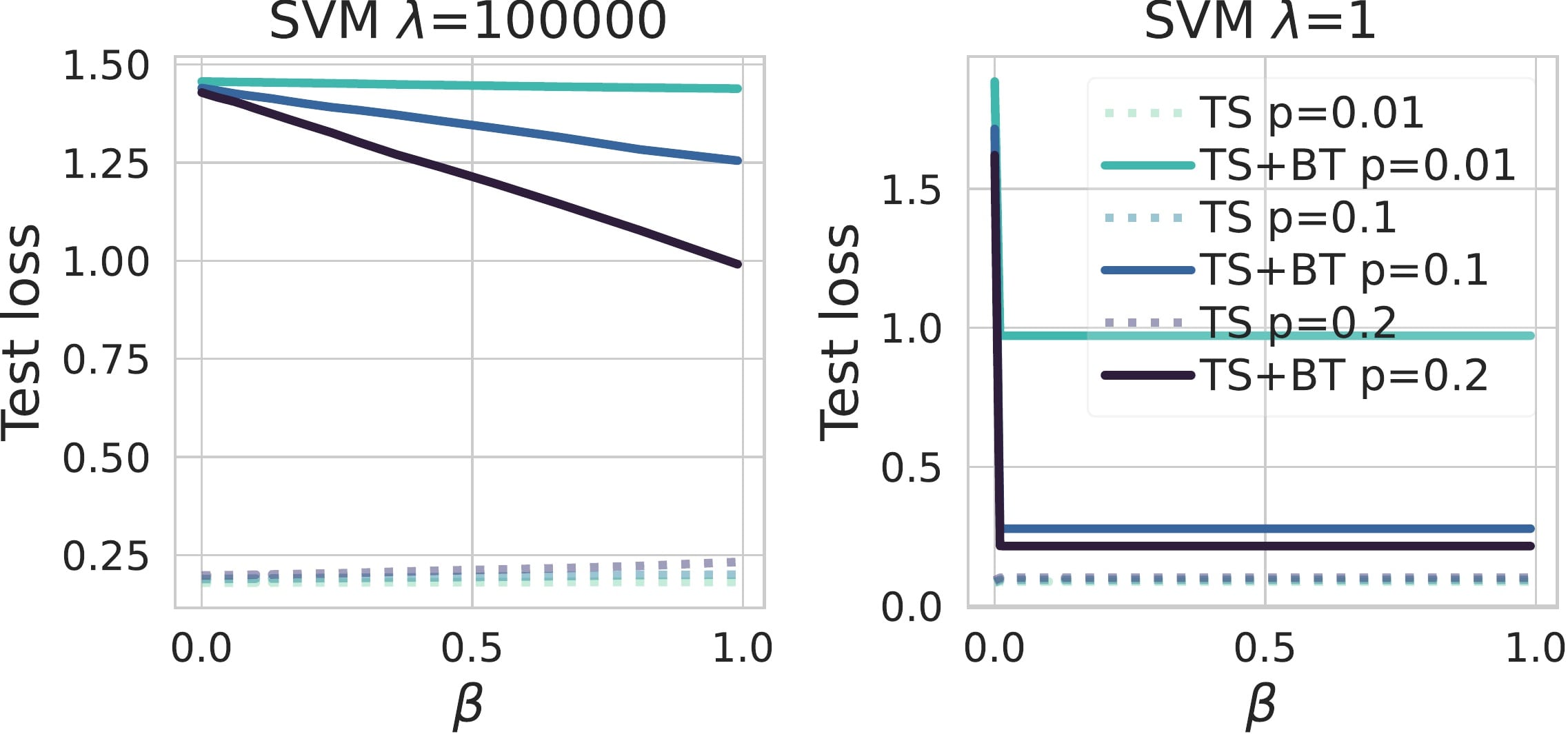}
\includegraphics[width=0.495\textwidth]{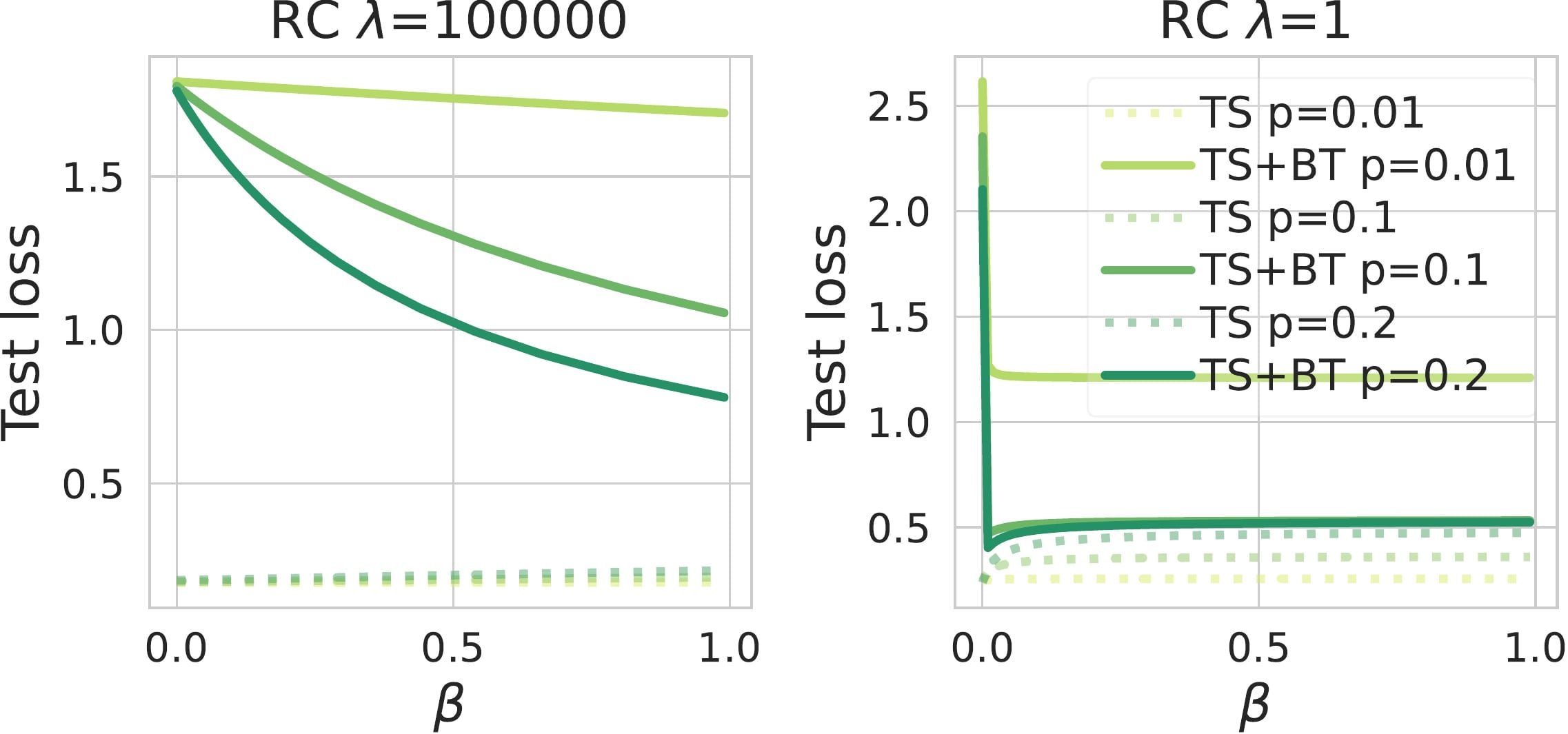}
\includegraphics[width=0.495\textwidth]{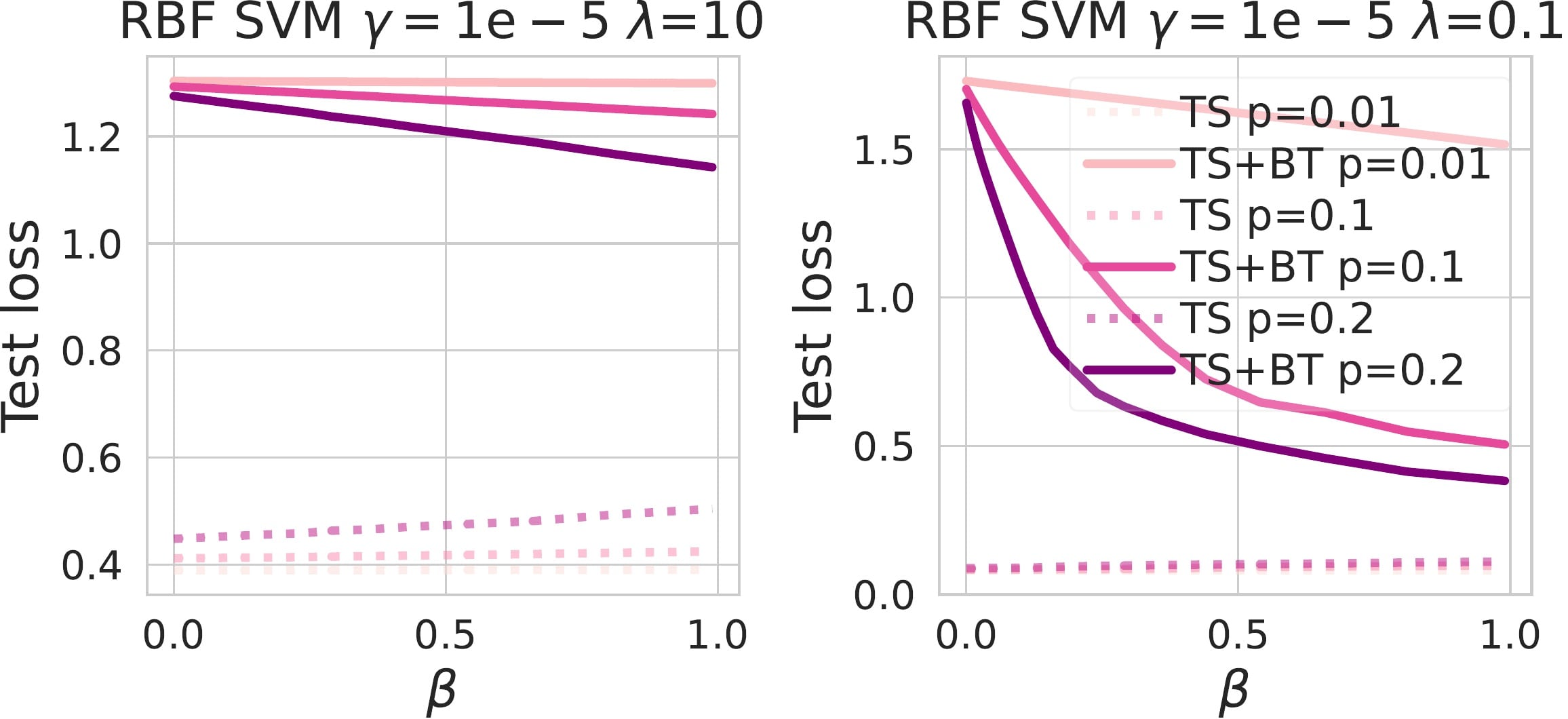}
\caption{Backdoor learning curves for different classifiers trained on Imagenette \imagenettetenchparachute. See the caption of Figure~\ref{fig:appendix_backdoor_learning_curves_mnist30} for further details.
}
\label{fig:appendix_backdoor_learning_curves_imagenette25}
\end{figure*}

\begin{figure*}[t]
  \centering
\includegraphics[width=0.242\textwidth]{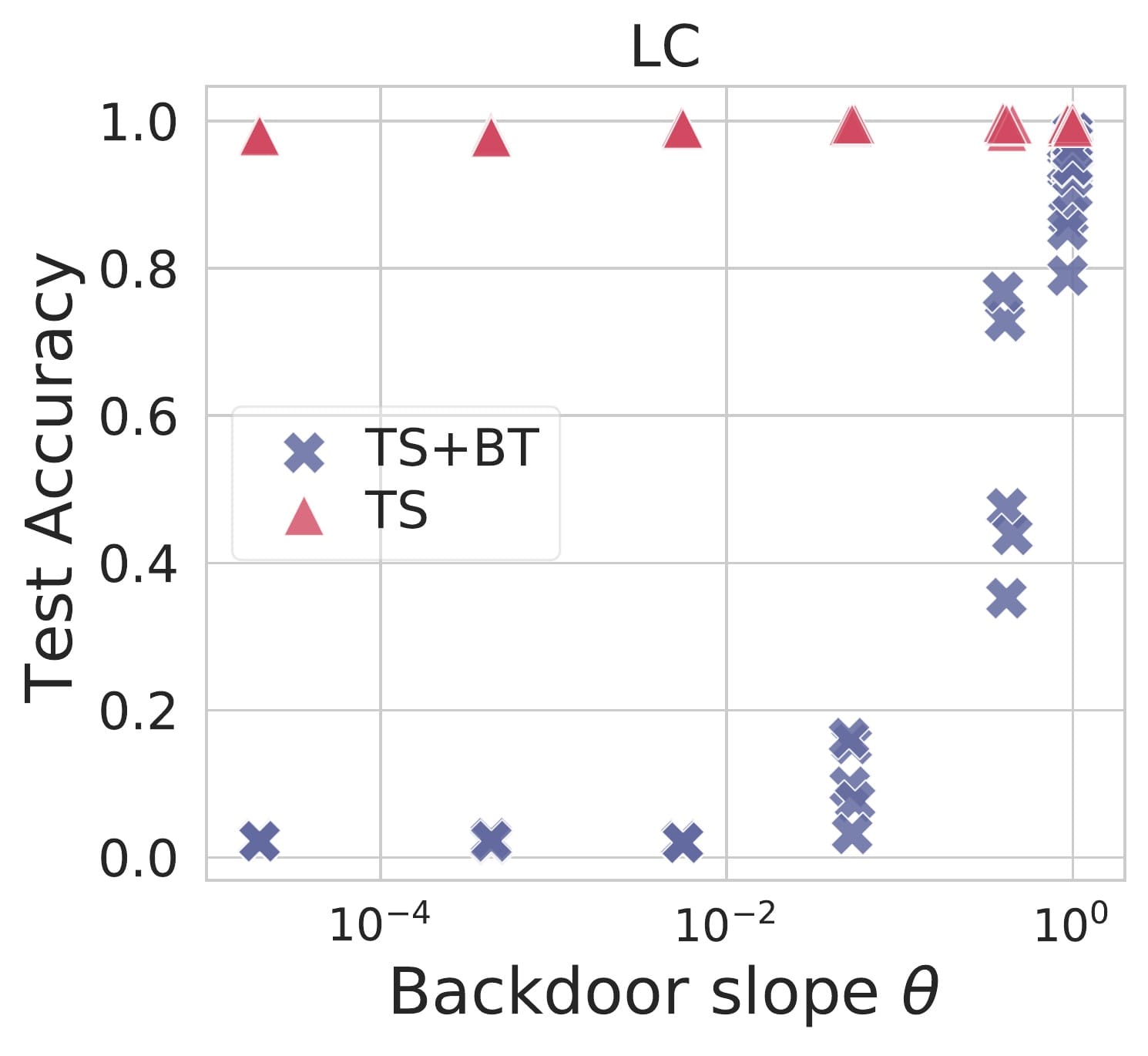}
\includegraphics[width=0.242\textwidth]{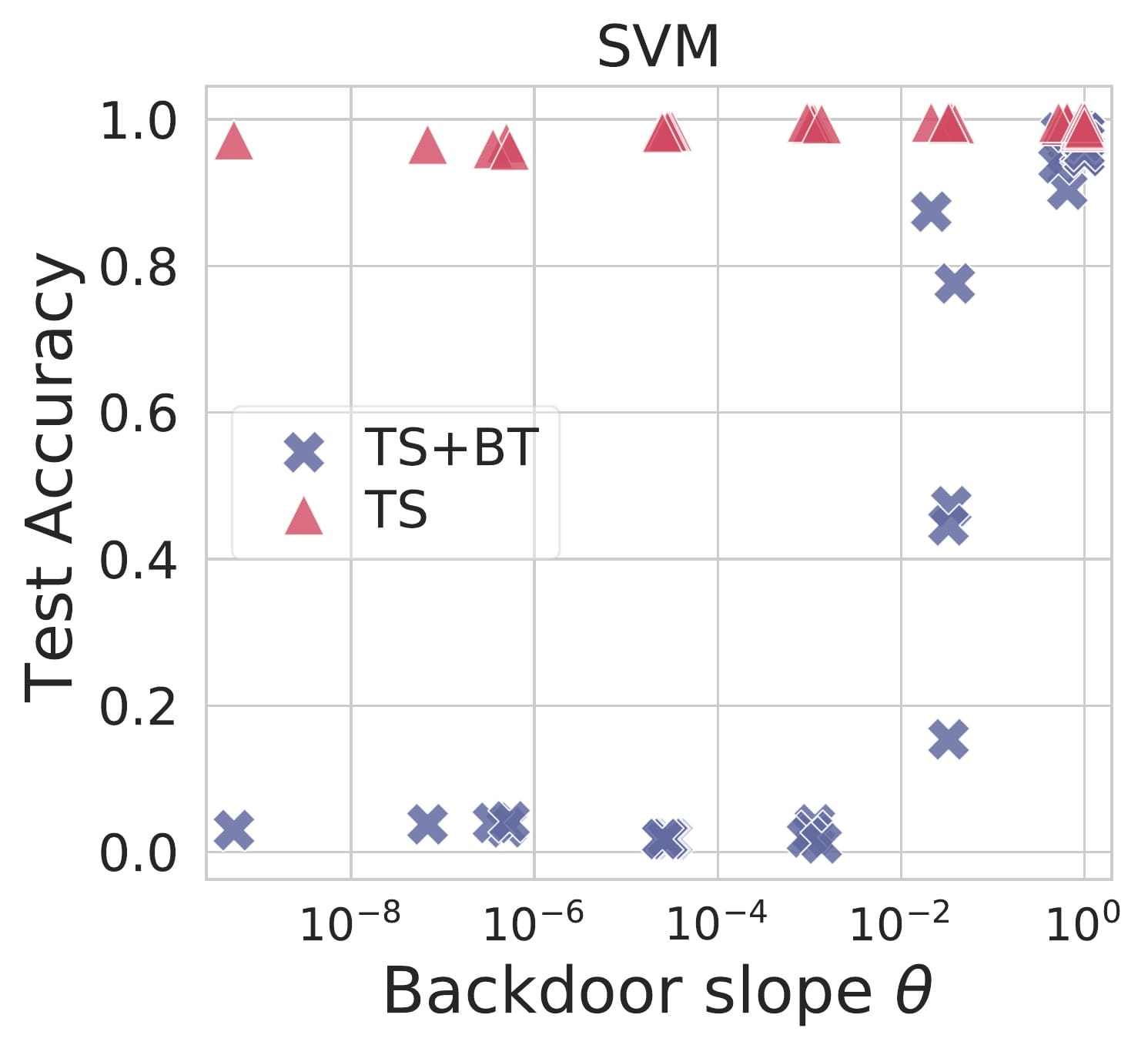}
\includegraphics[width=0.242\textwidth]{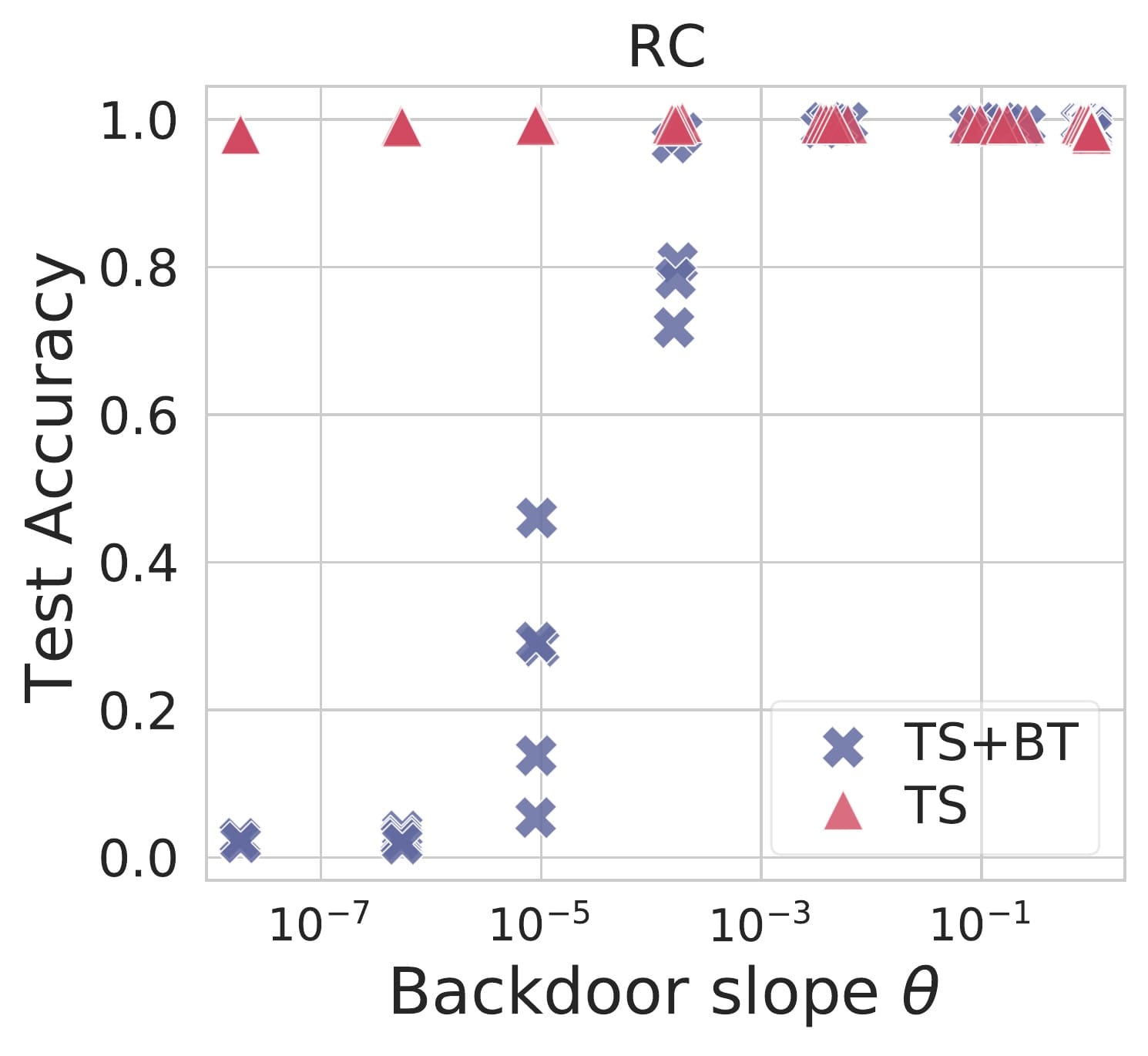}
\includegraphics[width=0.242\textwidth]{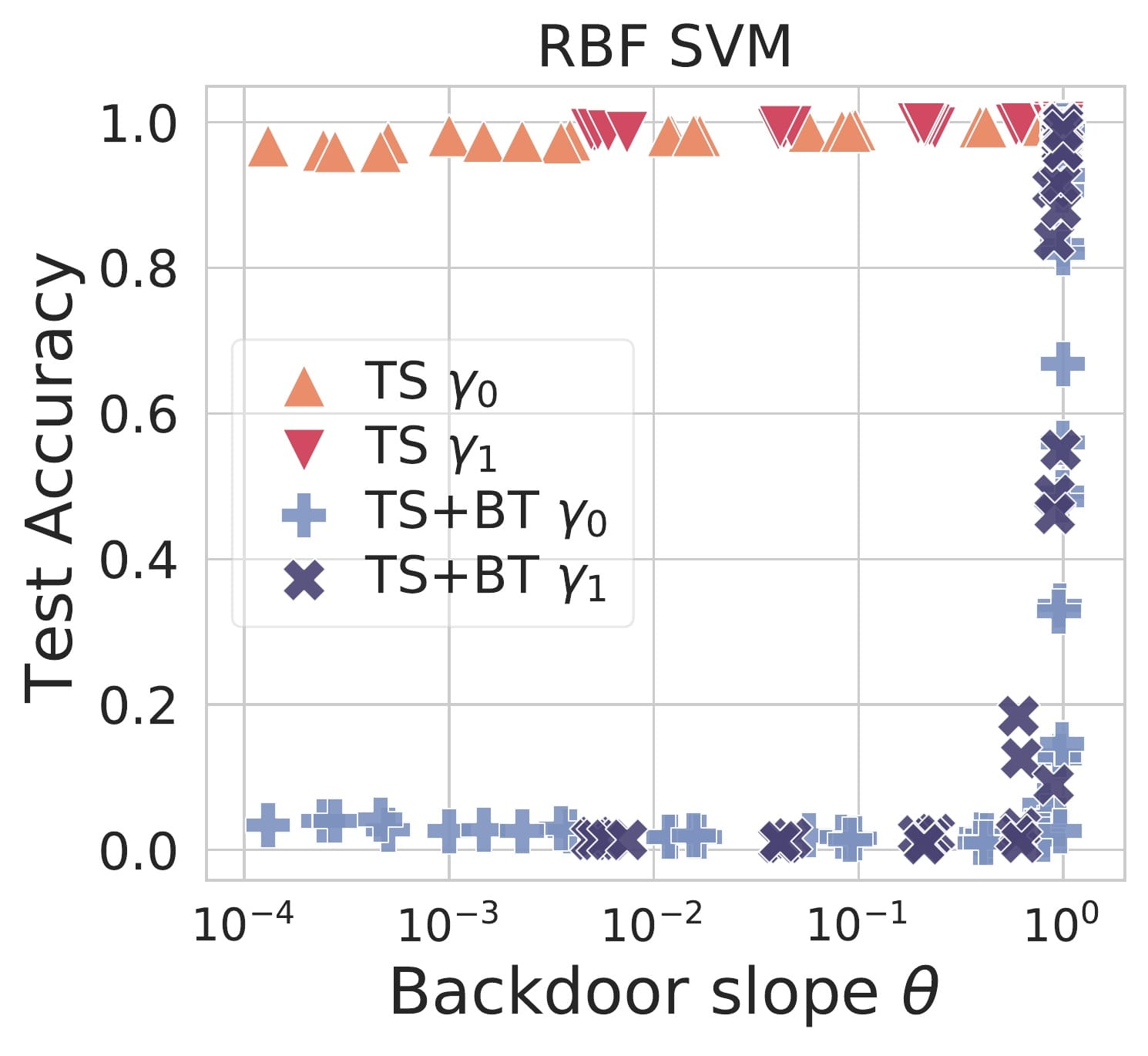}

\includegraphics[width=0.242\textwidth]{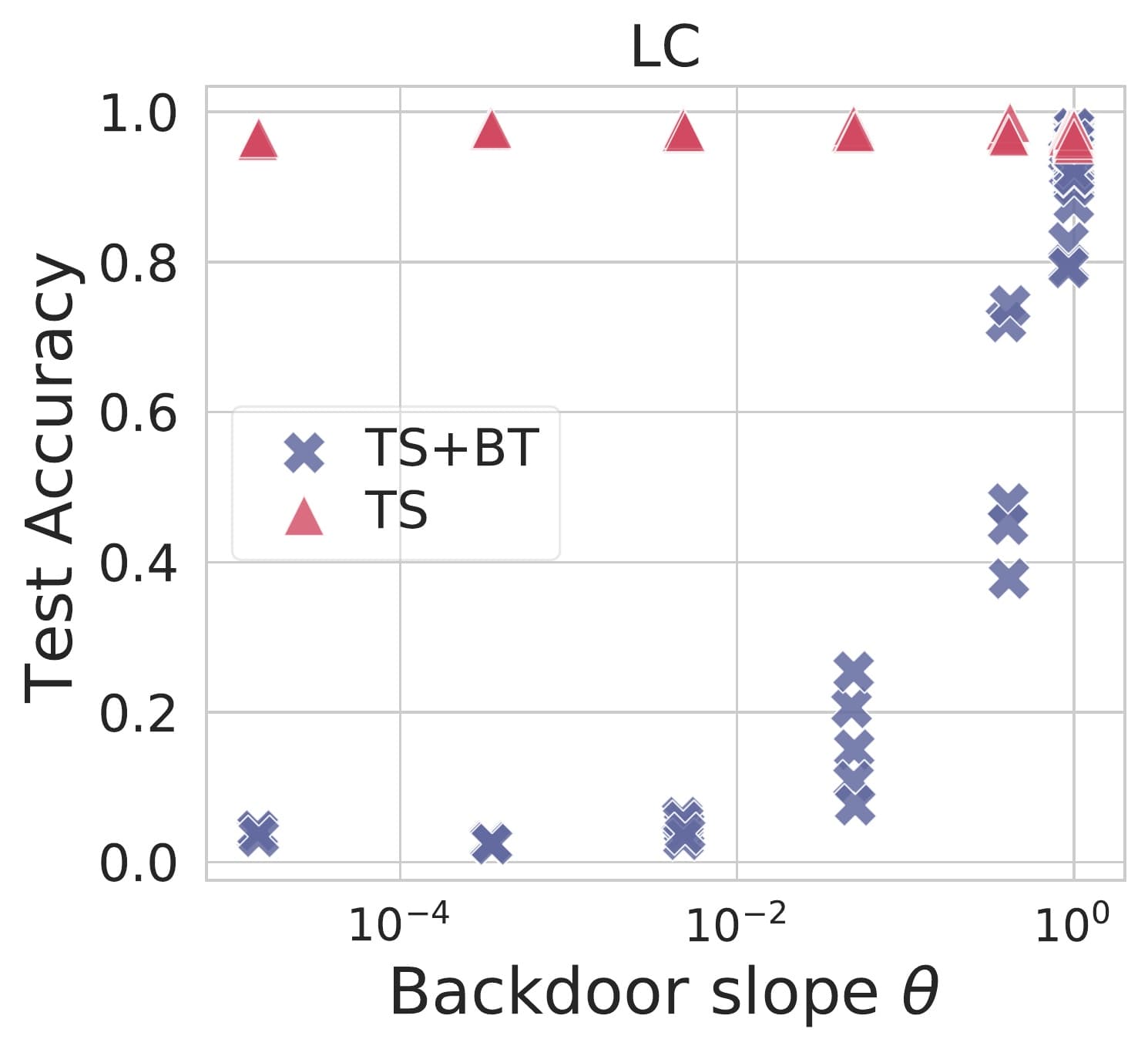}
\includegraphics[width=0.242\textwidth]{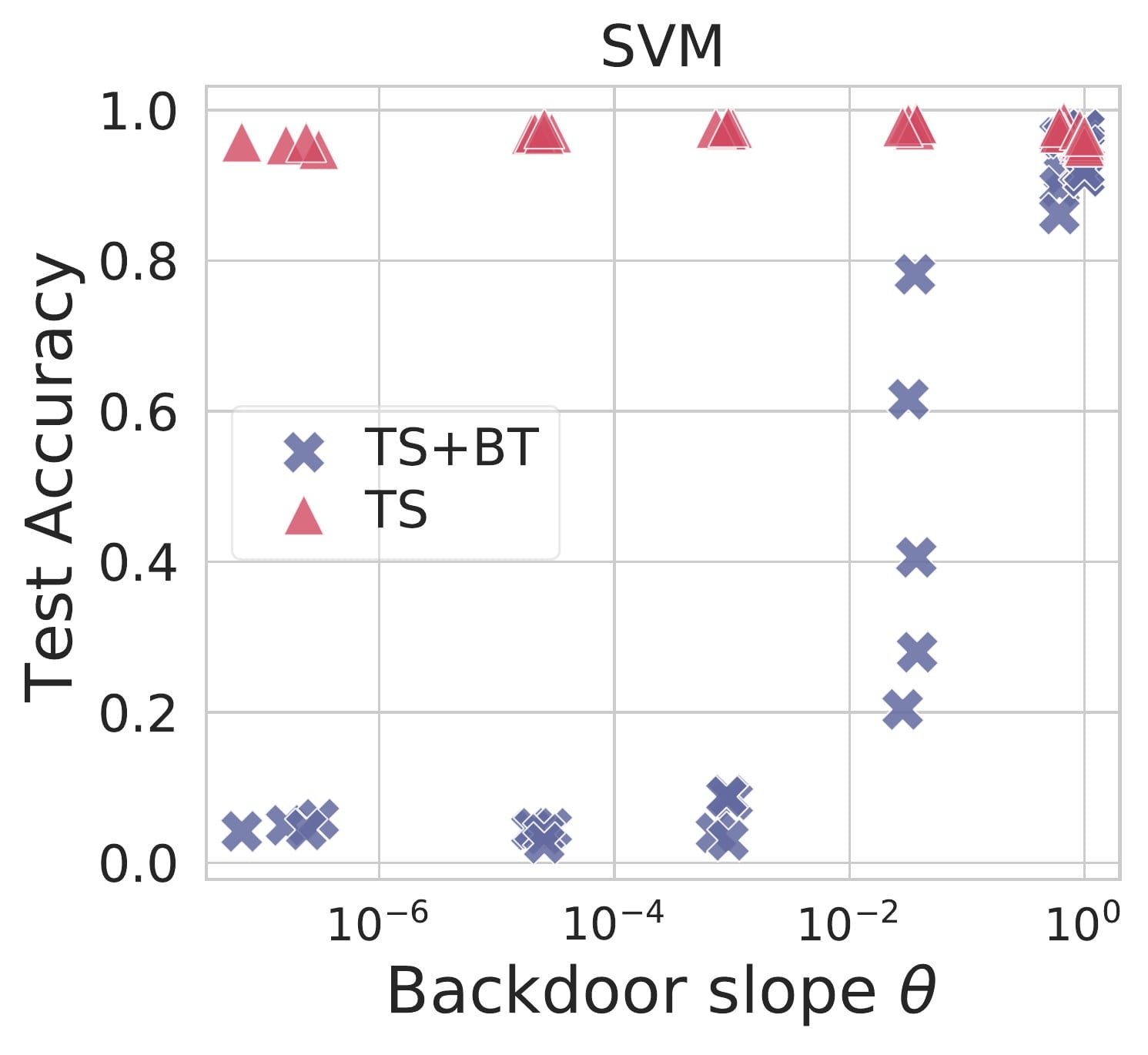}
\includegraphics[width=0.242\textwidth]{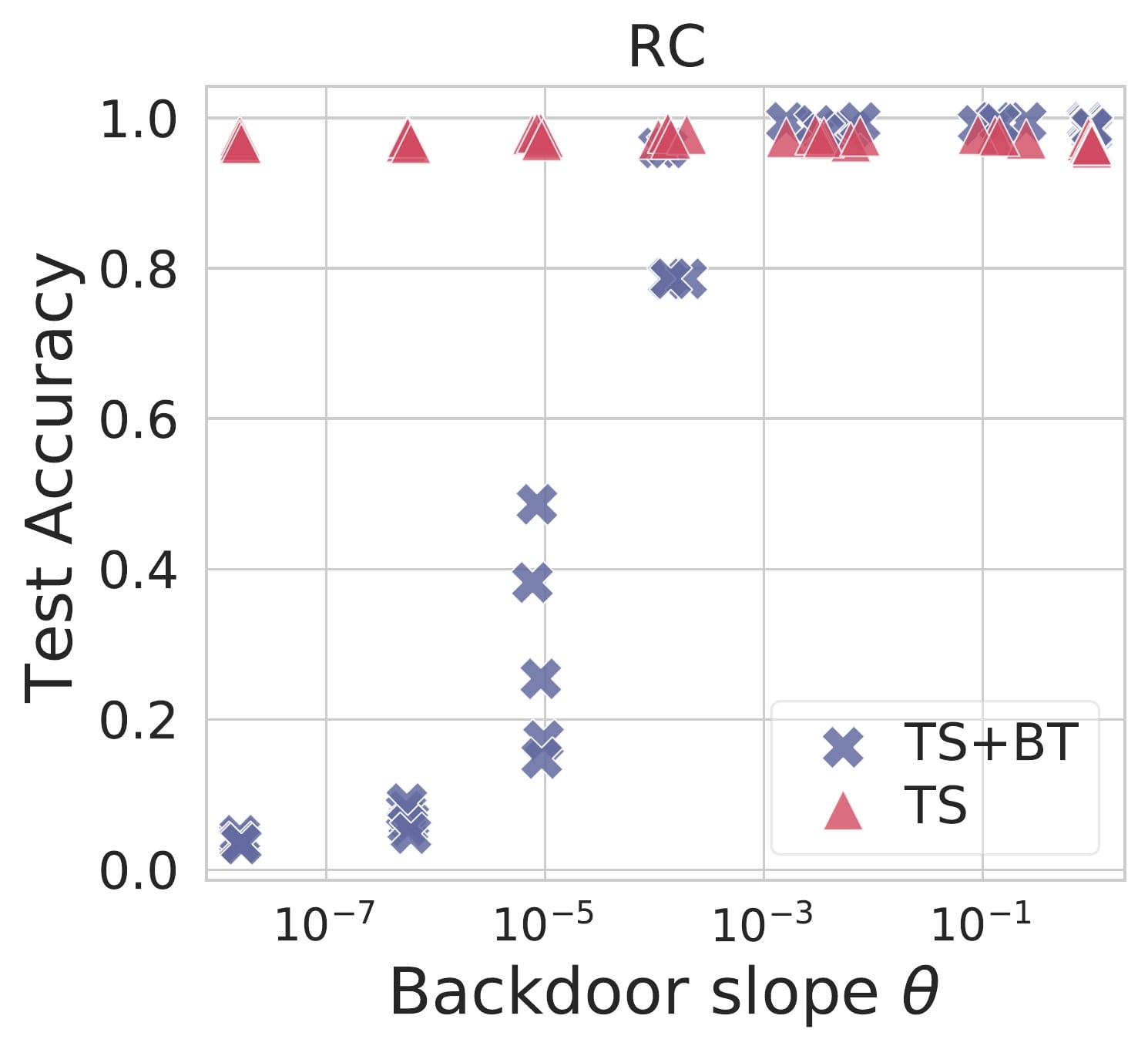}
\includegraphics[width=0.242\textwidth]{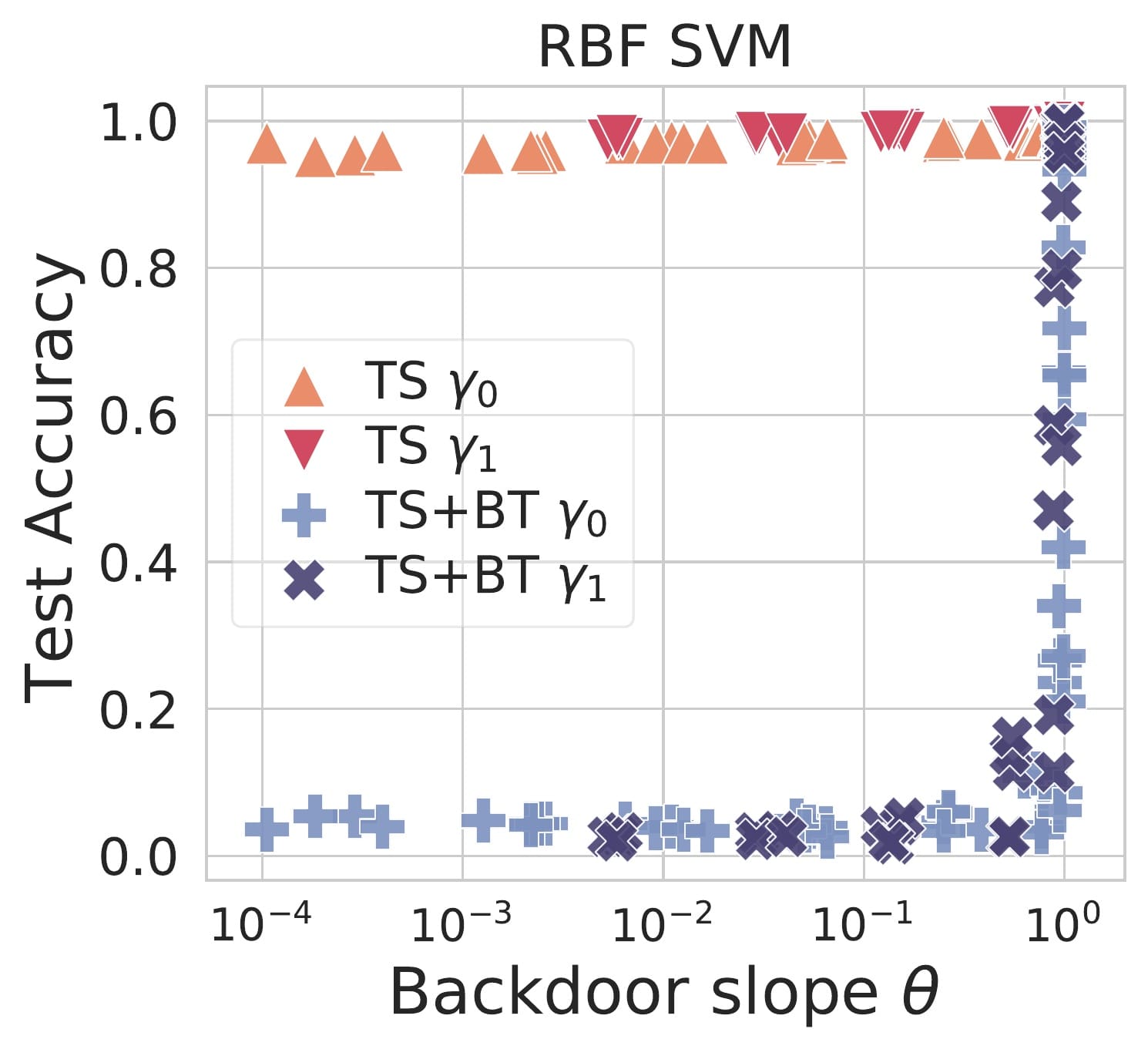}
 \caption{Backdoor slope $\theta$ vs clean accuracy (red) and backdoor effectiveness (blue) on MNIST 3vs.0 (top row) and 5vs.2 (bottom row). We measure the classification accuracy on the untainted test samples (TS), and on the same samples after adding the  $3\times 3$ backdoor trigger (TS+BT). We chose the $\gamma$ parameter for the RBF kernel as $\gamma_0=\expnumber{5}{-04}$ (orange triangle for clean data, light blue plus for data with trigger) and $\gamma_1=\expnumber{5}{-03}$ (red inverted triangle for clean data, dark blue x for data with trigger).}
  \label{fig:supplementary_resultsMNIST}
\end{figure*}

\begin{figure*}[h!]
  \centering
      \includegraphics[width=0.242\textwidth]{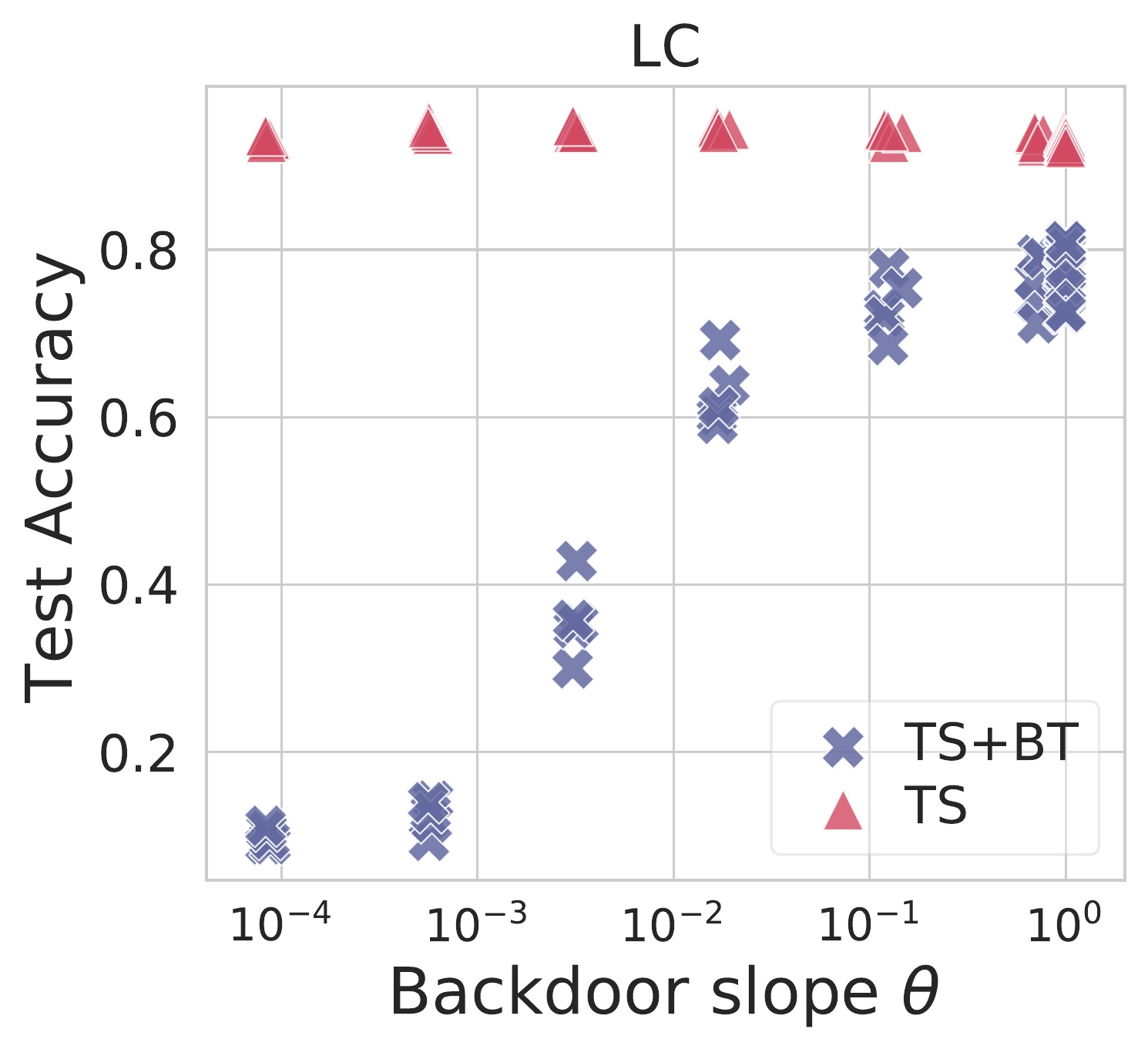}
    \includegraphics[width=0.242\textwidth]{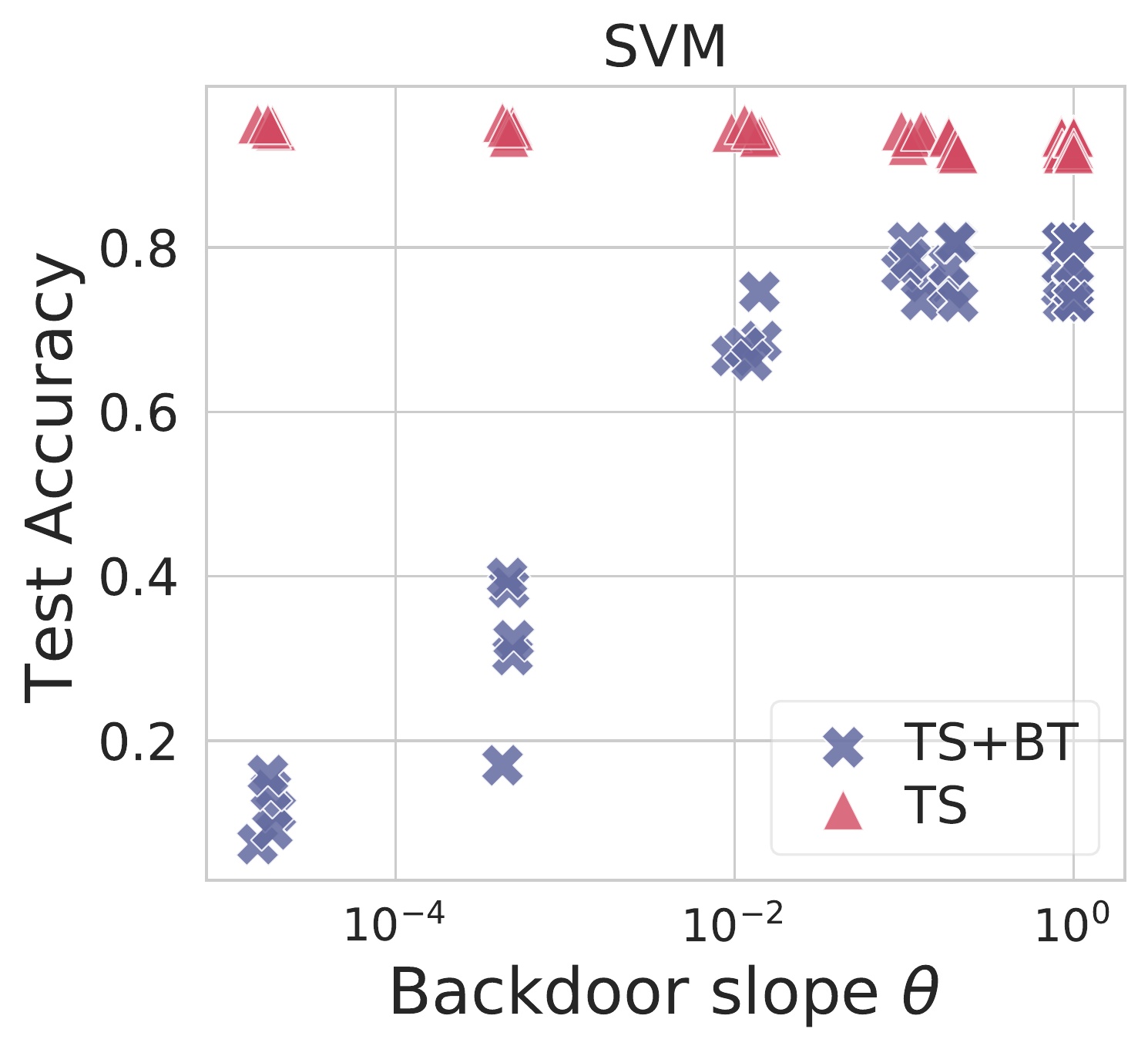}
    \includegraphics[width=0.242\textwidth]{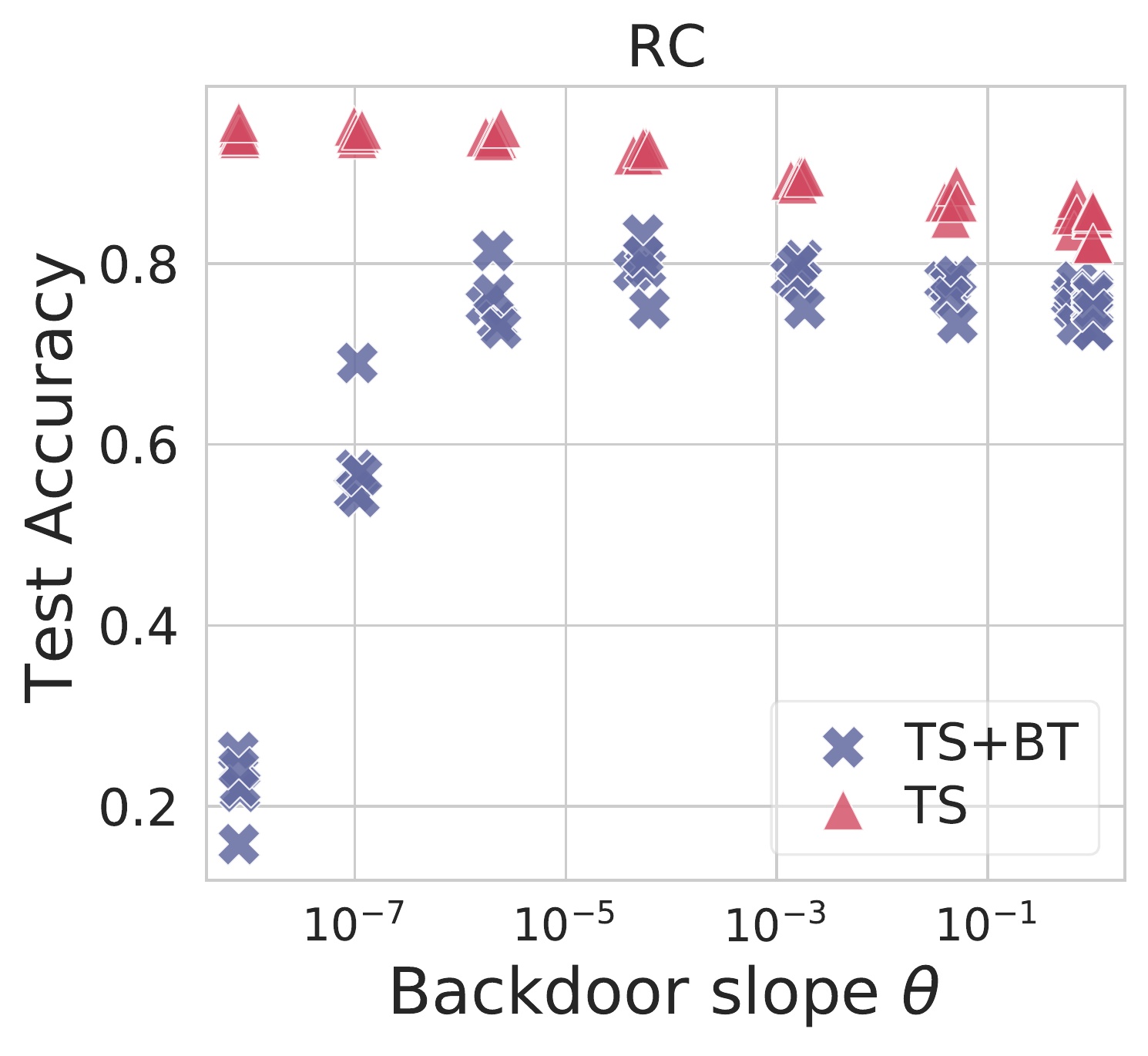}
    \includegraphics[width=0.242\textwidth]{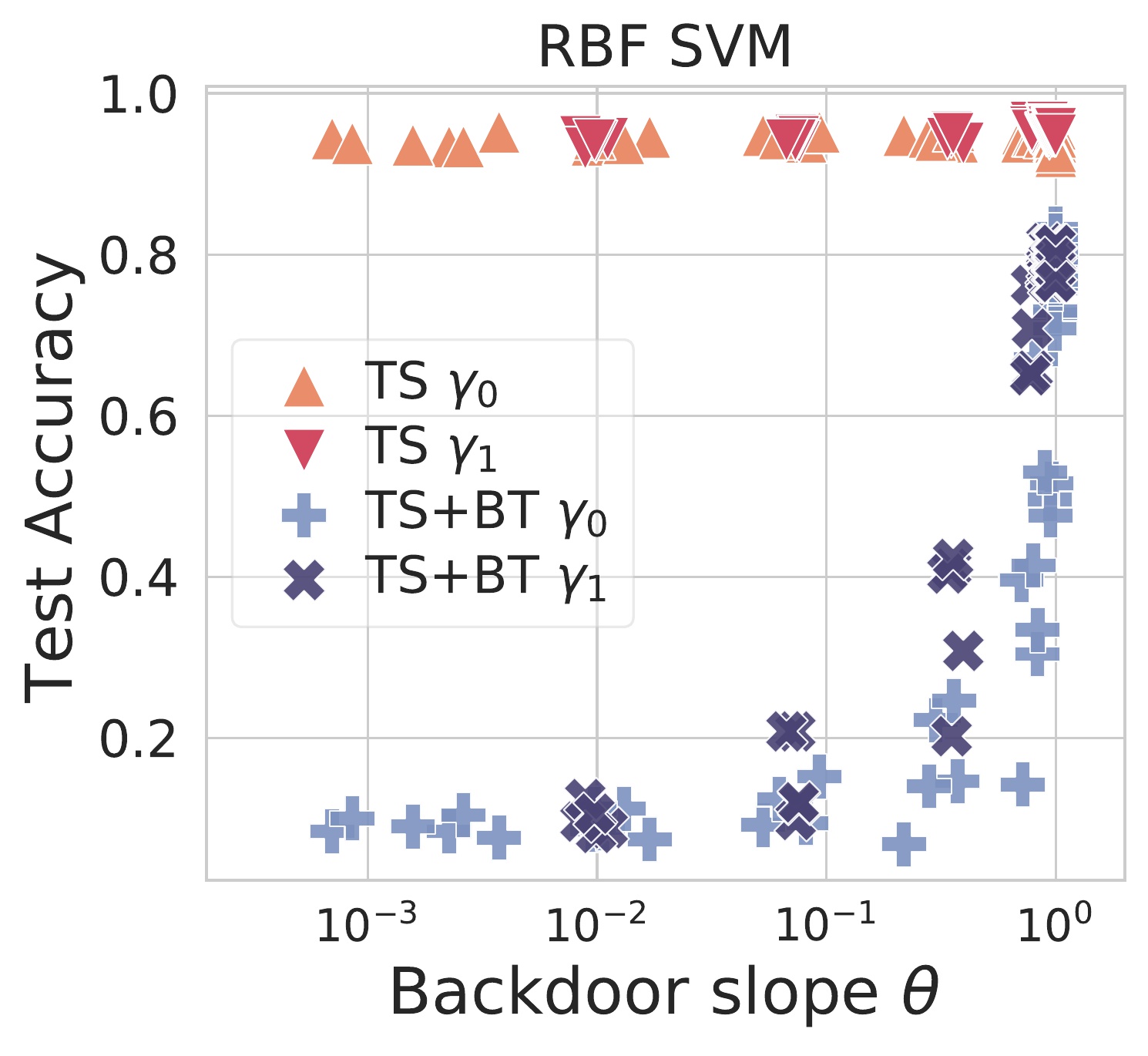}

    \includegraphics[width=0.242\textwidth]{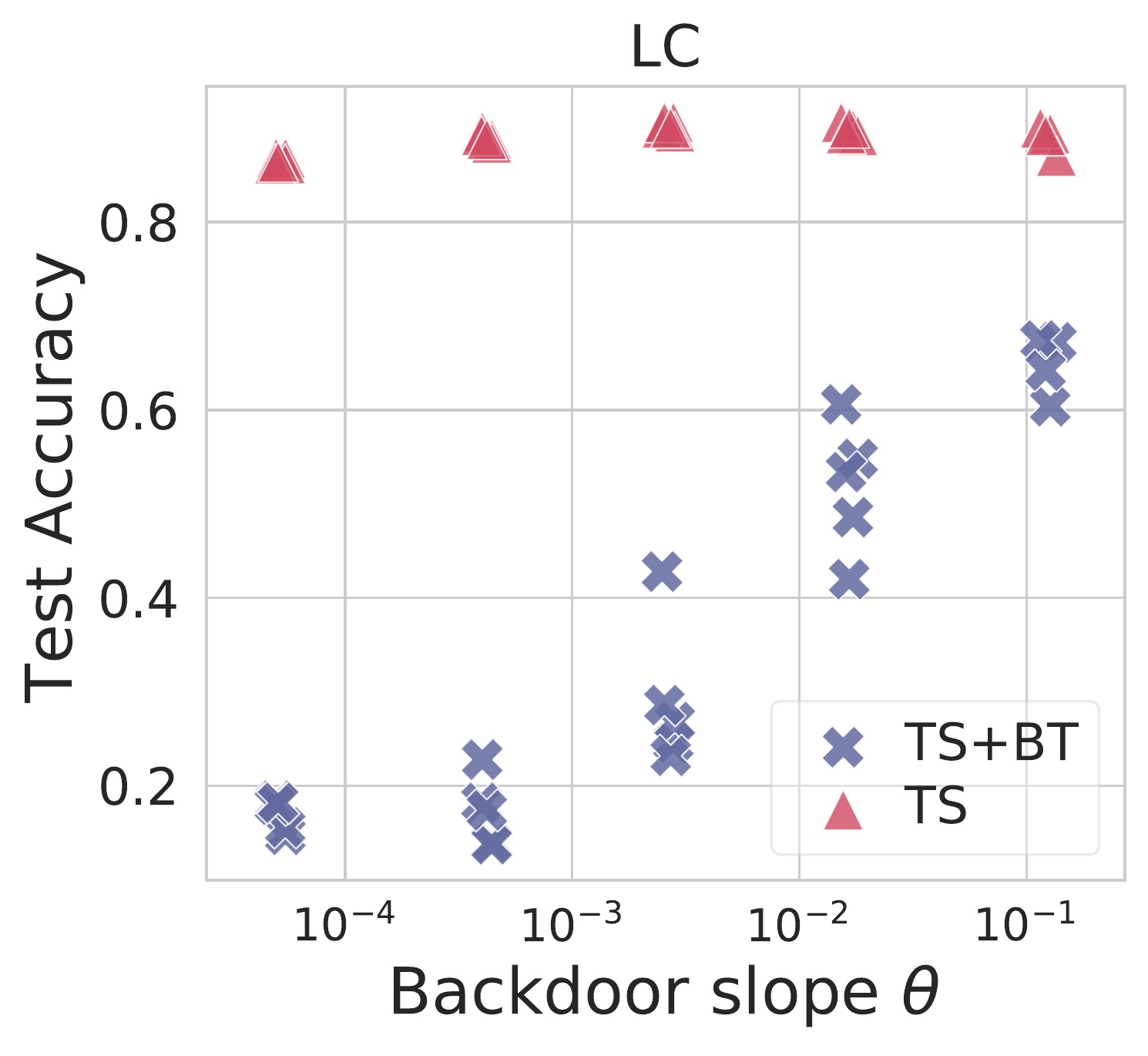}
    \includegraphics[width=0.242\textwidth]{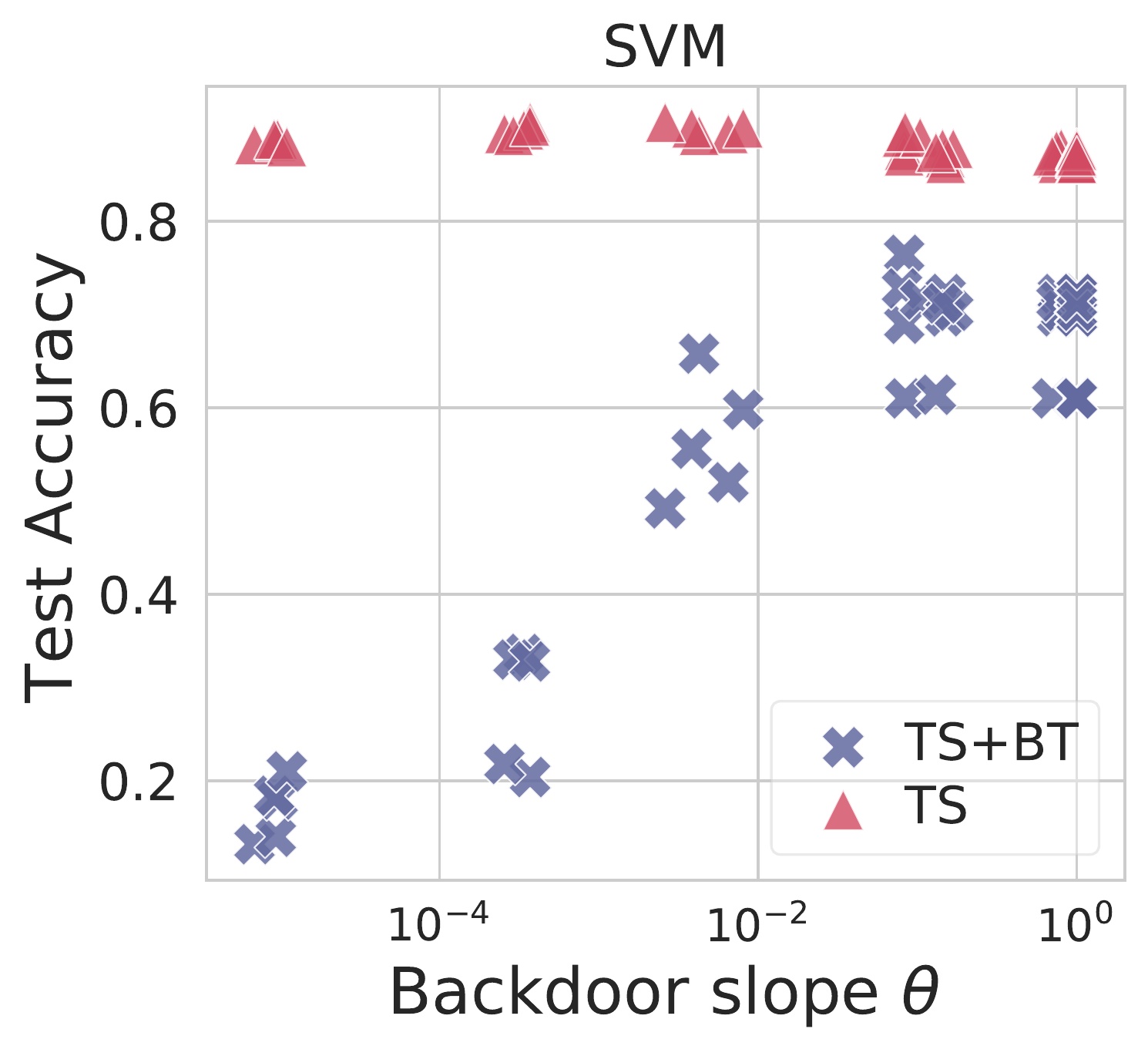}
    \includegraphics[width=0.242\textwidth]{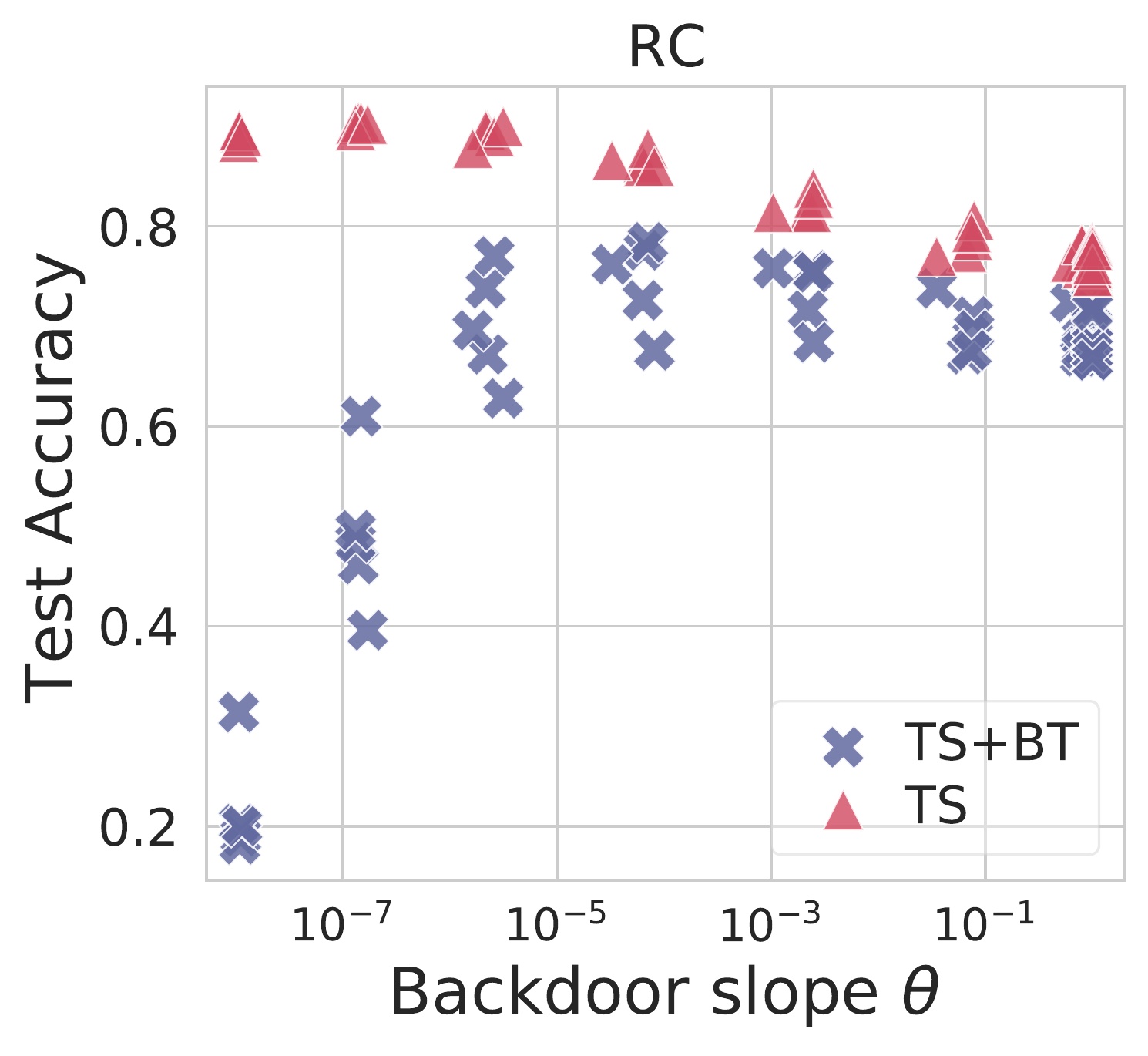}
    \includegraphics[width=0.242\textwidth]{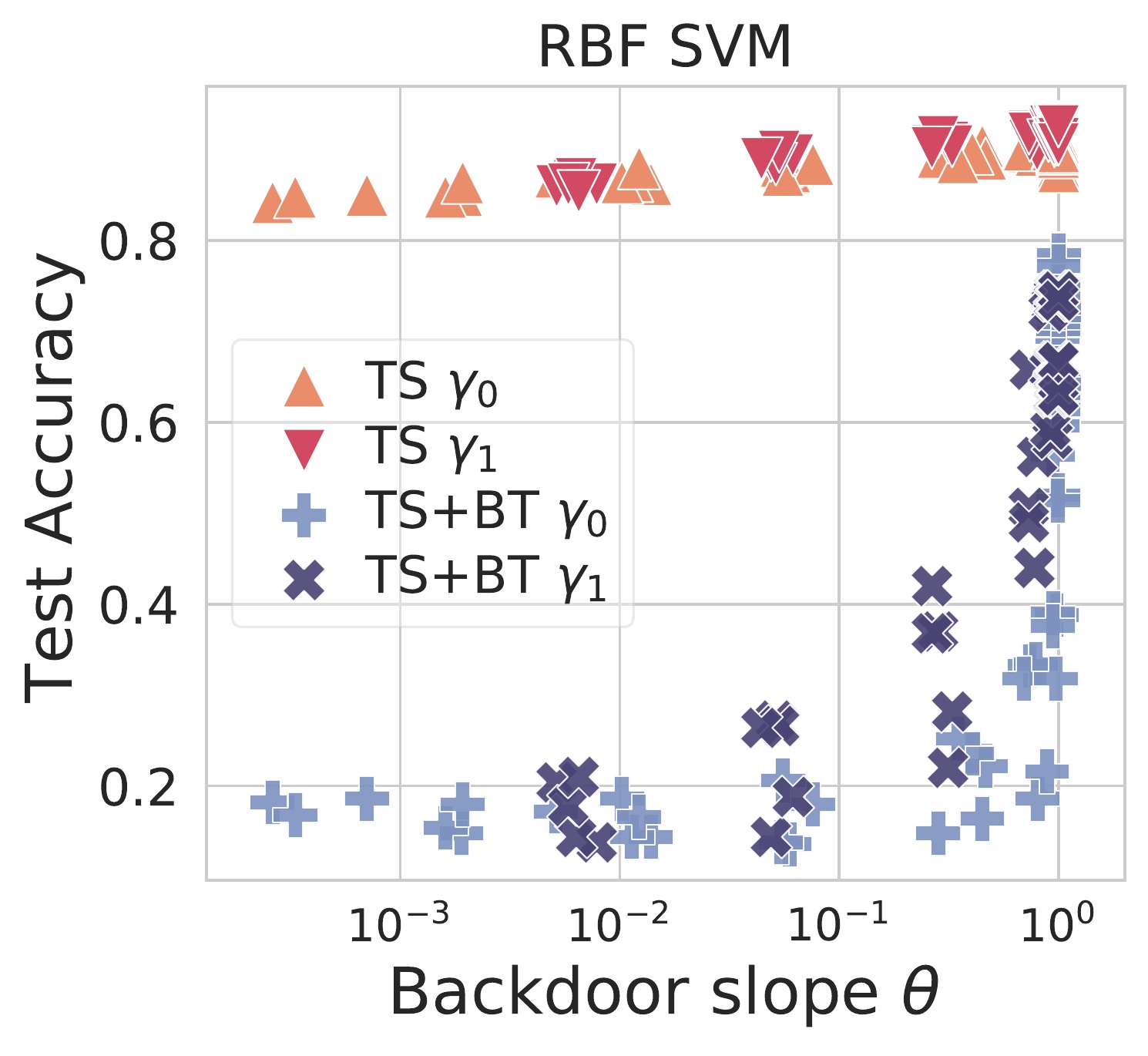}
 \caption{Backdoor slope vs backdoor (BK) effectiveness on CIFAR10 \cifarairplanetruck (top row) and \cifarbirddog (bottom row). See the caption of Figure~\ref{fig:supplementary_resultsMNIST} for further details.  The results are obtained considering a trigger size equal to $8$.}
  \label{fig:supplementary_resultsCIFAR}
\end{figure*}

\begin{figure*}[t]
  \centering
      \includegraphics[width=0.242\textwidth]{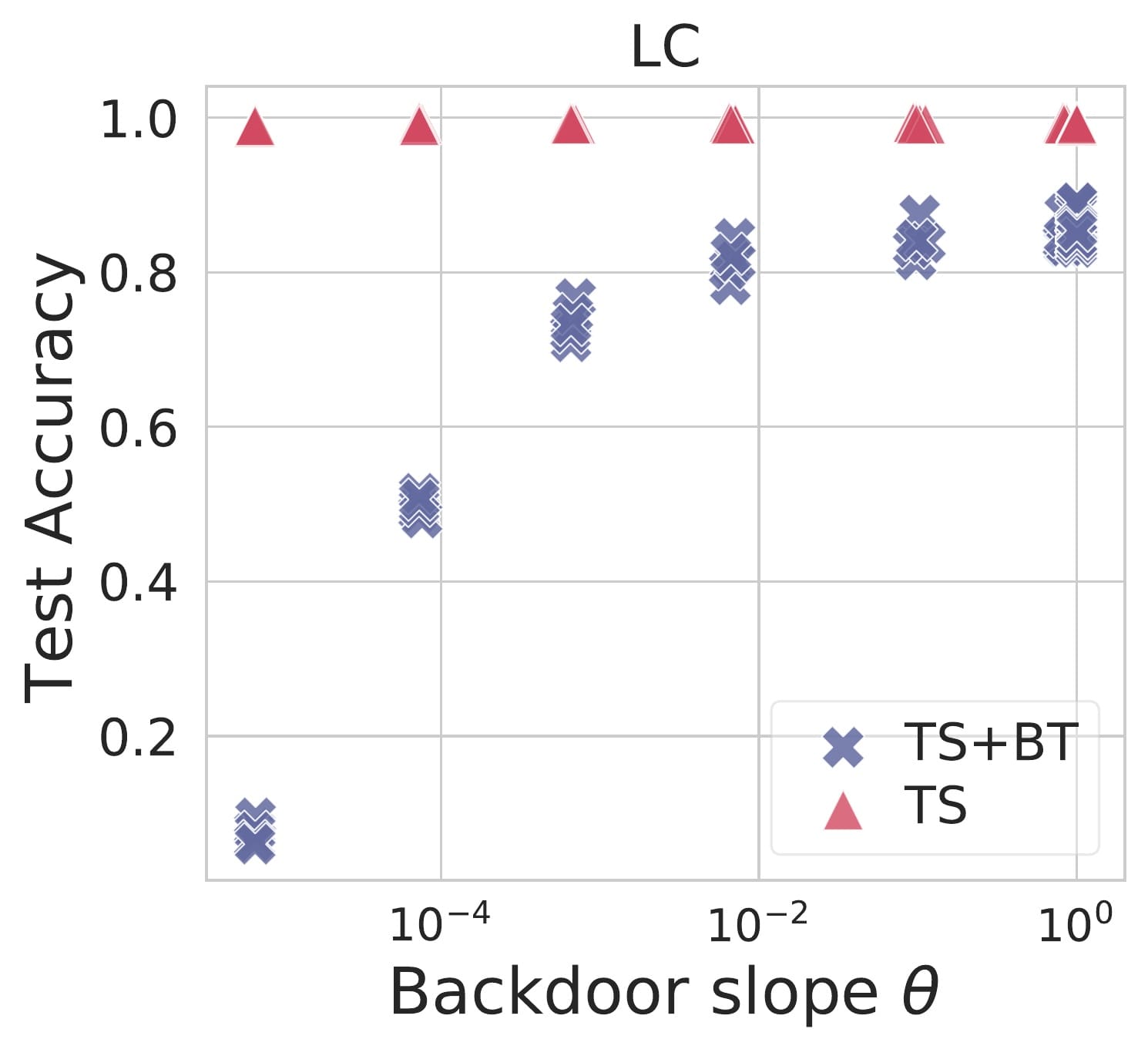}
    \includegraphics[width=0.242\textwidth]{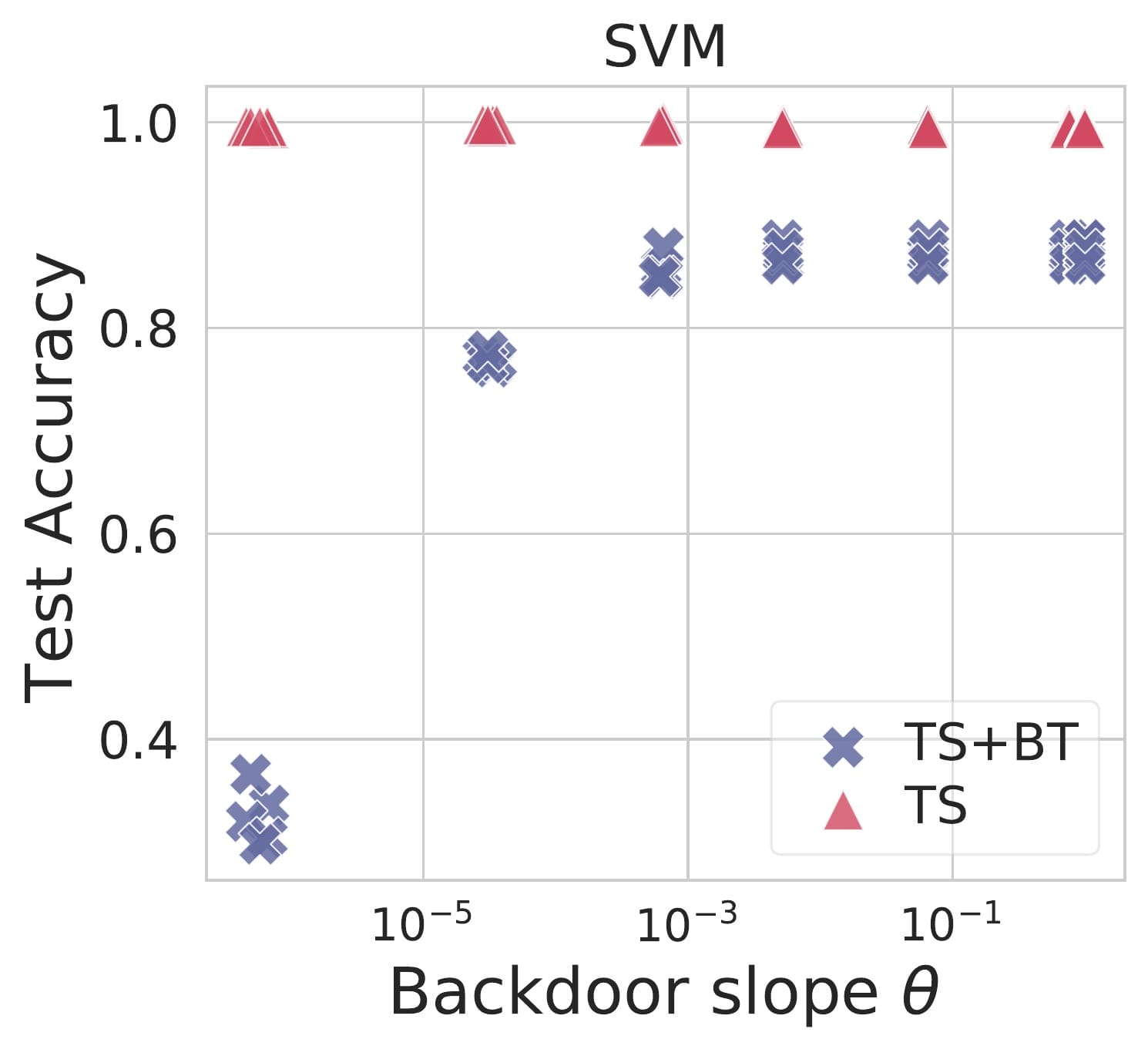}
    \includegraphics[width=0.242\textwidth]{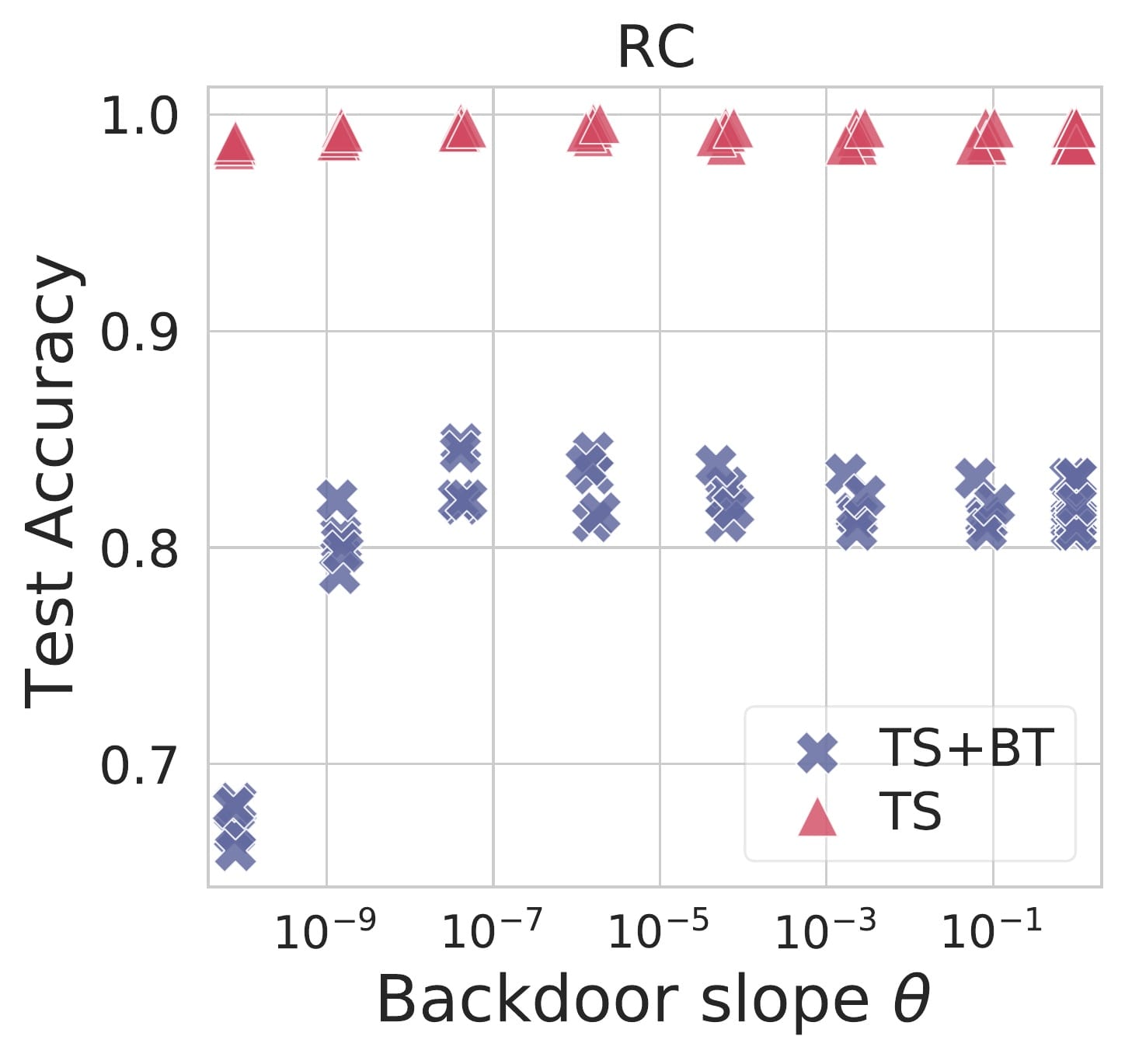}
    \includegraphics[width=0.242\textwidth]{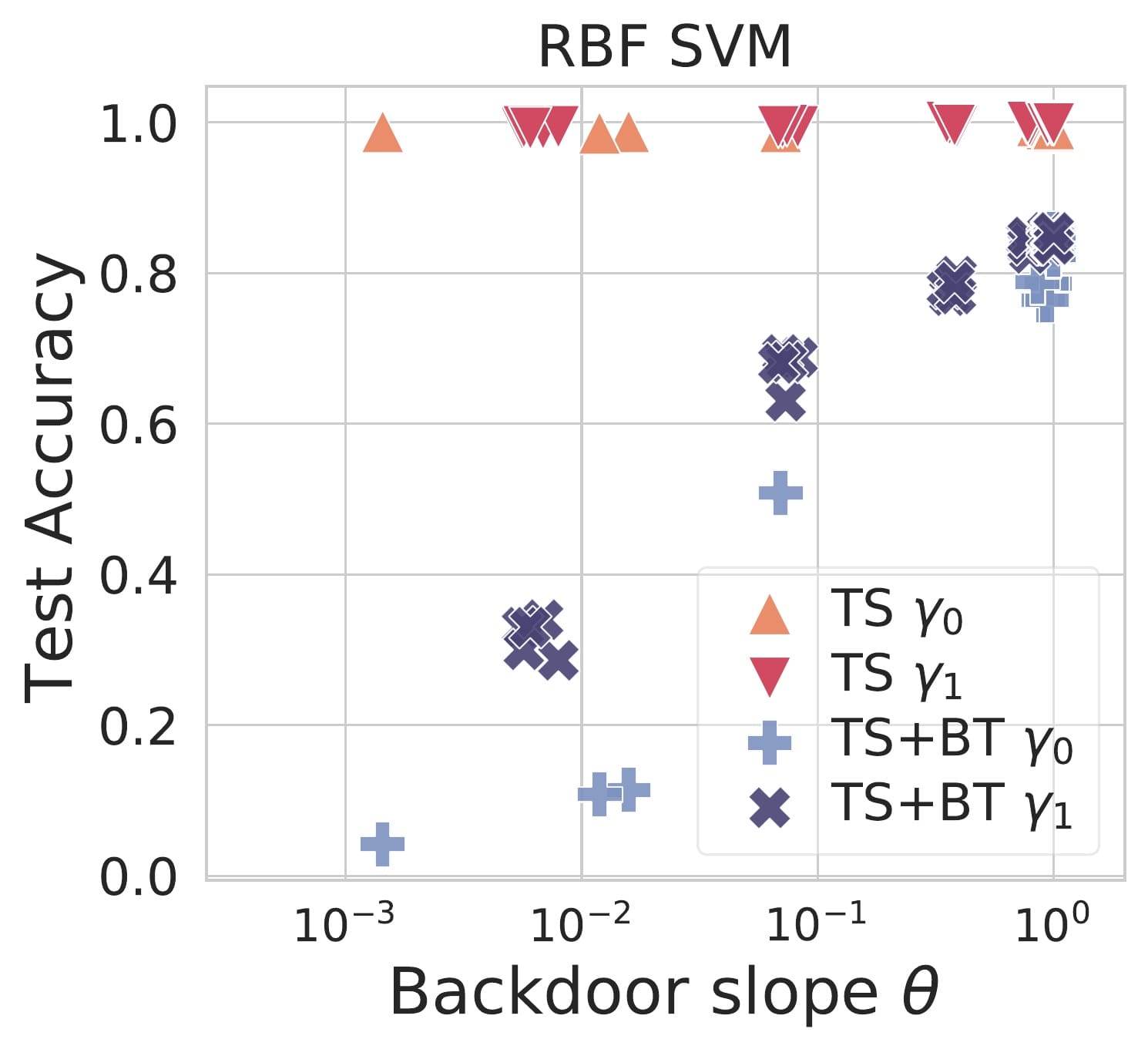}

    \includegraphics[width=0.242\textwidth]{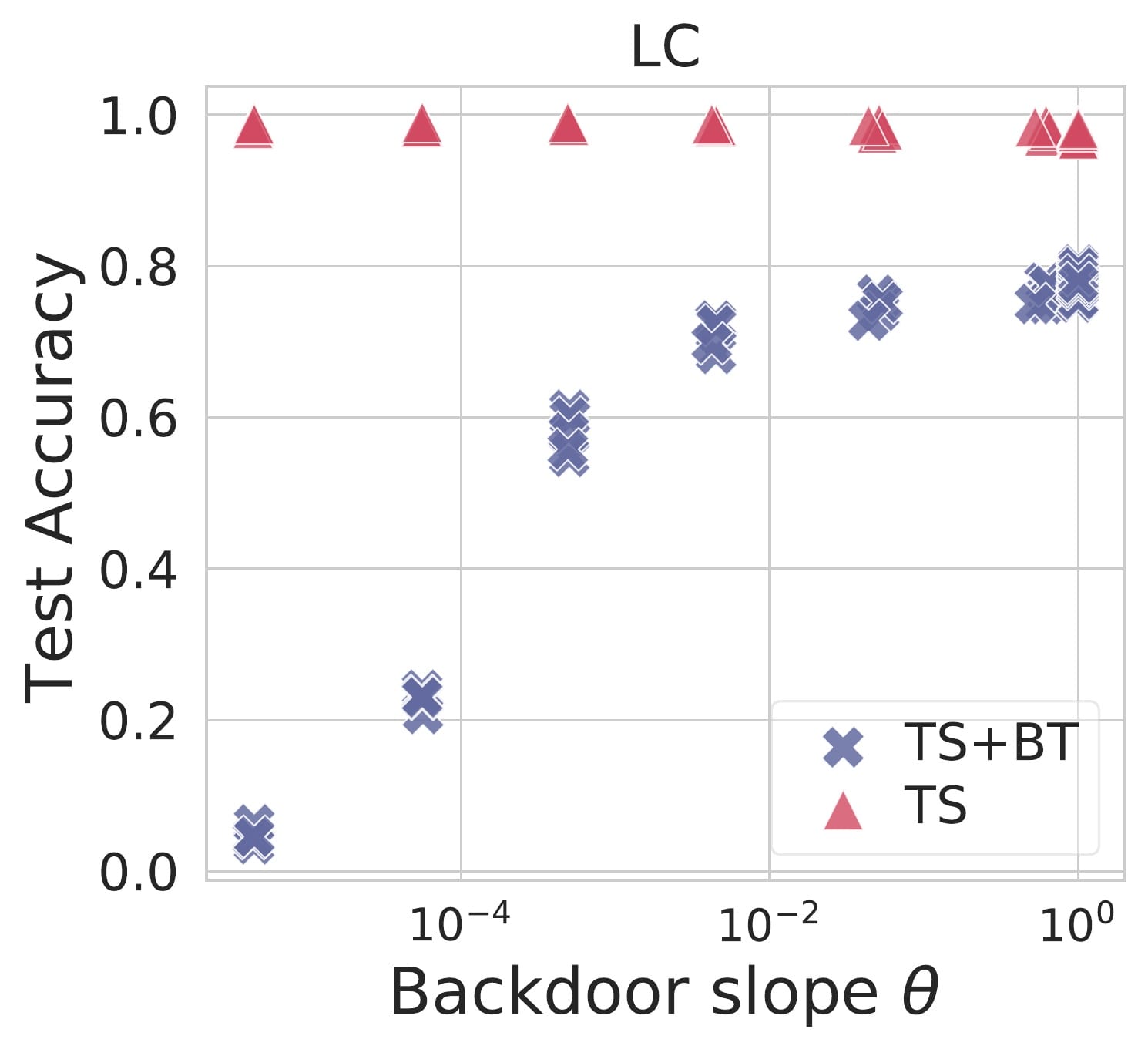}
    \includegraphics[width=0.242\textwidth]{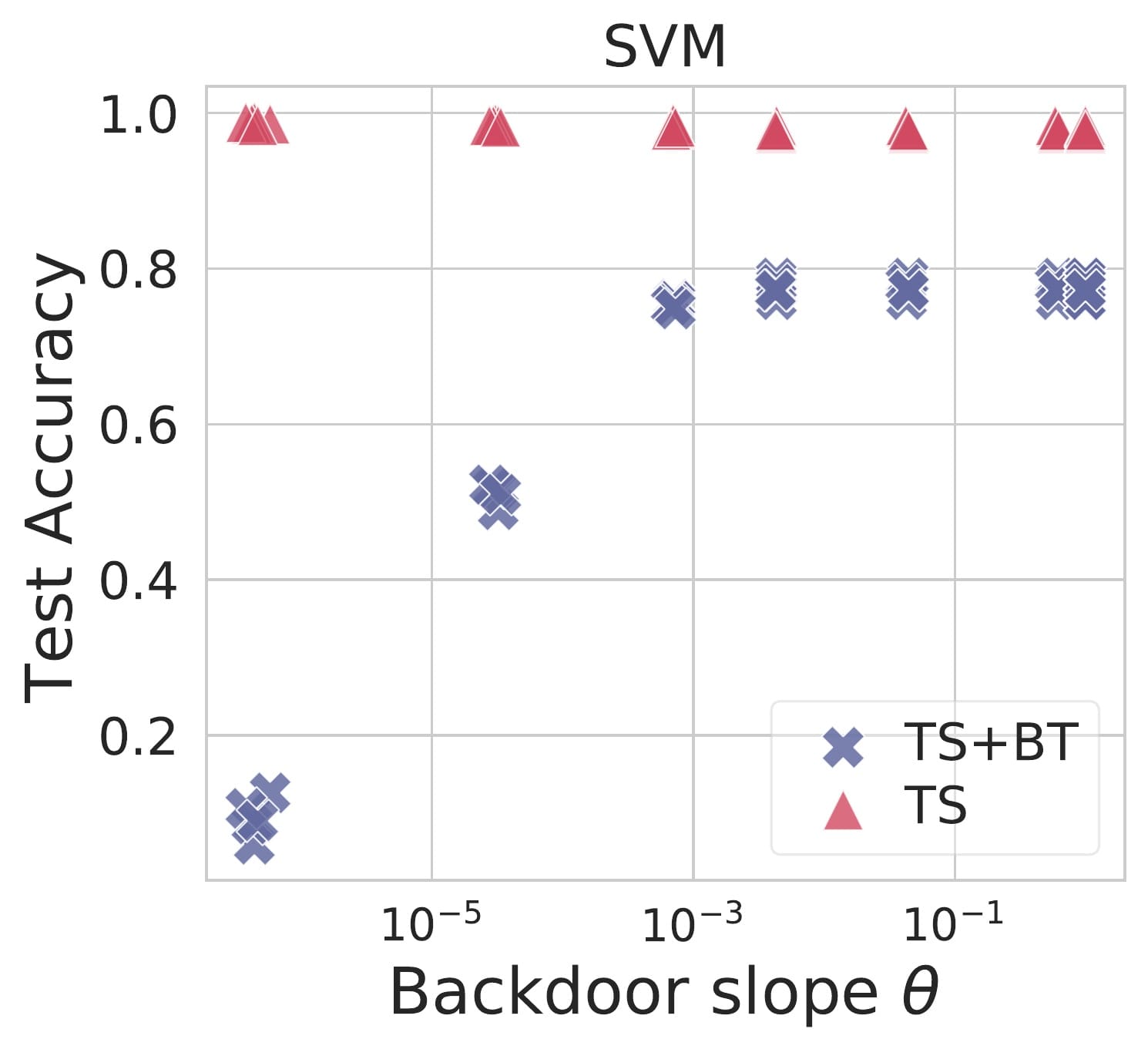}
    \includegraphics[width=0.242\textwidth]{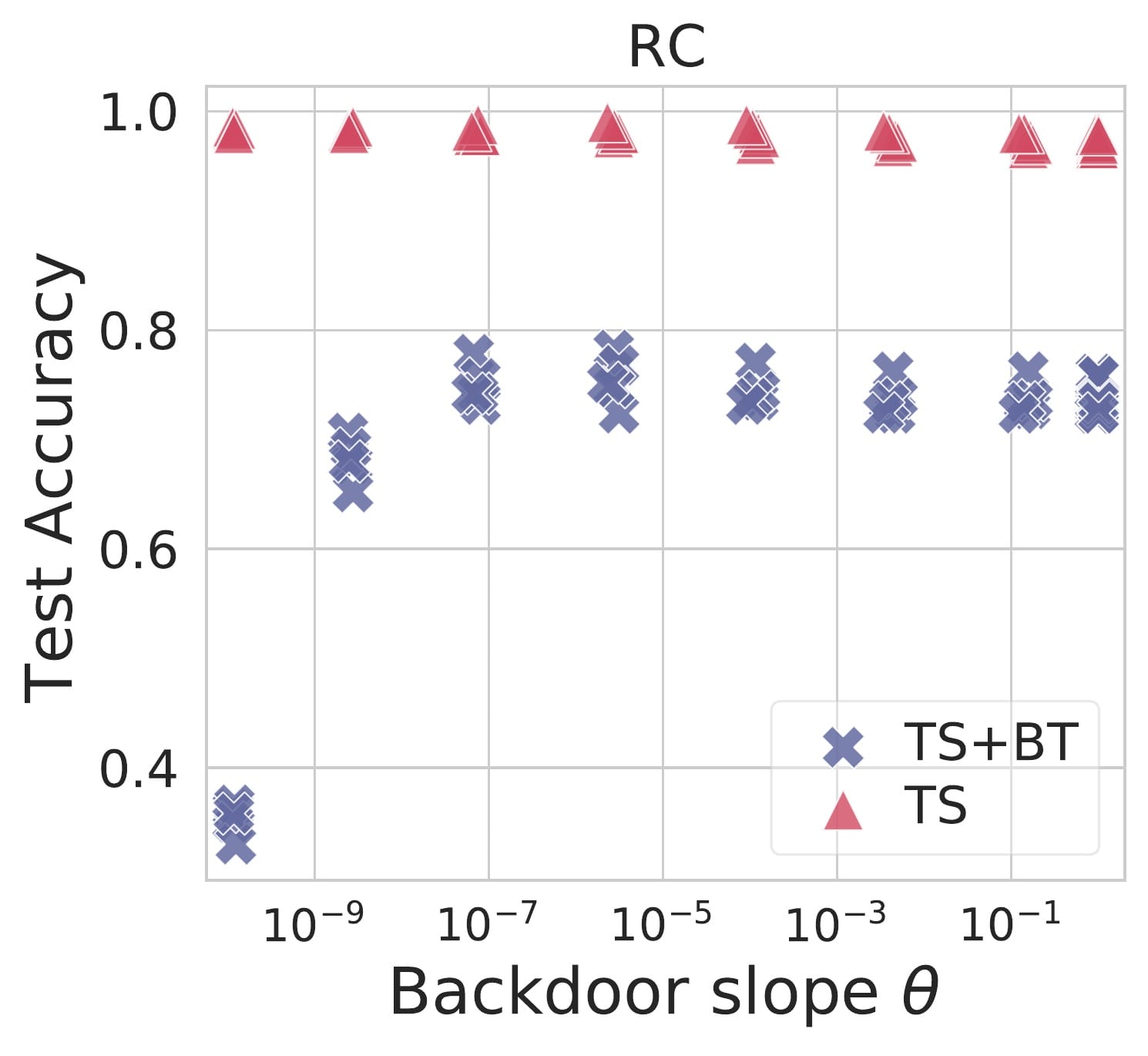}
    \includegraphics[width=0.242\textwidth]{ml_figs-imagenette-slope-svm_pair2-5_ttypeinvisible_tsizeNone.jpeg}
 \caption{Backdoor slope vs backdoor (BK) effectiveness on Imagenette \imagenetteplayerchurch (top row) and \imagenettetenchparachute (bottom row). See the caption of Figure~\ref{fig:supplementary_resultsMNIST} for further details.  The results are obtained considering a trigger size equal to $8$.}
  \label{fig:supplementary_resultsImagenette}
\end{figure*}

\myparagraph{Backdoor Learning Curves and Backdoor Learning Slope.} 
In Figure~\ref{fig:appendix_backdoor_learning_curves_mnist30}-\ref{fig:appendix_backdoor_learning_curves_imagenette25} we report the backdoor learning curves for each classifier and dataset pair. 
In Figure~\ref{fig:supplementary_resultsMNIST}-\ref{fig:supplementary_resultsImagenette}, we report the backdoor learning slope, computed with $p = 0.1$, for all the considered classifiers and all subset pairs. 
The results do not show significant variation with respect to the ones reported in the paper.\smallskip

\begin{figure*}[h!]
  \centering
    \includegraphics[width=0.235\textwidth]{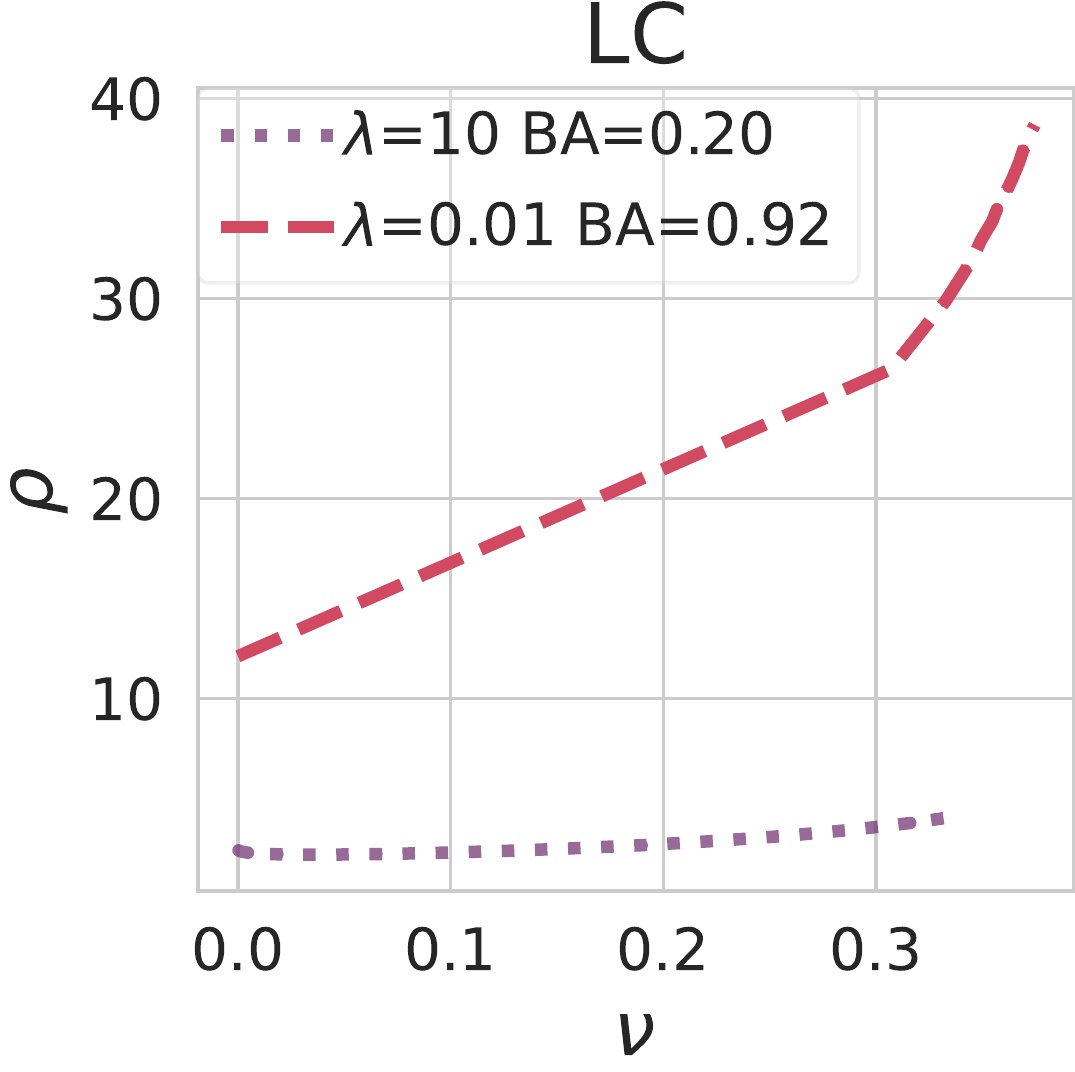}
  \includegraphics[width=0.235\textwidth]{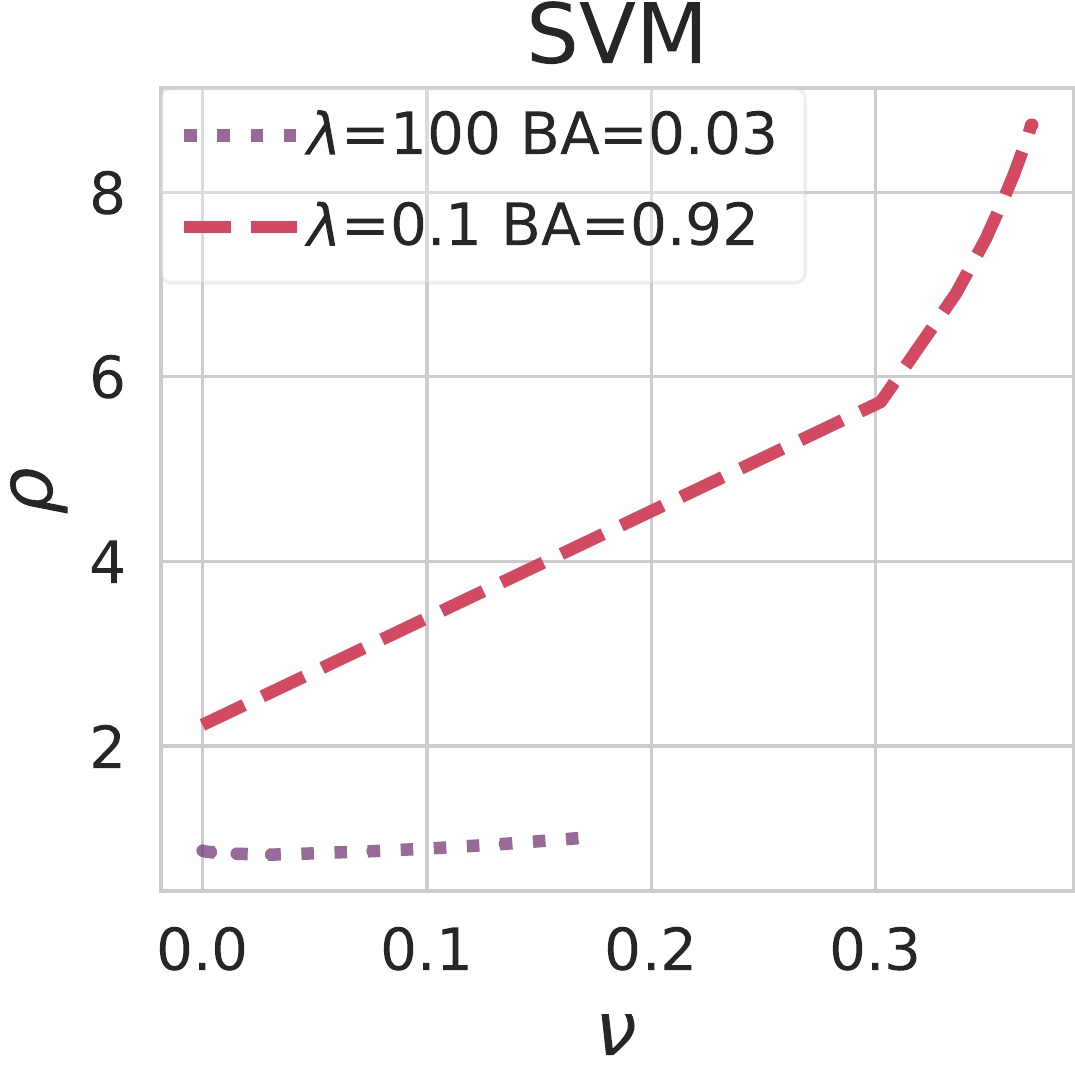}
  \includegraphics[width=0.235\textwidth]{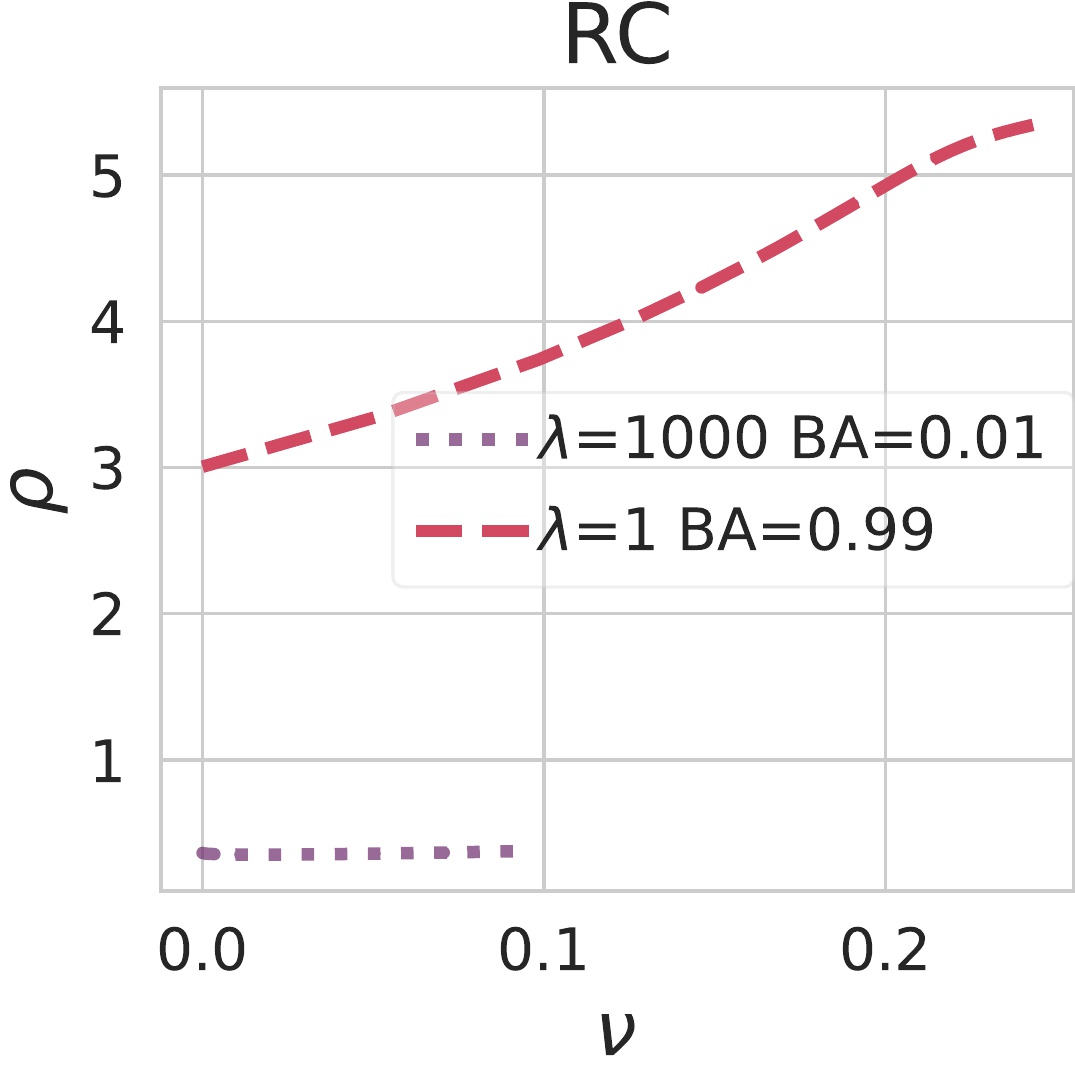}
  \includegraphics[width=0.235\textwidth]{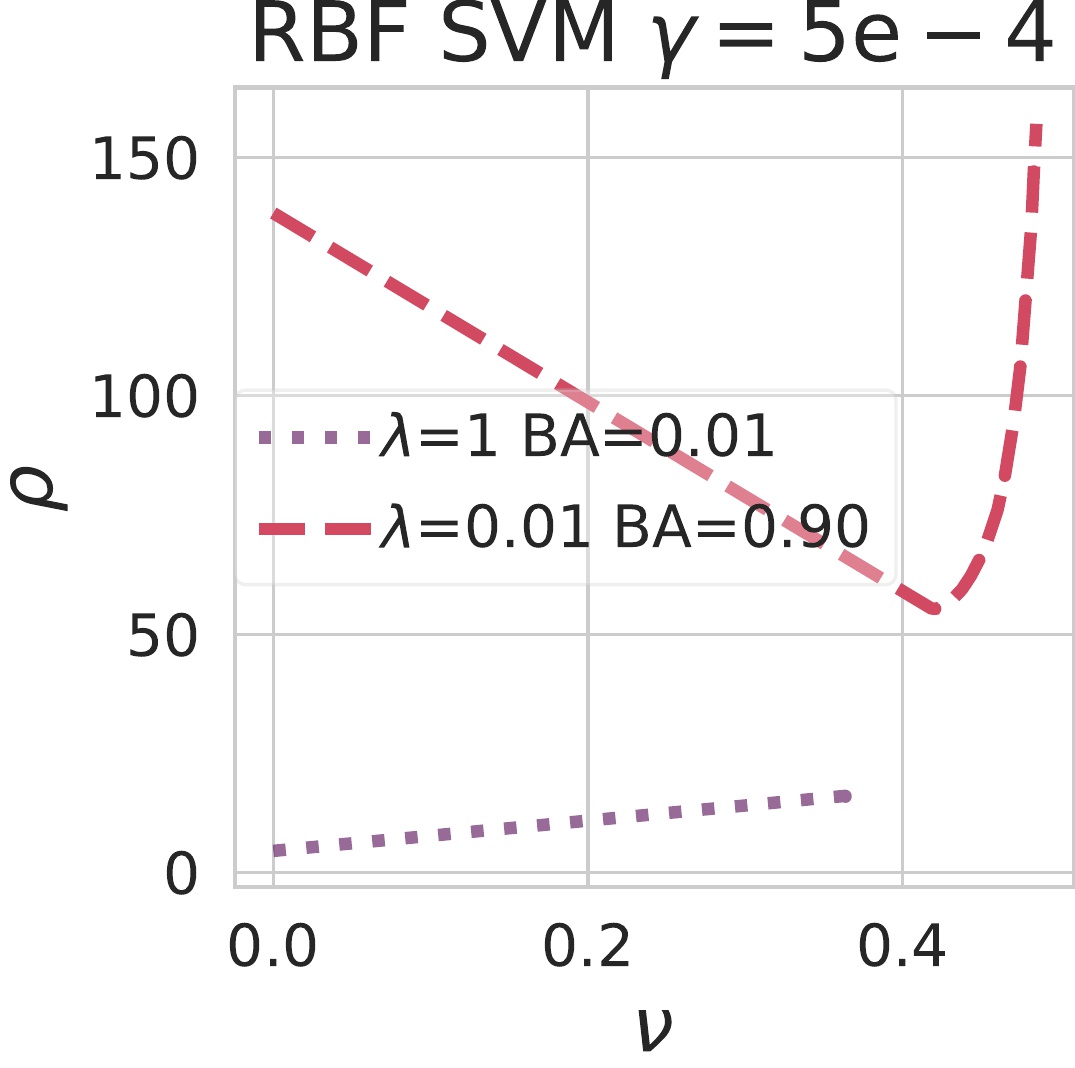}
  
      \includegraphics[width=0.235\textwidth]{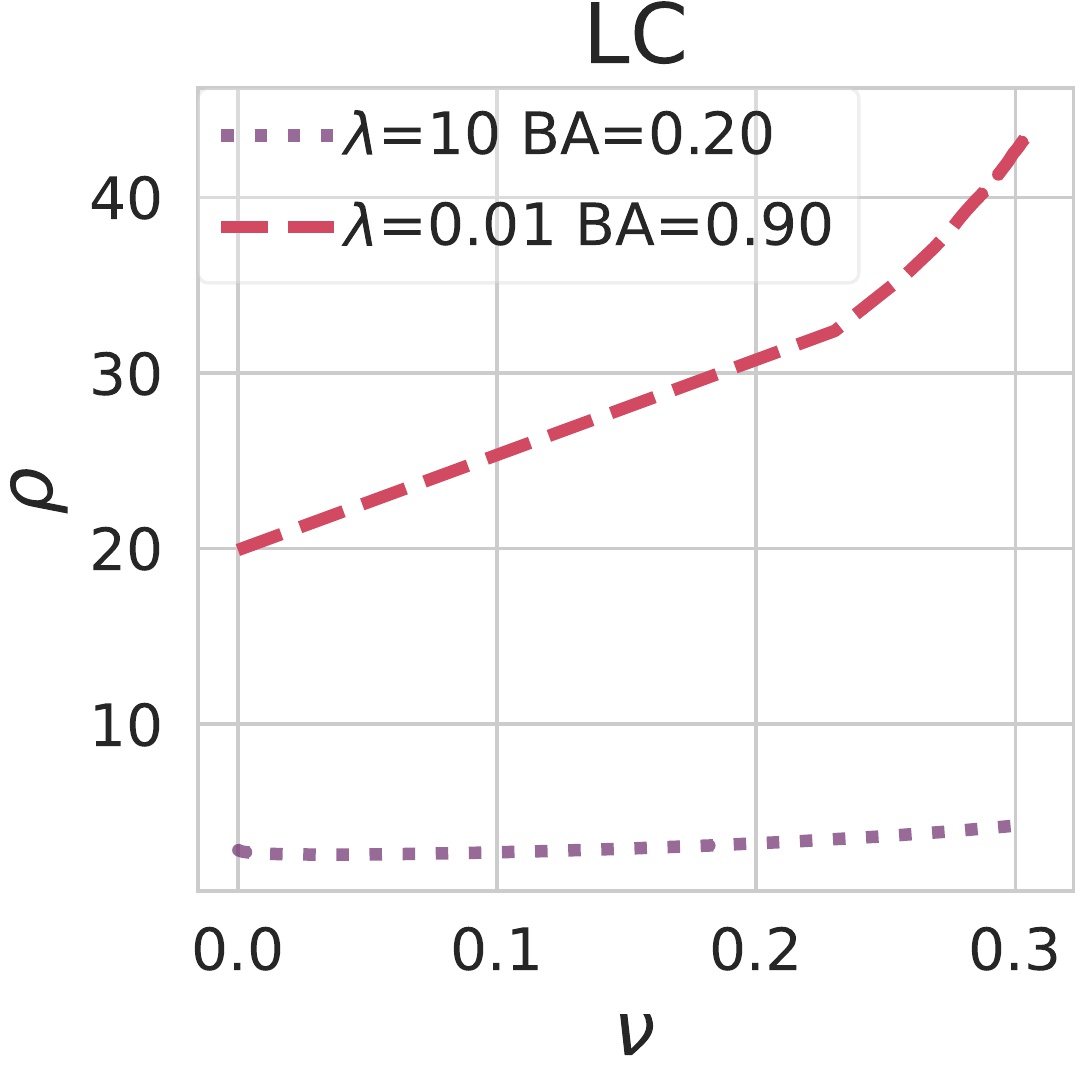}
  \includegraphics[width=0.235\textwidth]{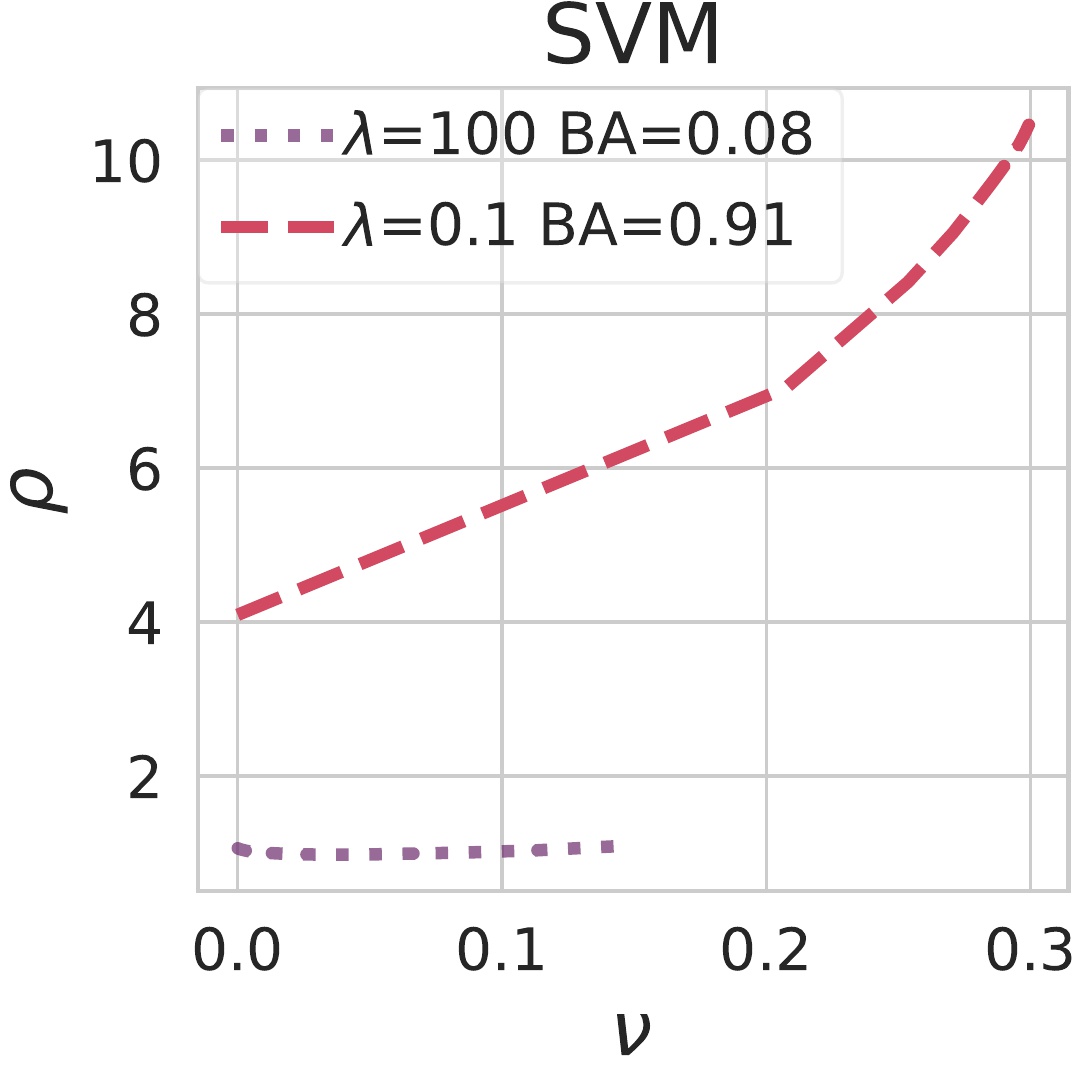}
  \includegraphics[width=0.235\textwidth]{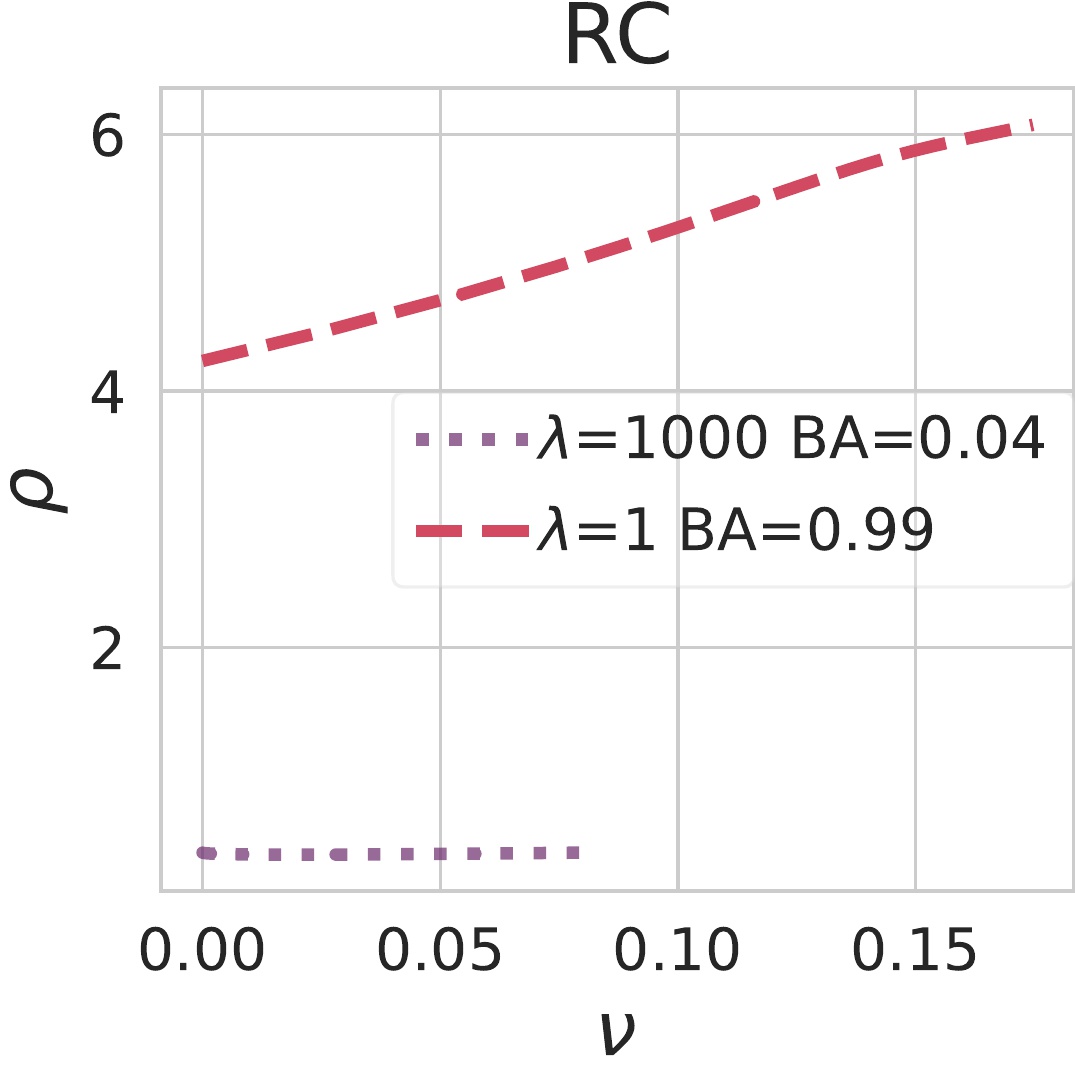}
  \includegraphics[width=0.235\textwidth]{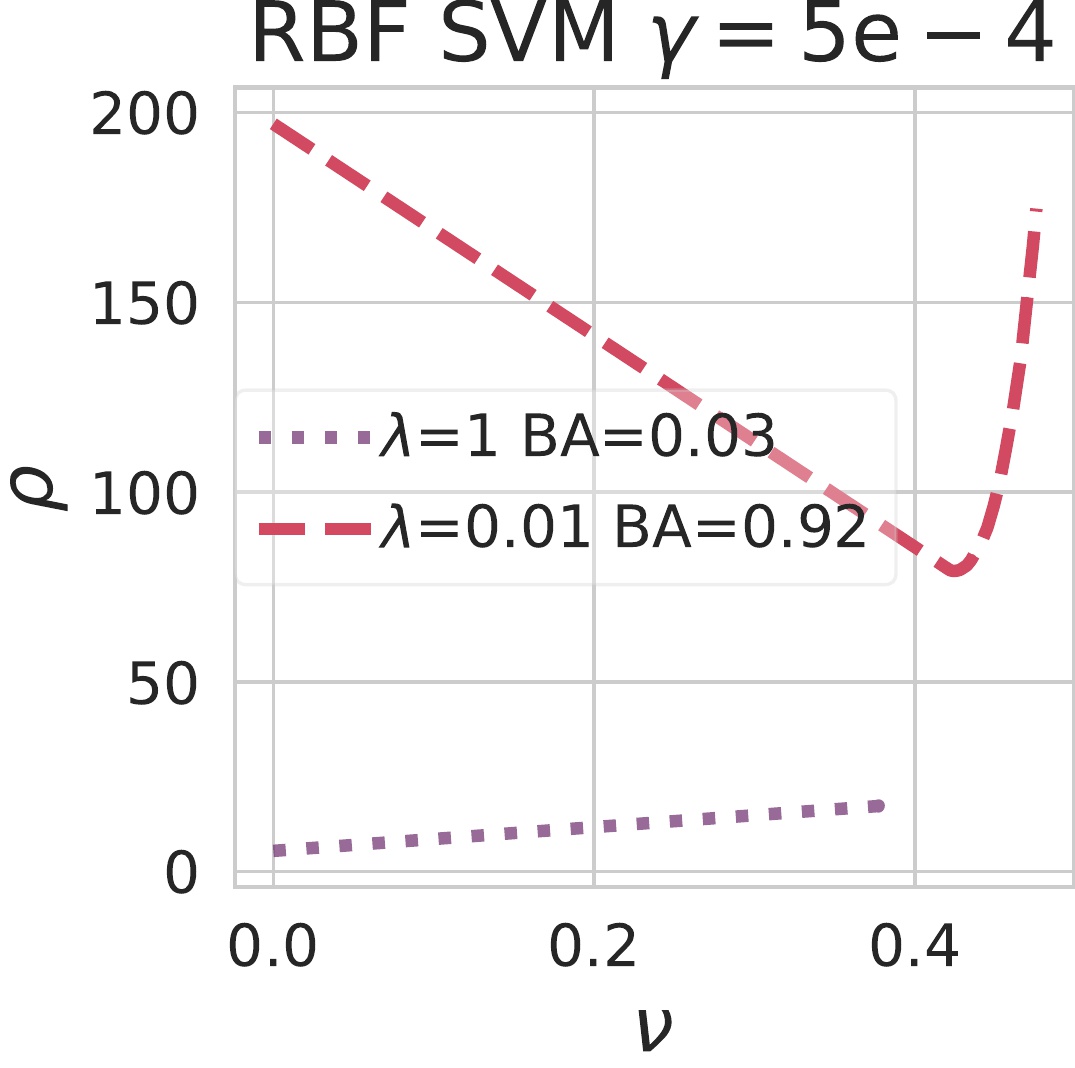}
  \caption{Backdoor weight deviation for different classifiers trained on MNIST $3~\rm{vs}~0$ (top row) and $5~\rm{vs}~2$ (bottom row).
  We specify regularization parameter $\lambda$ and backdoor (BK) accuracy for each setting in the legend of each plot.}
  \label{fig:supplementary_backdoorParametersDeviationMNIST}
\end{figure*}

\begin{figure*}[t]
  \centering
  \includegraphics[width=0.24\textwidth]{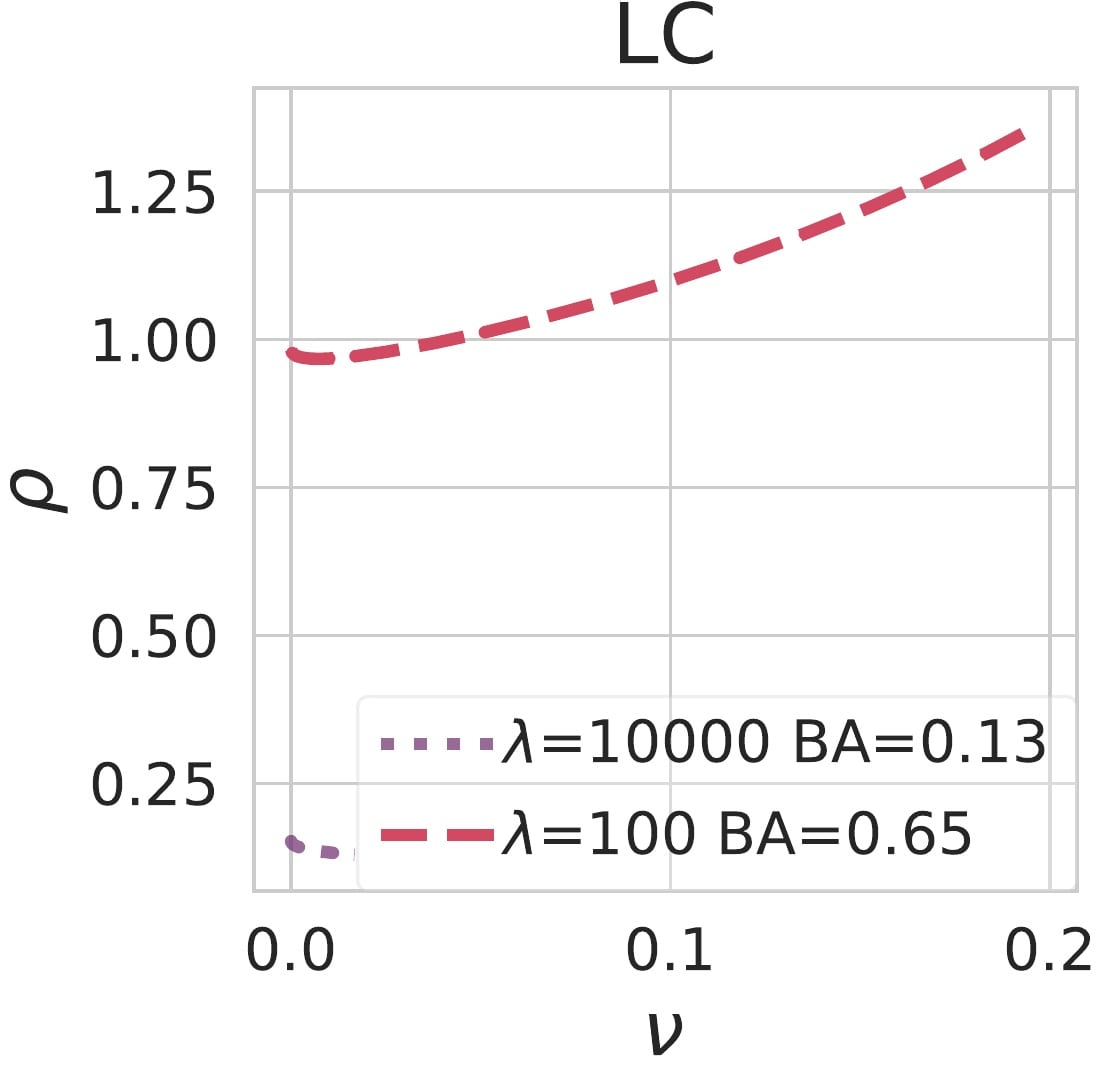}
  \includegraphics[width=0.235\textwidth]{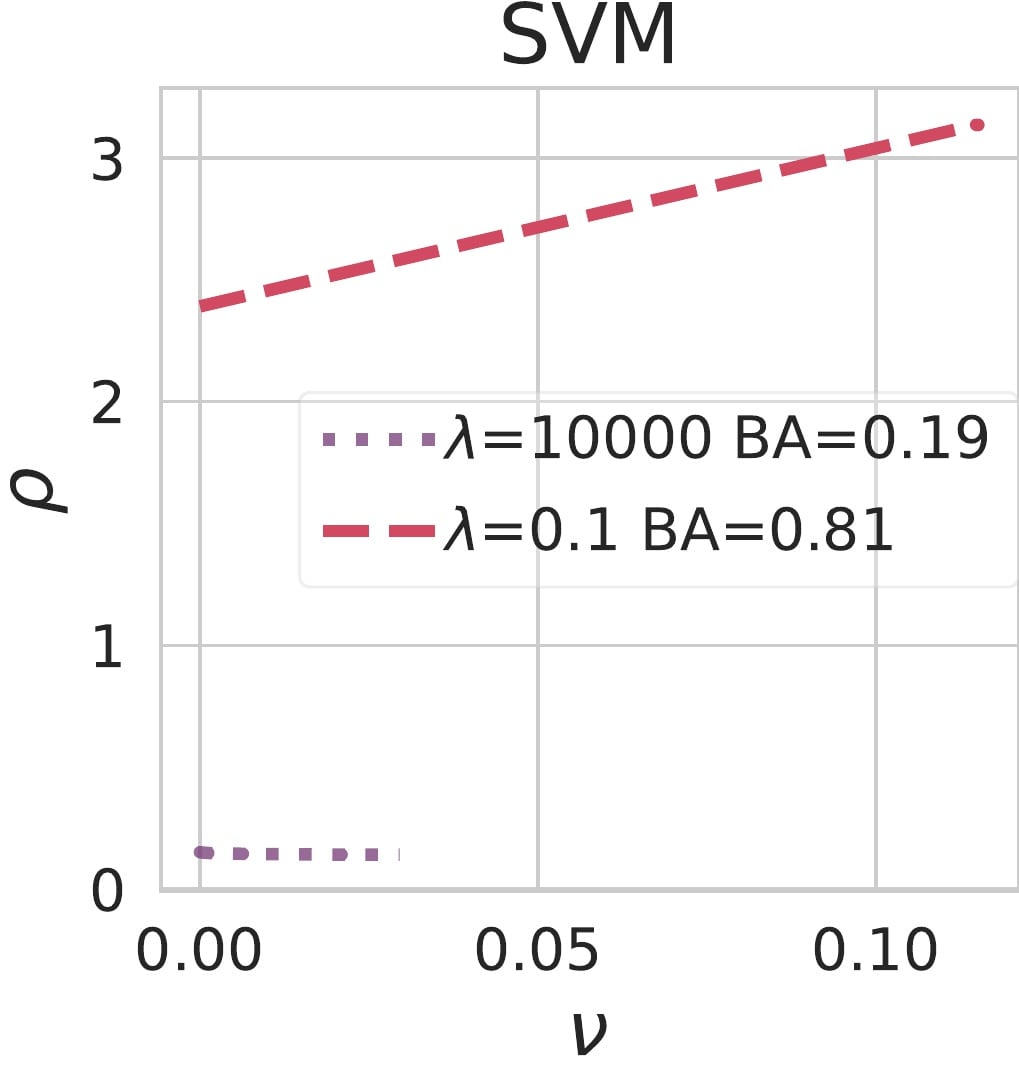}
  \includegraphics[width=0.235\textwidth]{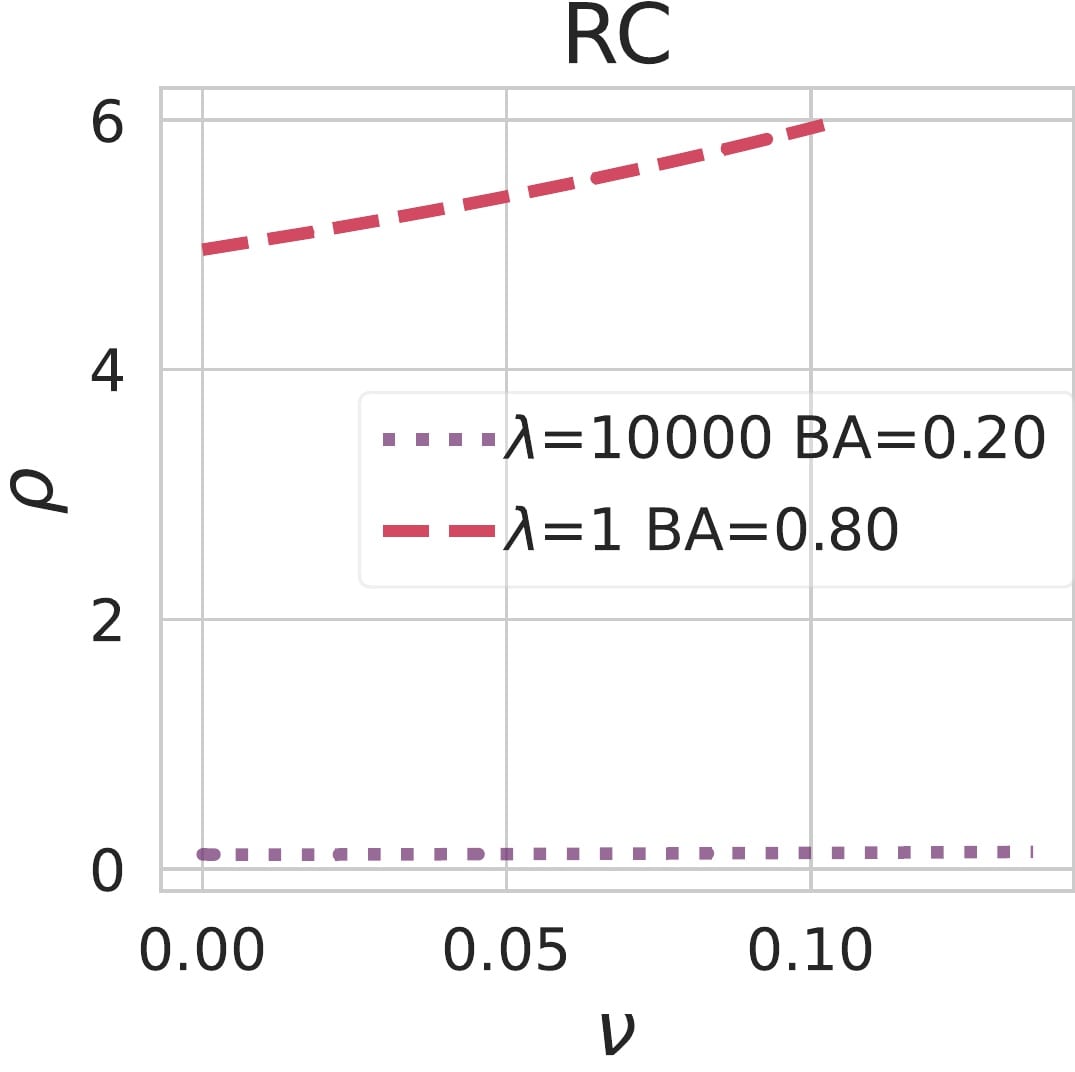}
  \includegraphics[width=0.225\textwidth]{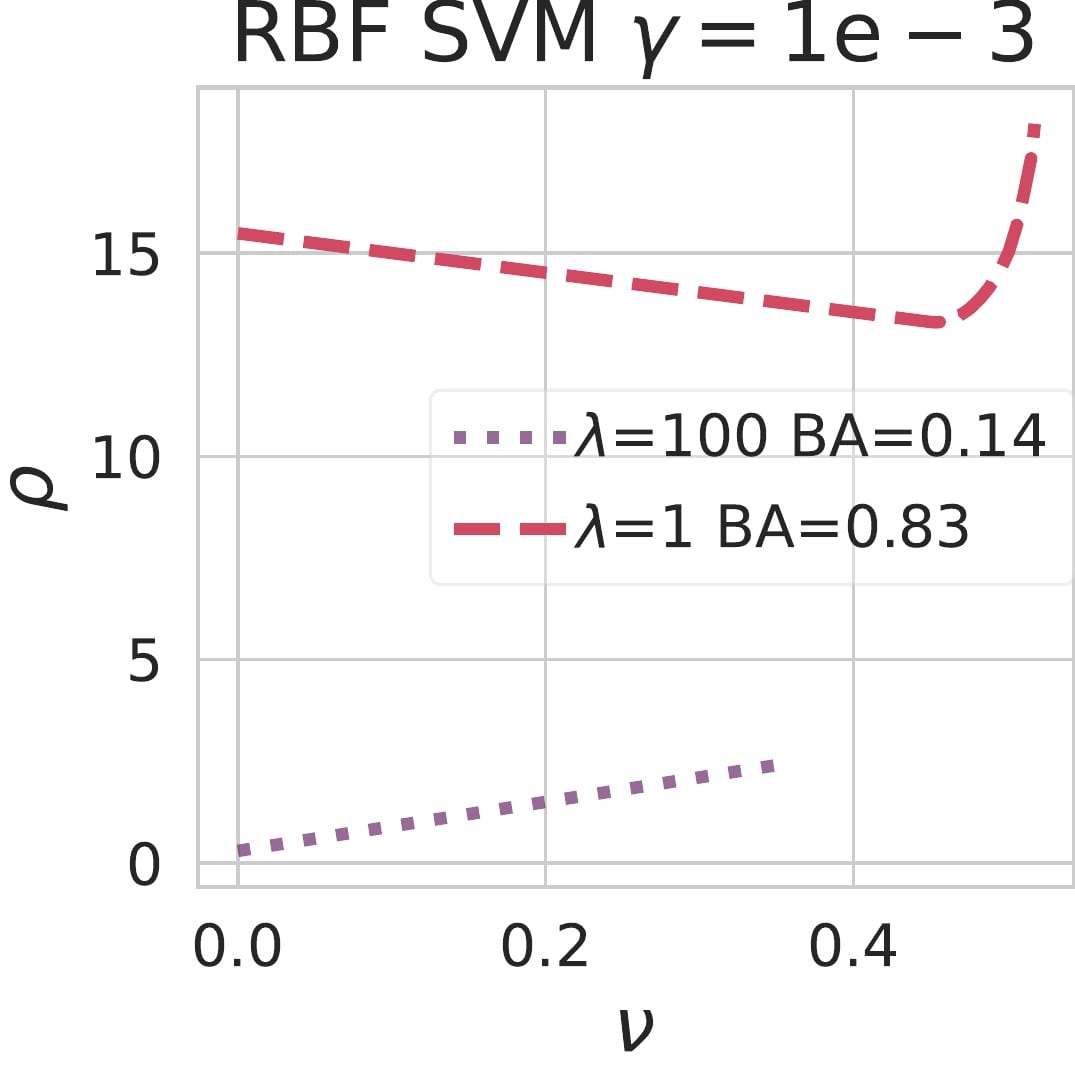}
 
  \includegraphics[width=0.24\textwidth]{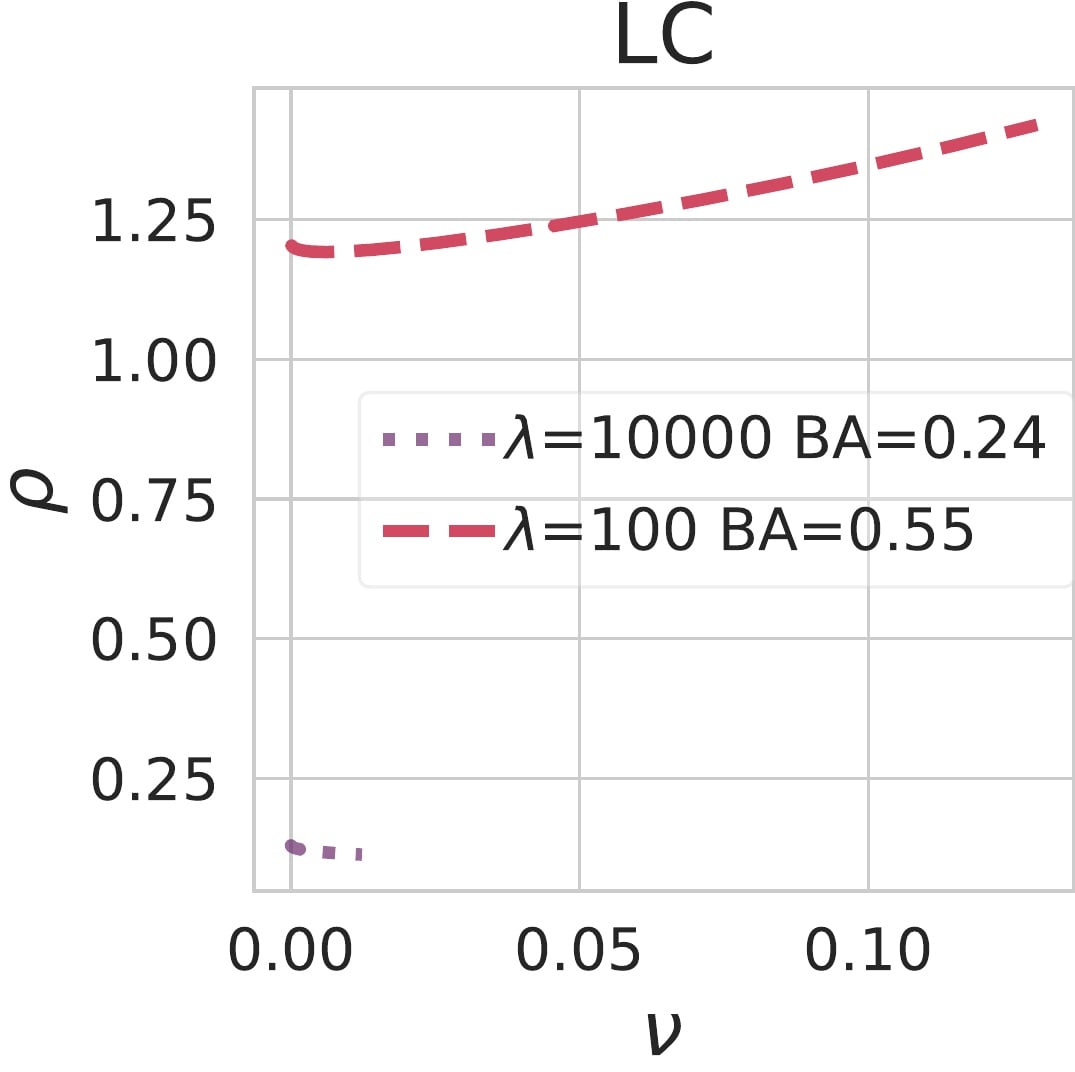}
  \includegraphics[width=0.235\textwidth]{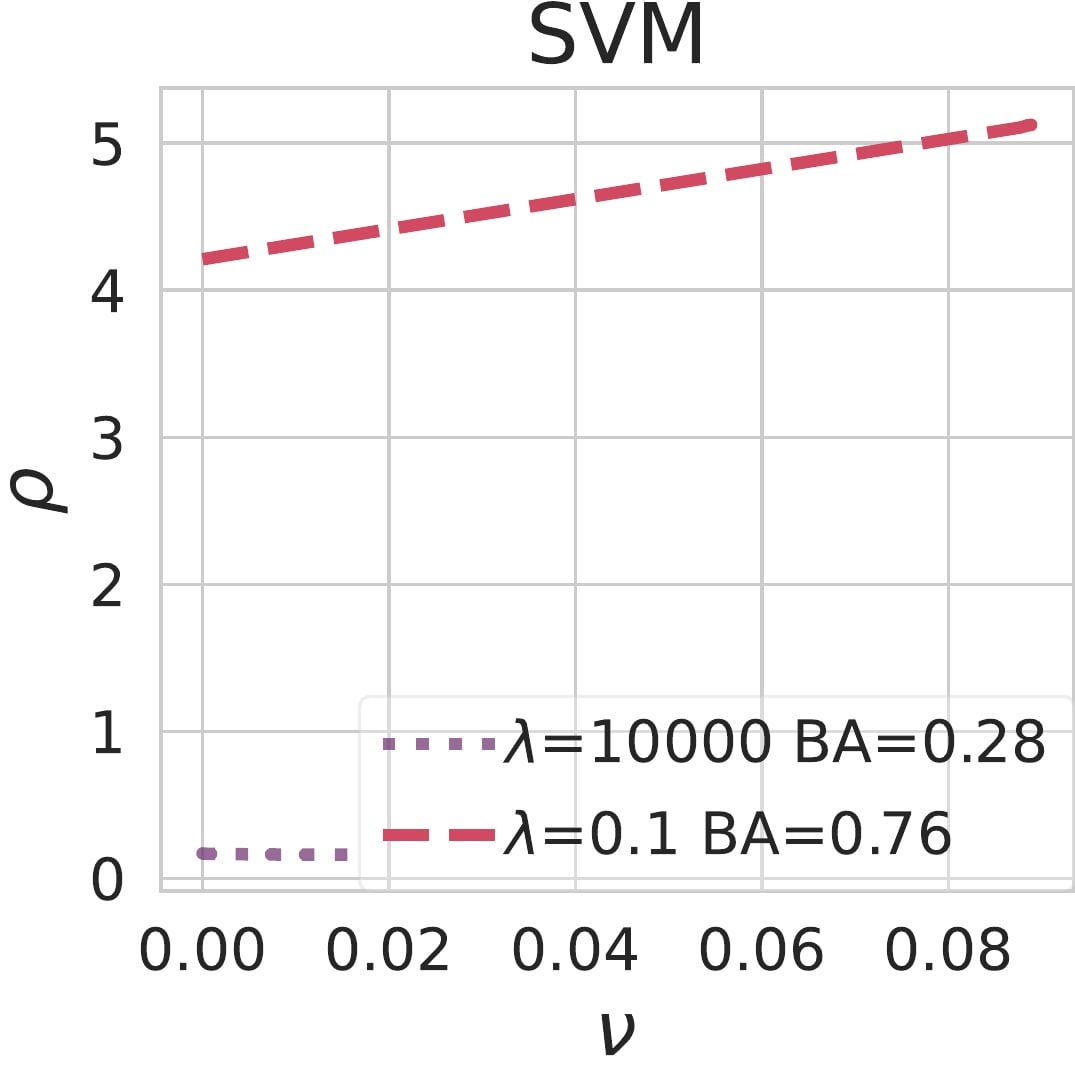}
  \includegraphics[width=0.235\textwidth]{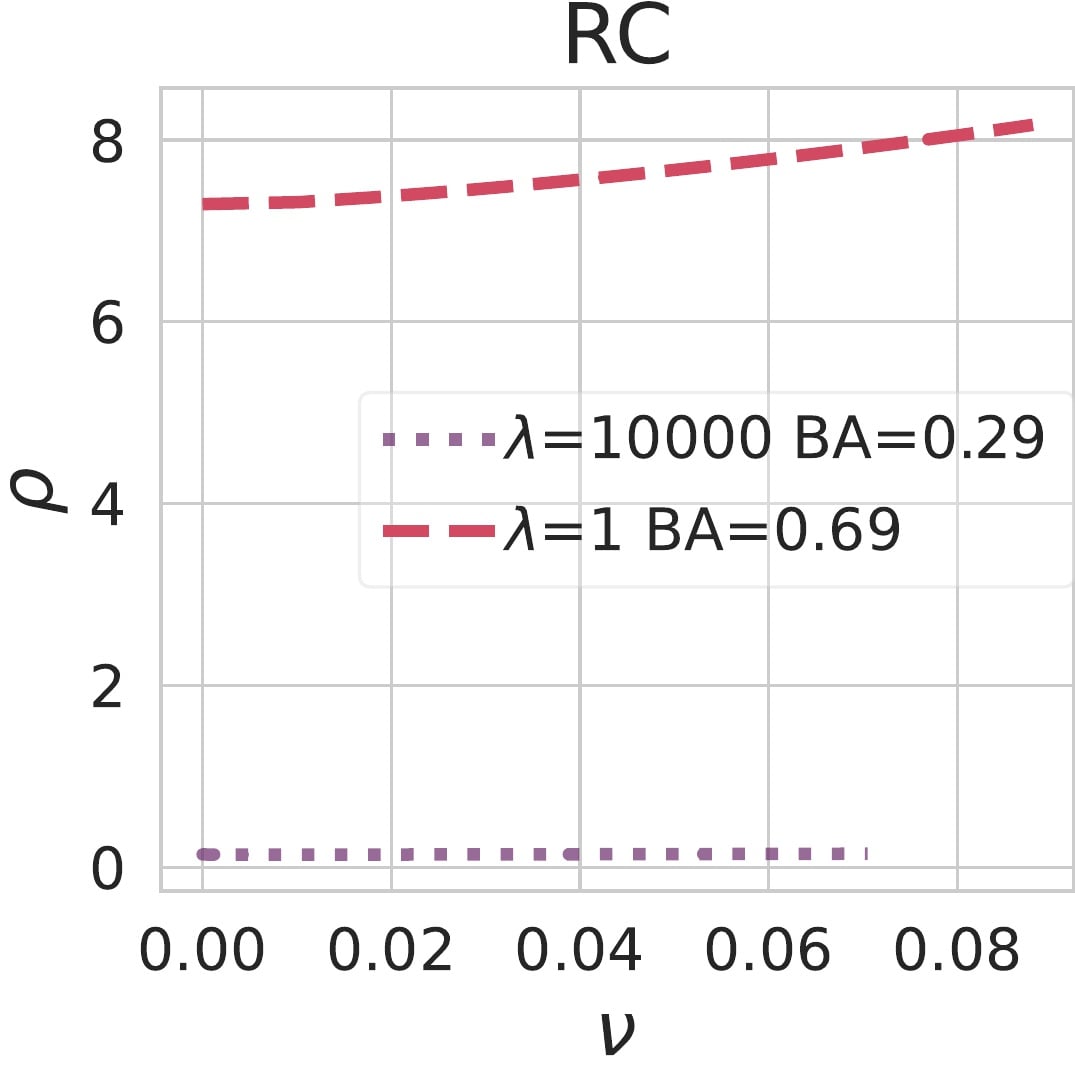}
  \includegraphics[width=0.225\textwidth]{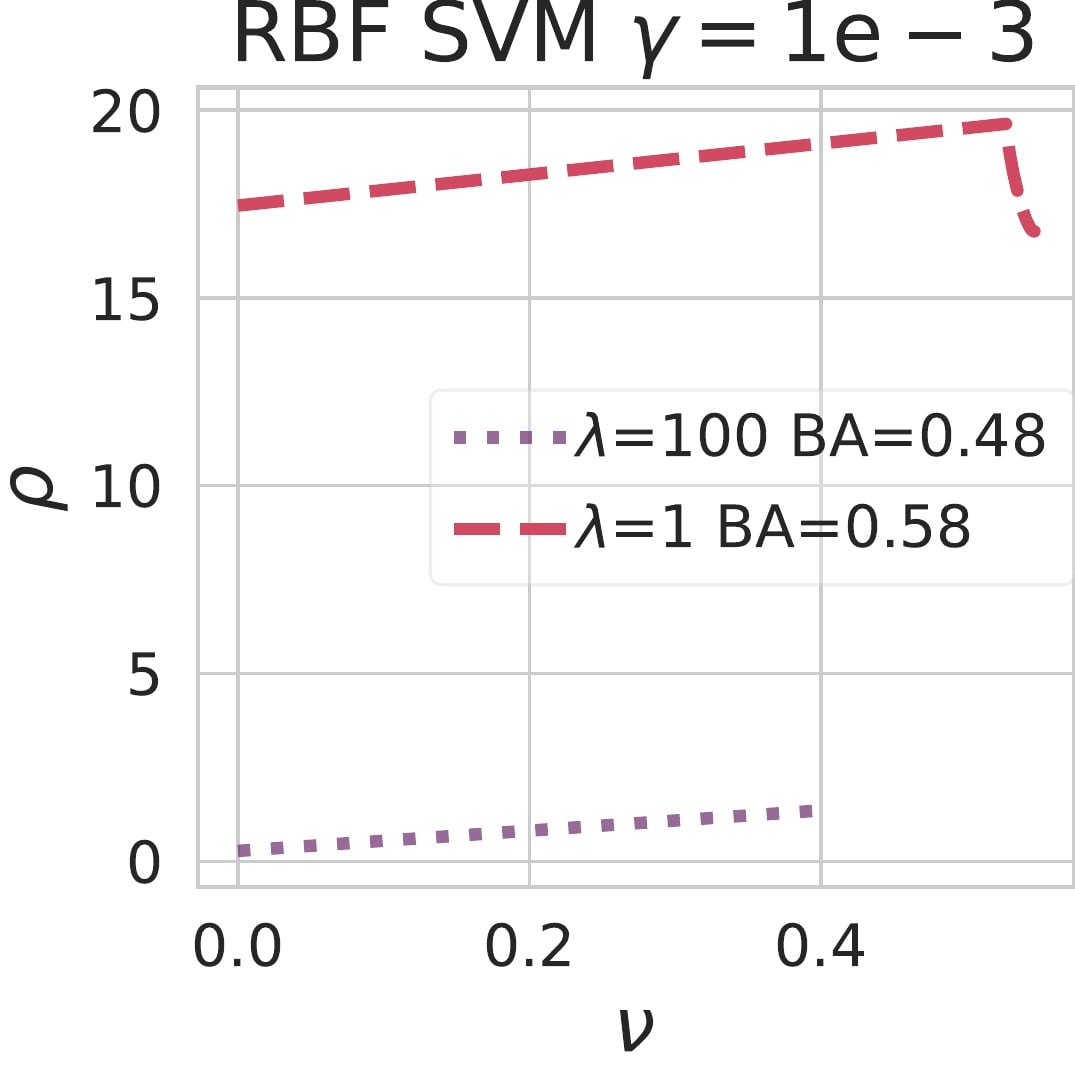}
  \caption{Backdoor weight deviation for different classifiers trained on CIFAR10 \cifarairplanetruck (top), and \cifarbirddog (bottom). 
   See Figure~\ref{fig:supplementary_backdoorParametersDeviationMNIST} for further details.}
  \label{fig:supplementary_backdoorParametersDeviationCIFAR}
\end{figure*}

\begin{figure*}[h!]
  \centering
  \includegraphics[width=0.24\textwidth]{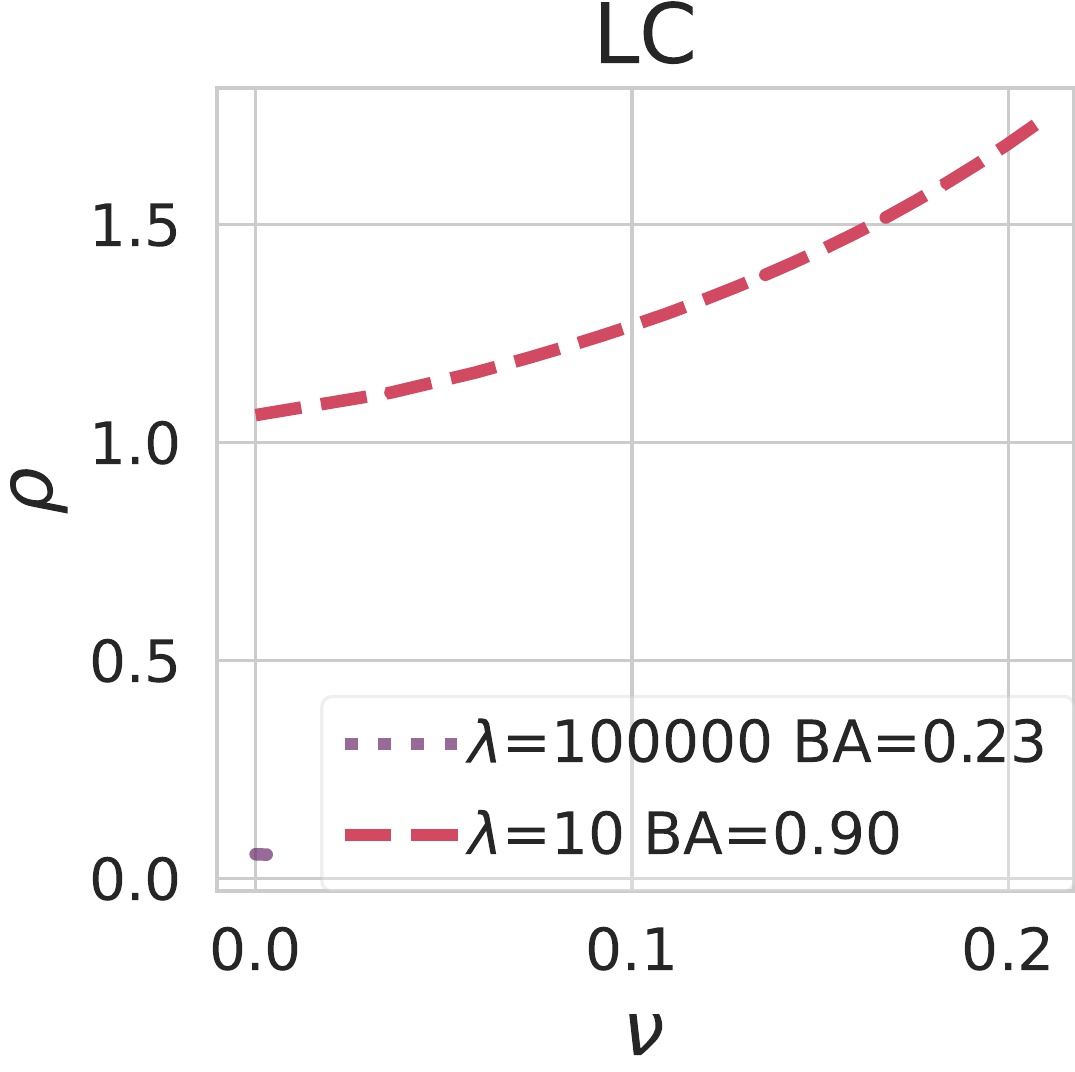}
  \includegraphics[width=0.235\textwidth]{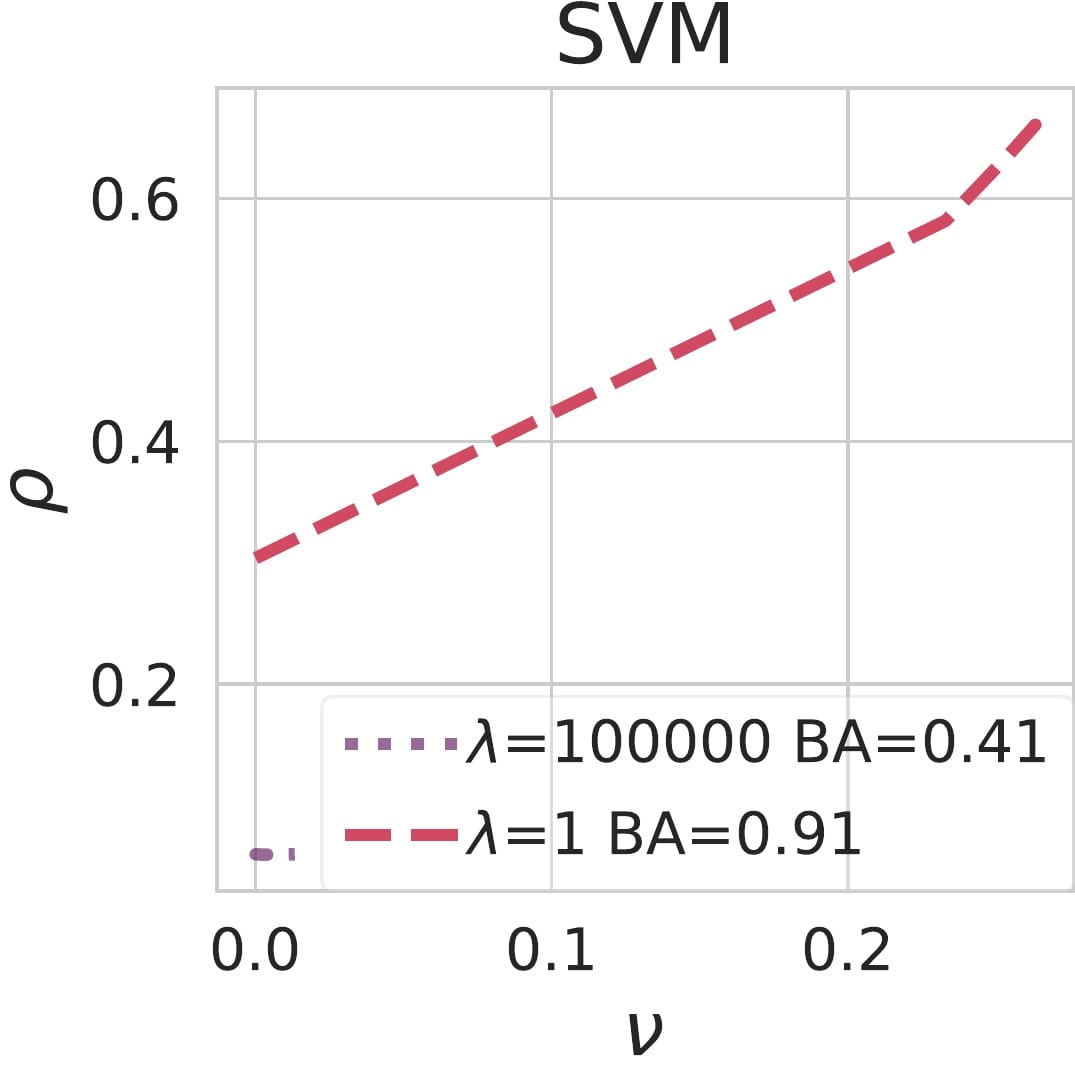}
  \includegraphics[width=0.235\textwidth]{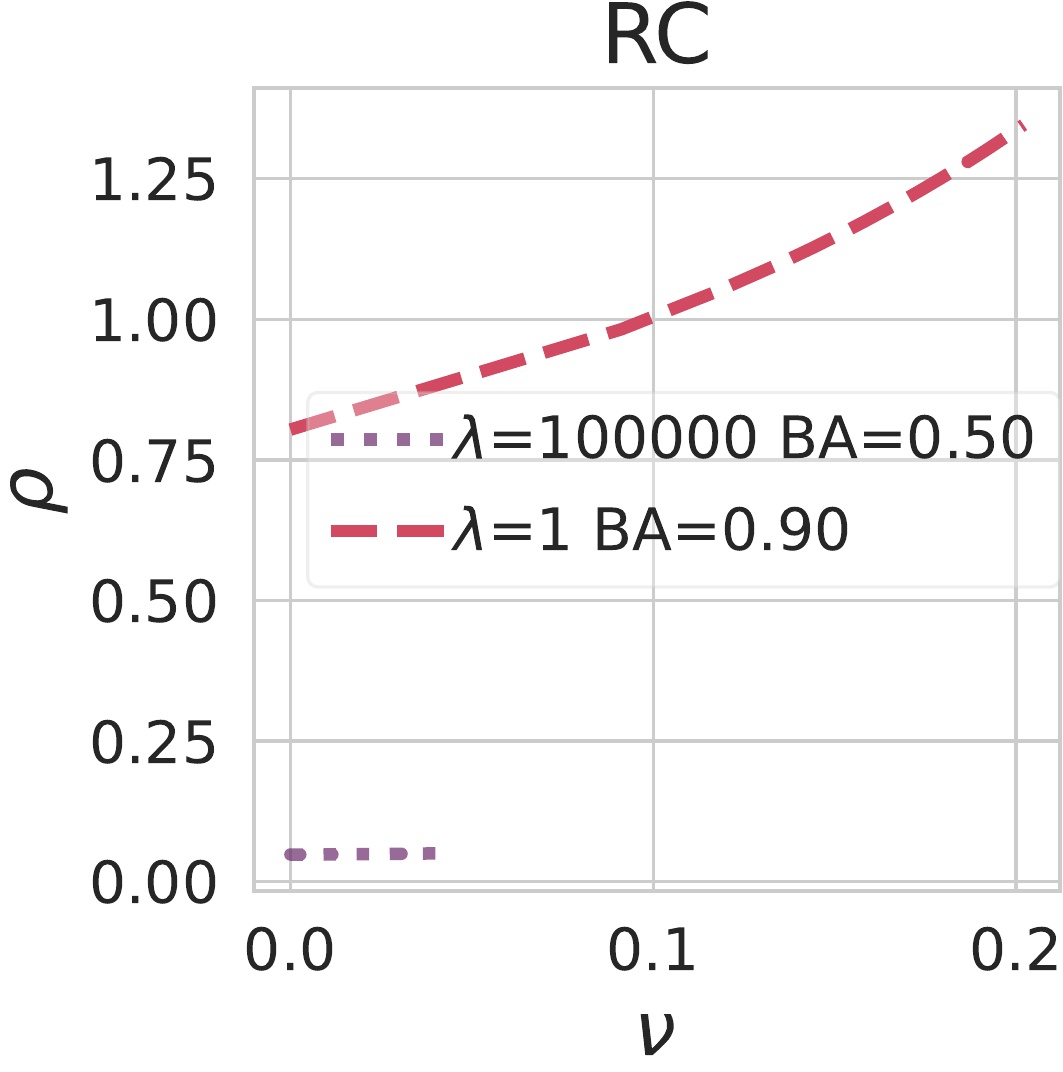}
  \includegraphics[width=0.225\textwidth]{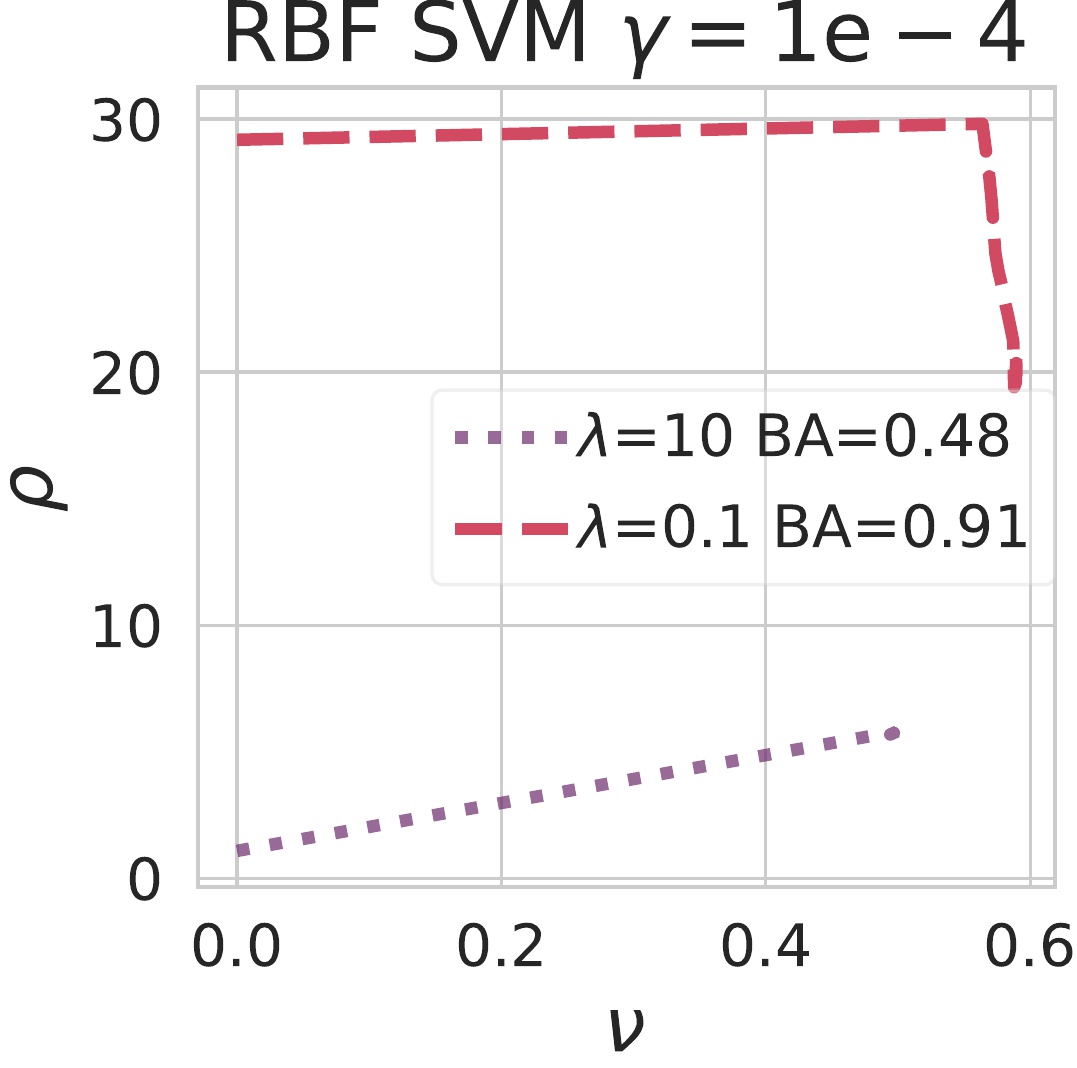}
 
  \includegraphics[width=0.24\textwidth]{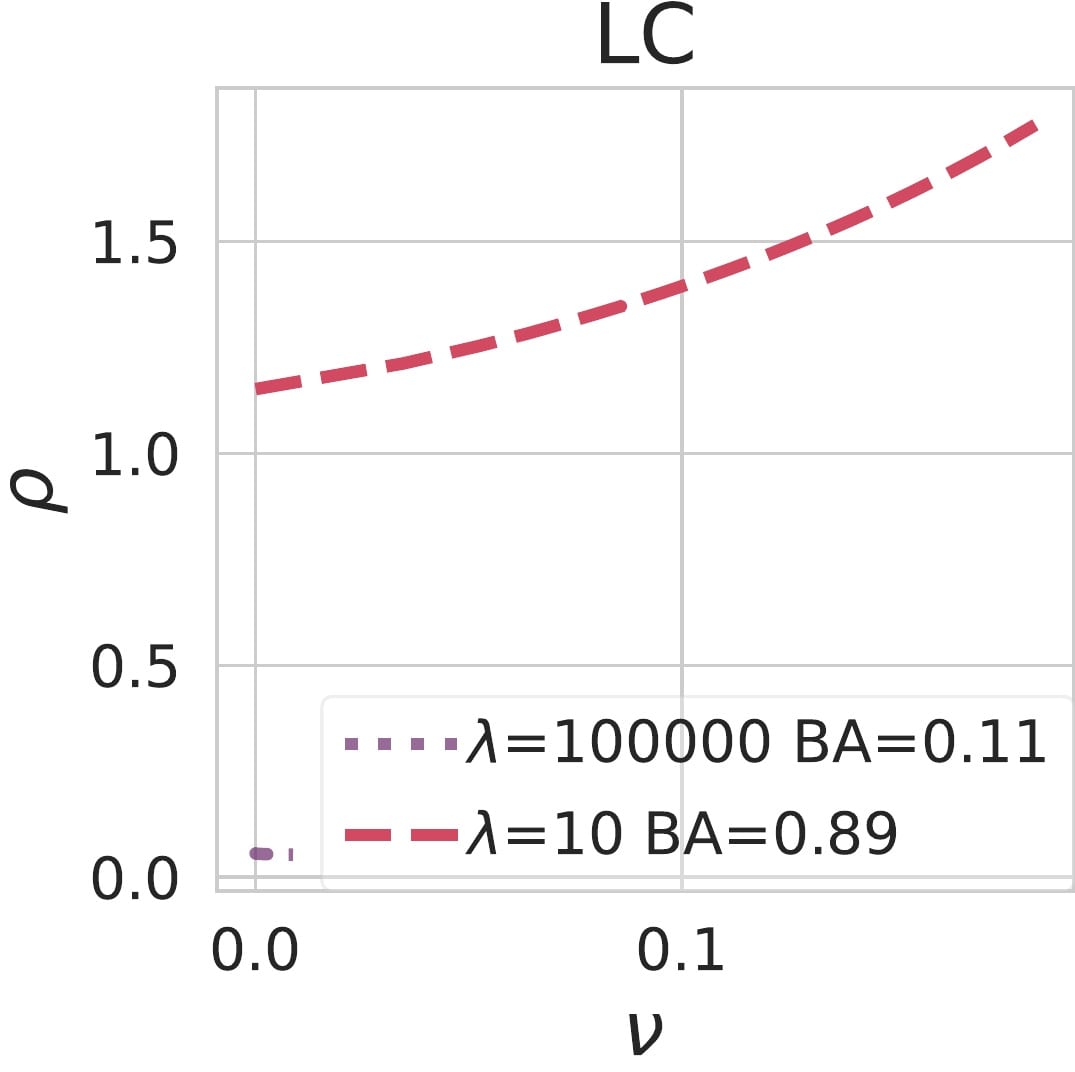}
  \includegraphics[width=0.235\textwidth]{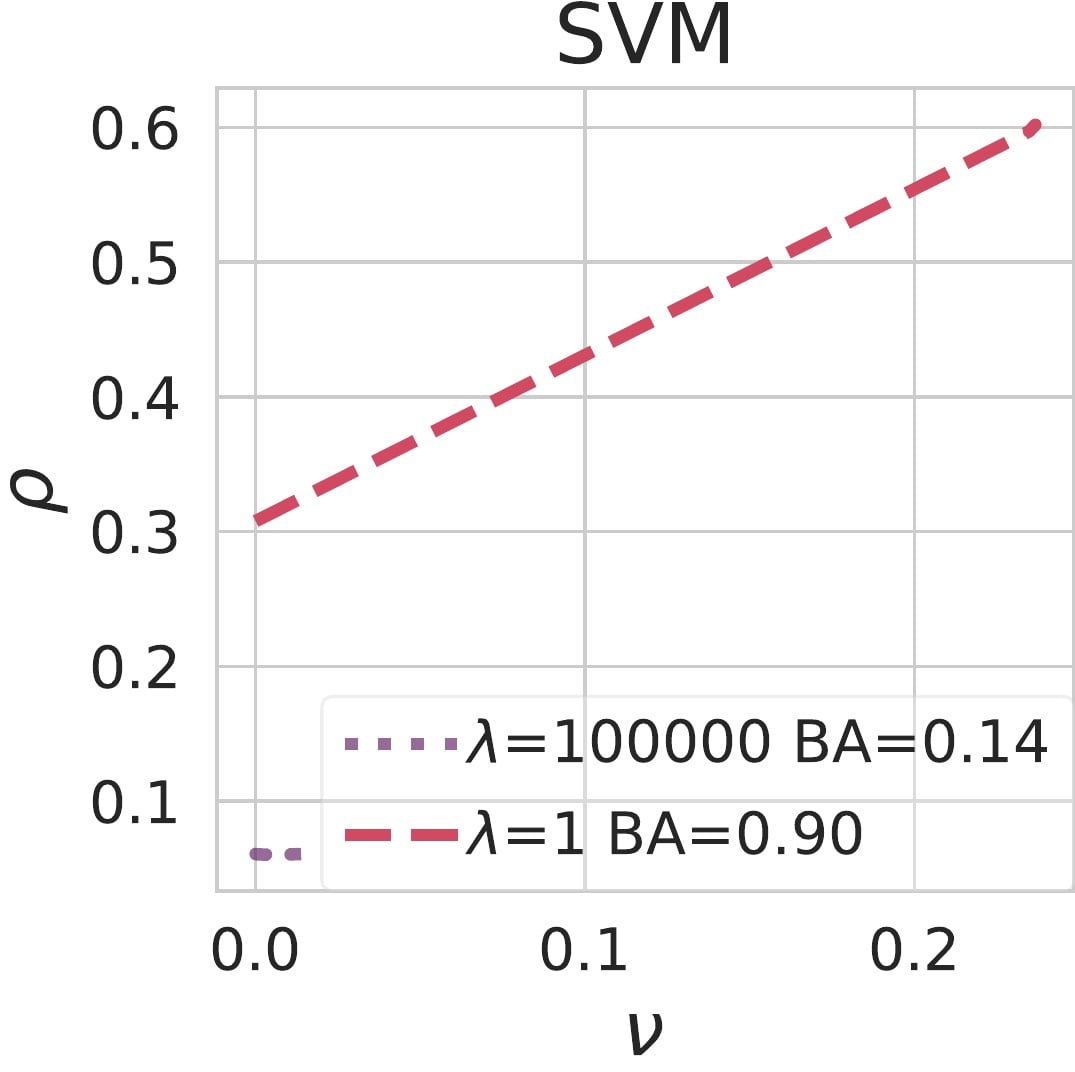}
  \includegraphics[width=0.235\textwidth]{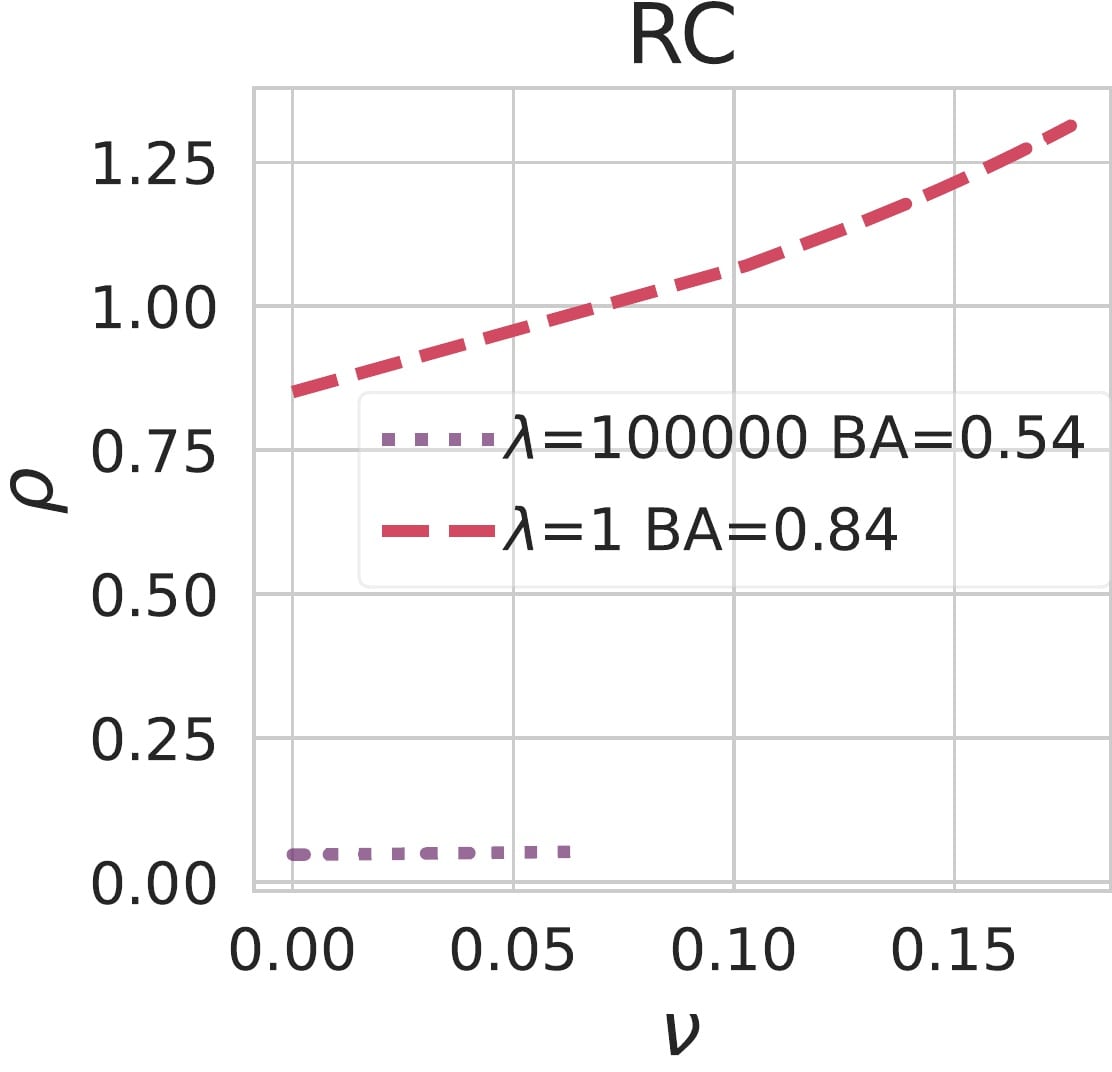}
  \includegraphics[width=0.225\textwidth]{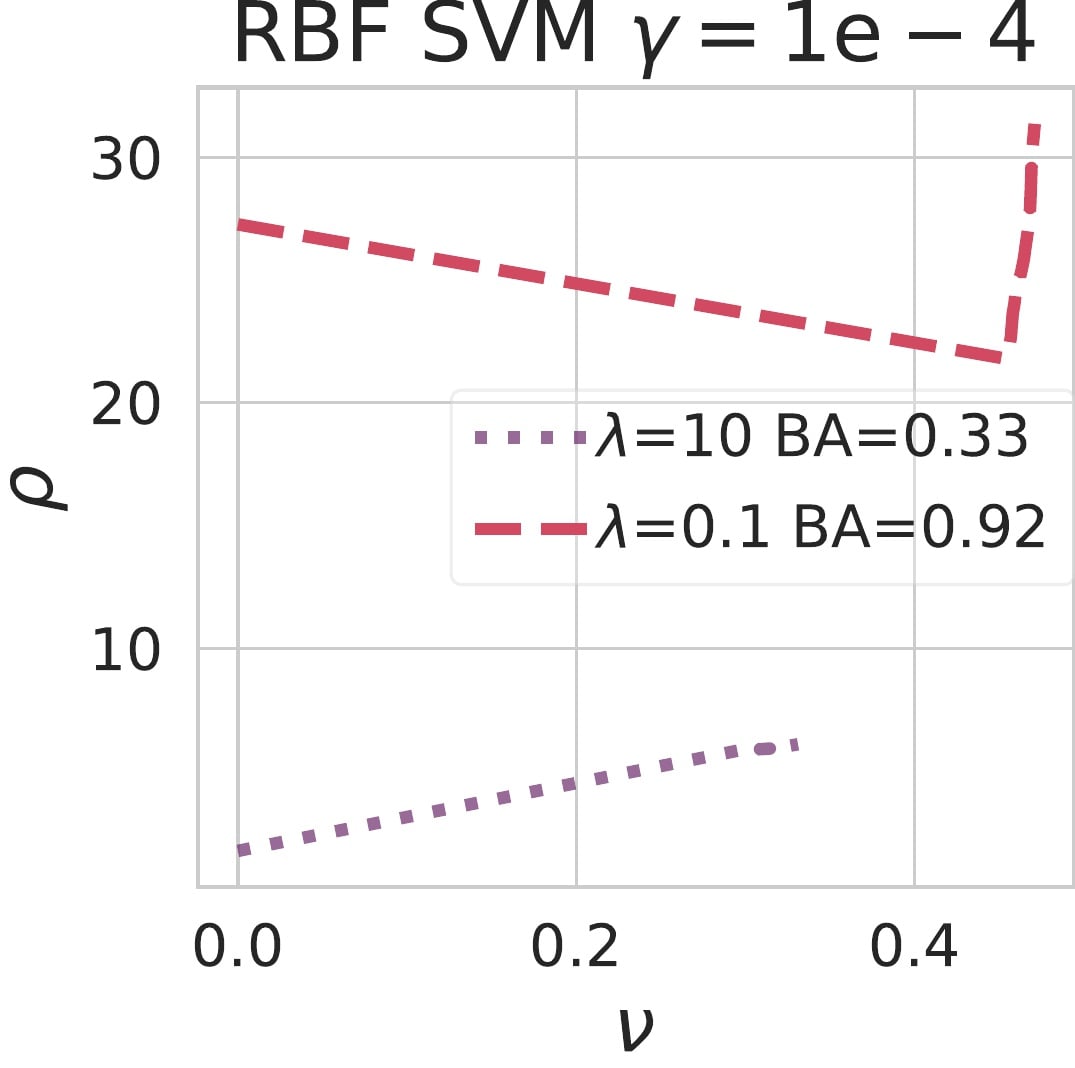}
  \caption{Backdoor weight deviation for different classifiers trained on Imagenette \imagenettetenchparachute (top), and \imagenetteplayerchurch (bottom). See Figure~\ref{fig:supplementary_backdoorParametersDeviationMNIST} for further details. }
  \label{fig:supplementary_backdoorParametersDeviationImagenette}
\end{figure*}

\myparagraph{Empirical Parameter Deviation Plots.}
In Figure~\ref{fig:supplementary_backdoorParametersDeviationMNIST}-\ref{fig:supplementary_backdoorParametersDeviationImagenette}, shows how the classifiers' parameters change when the classifiers learn the backdoors. This analysis is carried out with $p = 0.1$. The results do not vary significantly across different classifiers and class pairs. The only exception is MNIST $5~\rm{vs}~2$. The untainted classifier is already quite complex; therefore, it does not increase its complexity when it learns the backdoor.\smallskip

\begin{figure*}[t]
  \centering
  \begin{subfigure}{0.495\textwidth}
\includegraphics[width=0.99\textwidth]{ml_figs-mnist-incremental-3-0-incremental_backdoor_LC_resultsECML.jpeg}
      \caption{MNIST trigger size $3\times3$.}
  \end{subfigure}
    \begin{subfigure}{0.495\textwidth}
\includegraphics[width=0.99\textwidth]{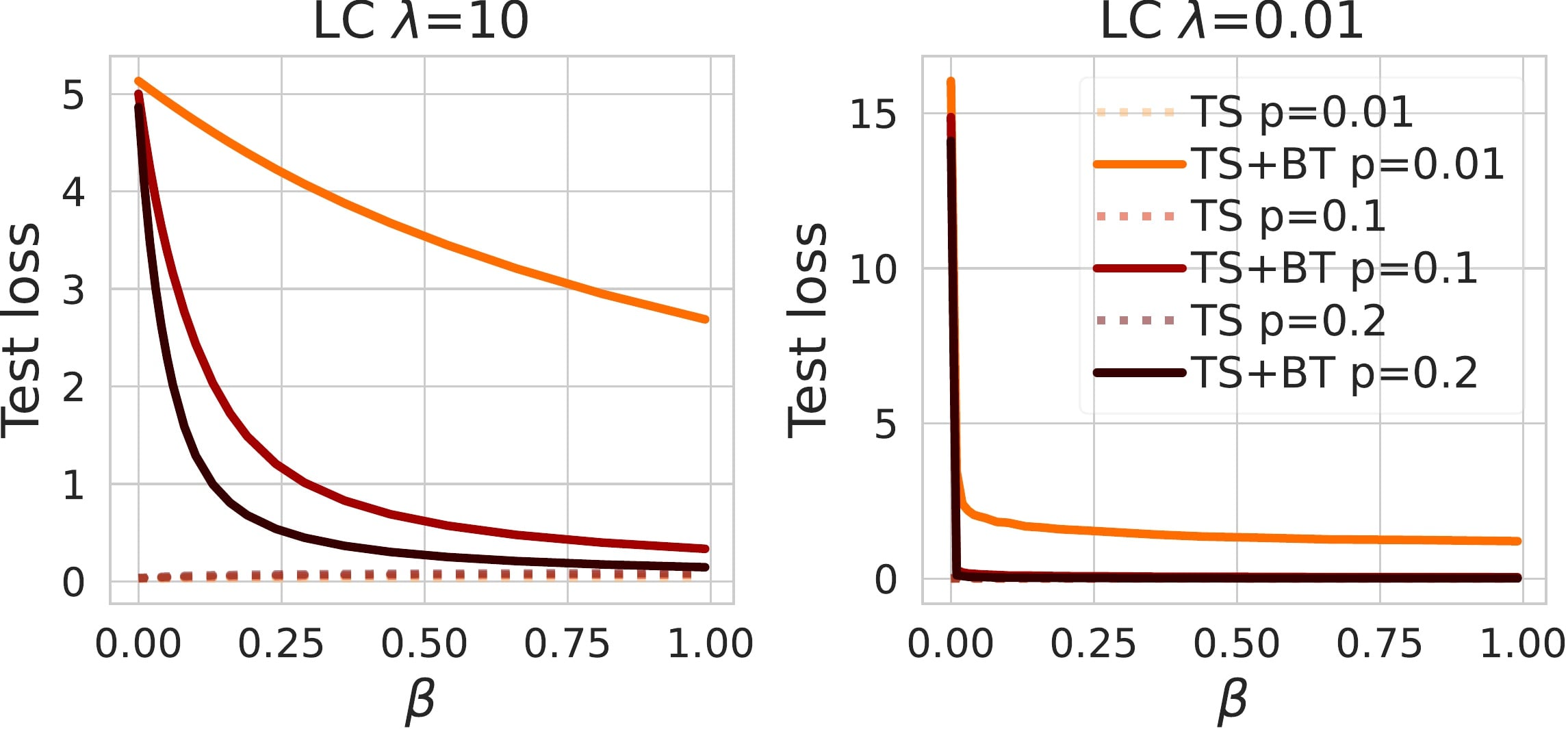}
      \caption{MNIST trigger size $6\times6$.}
  \end{subfigure}

  \begin{subfigure}{0.495\textwidth}
\includegraphics[width=1\textwidth]{ml_figs-cifar-incremental-0-9-incremental_backdoor_RC_resultsECML.jpeg}
      \caption{CIFAR10 trigger size $8\times8$.}
  \end{subfigure}
    \begin{subfigure}{0.495\textwidth}
\includegraphics[width=1\textwidth]{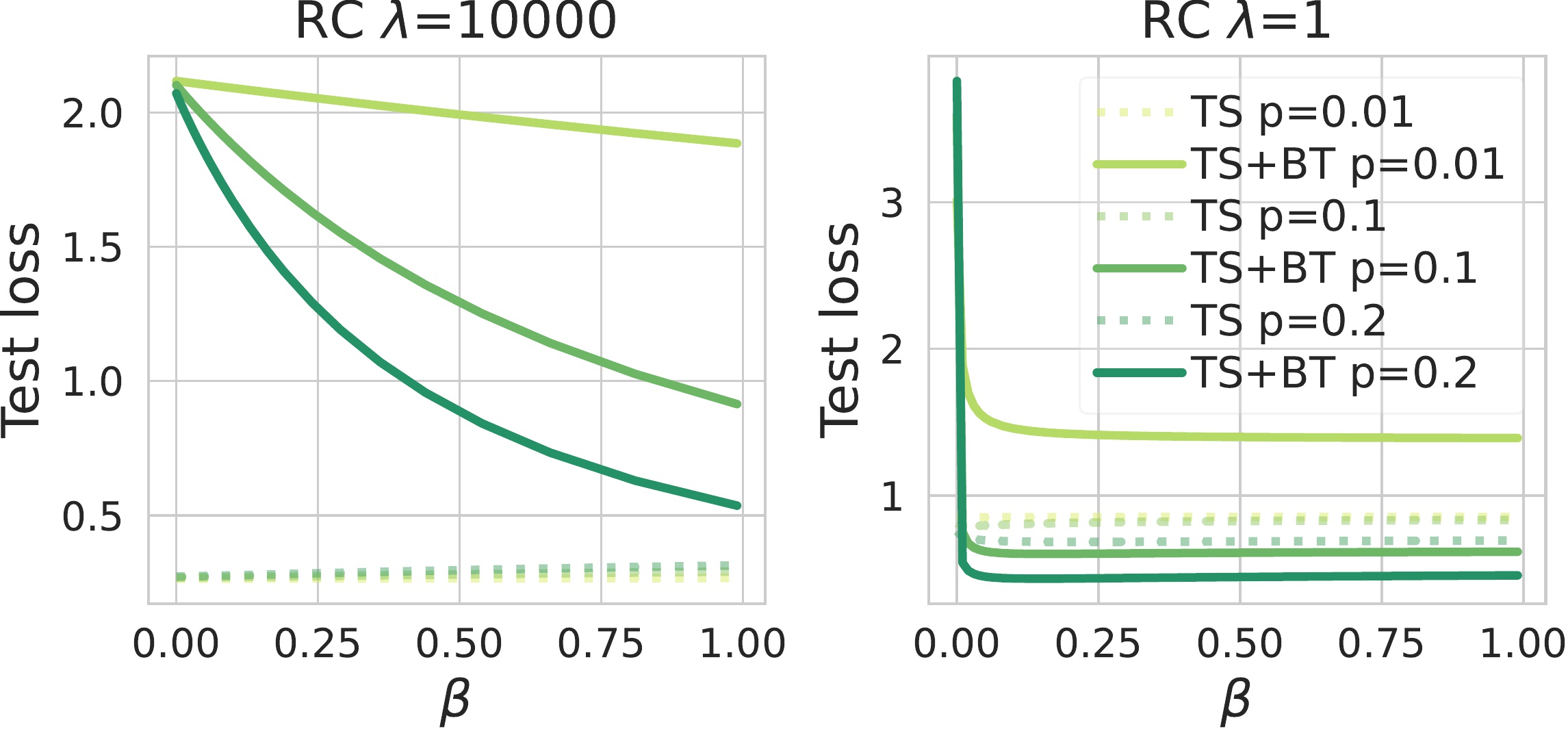}
      \caption{CIFAR10 trigger size $16\times 16$.}
  \end{subfigure}

  \begin{subfigure}{0.495\textwidth}
\includegraphics[width=0.99\textwidth]{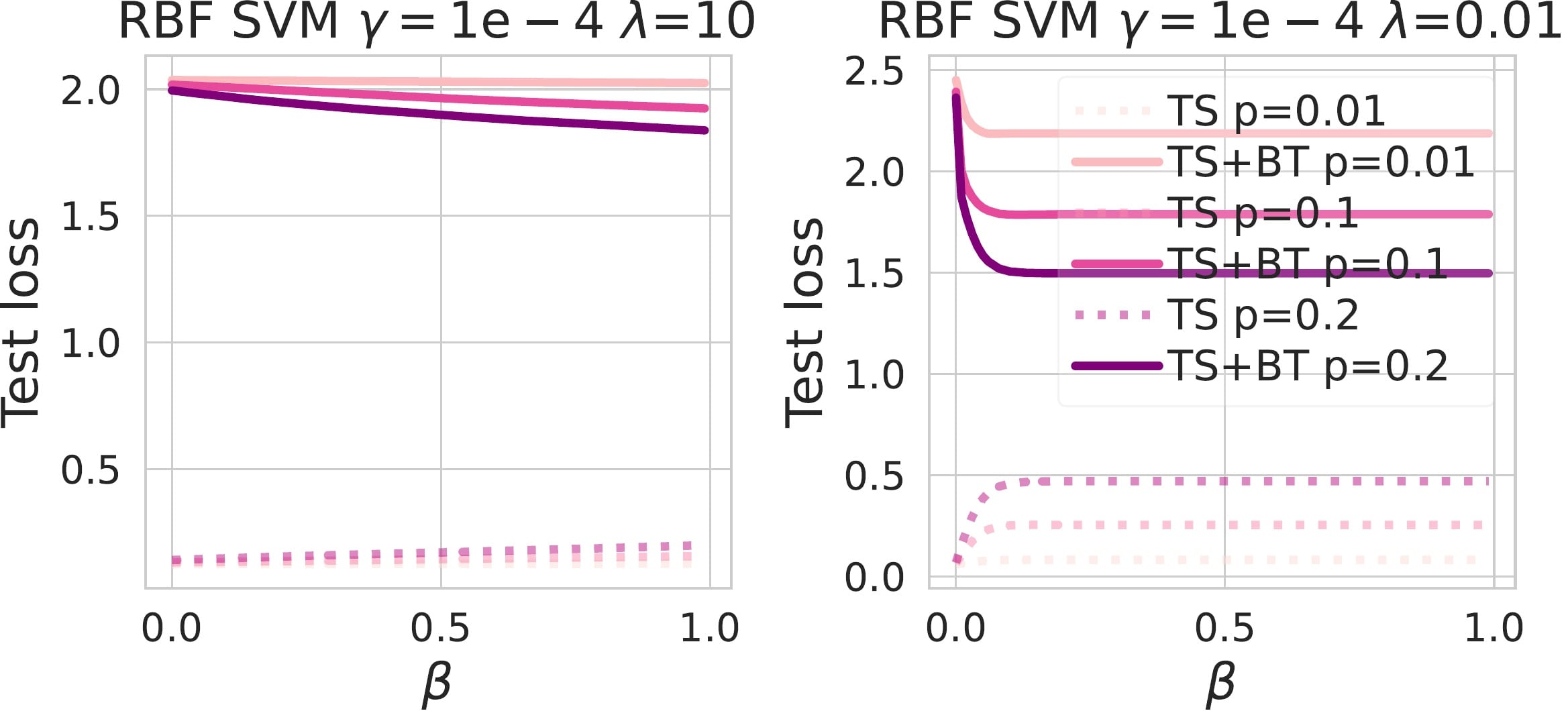}
      \caption{Imagenette trigger visibility $c_m=10$.}
  \end{subfigure}
    \begin{subfigure}{0.495\textwidth}
\includegraphics[width=0.99\textwidth]{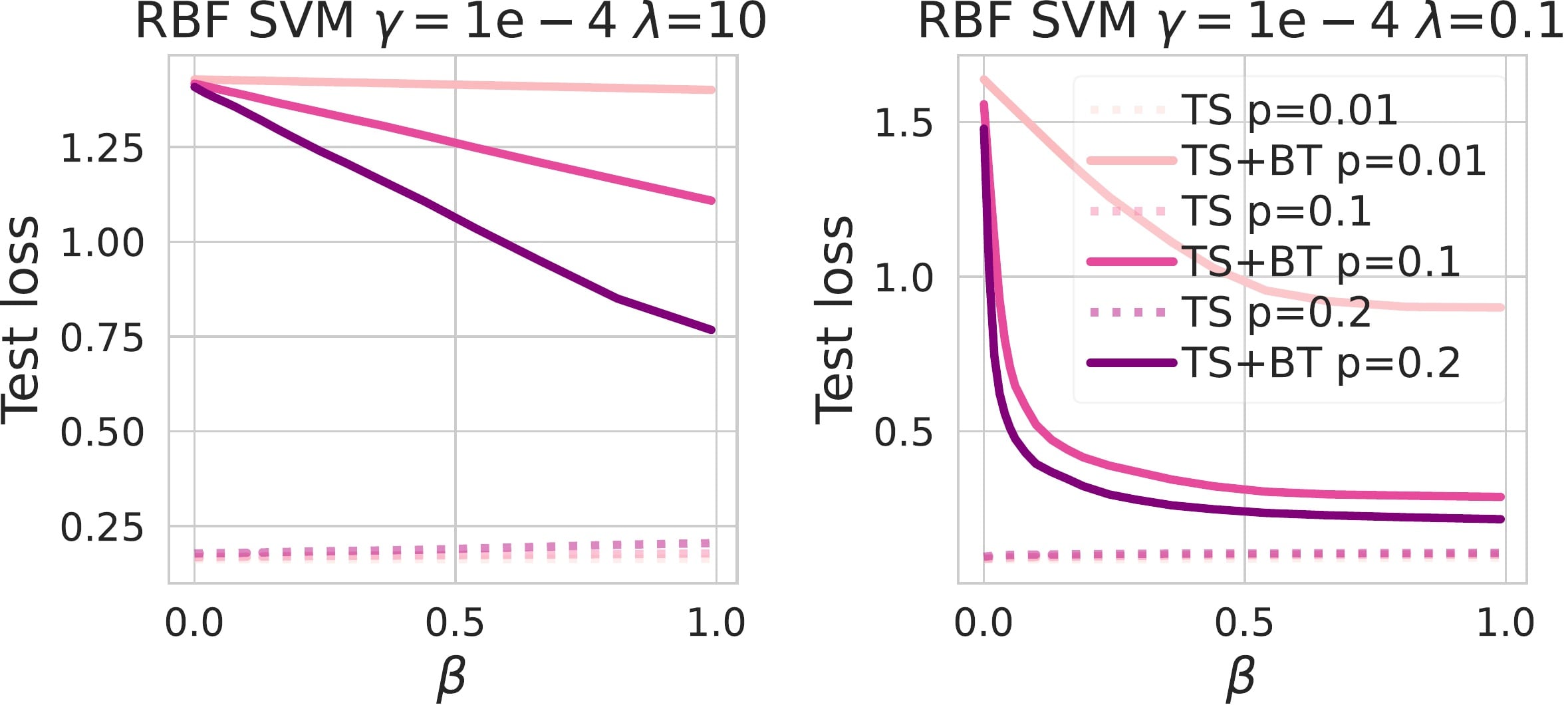}
      \caption{Imagenette trigger visibility $c_m=75$.}
  \end{subfigure}
   \caption{Backdoor learning curves for: (top row) LC on MNIST 3vs.0 with trigger size 3$\times$3 \textit{(left)} or 6$\times$6 \textit{(right)}; (middle row) RC on CIFAR10 \cifarairplanetruck with trigger size 8$\times$8 \textit{(left)} or 16$\times$16 \textit{(right)}; (bottom row) RBF SVM on Imagenette \imagenetteplayerchurch with trigger visibility $c_m$=10 \textit{(left)} or $c_m$=75 \textit{(right)}. 
   Further details in Figure~\ref{fig:appendix_backdoor_learning_curves_mnist30}.
   }
  \label{fig:trigger_size_learning_curves1}
\end{figure*}

\begin{figure*}[h!]
  \centering
  \begin{subfigure}[b]{0.495\textwidth}
\includegraphics[width=1\textwidth]{ml_figs-mnist-incremental-5-2-incremental_backdoor_LC_resultsECML.jpeg}
      \caption{MNIST trigger size $3\times3$.}
  \end{subfigure}
    \begin{subfigure}[b]{0.495\textwidth}
\includegraphics[width=1\textwidth]{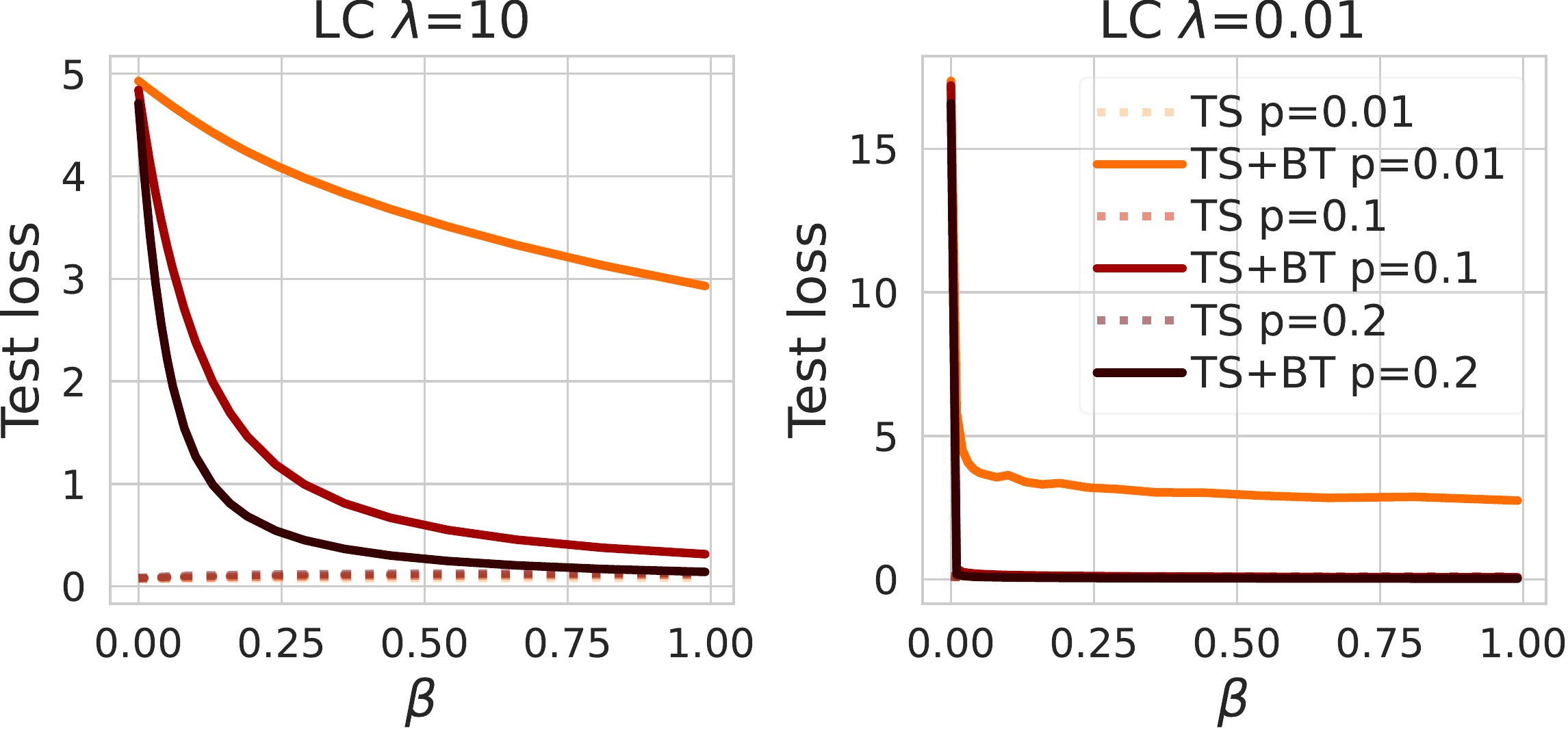}       
      \caption{MNIST trigger size $6\times6$.}
  \end{subfigure}

  \begin{subfigure}[b]{0.495\textwidth}
\includegraphics[width=1\textwidth]{ml_figs-cifar-incremental-2-5-incremental_backdoor_RC_resultsECML.jpeg}
      \caption{CIFAR10 trigger size $8\times8$.}
  \end{subfigure}
    \begin{subfigure}[b]{0.495\textwidth}
\includegraphics[width=1\textwidth]{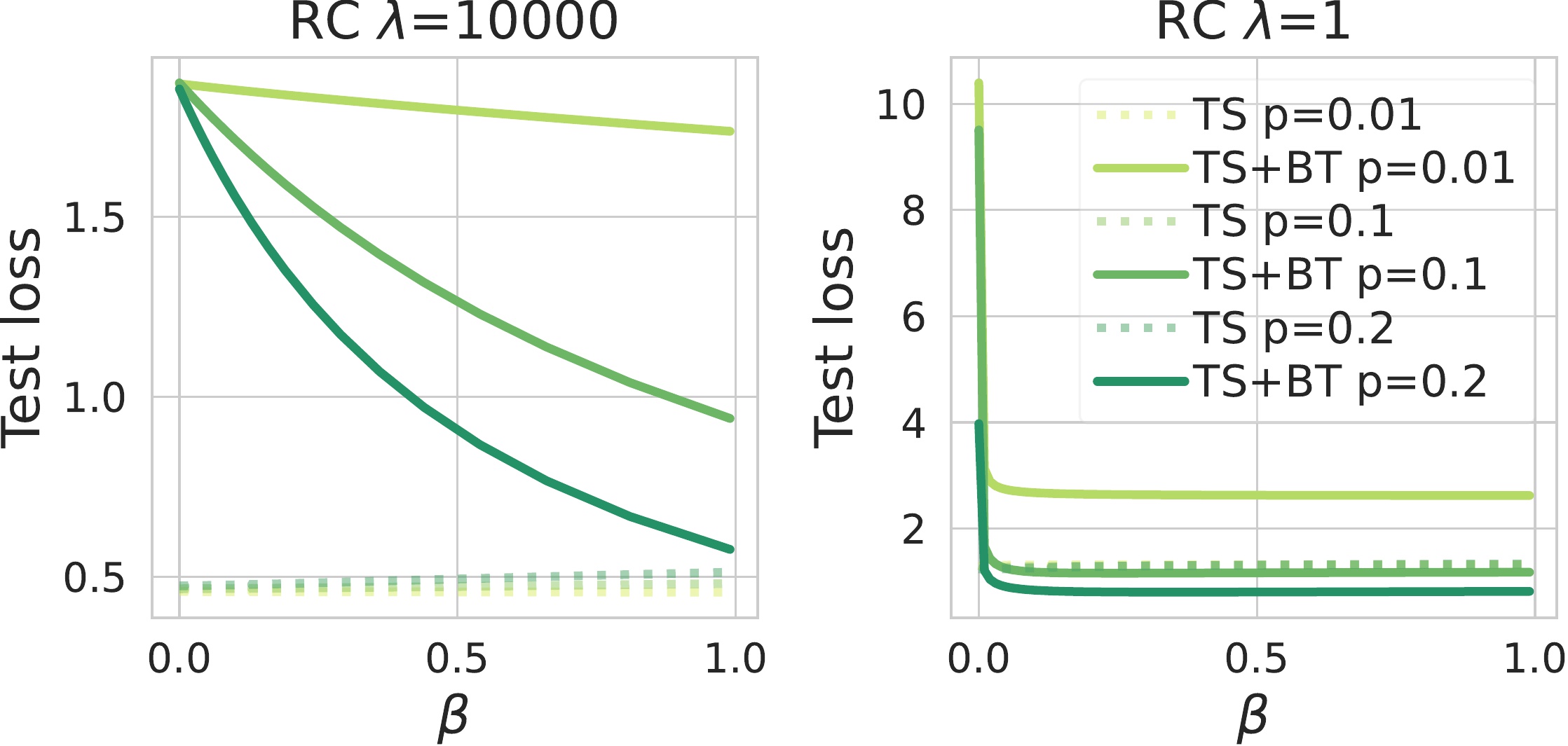}
      \caption{CIFAR10 trigger size $16\times 16$.}
  \end{subfigure}

  \begin{subfigure}[b]{0.495\textwidth}
\includegraphics[width=1\textwidth]{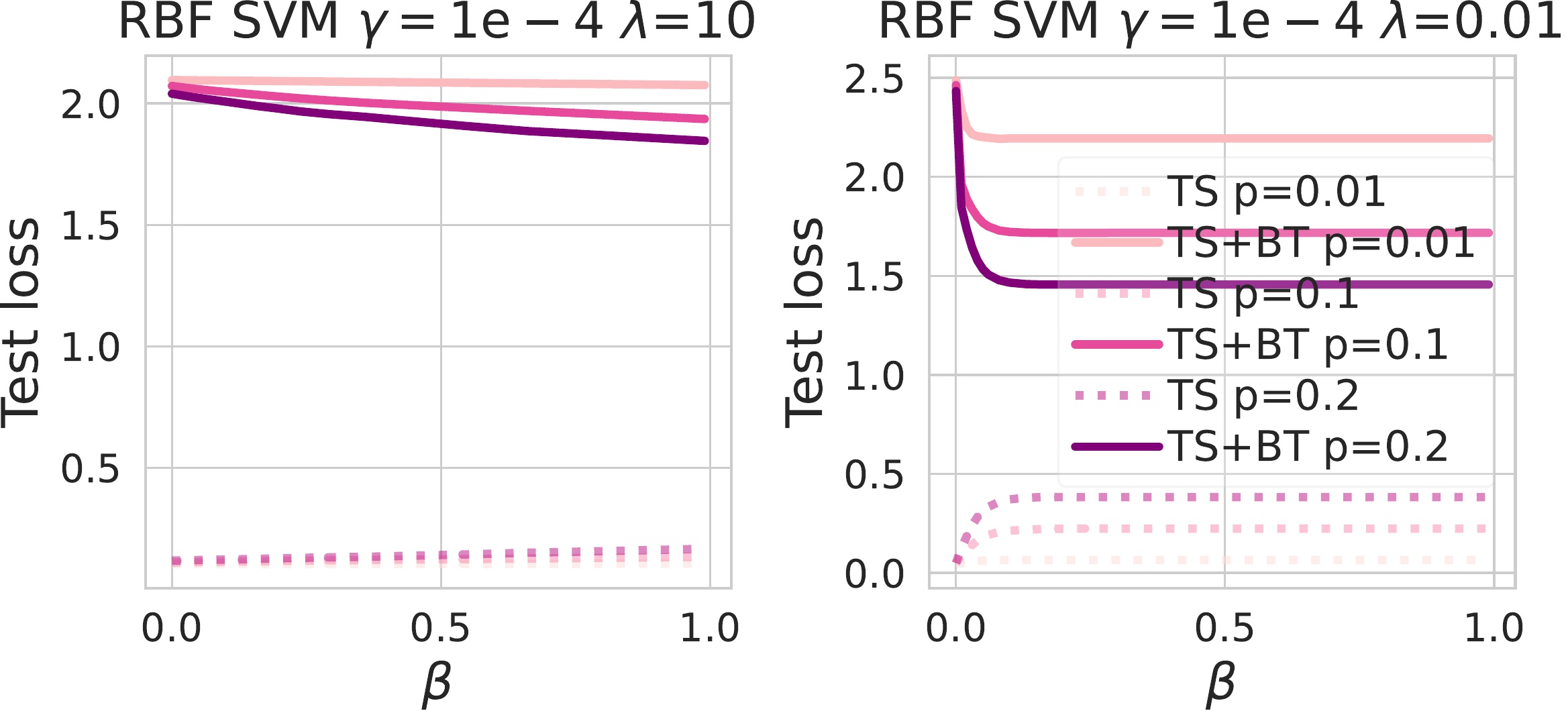}
      \caption{Imagenette trigger visibility $c_m=10$.}
  \end{subfigure}
    \begin{subfigure}[b]{0.495\textwidth}
\includegraphics[width=1\textwidth]{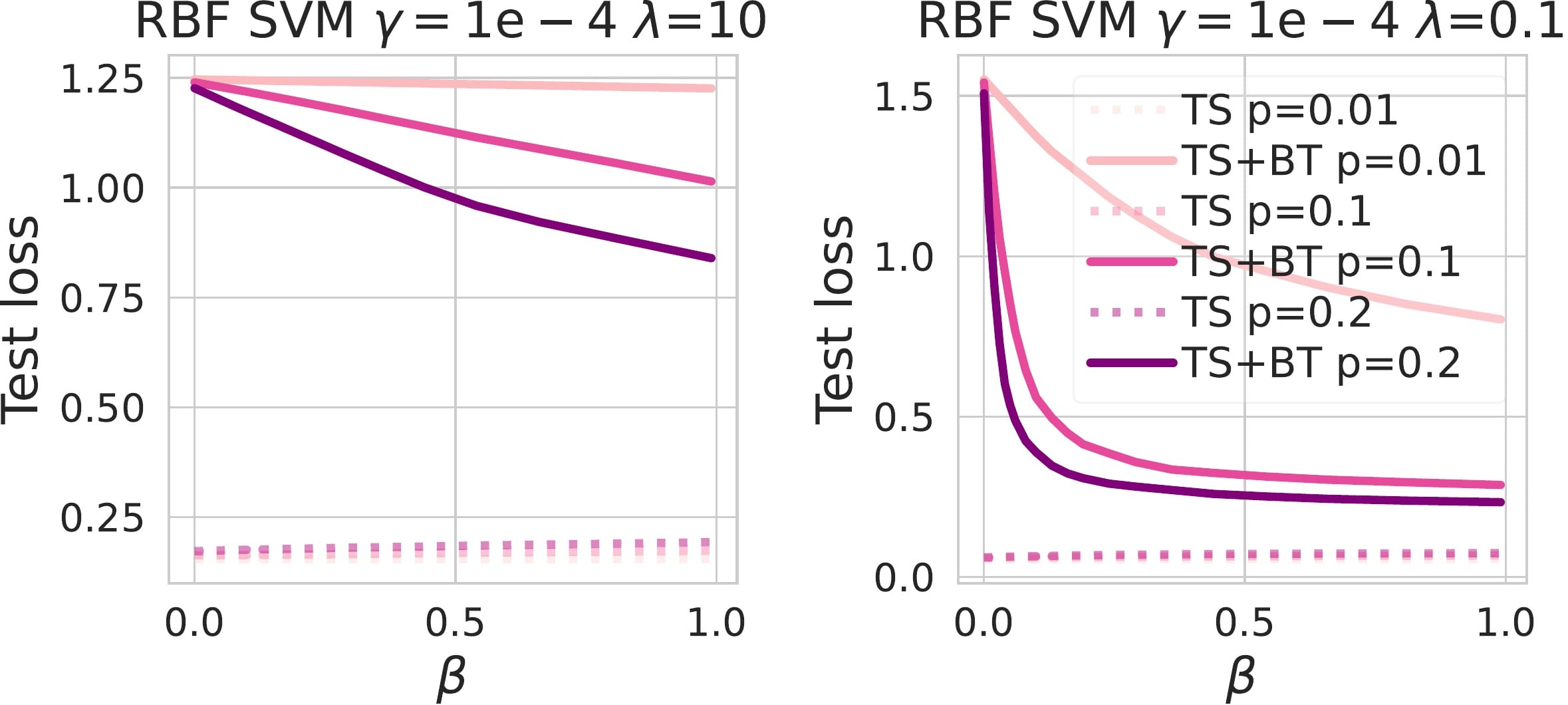}
      \caption{Imagenette trigger visibility $c_m=75$.}
  \end{subfigure}
   \caption{Backdoor learning curves for: (top row) LC on MNIST 5 vs. 2; (middle row) RC on CIFAR10 \cifarbirddog; (bottom row) RBF SVM on Imagenette \imagenettetenchparachute.
   }
  \label{fig:trigger_size_learning_curves2}
\end{figure*}

\myparagraph{Increasing the trigger size or visibility.}
Although it is a known result in the literature that the size of the trigger increases the effectiveness of the attack \cite{saha2020hidden, salem2020Dynamic}, here, for the first time to the best of our knowledge, we show how it interacts with other factors.
In this section we report further experimental results when increasing the trigger size or visibility.
As expected, the results in Fig.~\ref{fig:trigger_size_learning_curves1}-\ref{fig:trigger_size_learning_curves2} show that choosing a larger trigger enhances the effectiveness of the attack. Indeed, when the trigger is larger or more visible the backdoor learning curves go down faster. 
Using the proposed backdoor slope to analyze the effect of complexity, controlled via the hyperparameters, on the vulnerability against backdoors, we found a region of the hyperparameter space that leads to having desirable performances: an accuracy high on the clean test samples and low on the ones containing the backdoor trigger. \smallskip









\end{document}